\documentclass[]{fairmeta}

\title{MMGR: Multi-Modal Generative Reasoning Benchmark and Evaluation}

\author[*,1]{Zefan Cai}
\author[*,2]{Haoyi Qiu}
\author[*,3]{Tianyi Ma}
\author[*,4]{Haozhe Zhao}
\author[5]{Gengze Zhou}
\author[6]{Tingting Liao}
\author[7]{Xinyan Velocity Yu}
\author[8]{Kung-Hsiang Huang}
\author[9]{Ke Wan}
\author[10]{Shawn Lin}
\author[3]{Parisa Kordjamshidi}
\author[4]{Minjia Zhang}
\author[9]{Wen Xiao}
\author[11]{Jiuxiang Gu}
\author[2]{Nanyun Peng}
\author[1]{Junjie Hu}

\contribution[*]{Equal Contribution}

\affiliation[1]{University of Wisconsin--Madison}
\affiliation[2]{University of California, Los Angeles}
\affiliation[3]{Michigan State University}
\affiliation[4]{University of Illinois Urbana--Champaign}
\affiliation[5]{University of Adelaide}
\affiliation[6]{MBZUAI}
\affiliation[7]{University of Southern California}
\affiliation[8]{Salesforce AI Research}
\affiliation[9]{Microsoft}
\affiliation[10]{Kuma Stories Ltd}
\affiliation[11]{Adobe Research}

\abstract{Modern multimodal generative models can synthesize visually compelling images and videos, but it remains unclear whether this visual fluency reflects genuine \textit{reasoning}: when prompted to generate a solution, can a model preserve the physical, logical, spatial, and temporal constraints a task requires, or does it merely produce plausible-looking media? To answer this question, we introduce \textbf{MMGR} (\textbf{M}ulti-\textbf{M}odal \textbf{G}enerative \textbf{R}easoning Benchmark and Evaluation), a benchmark for evaluating generative reasoning across video, image, and language-based systems. MMGR covers \textit{10} tasks from \textit{three} domains (Abstract Reasoning, Embodied Navigation, and Physical Commonsense) and probes \textit{five} reasoning abilities: Physical, Logical, 2D Spatial, 3D Spatial, and Temporal. Its evaluation emphasizes answer-verifiable tasks and, for video generation, process-aware chain-of-frame reasoning, where intermediate frames must form valid steps toward the target outcome rather than visually smooth but incorrect transitions. Evaluating state-of-the-art video generators, image generators, and LLM/VLM baselines reveals a sharp gap between visual quality and reasoning correctness: video models perform best on Physical Commonsense, but remain weak on symbolic tasks such as Sudoku, ARC, and Math, and brittle in cross-view embodied navigation. Image generators often outperform video generators on embodied navigation despite lacking temporal outputs, showing that longer visual generation does not automatically yield stronger reasoning. MMGR reframes evaluation of multimodal generation from whether outputs look realistic to whether they solve the underlying reasoning problem.
}

\metadata[Website]{\url{https://zefan-cai.github.io/MMGR.github.io/}}
\metadata[Repo]{\url{https://github.com/Zefan-Cai/MMGR}}
\metadata[Dataset]{\url{https://huggingface.co/datasets/ZefanCai/MMGR}}
\metadata[Contact]{zefncai@gmail.com, haoyiqiu@g.ucla.edu, matiany3@msu.edu, haozhez6@illinois.edu}

\usepackage[utf8]{inputenc}
\usepackage{url}
\usepackage{array}
\usepackage{enumitem}
\usepackage{amsmath}
\usepackage{amssymb}
\usepackage{mathtools}
\usepackage{amsthm}
\usepackage{overpic}
\usepackage{textpos}
\usepackage{pifont}
\usepackage{adjustbox}
\usepackage{pgf}
\usepackage{tikz}
\usepackage{pgfplots}
\usepackage{makecell}
\usepackage{pgf-pie}
\usepackage{float}
\usepackage{wrapfig}
\usepackage{colortbl}
\usepackage{threeparttable}
\usepackage{xspace}
\usepackage[textsize=tiny]{todonotes}

\theoremstyle{plain}

\theoremstyle{definition}

\theoremstyle{remark}

\definecolor{citecolor}{HTML}{0071BC}
\definecolor{linkcolor}{HTML}{ED1C24}
\definecolor{acceptcolor}{HTML}{74C219}
\definecolor{rejectcolor}{HTML}{DE1616}
\definecolor{qcolor}{HTML}{536872}
\definecolor{demphcolor}{RGB}{100,100,100}
\definecolor{brightlavender}{rgb}{0.75, 0.58, 0.89}
\definecolor{palered}{rgb}{1.00, 0.70, 0.70}
\definecolor{palegreen}{rgb}{0.73, 0.96, 0.67}
\definecolor{paleblue}{rgb}{0.69, 0.84, 1.00}
\definecolor{paleorange}{rgb}{1.00, 0.86, 0.73}
\definecolor{palepurple}{rgb}{0.92, 0.85, 1.00}
\definecolor{paleyellow}{rgb}{1.00, 1.00, 0.50}

\newcommand{\hlblue}[1]{\setlength{\fboxsep}{2pt}\colorbox{paleblue}{#1}}
\newcommand{\hlyellow}[1]{\setlength{\fboxsep}{2pt}\colorbox{paleyellow}{#1}}
\newcommand{\hlgreen}[1]{\setlength{\fboxsep}{2pt}\colorbox{palegreen}{#1}}
\newcommand{\hlorange}[1]{\setlength{\fboxsep}{2pt}\colorbox{paleorange}{#1}}

\setlist[enumerate]{itemsep=-0.5mm,partopsep=0pt}

\begin{document}
\maketitle

\begin{figure*}[t!]
    \centering
    \includegraphics[width=\textwidth,clip,trim=10 800 10 0]{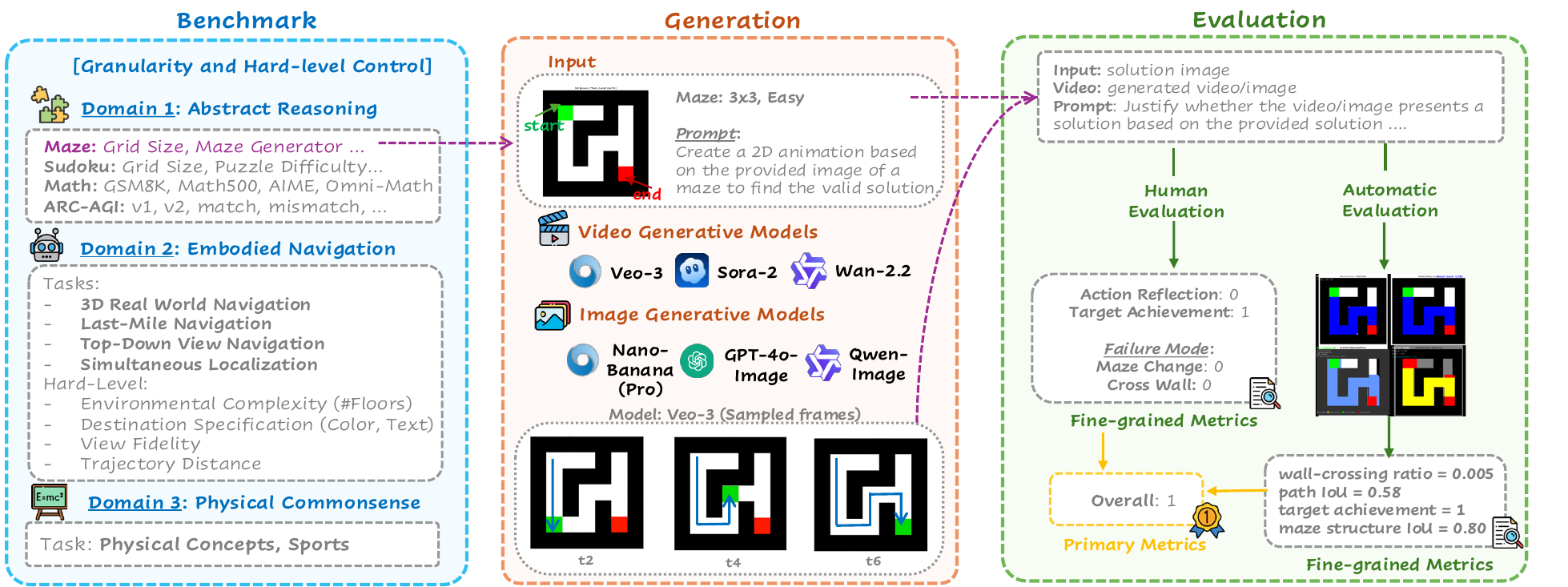}
    \vspace{-4mm}
    \caption{\textbf{Overview of MMGR.} \textbf{Benchmark:} MMGR assesses coherent reasoning across \textit{three} domains (Abstract Reasoning, Embodied Navigation, and Physical Commonsense), each with \textit{granular difficulty control}. \textbf{Generation:} given a task input and prompt (here, an Easy $3{\times}3$ maze), we prompt state-of-the-art video and image generative models (\textit{e.g.}, Veo-3) to produce a step-by-step solution. \textbf{Evaluation:} a hybrid pipeline scores each output through automatic evaluation (VLM-based judgment and deterministic rule-based checks) and human evaluation, yielding fine-grained metrics and a gated primary success score.
    }
    \label{fig:overview}
\end{figure*}

\section{Introduction}
\label{sec:intro}

Text-to-image and text-to-video systems now render photorealistic content directly from natural language prompts. Video generators synthesize coherent motion and temporally extended scenes~\citep{ho2022imagen,blattmann2023align,videoworldsimulators2024}, while image generators produce detailed compositions that often satisfy complex visual instructions~\citep{rombach2022high}. Yet visual fluency does not by itself establish reasoning ability. A generated video can look smooth while violating causal order, object permanence, or task constraints; likewise, a generated image can appear plausible while failing the logical or spatial relation required by the prompt. This motivates the central question of this paper: \textbf{can multimodal generative models \textit{reason} through the images and videos they produce, or do they mainly render plausible-looking outputs?}

Current evaluation protocols only partially answer this question. As summarized in \Cref{tab:benchmark_comparison}, video benchmarks emphasize visual quality, text-video alignment, compositional binding, or physical plausibility, while image-generation benchmarks test commonsense, spatial reasoning, editing, or generate-to-reason settings. These efforts are valuable but fragmented: most are modality-specific, outcome-only, and dependent on human preference, model-based judges, or checklist-style scoring. Even recent process-aware benchmarks remain video-only. As a result, existing evaluations struggle to tell whether a generated output solves the underlying reasoning problem, whether a video reaches the answer through valid intermediate states, and whether failures come from reasoning limits or modality-specific rendering constraints.\looseness=-1

\begin{table}[h!]
\centering
\small
\begin{adjustbox}{max width=\linewidth}
{
\setlength{\tabcolsep}{2pt}
\begin{tabular}{lcccccc}
\toprule
\textbf{Benchmark Task Domain} & \textbf{Phy.} & \textbf{Log.} & \textbf{3D Spa.} & \textbf{2D Spa.} & \textbf{Temp.} & \textbf{\# Samples} \\
\midrule
\multicolumn{7}{l}{\textit{\textbf{D1: Abstract Reasoning}}} \\
Maze (\Cref{apx:maze}) & & \checkmark & & \checkmark & \checkmark & $240$ \\
Sudoku (\Cref{apx:sudoku}) & & \checkmark & & \checkmark & & $300$ \\
ARC (\Cref{apx:arc}) & & \checkmark & & \checkmark & \checkmark & $456$ \\
Math (\Cref{apx:math}) & & \checkmark & & & & $327$ \\
\midrule
\multicolumn{7}{l}{\textit{\textbf{D2: Embodied Navigation}}} \\
Last-Mile Nav. (\Cref{apx:last_mile_navigation}) & \checkmark & & \checkmark & & \checkmark & $120$ \\
Top-down View Nav. (\Cref{apx:topdown}) & \checkmark & & & \checkmark & \checkmark & $120$ \\
3D R.-W. Nav. (\Cref{apx:3dnav}) & \checkmark & & \checkmark & & \checkmark & $120$ \\
SLAG (\Cref{apx:slag}) & \checkmark & & \checkmark & \checkmark & \checkmark & $120$ \\
\midrule
\multicolumn{7}{l}{\textit{\textbf{D3: Physical Commonsense}}} \\
Physical Concepts (\Cref{apx:physics}) & \checkmark & & \checkmark & & \checkmark & 25 \\
Sports (\Cref{apx:physics}) & \checkmark & & \checkmark & & \checkmark & 25 \\
\midrule
\multicolumn{6}{r}{\textbf{Total Evaluation Samples}} & \textbf{1,853} \\
\bottomrule
\end{tabular}
}
\end{adjustbox}
\caption{Overview of \textbf{MMGR}'s three task domains and their alignment with the five core reasoning abilities: Physical (Phy.), Logical (Log.), 3D Spatial (3D Spa.), 2D Spatial (2D Spa.), and Temporal (Temp.).}
\label{tab:benchmark_overview}
\end{table}

\begin{figure*}[t!]
    \centering
    \includegraphics[width=\textwidth,clip,trim=0 425 0 0]{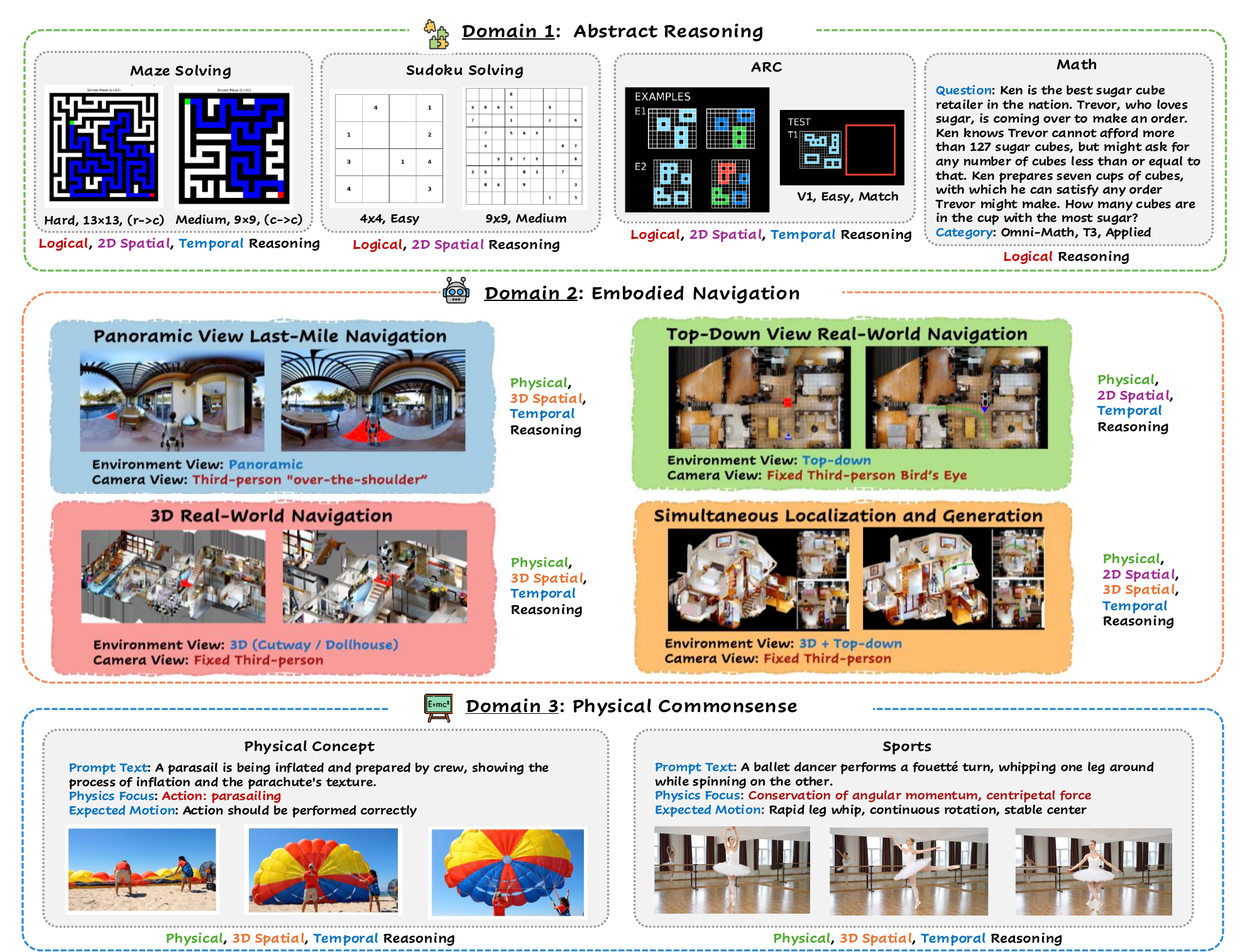}
    \vspace{-5mm}
    \caption{MMGR evaluates multimodal generative reasoning across: \textit{\textbf{D1: Abstract Reasoning}}, including Maze, Sudoku, ARC, and Math tasks. \textit{\textbf{D2: Embodied Navigation}}, spanning Panoramic View Last-Mile Navigation, Top-down View Real-World Navigation, 3D Real-World Navigation, and Simultaneous Localization and Generation. \textit{\textbf{D3: Physical Commonsense}}, covering Physical Concepts and Sports.}
    \label{fig:benchmark}
\end{figure*}

We introduce \textbf{MMGR} (\textbf{M}ulti-\textbf{M}odal \textbf{G}enerative \textbf{R}easoning Benchmark and Evaluation), a benchmark suite for evaluating video and image generators under a shared reasoning framework. MMGR spans \textbf{10} tasks across three domains (\Cref{tab:benchmark_overview,fig:overview,fig:benchmark}): \textbf{\underline{D1}: Abstract Reasoning}, covering \textit{Maze}, \textit{Sudoku}, \textit{Math}, and \textit{ARC}~\citep{chollet2019measure}; \textbf{\underline{D2}: Embodied Navigation}, covering panoramic, top-down, 3D real-world navigation, and \textit{Simultaneous Localization and Generation (SLAG)}~\citep{anderson2018vision,savva2019habitat,deitke2020robothor,HM3D}; and \textbf{\underline{D3}: Physical Commonsense}, covering basic physical interactions and sports scenarios~\citep{battaglia2013simulation,yi2019clevrer,bakhtin2019phyre,bear2021physion}. Together, these tasks probe five abilities: Physical, Logical, 2D Spatial, 3D Spatial, and Temporal reasoning.

MMGR is designed to be comprehensive along the axes missing from prior benchmarks. It compares video and image generators on the same task families and uses answer-verifiable scoring whenever reliable ground truth exists. For video generation, it further adopts a process-aware chain-of-frame criterion: intermediate frames must be valid steps toward the target outcome, not merely smooth transitions between plausible visual states. This lets MMGR ask whether temporal generation provides a real reasoning advantage or instead introduces new failures from maintaining cross-frame consistency.

Our evaluation of state-of-the-art video and image generators reveals a consistent gap between visual quality and reasoning correctness. Video models achieve their strongest performance on Physical Commonsense, with Sora-2 reaching 76.00\% on Physical Concepts and 64.00\% on Sports, but they remain weak on symbolic tasks, with all video models scoring 0.00\% on Sudoku, below 12\% on ARC, and below 11\% on Math. In Embodied Navigation, image generators obtain the strongest final task scores, with Nano-banana Pro leading all four navigation tasks, while video models remain brittle under cross-view alignment and long-horizon spatial grounding. These findings show that longer and more visually coherent generation does not automatically imply stronger reasoning. MMGR therefore reframes multimodal generation evaluation from asking whether generated images and videos look convincing to asking whether they solve the reasoning problems they depict.

\begin{table*}[t!]
\centering
\small
\setlength{\tabcolsep}{3pt}
\resizebox{0.9\textwidth}{!}{%
\begin{tabular}{@{}lcccccl@{}}
\toprule
\textbf{Benchmark family} & \textbf{Mod.} & \textbf{Process} & \textbf{Rule-ver.} & \textbf{V+I+L} & \textbf{Embod.} & \textbf{Primary scope} \\
\midrule
\rowcolor[gray]{0.92}\multicolumn{7}{@{}l}{\textit{Image-generation reasoning}} \\
WISE / Commonsense-T2I & I & $\times$ & $\times$ & $\times$ & $\times$ & Knowledge, physical, spatial T2I \\
RISEBench / KRIS-Bench & I & $\times$ & $\times$ & $\times$ & $\times$ & Reasoning-informed image editing \\
GIR-Bench / RBench-V & I & $\times$ & $\sim$ & $\times$ & $\times$ & Unified reason-and-generate \\
\midrule
\rowcolor[gray]{0.92}\multicolumn{7}{@{}l}{\textit{Generative-video evaluation}} \\
VBench / EvalCrafter & V & $\times$ & $\times$ & $\times$ & $\times$ & Quality, alignment, human preference \\
T2V-CompBench / VideoPhy & V & $\times$ & $\sim$ & $\times$ & $\times$ & Compositional, physical, world-model fidelity \\
\midrule
\rowcolor[gray]{0.92}\multicolumn{7}{@{}l}{\textit{Concurrent process-aware video reasoning}} \\
VIPER & V & \checkmark & $\times$ & $\times$ & $\times$ & Process--outcome consistency \\
VBVR & V & \checkmark & \checkmark & $\times$ & $\times$ & Rule-scored synthetic video reasoning \\
\midrule
\textbf{MMGR (ours)} & \textbf{V+I+L} & \checkmark & \checkmark$^\dagger$ & \checkmark & \textbf{\checkmark} & Abstract, embodied, physical; cross-modal \\
\bottomrule
\end{tabular}%
}
\vspace{-1mm}
\caption{MMGR versus representative and \textit{concurrent} generative-video and image-reasoning benchmarks. \emph{Mod.}: \textbf{V}ideo / \textbf{I}mage / \textbf{L}LM-VLM generators. \emph{Process}: scores intermediate steps, not only the final clip/image. \emph{Rule-ver.}: deterministic rule-based scoring vs.\ only human/VLM judgment. \emph{V+I+L}: covers all three generator types in one framework. \emph{Embod.}: real-world agent-centric navigation. \checkmark{} yes, $\times$ no, $\sim$ partial; $^\dagger$deterministic checkers where reliable rules/geometry exist, VLM judges otherwise.}
\vspace{-1mm}
\label{tab:benchmark_comparison}
\end{table*}

\section{Related Work}
\vspace{-1mm}

\paragraph{Evaluating Generative Models.}
Modern image and video generators can produce fluent, high-fidelity visual content from natural language prompts~\citep{rombach2022high,ho2022video,hong2022cogvideo,videoworldsimulators2024}, yet most benchmarks for these models evaluate only the final output. For video, early metrics such as FVD measure distributional similarity~\citep{unterthiner2019fvd}, while newer suites such as VBench, EvalCrafter, and T2V-CompBench score text--video alignment, motion quality, temporal consistency, and compositional binding~\citep{huang2024vbench,huang2024vbenchpp,zheng2025vbench2,liu2023evalcrafter,he2024videoscore,sun2025t2vcompbench}; a closer line probes physical plausibility and world-model fidelity, e.g.\ VideoPhy~\citep{videophy2024,bansal2025videophy2,meng2024phygenbench,li2025worldmodelbench,qin2024worldsimbench,ewmbench}, showing that visual realism and physical understanding can diverge, as Physics-IQ makes explicit~\citep{motamed2025physicsiq}. Image-generation benchmarks reveal the same limitation in another modality: commonsense, physical, world-knowledge, and spatial text-to-image benchmarks such as WISE and Commonsense-T2I ask whether an image satisfies facts or relations implied by the prompt~\citep{fu2024commonsenset2i,meng2024phybench,niu2025wise,chen2025r2ibench,zhang2025worldgenbench,sun2025t2ireasonbench,luo2025mmmg,wang2026spatialgeneval}, editing benchmarks such as RISEBench and KRIS-Bench require inferring the intended transformation before producing the target~\citep{zhao2025risebench,wu2025krisbench,yang2025complexedit,he2025reasontoedit}, and interleaved or reason-and-generate benchmarks such as GIR-Bench and RBench-V connect synthesis with problem solving~\citep{chen2024isg,zhou2024opening,liu2024interleavedbench,li2025girbench,guo2025rbenchv,li2025tirbench}. Across both modalities, however, evaluation collapses to a final clip, image, edit, or answer scored by holistic human, VLM, or checklist judgments, leaving the validity of the intermediate steps unchecked. A recent line instead evaluates the generation \textit{process} itself, not just its final output: chain-of-frame and video-as-reasoning studies ask whether intermediate frames behave like reasoning steps rather than decorative interpolation~\citep{wiedemer2025video,guo2025mmecof,tong2025thinking}, and concurrent benchmarks add rule-verifiable scoring, with VIPER checking process--outcome consistency beyond the last frame~\citep{li2025beyond} and VBVR building rule-checkable tasks from synthetic, parameterized generators~\citep{wang2026very}.

\vspace{-1mm}
\paragraph{From Visual Understanding to Generative Reasoning.}
A separate tradition evaluates visual reasoning in the \textit{discriminative} setting: video-understanding benchmarks test temporal, causal, and long-form reasoning over \textit{given} inputs~\citep{goyal2017something,girdhar2020cater,li2023mvbench,fu2024videomme,mangalam2023egoschema,wang2024lvbench,yu2025vrbench}, and embodied-AI benchmarks evaluate perception, navigation, and decision-making in fixed or simulator-grounded environments~\citep{anderson2018vision,savva2019habitat,HM3D,deitke2020robothor}. These probe abilities close to MMGR's, but in the opposite direction: the model observes an input and answers a question or selects an action, whereas MMGR asks whether a generator can make its reasoning visible through the output it produces; recognizing a correct trajectory in an existing video is not the same as generating one whose intermediate states stay valid under task rules. MMGR shares the process-aware, rule-verifiable scoring of the concurrent generative benchmarks above (deterministic symbolic, physical, or geometric checks, with a \textit{chain-of-frame} criterion for video), but differs in scope (\Cref{tab:benchmark_comparison}). First, it is cross-modal: the same tasks compare video, image, and LLM/VLM generators, which helps separate reasoning failures from modality-specific rendering failures, something a video-only suite cannot do. Second, it adds a real-world embodied-navigation domain that neither VIPER nor VBVR covers: panoramic, top-down, and 3D real-world navigation, together with SLAG. Together, these position MMGR as a cross-domain, cross-modal evaluation spanning \textit{Abstract Reasoning}, \textit{Embodied Navigation}, and \textit{Physical Commonsense}.\looseness=-1

\section{MMGR Benchmark}
\label{sec:benchmark_overview}
\vspace{-1mm}

MMGR organizes multimodal generative reasoning into \textit{three} complementary domains. \textbf{Abstract Reasoning} tests whether a model can preserve symbolic rules and spatial layouts while generating a solution. \textbf{Embodied Navigation} tests whether generated trajectories remain consistent with an environment, goal, and viewpoint. \textbf{Physical Commonsense} tests whether generated motion respects intuitive object dynamics. Together, these domains exercise the five reasoning abilities outlined in \Cref{sec:intro}: Physical, Logical, 2D Spatial, 3D Spatial, and Temporal reasoning. \Cref{tab:benchmark_overview} summarizes the mapping, and \Cref{fig:benchmark} provides examples.

\begin{table*}[!t]
\centering
\small
\begin{adjustbox}{max width=\textwidth}
{
\renewcommand{\arraystretch}{1.3}
\begin{tabular}{l m{19cm}}
\toprule
\textbf{Benchmark Task} & \multicolumn{1}{c}{\textbf{Success Criteria}} \\
\midrule
\multicolumn{2}{l}{\textit{\textbf{Domain 1: Abstract Reasoning}}} \\
Maze (\Cref{apx:maze_evaluation}) & A successful solution should \hlblue{not alter the maze}, \hlblue{cross no walls}, \hlgreen{reach the target}, and exhibit \hlorange{action reflection}. \\
Sudoku (\Cref{apx:sudoku_evaluation}) & A successful solution should \hlblue{keep the given clues unchanged}, \hlblue{violate no constraints}, \hlgreen{complete the grid accurately}, and exhibit \hlorange{action reflection}. \\
ARC (\Cref{apx:arc_evaluation}) & A successful solution should show correct \hlorange{pattern recognition}, maintain \hlblue{grid integrity}, and apply \hlgreen{accurate colors}, exactly matching the ground-truth output. \\
Math (\Cref{apx:math_evaluation}) & A successful solution should be correct in both its \hlorange{intermediate steps} and \hlgreen{final answer}, and exhibit \hlorange{action reflection}. \\
\midrule
\multicolumn{2}{l}{\textit{\textbf{Domain 2: Embodied Navigation}}} \\
Last-Mile Nav. (\Cref{apx:last_mile_navigation}) & \multirow{4}{*}{\parbox[t]{12cm}{A successful episode must achieve \hlgreen{task completeness} (success, oracle success, and trajectory alignment), demonstrate \hlblue{physical understanding} (object semantics, agent consistency, and spatial alignment), and maintain \hlblue{instruction following} (destination integrity and scene consistency).}} \\
Top-down View Nav. (\Cref{apx:topdown}) & \\
3D R.-W. Nav. (\Cref{apx:3dnav}) & \\
SLAG (\Cref{apx:slag}) & \\
\midrule
\multicolumn{2}{l}{\textit{\textbf{Domain 3: Physical Commonsense}}} \\
\makecell[l]{Physical Concepts \& \\ Sports (\Cref{apx:physics})} & A successful generation should exhibit \hlgreen{physics accuracy}, \hlblue{motion quality}, \hlblue{visual realism}, and \hlblue{prompt adherence}. \\
\bottomrule
\end{tabular}
}
\end{adjustbox}
\vspace{-2mm}
\caption{Evaluation criteria for each task in MMGR. Each entry states what a successful sample must satisfy; highlighted phrases mark the \textit{fine-grained} metrics, color-coded by role: \hlblue{integrity/constraint}, \hlgreen{outcome/correctness}, and \hlorange{reasoning process}, and a sample receives the binary \textit{gated primary} score of $1$ only when all required criteria hold. Full metric definitions appear in the cited appendices.}
\vspace{-2mm}
\label{tab:evaluation_metrics}
\end{table*}

\vspace{-1mm}
\subsection{Domain 1: Abstract Reasoning}
\vspace{-1mm}

\textbf{\underline{T1}: Maze} (\Cref{apx:maze}) evaluates 2D spatial, logical, and temporal reasoning by requiring an obstacle-free path from start to goal, using 240 DFS and Wilson mazes across three difficulty levels and four start-goal configurations~\citep{maze-dataset}. \textbf{\underline{T2}: Sudoku} (\Cref{apx:sudoku}) evaluates constraint satisfaction on 300 puzzles across 4x4 and 9x9 grids with difficulty controlled by clue sparsity~\citep{seely2025sudoku}. \textbf{\underline{T3}: ARC} (\Cref{apx:arc}) evaluates abstract rule induction from input-output demonstrations using 456 ARC v1 and v2 tasks~\citep{chollet2019measure}. \textbf{\underline{T4}: Math} (\Cref{apx:math}) evaluates mathematical reasoning on 327 problems from GSM8K, MATH500, AIME 2024/2025, and Omni-MATH~\citep{cobbe2021gsm8k,hendrycks2021math,aime2024,aime2025,omnimath2024}.

\vspace{-1mm}
\subsection{Domain 2: Embodied Navigation}
\vspace{-1mm}

\textbf{\underline{T5}: Panoramic View Last-Mile Navigation} (Last-Mile Nav.; \Cref{apx:last_mile_navigation}) uses Matterport3D panoramas to test short-range goal-directed navigation from a 360$^\circ$ over-the-shoulder view~\citep{Matterport3D}. \textbf{\underline{T6}: Top-down View Real-World Navigation} (Top-down View Nav.; \Cref{apx:topdown}) uses HM3D bird's-eye maps to test global 2D planning~\citep{HM3D}. \textbf{\underline{T7}: 3D Real-World Navigation} (3D R.-W. Nav.; \Cref{apx:3dnav}) uses HM3D dollhouse views to test multi-room and multi-level spatial reasoning~\citep{HM3D}. \textbf{\underline{T8}: Simultaneous Localization and Generation} (SLAG; \Cref{apx:slag}) combines 3D and top-down views, requiring cross-view localization and scene generation. Each task provides 120 samples spanning 24 hard-level configurations that vary floor complexity, view fidelity, path length, and goal specification.

\vspace{-1mm}
\subsection{Domain 3: Physical Commonsense}
\vspace{-1mm}

\textbf{\underline{T9}: Physical Concepts} (\Cref{apx:physics}) samples 25 VideoPhy examples spanning solid-solid, solid-fluid, and fluid-fluid interactions~\citep{videophy2024}. \textbf{\underline{T10}: Sports} (\Cref{apx:physics}) samples 25 compositional physical scenarios from activities such as ballet, skiing, diving, and swimming, where motion depends on balance, momentum, projectile motion, and fluid dynamics.

\section{Evaluation}
\vspace{-1mm}

To systematically evaluate \textbf{zero-shot reasoning} capabilities~\citep{wiedemer2025video,guo2025mmecof,tong2025thinking}, we benchmark state-of-the-art generative models across the \textbf{10} tasks outlined in \Cref{sec:benchmark_overview}, aiming to quantify overall performance while identifying strengths and weaknesses across the five core reasoning dimensions.

\subsection{Evaluated Models and Setup}
\vspace{-1mm}

We evaluate video generators (\textit{Veo-3}, \textit{Sora-2}, and \textit{Wan-2.2}) against image generators (\textit{Nano-banana}, \textit{Nano-banana Pro}, \textit{GPT-4o-image}, \textit{GPT-image-1.5}, and \textit{Qwen-image}). On the applicable abstract-reasoning tasks, we additionally report \textit{Gemini-3-Flash} and \textit{Gemini-3-Pro} as text-solution baselines. To account for sampling variability, we draw five outputs per prompt from each generative model, using default settings for closed models and recommended settings for open ones, without task-specific tuning.

\begin{table*}[!t]
\centering
\small
\setlength{\tabcolsep}{4.5pt}
\begin{adjustbox}{max width=\textwidth}
{
\begin{tabular}{lcccccccccc}
\toprule
& \multicolumn{3}{c}{\textbf{Video Models}} & \multicolumn{5}{c}{\textbf{Image Models}} & \multicolumn{2}{c}{\textbf{LLM Baselines}} \\
\cmidrule(lr){2-4} \cmidrule(lr){5-9} \cmidrule(lr){10-11}
\textbf{Task} & Veo-3 & Sora-2 & Wan-2.2 & Nano-banana & Nano-banana Pro & GPT-4o-image & GPT-image-1.5 & Qwen-image & Gemini-3-Flash & Gemini-3-Pro \\
\midrule
\multicolumn{11}{l}{\textbf{\textit{D1: Abstract Reasoning}}} \\
Maze & 3.67 & 0.00 & 0.00 & 3.67 & 9.17 & 0.00 & 0.97 & 0.14 & 55.67 & \textbf{59.33} \\
Sudoku & 0.00 & 0.00 & 0.00 & 8.60 & 7.00 & 1.33 & 6.78 & 0.00 & \textbf{65.53} & 63.75 \\
ARC & 4.80 & 11.67 & 0.15 & 8.15 & \textbf{28.07} & 0.00 & 12.72 & 2.15 & -- & -- \\
Math & 10.61 & 10.56 & 0.00 & 11.47 & 72.69 & 26.24 & 25.52 & 11.22 & 71.38 & \textbf{74.14} \\
\midrule
\multicolumn{11}{l}{\textbf{\textit{D2: Embodied Navigation}}} \\
Last-Mile Nav. & 60.00 & 0.00 & 14.17 & 74.17 & \textbf{75.83} & 0.00 & 55.84 & 16.67 & -- & -- \\
Top-down View Nav. & 19.49 & 3.39 & 5.09 & 11.11 & \textbf{33.05} & 3.39 & 26.27 & 5.08 & -- & -- \\
3D R.-W. Nav. & 22.50 & 0.00 & 24.17 & 79.17 & \textbf{85.00} & 13.33 & 77.92 & 38.33 & -- & -- \\
SLAG & 11.02 & 12.50 & 0.85 & 28.79 & \textbf{37.29} & 16.67 & 31.36 & 6.78 & -- & -- \\
\midrule
\multicolumn{11}{l}{\textbf{\textit{D3: Physical Commonsense}}} \\
Physical Concepts & 41.67 & \textbf{76.00} & 26.67 & -- & -- & -- & -- & -- & -- & -- \\
Sports & 60.00 & \textbf{64.00} & 21.33 & -- & -- & -- & -- & -- & -- & -- \\
\bottomrule
\end{tabular}
}
\end{adjustbox}
\vspace{-2mm}
\caption{\textbf{Zero-shot reasoning performance of state-of-the-art generative models on MMGR.} Each cell reports a single task-level final score (\%), aggregated over all difficulty levels and subsets and computed with the primary metric in \Cref{tab:evaluation_metrics}; Maze and Sudoku generation scores use deterministic pixel- and OCR-based evaluators, respectively. The best score per task is shown in bold. Gemini baselines are direct text-solver scores on the abstract reasoning tasks, and ``--'' marks a model that was not evaluated or for which no usable result file was available. Fine-grained results are provided in the Appendix.}
\vspace{-2mm}
\label{tab:main_results}
\end{table*}

\subsection{Evaluation Protocol}
\vspace{-1mm}

Each task receives a primary success score from the criteria in \Cref{tab:evaluation_metrics}. Rather than one holistic judgment, each task's criteria are decomposed into \textit{fine-grained} metrics with three roles: integrity and constraint preservation, outcome correctness, and validity of the reasoning process. The primary score is \textit{gated}: it equals $1$ only when \emph{all} required metrics hold, so an output that looks plausible but violates a single constraint still scores $0$. The metrics are task-specific. In Abstract Reasoning, a successful Maze solution must leave the input maze unaltered, cross no walls, and reach the target, while a Sudoku solution must keep the given clues fixed, satisfy every row, column, and box constraint, and complete the grid. In Embodied Navigation, an episode must jointly satisfy task completeness (success, oracle success, and trajectory alignment), physical understanding (object semantics, agent consistency, and spatial alignment), and instruction following (destination integrity and scene consistency). In Physical Commonsense, a generation must combine physics accuracy with motion quality, visual realism, and prompt adherence. Video outputs are scored for both final outcome and, where relevant, intermediate-frame validity, whereas image outputs are scored against the same task-level constraints at the final output alone.

\vspace{-1mm}
\paragraph{Deterministic Evaluation.} For tasks with exact ground truth, we use rule-based checks as the main generation scores in \Cref{tab:main_results}. Maze uses pixel-based path reconstruction to check whether the generated path preserves the maze layout, avoids walls, and reaches the target. Sudoku uses OCR followed by exact-match and constraint checks to verify clue preservation and grid validity. For videos, these checks are applied frame by frame; for images, they are applied to the single generated output.

\vspace{-1mm}
\paragraph{VLM-Based Evaluation.} For tasks without reliable pixel- or symbol-level verification, including embodied navigation and physical commonsense, we use Gemini-2.5-Pro as an automated judge with task-specific rubrics. The judge scores the intermediate and final criteria specified by each rubric, which combine into the gated primary score defined above. Because VLM-based judging is imperfect, we report human-evaluator agreement in \Cref{sec:human_eval} and interpret these results with task-specific reliability caveats. Text-solution baselines are scored directly against the corresponding ground-truth answers or action trajectories where available.\looseness=-1



\vspace{-1mm}

\section{Evaluation Results}

\subsection{Automatic Evaluation}

\Cref{tab:main_results} reveals a clear gap between visual generation and task-level reasoning. Each score is zero-shot performance under the task's primary success metric: Maze and Sudoku use deterministic generation checks, while tasks without exact verification use the rubric-based VLM evaluator described above. Text-only Gemini results appear only as diagnostic baselines for abstract reasoning, not as multimodal generation systems.

\vspace{-1mm}
\paragraph{Abstract Reasoning.} Generative models struggle most where tasks impose explicit symbolic or spatial constraints. In Maze, the best generator is Nano-banana Pro (9.17\%), with Veo-3 and Nano-banana at 3.67\% and all others below 1\%. In Sudoku, every video model scores 0.00\%, and the best image generator reaches only 8.60\%. ARC is similarly hard: Nano-banana Pro leads at 28.07\%, while the best video model, Sora-2, reaches just 11.67\%. Math is the lone exception: Nano-banana Pro reaches 72.69\%, close to the Gemini text baselines (71.38\% and 74.14\%), though video models stay far lower (0.00--10.61\%). Across these tasks, visually coherent generation does not reliably preserve the state, rules, and intermediate deductions that abstract reasoning requires.

\vspace{-1mm}
\paragraph{Embodied Navigation.} Image generators obtain the strongest task-level scores in all four navigation settings: Nano-banana Pro leads Last-Mile (75.83\%), Top-down View (33.05\%), 3D Real-World (85.00\%), and SLAG (37.29\%). Video models are competitive only in short-horizon local navigation, where Veo-3 reaches 60.00\% on Last-Mile, but they degrade sharply once the viewpoint or horizon changes, falling to 19.49\% on Top-down View and 11.02\% on SLAG. This pattern speaks directly to our central question of whether temporal generation aids reasoning: it does not do so automatically, and often exposes failures in maintaining geometry, identity, and goal state across frames.

\vspace{-1mm}
\paragraph{Physical Commonsense.} Video models perform best where the task most closely resembles natural video dynamics. Sora-2 leads with 76.00\% on Physical Concepts and 64.00\% on Sports, and Veo-3 follows at 41.67\% and 60.00\%. These results indicate that current video generators are strongest on visually grounded physical plausibility; because this domain relies on VLM-based evaluation rather than a deterministic checker, however, the scores should not be read as proof of genuine physical reasoning.\looseness=-1

\subsection{Evaluation Reliability Analysis}
\label{sec:human_eval}
\vspace{-4mm}

\begin{table}[H]
\centering
\small
\setlength{\tabcolsep}{2.5pt}
\begin{adjustbox}{max width=\columnwidth}
\begin{tabular}{@{}lllc@{}}
\toprule
\textbf{Task} & \textbf{Evaluator} & \textbf{Model} & \textbf{Agreement acc. (\%)} \\
\midrule
\multirow{3}{*}{Maze} & \multirow{3}{*}{Pixel} & Sora-2 & 100.00 \\
 & & Veo-3 & 95.56 \\
 & & Wan-2.2 & 100.00 \\
\midrule
\multirow{3}{*}{Sudoku} & \multirow{3}{*}{OCR} & Sora-2 & 100.00 \\
 & & Veo-3 & 100.00 \\
 & & Wan-2.2 & 100.00 \\
\midrule
ARC & VLM & Veo-3 & 94.90 \\
\midrule
\multirow{2}{*}{3D R.-W. Nav.} & \multirow{2}{*}{VLM} & Veo-3 & 75.33 \\
 & & Nano-banana Pro & 21.67 \\
\addlinespace
\multirow{2}{*}{Last-Mile Nav.} & \multirow{2}{*}{VLM} & Veo-3 & 53.50 \\
 & & Nano-banana Pro & 30.00 \\
\addlinespace
\multirow{2}{*}{Top-down Nav.} & \multirow{2}{*}{VLM} & Veo-3 & 79.15 \\
 & & Nano-banana Pro & 51.69 \\
\addlinespace
\multirow{2}{*}{SLAG} & \multirow{2}{*}{VLM} & Veo-3 & 88.45 \\
 & & Nano-banana Pro & 62.39 \\
\midrule
Physics & VLM & Veo-3 & 65.31 \\
\bottomrule
\end{tabular}
\end{adjustbox}
\vspace{-2mm}
\caption{\textbf{Evaluation reliability analysis.} Agreement between human annotations and the evaluator check reported for each reliability subset. Pixel-based Maze and OCR-based Sudoku are deterministic checks used for the main generation scores; VLM denotes tasks or subsets without deterministic verification.
}
\label{tab:evaluation_reliability_primary_accuracy}
\end{table}

\Cref{tab:evaluation_reliability_primary_accuracy} indicates how much confidence each conclusion deserves. The deterministic checks used for Maze and Sudoku are the most reliable: pixel-based Maze agreement ranges from 95.56\% to 100.00\%, and OCR-based Sudoku agreement reaches 100.00\% across all three video models checked. ARC also shows high agreement (94.90\%), though this partly reflects agreement on the many failed cases rather than on fine-grained correctness. VLM-based evaluation is less stable for embodied navigation, ranging from 21.67\% to 88.45\%, so exact model rankings in that domain should be read cautiously even where the aggregate trend is informative. Physics reaches only moderate agreement (65.31\%), so its scores are best read as judged physical plausibility rather than verified physical correctness. Overall, the table supports our main qualitative conclusions while bounding their interpretation: the strongest evidence comes from deterministic abstract-reasoning checks, whereas embodied-navigation and physical-commonsense tasks admit no exact rule-based verifier. Lower agreement in these domains therefore bounds our confidence rather than rendering the results uninformative; absent a deterministic checker, VLM-based judgment remains the best available proxy.

\begin{figure*}[t!]
    \centering
    \includegraphics[width=\textwidth,clip,trim=0 255 0 0]{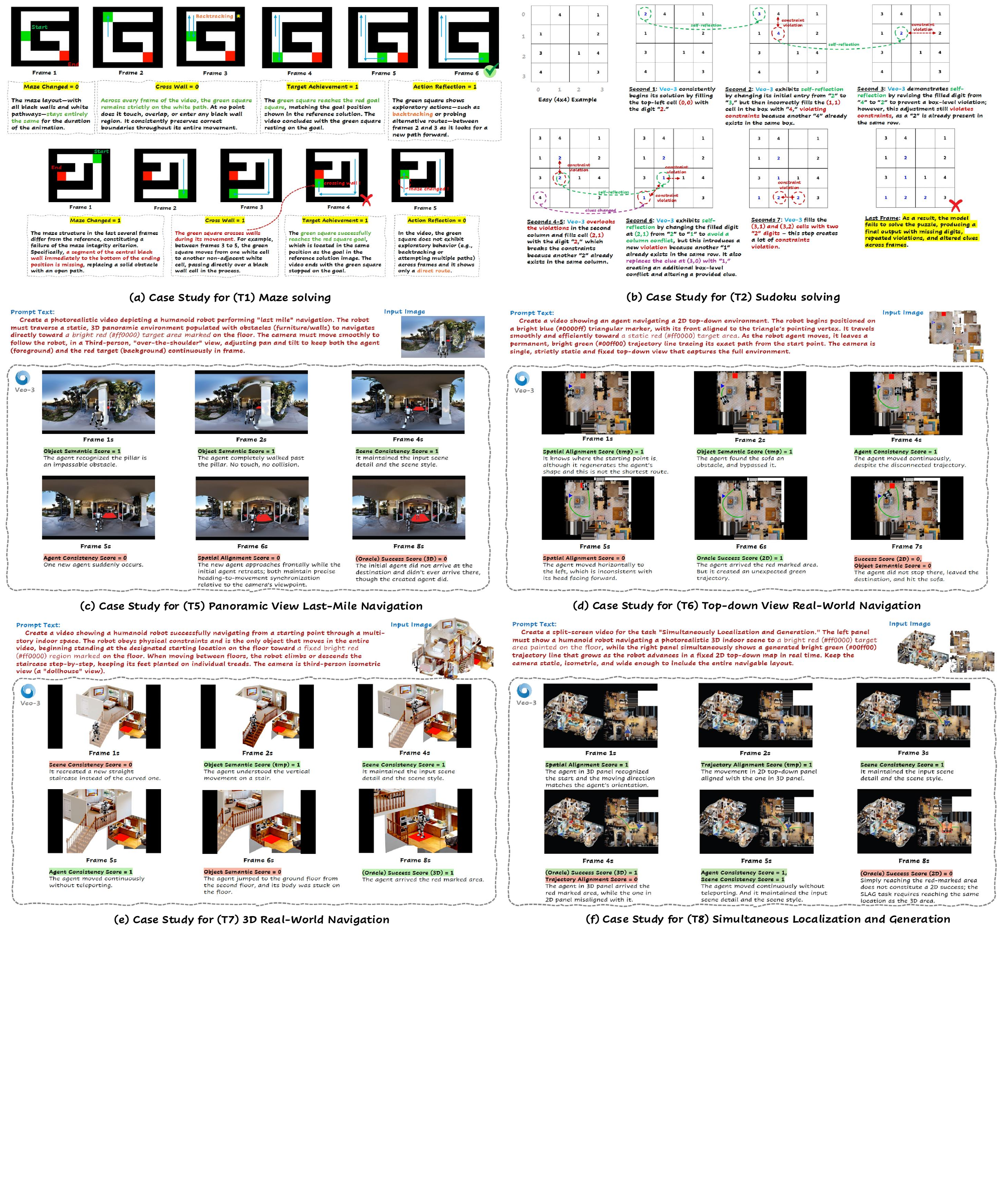}
    \vspace{-7mm}
    \caption{\textbf{Qualitative case studies on MMGR}, all generated by Veo-3. (a) \textit{Maze} (T1): success and failure cases. (b) \textit{Sudoku} (T2): failure cases. (c) \textit{Panoramic View Last-Mile Nav.} (T5): agent inconsistency, where Veo-3 hallucinates a second agent. (d) \textit{Top-down View Real-World Nav.} (T6): Veo-3 reaches the destination but traces an abnormal trajectory. (e) \textit{3D Real-World Nav.} (T7): Veo-3 reaches the destination despite physics violations (floor jumping) and scene alterations (staircase geometry). (f) \textit{SLAG} (T8): Veo-3 achieves 3D target success but suffers from 2D trajectory misalignment.}
    \vspace{-3mm}
    \label{fig:case_study}
\end{figure*}

\section{Failure Cases and Future Directions}
\vspace{-1mm}

\paragraph{Failure Patterns.} MMGR exposes a recurring gap between plausible media generation and valid task execution (\Cref{fig:case_study}). In abstract reasoning, models often alter the input state or break global constraints: maze paths cross walls or stop short of the goal, Sudoku generations modify clues or violate row and column rules, ARC outputs corrupt the demonstration context, and video math solutions drift away from valid symbolic steps. In embodied navigation, image generators satisfy final-state criteria more often than video generators, whose outputs fail through agent identity changes, inconsistent geometry, and goal-state drift across frames. Physical Commonsense is the strongest domain for video models, but because its scores rely on model-based judging, they reflect physical plausibility rather than fully verified physical correctness. These failures point to a single diagnosis: current generators are better at local visual plausibility than at maintaining a persistent task state. That distinction is central to multimodal generative reasoning, where solving a task requires preserving the input, applying the right rule or plan, and keeping intermediate states consistent through to the final output. Closing the gap therefore calls for progress on two fronts: generators that explicitly optimize for task state, and evaluation that can verify it.

\vspace{-1mm}
\paragraph{Future Directions.} Progress will require advances on both fronts. On the modeling side, generators need training signals and architectures that reward state preservation, constraint satisfaction, cross-view consistency, and completed trajectories rather than only realistic appearance, together with longer, more controllable temporal horizons and training data that moves beyond salient physical-commonsense clips toward embodied, navigational, and interaction-heavy scenarios. On the evaluation side, benchmarks should pair deterministic and process-level metrics with task-state checks, trajectory-consistency tests, and failure-specific measures that verify whether an output actually solves the reasoning problem rather than merely looking plausible.

\section{Conclusion}
\vspace{-2mm}

We introduce MMGR, a benchmark that shifts multimodal generative reasoning from visual realism toward answer-verifiable, process-aware evaluation. Across Abstract Reasoning, Embodied Navigation, and Physical Commonsense, with chain-of-frame validity for video, it tests whether generative models respect physical, logical, spatial, and temporal constraints. Our results reveal a clear gap: modern models produce visually compelling outputs yet fail on tasks requiring global consistency, symbolic abstraction, and long-horizon reasoning, exposing reliance on perceptual shortcuts rather than grounded inference.

\section*{Limitations}

\paragraph{Evaluation Scope and Model Coverage.} MMGR evaluates zero-shot generative reasoning and excludes fine-tuned or task-specific models. This design isolates the reasoning capabilities of pre-trained models, but it may underestimate the performance reachable through task adaptation. Our model pool covers leading video and image generators, yet the pace of releases means some recent systems are absent. We also focus on text-to-video and text-to-image generation; conditioning on other input modalities (\textit{e.g.}, video-to-video editing) is an orthogonal capability we do not assess.

\paragraph{Deterministic Evaluation Limitations.} Only some tasks admit deterministic verification. Maze (pixel-based) and Sudoku (OCR-based) use exact checks for their main generation scores, but Physical Commonsense and several embodied navigation tasks rely solely on VLM-based evaluation, inheriting VLMs' known weaknesses in spatial reasoning and physical understanding.

\paragraph{Benchmark Design Constraints.} Our difficulty tiers are human-designed and may not match model-specific difficulty gradients. Some tasks show floor effects for current video models, especially Sudoku and ARC, which limits discriminative power among weaker systems. Our focus on discrete reasoning tasks also leaves out continuous control, long-horizon planning, and open-ended creative generation, where generative models may excel. Finally, prompting strongly affects performance, yet our standardized prompts prioritize consistency over per-model optimization.

\paragraph{Generalization and Ecological Validity.} MMGR emphasizes systematic reasoning with clear ground-truth answers, so it may underrepresent real-world settings with ambiguous objectives, partial observability, or subjective quality. Its structured tasks (finite mazes, fixed Sudoku grids, standardized math problems) do not capture the open-ended complexity of natural environments. Because video models are trained on internet-scale data, they may perform well on distribution-matched scenarios our controlled tasks omit, leaving an incomplete picture of their capabilities.

\paragraph{Reliability Scope.} Our reliability analysis rests on sampled manual annotations rather than exhaustive labeling. The agreement analysis in \Cref{tab:evaluation_reliability_primary_accuracy} offers a compact check against manual labels, but raw accuracy can be inflated by class imbalance and does not replace fine-grained agreement metrics. Because annotating every generated output is infeasible, comprehensive assessment depends on automatic metrics, especially for the VLM-only tasks of embodied navigation and physical commonsense.

\section*{Ethical Considerations}

This paper presents MMGR, a benchmark that evaluates multimodal generative reasoning across image, video, language, and vision-language models using answer-verifiable tasks. Its goal is to advance the evaluation of reasoning in modern generative models, particularly their ability to connect perception with structured, multi-step inference. By emphasizing tasks with deterministic or verifiable outcomes, MMGR encourages models whose outputs can be audited and validated, which matters for applications in decision-making, education, and scientific reasoning, and we expect it to support more reliable, interpretable, and trustworthy multimodal AI systems. At the same time, stronger reasoning in generative models may amplify existing risks of misuse, over-reliance, or deployment in high-stakes settings without adequate safeguards. MMGR introduces no new modeling techniques or deployment mechanisms; it is a diagnostic framework for surfacing the limitations and failure modes of current systems, which we hope will help practitioners understand model behavior and guide responsible development. Overall, its societal implications align with established goals in machine learning research, namely improving robustness, transparency, and evaluation rigor, and we foresee no significant negative impacts arising uniquely from this benchmark.

\clearpage
\bibliographystyle{iclr2025_conference}
\bibliography{main}

\newpage
\appendix

\section{Computational Experiments}
\label{apx:compute}

Because MMGR evaluates pre-trained models without any task-specific training, its cost is incurred entirely at inference time: every model generates outputs for all task instances, which are then scored by the automatic evaluators and, on sampled subsets, by human annotators. We query the closed-source models (\textit{Veo-3}, \textit{Sora-2}, \textit{Nano-banana}, \textit{Nano-banana Pro}, \textit{GPT-4o-image}, \textit{GPT-image-1.5}, \textit{Gemini-3-Flash}, \textit{Gemini-3-Pro}, and the \textit{Gemini-2.5-Pro} judge) through their official paid APIs, and run the two open-source generators, \textit{Qwen-Image} and \textit{Wan-2.2}, locally on NVIDIA A100 80GB GPUs.

\subsection{Model Size and Budget}
\label{apx:compute_budget}

\paragraph{Model Sizes.}
The closed-source models are proprietary API endpoints whose parameter counts are not disclosed, so we treat them as black boxes. For the open-source models, \textit{Qwen-Image} is a 20B-parameter Multimodal Diffusion Transformer generating at $1024{\times}1024$ resolution~\citep{qwen2024image}, and \textit{Wan-2.2} is used in its image-to-video \texttt{I2V-A14B} Mixture-of-Experts configuration, with 27B total parameters but only about 14B active per denoising step~\citep{wan2025video}.

\paragraph{Budget.}
MMGR comprises \textbf{1{,}853} task instances (1{,}323 in Abstract Reasoning: Maze 240, Sudoku 300, ARC 456, Math 327; 480 in Embodied Navigation: 4 tasks of 120; and 50 in Physical Commonsense: 2 tasks of 25), and we draw five samples per instance from each model. A video model therefore produces 9{,}265 generations, an image model 9{,}015 (the eight non-physics tasks), and a Gemini text baseline 4{,}335 (Maze, Sudoku, and Math). On a single A100 80GB GPU, \textit{Qwen-Image} renders one image in about 40\,s and \textit{Wan-2.2} produces one clip in roughly 520\,s, so the two open-source models consume about 100 and 1{,}340 A100 GPU-hours, or roughly \textbf{1{,}440 GPU-hours} in total. The eight closed-source systems are accessed through paid commercial APIs and billed per output. Assuming a rental rate of about \$1.5 per A100 GPU-hour and per-output API prices on the order of \$0.05 per image and \$0.40 and \$0.10 per second of generated video for Veo-3 and Sora-2, together with metered token usage for the Gemini baselines and the Gemini-2.5-Pro judge, we estimate the total inference cost of the benchmark at roughly \textbf{\$46k}, dominated by video generation; \Cref{tab:compute_budget} gives the per-model breakdown. The deterministic Maze and Sudoku checks run on CPU at negligible cost.

\begin{table}[h]
\centering
\small
\setlength{\tabcolsep}{6pt}
\renewcommand{\arraystretch}{1.2}
\begin{adjustbox}{max width=\columnwidth}
\begin{tabular}{@{}lll@{}}
\toprule
\textbf{Model(s)} & \textbf{Access} & \textbf{Est.\ cost (USD)} \\
\midrule
Veo-3                    & API                       & $\sim$30k \\
Sora-2                   & API                       & $\sim$9k \\
Four closed image models & API                       & $\sim$1.8k \\
Gemini baselines + judge & API                       & $\sim$3k \\
Wan-2.2                  & A100 ($\sim$1{,}340\,h) & $\sim$2.0k \\
Qwen-Image               & A100 ($\sim$100\,h)     & $\sim$0.15k \\
\midrule
\textbf{Total}           &                           & \textbf{$\sim$46k} \\
\bottomrule
\end{tabular}
\end{adjustbox}
\vspace{-2mm}
\caption{\textbf{Estimated inference cost.} Figures are estimated from the local GPU rental cost and per-output API pricing, with video generation dominating the total. Closed-source models are accessed through paid commercial APIs; the two open-source models run locally on A100 80GB GPUs. The four closed image models are Nano-banana, Nano-banana Pro, GPT-4o-image, and GPT-image-1.5.}
\label{tab:compute_budget}
\end{table}

\section{Human Subjects Including Annotators}
\label{apx:human_subjects}

The human evaluation underlying our reliability analysis (\Cref{sec:human_eval}) was carried out by three graduate-level students whom we recruited and paid for this task; we used no anonymous crowdsourcing platform. All three were familiar with the task definitions and independently scored balanced subsets of generated outputs, ranging from several dozen to several hundred outputs per task, to validate the automatic evaluators. Each output was judged by at least two annotators with disagreements resolved by discussion, and we report the resulting agreement between the human and automatic labels in \Cref{tab:evaluation_reliability_primary_accuracy}. These subsets span all task families: Maze and Sudoku were checked against the deterministic pixel and OCR evaluators across the three video models, whereas ARC, the four embodied-navigation tasks, and Physical Commonsense were checked against the VLM judge, with Veo-3 annotated throughout and Nano-banana Pro additionally annotated on the embodied tasks. Annotators were compensated at an hourly rate of \$25, in line with the local average wage, and took part with informed consent. Because all source materials are public research datasets, the generated outputs contain no personally identifying information, and the work involved only judging the correctness of synthetic images and videos, the study posed no more than minimal risk to participants.

\subsection{Instructions Given to Participants}
\label{apx:human_instructions}

Annotators applied the same fine-grained, gated rubric used by the automatic evaluators, whose full per-task definitions are given in the \textit{Evaluation and Metrics} section of each task appendix (\Cref{apx:maze_evaluation,apx:sudoku_evaluation,apx:arc_evaluation,sec:embodied_eval_metrics,sec:physics_eval}) and summarized in \Cref{tab:evaluation_metrics}. For each item, an annotator was shown the task input, the model output (a video played frame by frame, or a single image), and the ground-truth solution where one existed, and assigned a binary $0/1$ label to every applicable criterion. They were instructed that the holistic overall score is $1$ only when all required criteria hold, that video outputs must be inspected across intermediate frames rather than at the final frame alone, and that judgments must follow the literal task constraints rather than overall visual quality. Concretely, the criteria are layout preservation, the absence of wall crossings, and arrival at the goal for Maze; clue preservation, constraint satisfaction, and completion for Sudoku; transformation correctness and exact-match validity for ARC; task completeness, physical consistency, and instruction following (destination integrity and a static, coherent scene) for embodied navigation; and physics accuracy, motion quality, visual realism, and prompt adherence for Physical Commonsense.

\section{Maze}
\label{apx:maze}

\subsection{Task Description}

We introduce the \textbf{2D Maze} task to evaluate a model's foundational reasoning capabilities. This task is a direct probe for \textbf{2D spatial reasoning}, as the model must understand the topology of the maze (\textit{i.e.}, the white path vs. the black walls). Furthermore, it challenges \textbf{logical reasoning} by requiring the model to generate a valid plan (the solution path) from a start state (green square) to a goal state (red square). Finally, it tests \textbf{temporal reasoning} by requiring the model to execute this plan sequentially over time, moving the agent along the path without deviation.

\begin{figure*}[h!]
\centering
\small
\setlength{\tabcolsep}{1pt}
\begin{tabular}{cccc}

\subcaptionbox{Easy 5$\times$5 (c$\rightarrow$c)}{
    \includegraphics[width=0.15\textwidth]{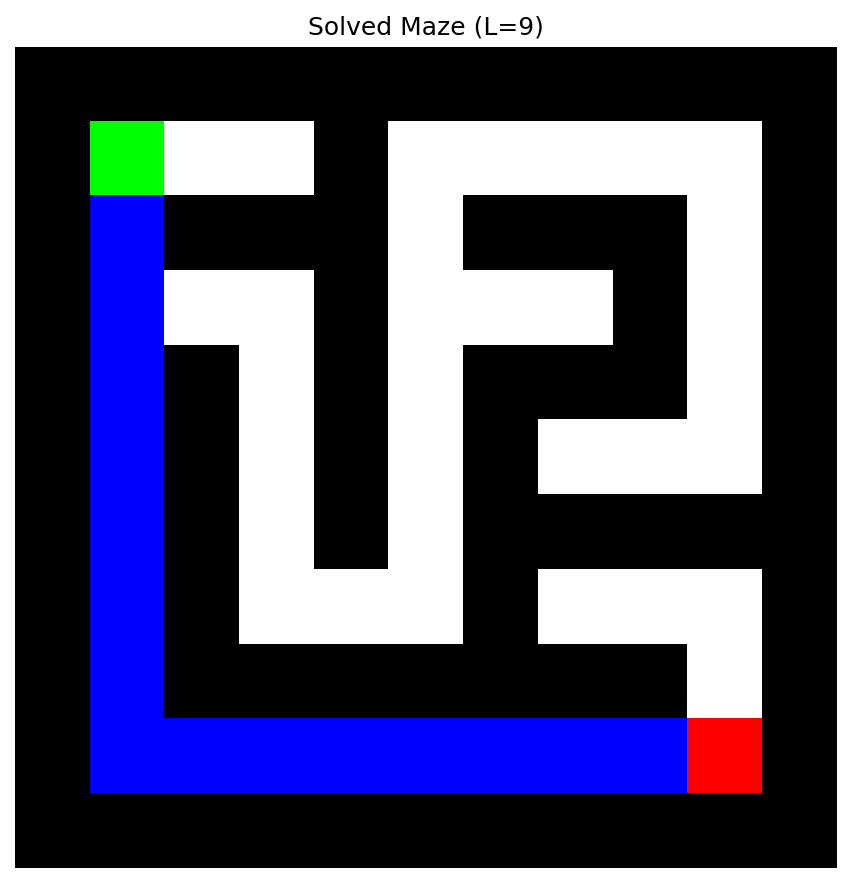}
} &
\subcaptionbox{Easy 3$\times$3 (r$\rightarrow$c)}{
    \includegraphics[width=0.15\textwidth]{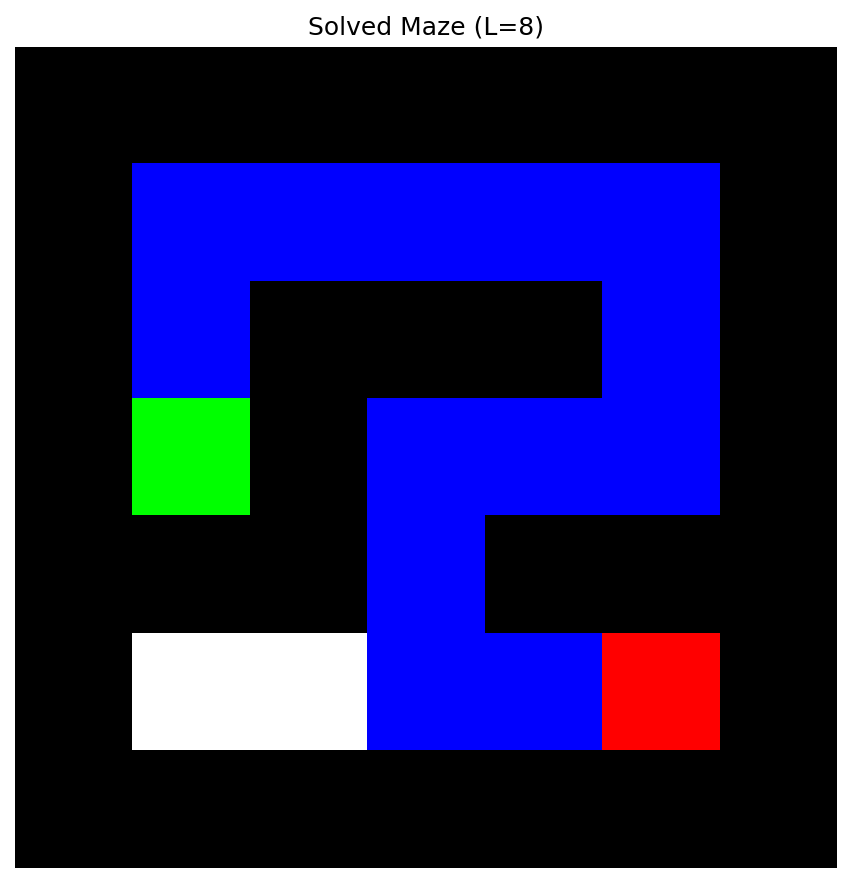}
} &
\subcaptionbox{Easy 5$\times$5 (c$\rightarrow$r)}{
    \includegraphics[width=0.15\textwidth]{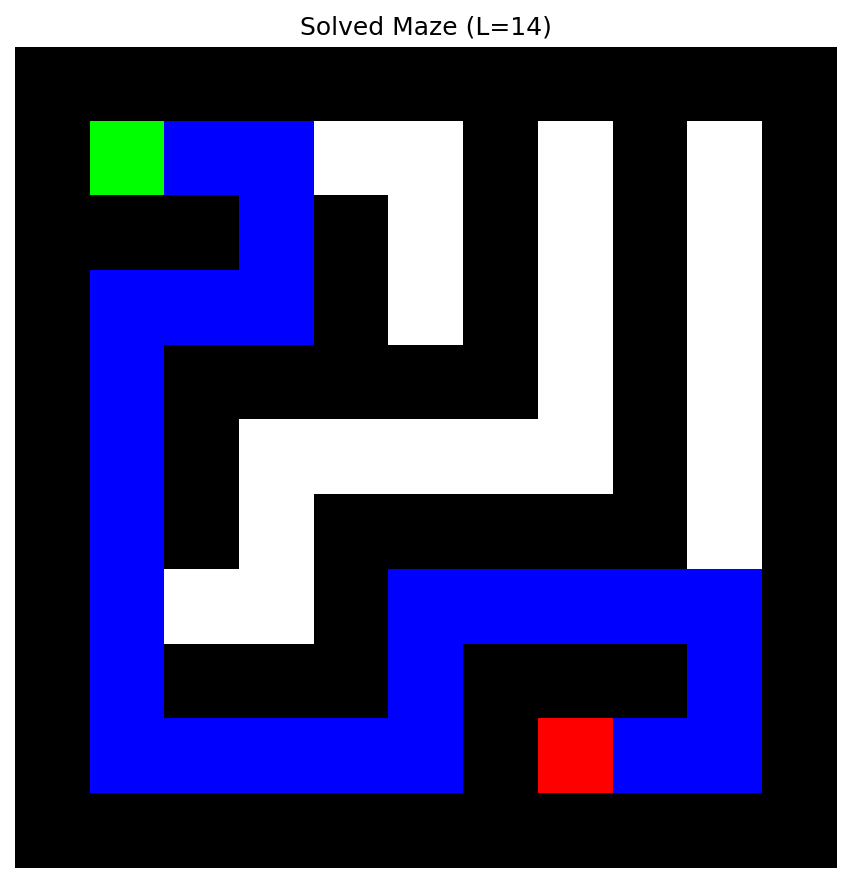}
} &
\subcaptionbox{Easy 4$\times$4 (r$\rightarrow$r)}{
    \includegraphics[width=0.15\textwidth]{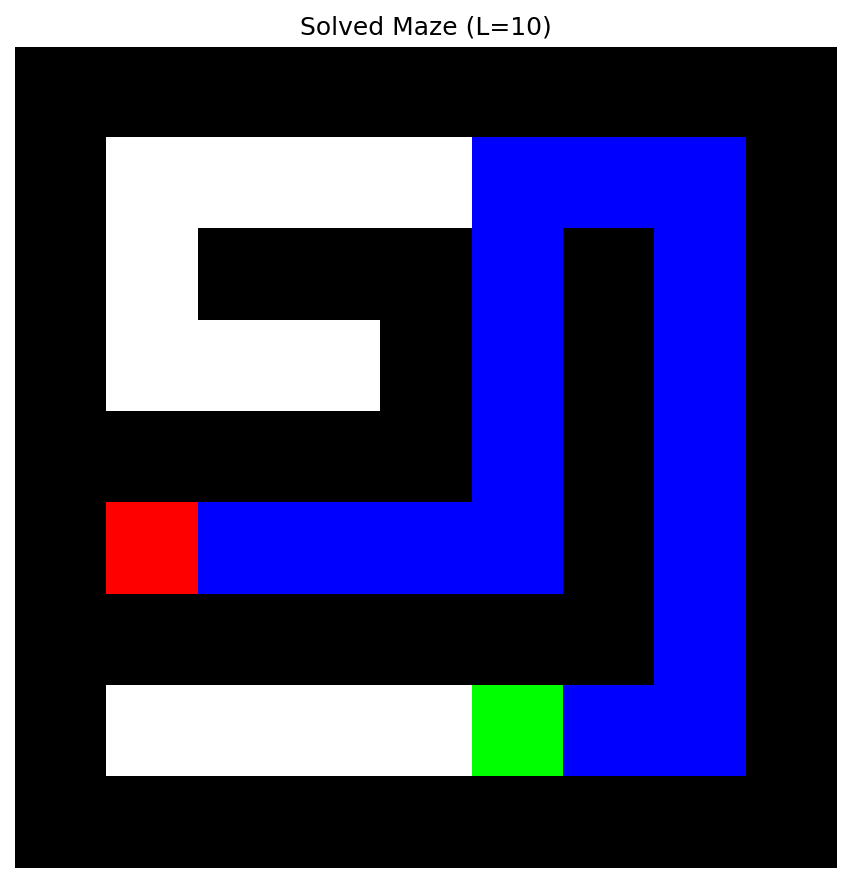}
}
\\[2em]

\subcaptionbox{Medium 9$\times$9 (c$\rightarrow$c)}{
    \includegraphics[width=0.21\textwidth]{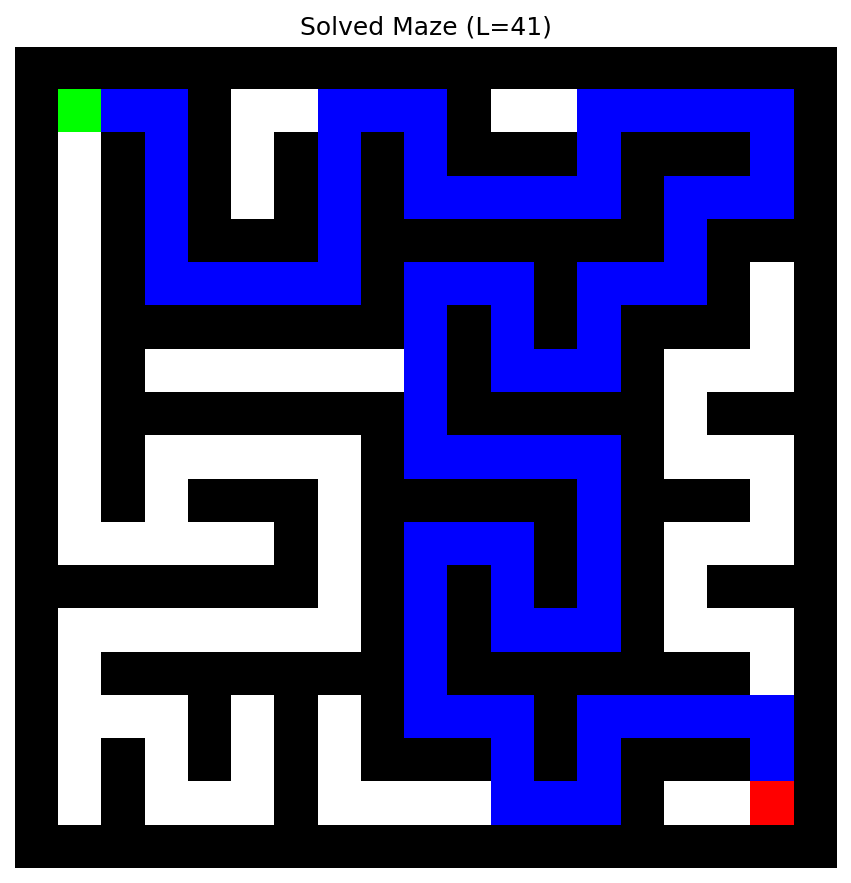}
} &
\subcaptionbox{Medium 9$\times$9 (r$\rightarrow$c)}{
    \includegraphics[width=0.21\textwidth]{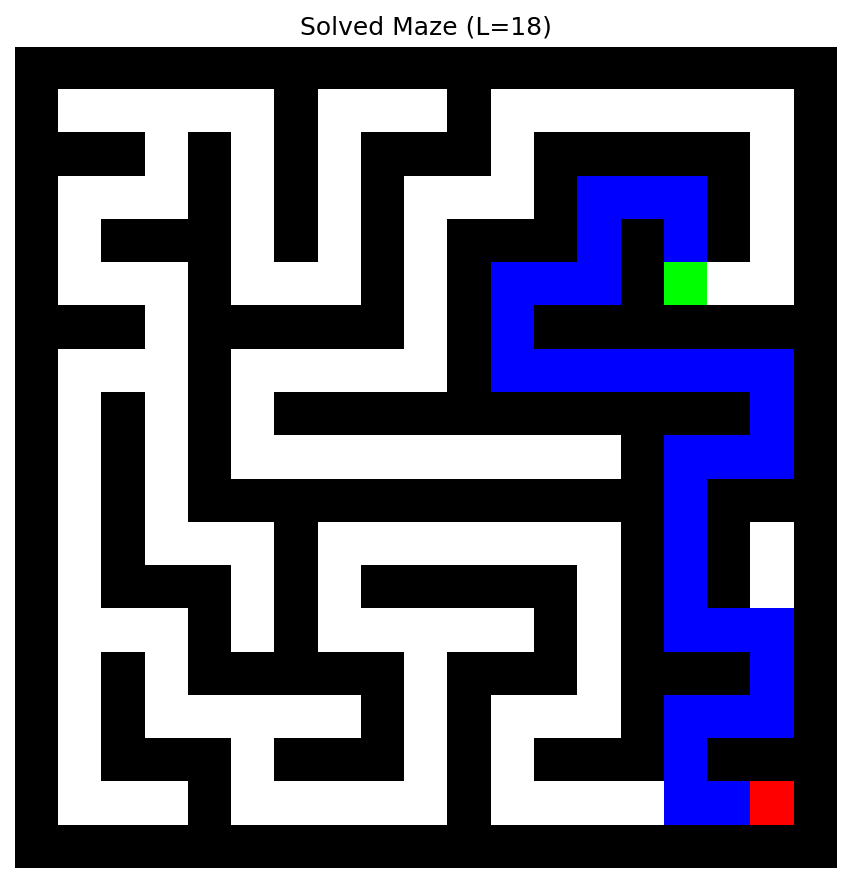}
} &
\subcaptionbox{Medium 9$\times$9 (c$\rightarrow$r)}{
    \includegraphics[width=0.21\textwidth]{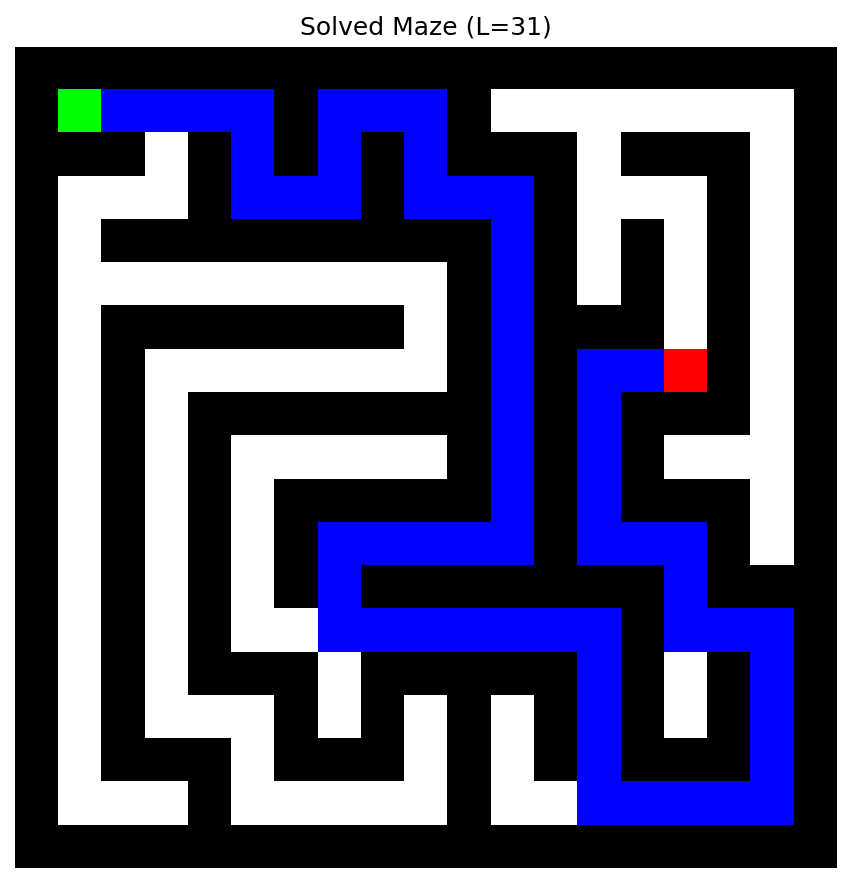}
} &
\subcaptionbox{Medium 8$\times$8 (r$\rightarrow$r)}{
    \includegraphics[width=0.21\textwidth]{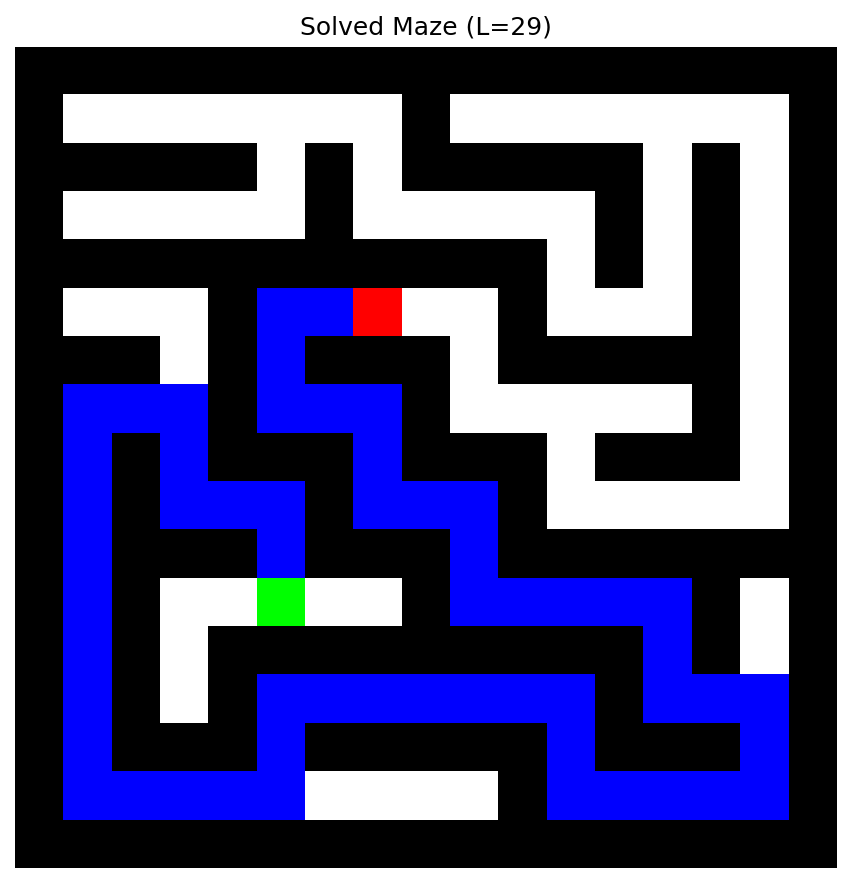}
}
\\[2em]

\subcaptionbox{Hard 11$\times$11 (c$\rightarrow$c)}{
    \includegraphics[width=0.21\textwidth]{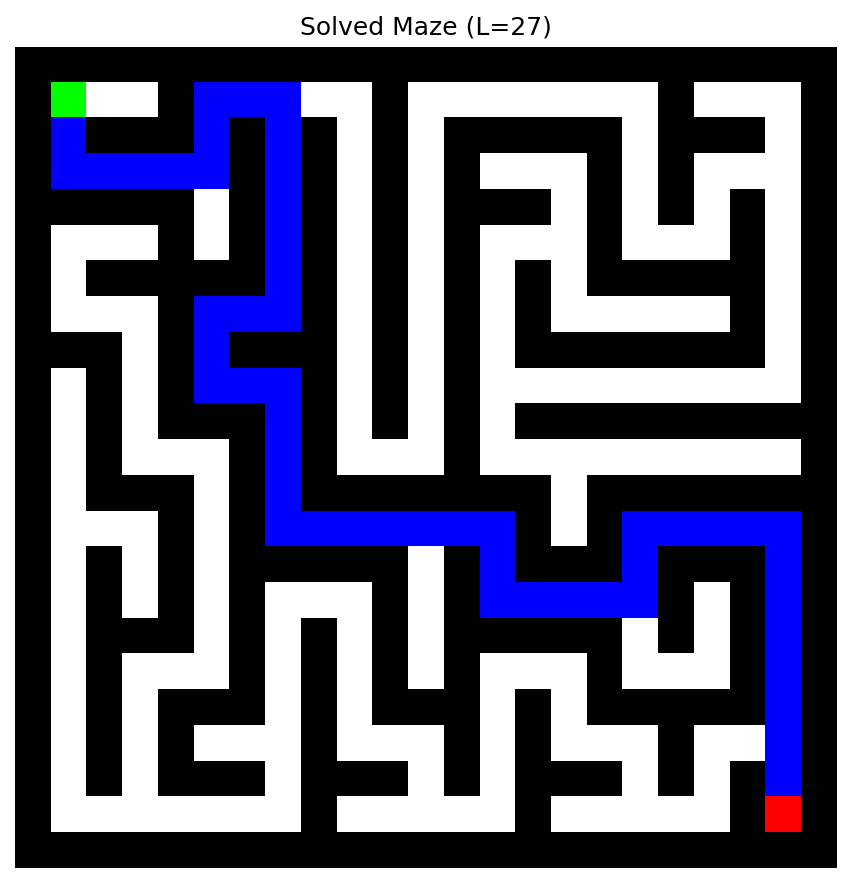}
} &
\subcaptionbox{Hard 13$\times$13 (r$\rightarrow$c)}{
    \includegraphics[width=0.21\textwidth]{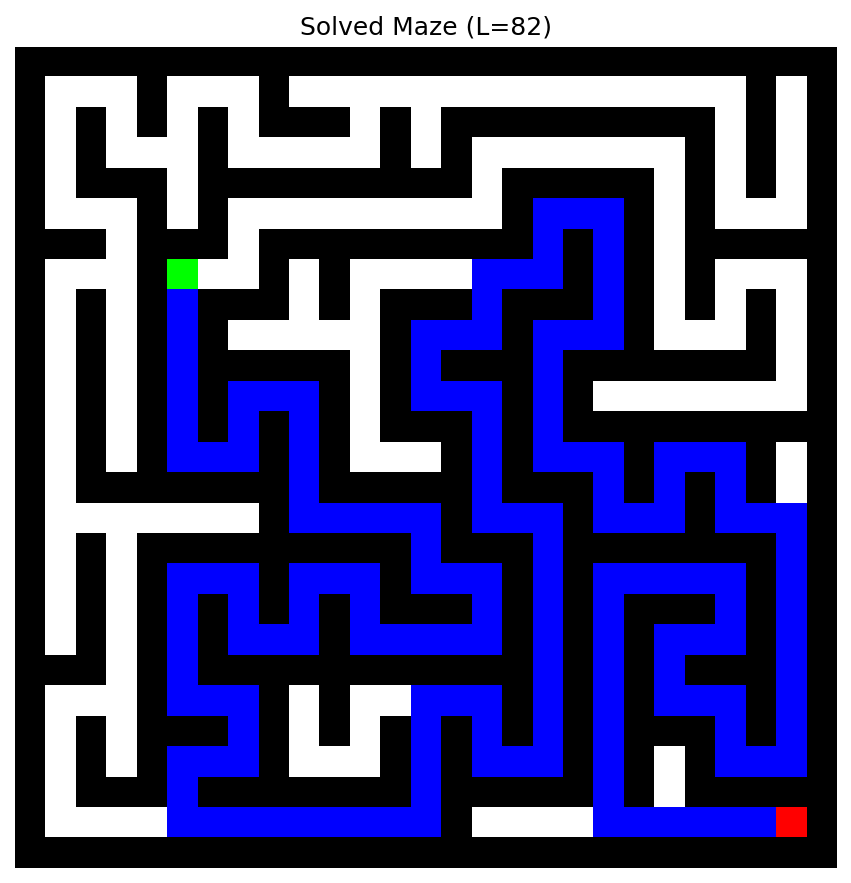}
} &
\subcaptionbox{Hard 13$\times$13 (c$\rightarrow$r)}{
    \includegraphics[width=0.21\textwidth]{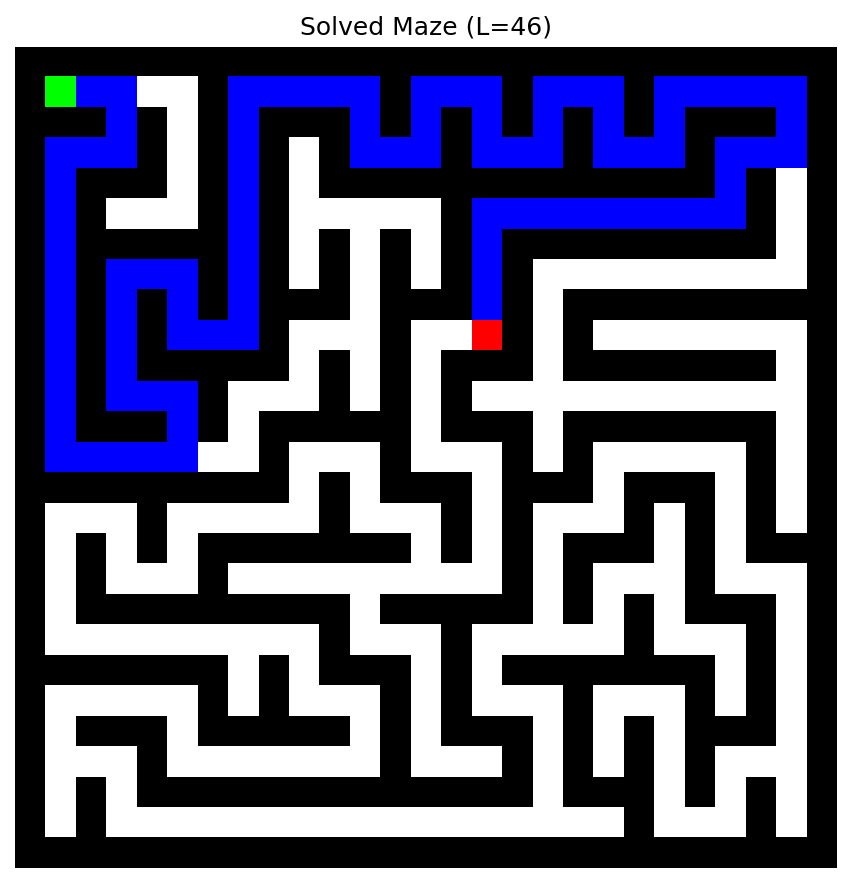}
} &
\subcaptionbox{Hard 13$\times$13 (r$\rightarrow$r)}{
    \includegraphics[width=0.21\textwidth]{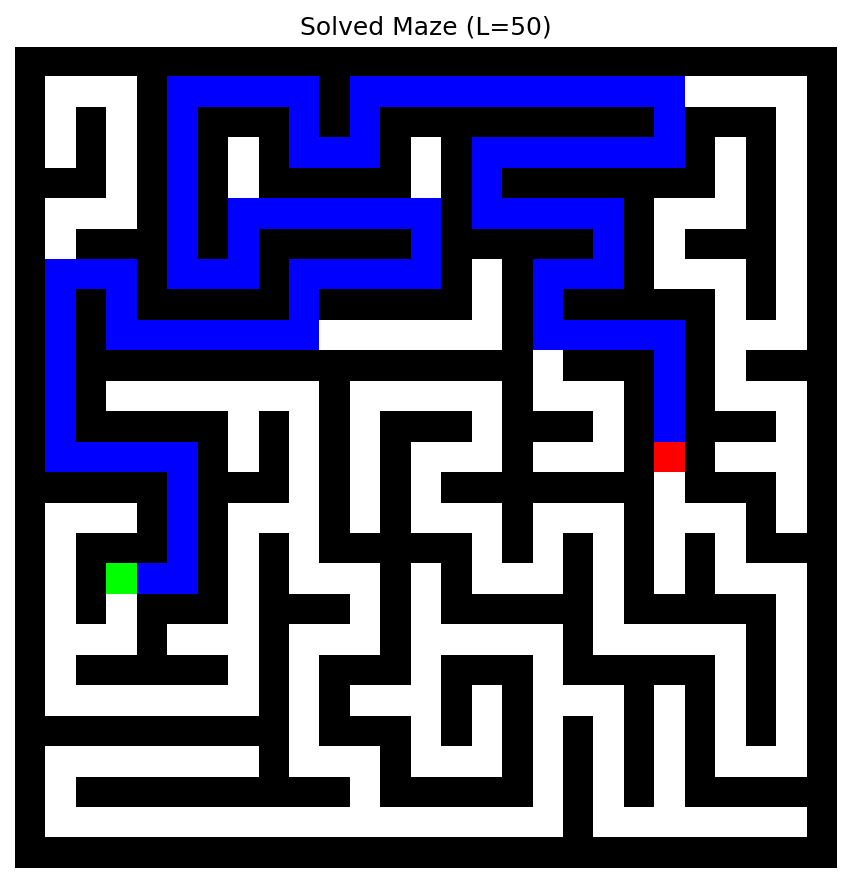}
}
\end{tabular}
\caption{Maze-solving results across three difficulty levels (Easy, Medium, Hard) and four start-end configurations: \textbf{corner→corner}, \textbf{random→corner}, \textbf{corner→random}, and \textbf{random→random}. Each subfigure shows a unique DFS solution illustrating how maze size and start/end randomness affect traversal patterns. Solution paths are highlighted blue.}
\label{fig:maze_examples}
\end{figure*}

\subsection{Hard-Level Control}
\label{sec:maze_hardlevel}

To ensure a diverse yet controllable set of evaluation cases, we leverage the open-source Python library \texttt{maze-dataset} \citep{maze-dataset} to programmatically generate mazes with systematically varied structure and difficulty. Maze complexity is explicitly controlled by varying task parameters along \textit{three} axes:
\vspace{-2mm}
\begin{itemize}
    \item \textbf{Generators (2 types):} We employ two maze-generation algorithms—Depth-First Search (DFS) and Wilson's Algorithm—to produce topologically diverse maze layouts.
    \item \textbf{Grid Sizes (10 levels):} Maze difficulty is scaled across ten grid sizes, ranging from 3$\times$3 to 13$\times$13.
    \item \textbf{Start–Goal Placement (4 schemes):} Each maze is instantiated under four placement schemes—corner-to-corner, corner-to-random, random-to-corner, and random-to-random—to prevent models from overfitting to a single trajectory pattern. A minimum start–goal distance is enforced to rule out trivial solutions.
\end{itemize}

For each generator, we produce 120 mazes, comprising 40 \textit{Easy} (3$\times$3-5$\times$5), 40 \textit{Medium} (6$\times$6-9$\times$9), and 40 \textit{Hard} (10$\times$10-13$\times$13) instances. Overall, this yields \textbf{240} mazes across the two generators. With our generation algorithm, each maze \textit{only} has one solution path. \Cref{fig:maze_examples} presents representative examples across difficulty levels and start–end configurations, along with their corresponding solutions.

\subsection{Evaluation and Metrics}
\label{apx:maze_evaluation}

We evaluate generated videos and images using a Vision–Language Model (VLM)–based evaluator, Gemini-2.5-Pro \citep{comanici2025gemini}. The evaluator receives \textit{three} inputs: (i) the model-generated video or image, (ii) the ground-truth maze solution image, and (iii) a structured evaluation prompt (with modality-specific variants for video vs.\ image). Given these inputs, the VLM judges whether the model solved the task and provides fine-grained feedback across multiple failure modes. The evaluation prompt asks: (i) Does the green square (start) reach and stop on the red square (end)? (ii) Does the green square ever touch or cross a black wall? (iii) Does the layout of the black walls or the position of the red square change at any time? Using the VLM's responses, we compute \textit{four} fine-grained metrics and \textit{one} primary metric:
\vspace{-2mm}
\begin{itemize}
    \item \textbf{Maze Changed (MC, Failure Mode)}: Indicates whether the maze layout is altered. (i) \textit{Video}: 1 if the maze structure changes in \emph{any} frame; 0 if it remains identical throughout the video. (ii) \textit{Image}: 1 if the maze structure differs from the reference solution; 0 otherwise.
    \item \textbf{Cross Wall (CW, Failure Mode)}: Captures violations of maze constraints.  (i) \textit{Video}: 1 if the green square touches or crosses a black wall in \emph{any} frame; 0 if it stays entirely within white corridors.  (ii) \textit{Image}: 1 if the blue path touches or crosses black walls; 0 if fully contained within white corridors.
    \item \textbf{Action Reflection (AR):} Measures exploratory behavior during navigation. (i) \textit{Video}: 1 if the trajectory exhibits exploration (e.g., backtracking or trying multiple paths); 0 if it follows a single, direct route. (ii) \textit{Image}: Not applicable.
    \item \textbf{Target Achievement (TA)}: Measures task completion. (i) \textit{Video}: 1 if the green square reaches and stops on the red target in \emph{any} frame; 0 otherwise. (ii) \textit{Image}: 1 if a continuous, valid blue path connects the start and end points; 0 otherwise.\looseness=-1
    \item \textbf{Overall Score}: A binary success indicator. Set to 1 if \textbf{MC = 0}, \textbf{CW = 0}, and \textbf{TA = 1}; 0 otherwise.
\end{itemize}

\begin{figure*}[h!]
\centering

\begin{subfigure}[t]{0.8\textwidth}
    \centering
    \includegraphics[width=\textwidth, trim=0 180 0 0, clip]{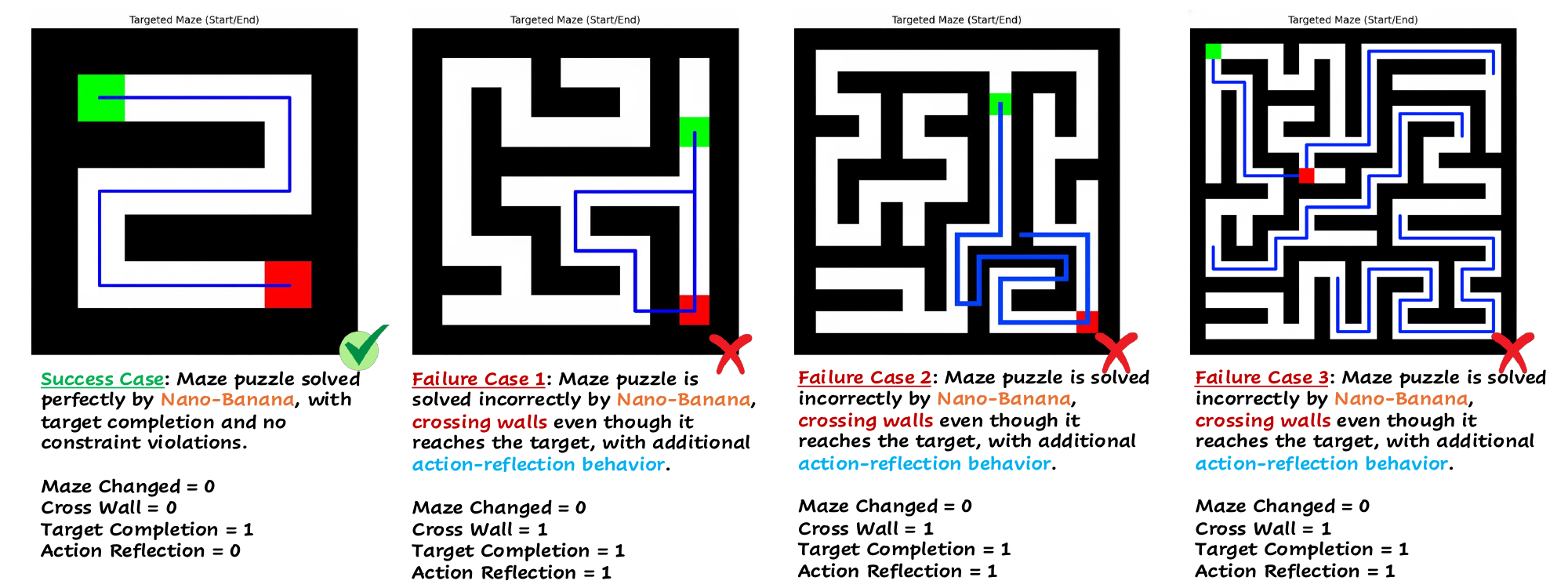}
    \caption{Success and failure cases generated by \textbf{Nano-banana}. Solution paths are highlighted in blue.}
    \label{fig:maze_case_study_3}
\end{subfigure}

\begin{subfigure}[t]{\textwidth}
    \centering
    \includegraphics[width=\textwidth, trim=0 0 0 0, clip]{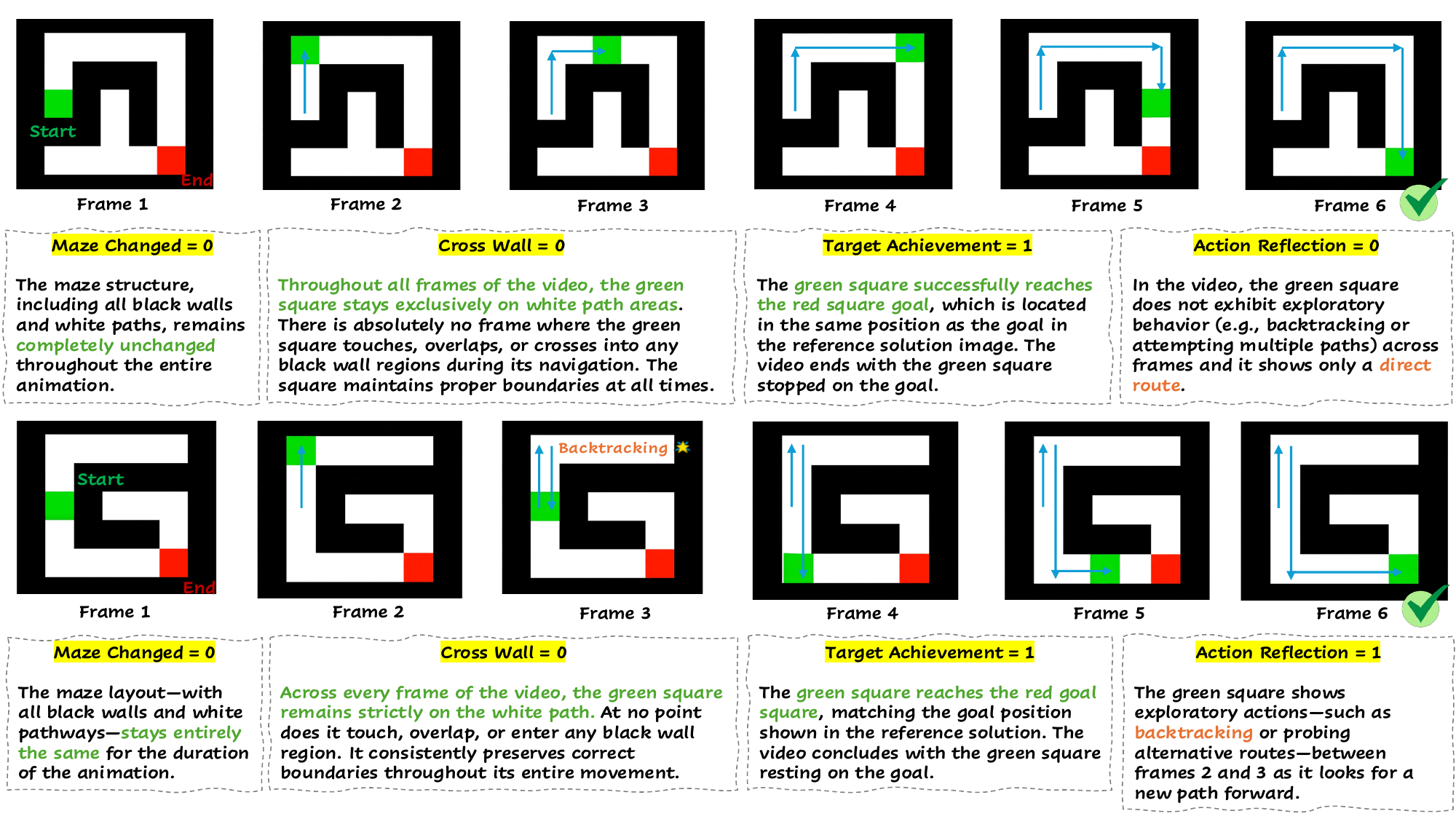}
    \caption{Success cases generated by \textbf{Veo-3}. Solution paths are highlighted in blue.}
    \label{fig:maze_case_study_1}
\end{subfigure}

\begin{subfigure}[t]{0.95\textwidth}
    \centering
    \includegraphics[width=\textwidth, trim=0 250 0 0, clip]{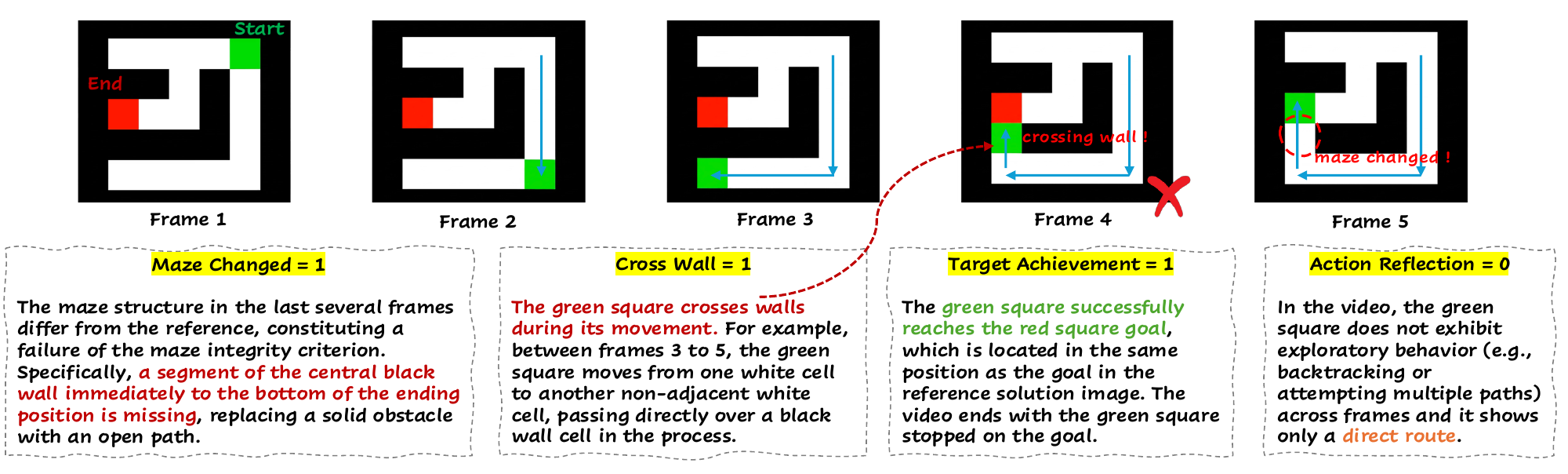}
    \caption{Failure cases generated by \textbf{Veo-3}. Solution paths are highlighted in blue.}
    \label{fig:maze_case_study_2}
\end{subfigure}

\caption{Case studies comparing success and failure behaviors across \textbf{Nano-banana} and \textbf{Veo-3}.}
\label{fig:maze_case_study_all}
\end{figure*}

\vspace{-2mm}
\subsection{Case Study}

\textbf{Image Generation.} \Cref{fig:maze_case_study_3} illustrates Nano-banana's behaviors, showing both perfectly solved outputs and common failure modes. These include wall-crossing artifacts in the final predicted trajectory, slight distortions of maze geometry, and ``action-reflection'' artifacts—cases where the rendered trajectory contains redundant loops or implausible detours even though no temporal dynamics are involved. These artifacts reflect the model’s uncertainty when inferring long-range paths from a single static instruction, resulting in inconsistent or physically implausible solution traces.

\textbf{Video Generation.} \Cref{fig:maze_case_study_1} shows two successful generations by Veo-3. Frame-by-frame annotations reveal that Veo-3 can preserve the maze’s topology, keeps the green square strictly on valid white paths, and maintains consistent wall boundaries from start to finish. The model occasionally performs mild \textit{exploratory} behaviors—such as brief backtracking or short directional adjustments between early frames—before ultimately converging on the correct route. Notably, these explorations remain structurally valid: the agent never crosses walls or distorts the environment, and the maze remains unchanged throughout the entire sequence. In contrast, \Cref{fig:maze_case_study_2} highlights Veo-3’s failure cases. Here, the model sometimes introduces structural inconsistencies—removing or altering wall segments, inserting open passages, or shifting the geometry of the target region. In other cases, the green square traverses invalid regions (\textit{e.g.}, sliding across black walls during transitions) or produces contradictory intermediate frames despite ending at the correct goal cell. The examples also show how subtle frame-to-frame drifts, such as disappearing wall pixels or morphing corridors, can accumulate into integrity violations not captured by coarse success metrics. Collectively, these case studies show that both models may successfully reach the red goal but still differ dramatically in \textit{path fidelity}, \textit{wall adherence}, and \textit{temporal consistency}. The frame-level evidence—ranging from clean, stable trajectories to structurally inconsistent or wall-violating behaviors—underscores the necessity of a fine-grained maze-evaluation framework capable of capturing these nuanced, multimodal failure modes that simple goal-achievement metrics overlook.

\begin{table*}[t!]
\centering
\small
\caption{Quantitative results for the \textbf{2D Maze} task using \textbf{VLM-based} evaluation method. We compare the performance of video models (Veo-3, Sora-2, and Wan-2.2) and image models (Nano-banana, Nano-banana Pro, GPT-4o-image, GPT-image-1.5, and Qwen-image) across two maze generation algorithms (DFS and Wilson) and three difficulty levels (Easy, Medium, and Hard). Because image outputs do not support frame-by-frame reasoning, action-reflection–based metrics are omitted for image generative models (marked with ``N/A''). The best results among video models are highlighted with \colorbox{paleblue}{blue}, and the best results among image models are highlighted with \colorbox{paleyellow}{yellow}.}
\label{tab:maze_all_results}
\begin{adjustbox}{max width=0.75\textwidth}
{
\begin{tabular}{@{}lccccc@{}}
\toprule
& \multicolumn{4}{c}{\textbf{Fine-grained Metrics}} & \multicolumn{1}{c}{\textbf{Primary Metric}} \\
\cmidrule(lr){2-5} \cmidrule(lr){6-6}

\textbf{Model} & \textbf{Maze Changed} $\downarrow$ & \textbf{Cross Wall} $\downarrow$ & \textbf{Target Achievement} $\uparrow$ & \textbf{Action Reflection} $\uparrow$ & \textbf{Overall} $\uparrow$ \\
\midrule

\multicolumn{6}{@{}l}{\textbf{Generator: Depth-First Search}} \\

\multicolumn{6}{@{}l}{\quad \textit{Level: Easy (3$\times$3–5$\times$5)}} \\
\multicolumn{6}{@{}l}{\quad \quad \textbf{Video Models}} \\
\quad \quad \quad Veo-3 & \hlblue{15.50\%} & 25.50\% & \hlblue{60.50\%} & 1.00\% & \hlblue{\textbf{42.00\%}} \\
\quad \quad \quad Sora-2 & 67.50\% & \hlblue{7.50\%} & 12.50\% & \hlblue{77.50\%} & 2.50\% \\
\quad \quad \quad Wan-2.2 & 35.00\% & 79.17\% & 10.00\% & 7.50\% & 1.67\% \\
\multicolumn{6}{@{}l}{\quad \quad \textbf{Image Models}} \\
\quad \quad \quad Nano-banana & 5.00\% & 30.00\% & 85.00\% & N/A & 15.50\% \\
\quad \quad \quad Nano-banana Pro & 5.00\% & 42.50\% & \hlyellow{90.00\%} & N/A & \hlyellow{\textbf{17.50\%}} \\
\quad \quad \quad GPT-4o-image & 95.00\% & \hlyellow{5.00\%} & 72.50\% & N/A & 0.00\% \\
\quad \quad \quad GPT-image-1.5 & \hlyellow{3.33\%} & 13.33\% & \hlyellow{90.00\%} & N/A & 13.33\% \\
\quad \quad \quad Qwen-image & 5.00\% & 23.33\% & 65.00\% & N/A & 11.67\% \\

\multicolumn{6}{@{}l}{\quad \textit{Level: Medium (6$\times$6–9$\times$9)}} \\
\multicolumn{6}{@{}l}{\quad \quad \textbf{Video Models}} \\
\quad \quad \quad Veo-3 & \hlblue{0.50\%} & 25.64\% & \hlblue{50.78\%} & 0.00\% & \hlblue{\textbf{38.72\%}} \\
\quad \quad \quad Sora-2 & 47.50\% & \hlblue{12.50\%} & 10.00\% & \hlblue{60.00\%} & 7.50\% \\
\quad \quad \quad Wan-2.2 & 10.83\% & 90.83\% & 23.33\% & 28.33\% & 1.67\% \\
\multicolumn{6}{@{}l}{\quad \quad \textbf{Image Models}} \\
\quad \quad \quad Nano-banana & 0.63\% & 30.63\% & 71.25\% & N/A & 4.38\% \\
\quad \quad \quad Nano-banana Pro & \hlyellow{0.00\%} & 25.00\% & \hlyellow{82.50\%} & N/A & 2.50\% \\
\quad \quad \quad GPT-4o-image & 72.50\% & 10.00\% & \hlyellow{82.50\%} & N/A & 5.00\% \\
\quad \quad \quad GPT-image-1.5 & 0.83\% & \hlyellow{5.83\%} & 80.83\% & N/A & \hlyellow{\textbf{9.17\%}} \\
\quad \quad \quad Qwen-image & 2.50\% & 28.33\% & 44.17\% & N/A & 0.00\% \\

\multicolumn{6}{@{}l}{\quad \textit{Level: Hard (10$\times$10–13$\times$13)}} \\
\multicolumn{6}{@{}l}{\quad \quad \textbf{Video Models}} \\
\quad \quad \quad Veo-3 & \hlblue{0.00\%} & 18.50\% & \hlblue{60.00\%} & 1.50\% & \hlblue{\textbf{51.50\%}} \\
\quad \quad \quad Sora-2 & 57.50\% & \hlblue{7.50\%} & 25.00\% & \hlblue{60.00\%} & 10.00\% \\
\quad \quad \quad Wan-2.2 & 6.67\% & 80.83\% & 20.00\% & 35.00\% & 5.00\% \\
\multicolumn{6}{@{}l}{\quad \quad \textbf{Image Models}} \\
\quad \quad \quad Nano-banana & \hlyellow{0.00\%} & 24.17\% & 60.00\% & N/A & 0.83\% \\
\quad \quad \quad Nano-banana Pro & \hlyellow{0.00\%} & 12.50\% & 80.00\% & N/A & \hlyellow{\textbf{5.00\%}} \\
\quad \quad \quad GPT-4o-image & 62.50\% & \hlyellow{5.00\%} & 77.50\% & N/A & 0.00\% \\
\quad \quad \quad GPT-image-1.5 & 2.50\% & 9.17\% & \hlyellow{86.67\%} & N/A & 3.33\% \\
\quad \quad \quad Qwen-image & 7.50\% & 11.67\% & 42.50\% & N/A & 1.67\% \\

\midrule
\multicolumn{6}{@{}l}{\textbf{Generator: Wilson's Algorithm}} \\

\multicolumn{6}{@{}l}{\quad \textit{Level: Easy (3$\times$3–5$\times$5)}} \\
\multicolumn{6}{@{}l}{\quad \quad \textbf{Video Models}} \\
\quad \quad \quad Veo-3 & \hlblue{3.50\%} & 21.50\% & \hlblue{61.50\%} & 2.50\% & \hlblue{\textbf{46.50\%}} \\
\quad \quad \quad Sora-2 & 67.50\% & \hlblue{10.00\%} & 15.00\% & \hlblue{40.00\%} & 5.00\% \\
\quad \quad \quad Wan-2.2 & 32.50\% & 84.17\% & 15.00\% & 8.33\% & 1.67\% \\
\multicolumn{6}{@{}l}{\quad \quad \textbf{Image Models}} \\
\quad \quad \quad Nano-banana & 10.50\% & 47.50\% & 81.00\% & N/A & 6.50\% \\
\quad \quad \quad Nano-banana Pro & 10.00\% & 32.50\% & 85.00\% & N/A & 12.50\% \\
\quad \quad \quad GPT-4o-image & 82.50\% & 15.00\% & \hlyellow{90.00\%} & N/A & 2.50\% \\
\quad \quad \quad GPT-image-1.5 & \hlyellow{5.00\%} & 16.67\% & 86.67\% & N/A & 5.00\% \\
\quad \quad \quad Qwen-image & 5.83\% & \hlyellow{12.50\%} & 67.50\% & N/A & \hlyellow{\textbf{20.00\%}} \\

\multicolumn{6}{@{}l}{\quad \textit{Level: Medium (6$\times$6–9$\times$9)}} \\
\multicolumn{6}{@{}l}{\quad \quad \textbf{Video Models}} \\
\quad \quad \quad Veo-3 & \hlblue{1.25\%} & 15.62\% & \hlblue{55.63\%} & 1.25\% & \hlblue{\textbf{47.50\%}} \\
\quad \quad \quad Sora-2 & 62.50\% & \hlblue{12.50\%} & 20.00\% & \hlblue{47.50\%} & 10.00\% \\
\quad \quad \quad Wan-2.2 & 14.17\% & 85.83\% & 14.17\% & 15.83\% & 1.67\% \\
\multicolumn{6}{@{}l}{\quad \quad \textbf{Image Models}} \\
\quad \quad \quad Nano-banana & \hlyellow{0.00\%} & 36.88\% & 70.00\% & N/A & 1.25\% \\
\quad \quad \quad Nano-banana Pro & 2.50\% & 17.50\% & 80.00\% & N/A & 2.50\% \\
\quad \quad \quad GPT-4o-image & 75.00\% & \hlyellow{5.00\%} & 80.00\% & N/A & \hlyellow{\textbf{5.00\%}} \\
\quad \quad \quad GPT-image-1.5 & \hlyellow{0.00\%} & 11.67\% & \hlyellow{83.33\%} & N/A & 4.17\% \\
\quad \quad \quad Qwen-image & 4.17\% & 25.00\% & 47.50\% & N/A & 0.00\% \\

\multicolumn{6}{@{}l}{\quad \textit{Level: Hard (10$\times$10–13$\times$13)}} \\
\multicolumn{6}{@{}l}{\quad \quad \textbf{Video Models}} \\
\quad \quad \quad Veo-3 & \hlblue{1.25\%} & 18.75\% & \hlblue{58.75\%} & 0.63\% & \hlblue{\textbf{45.62\%}} \\
\quad \quad \quad Sora-2 & 45.00\% & \hlblue{10.00\%} & 10.00\% & \hlblue{52.50\%} & 2.50\% \\
\quad \quad \quad Wan-2.2 & 10.00\% & 89.17\% & 13.33\% & 28.33\% & 0.83\% \\
\multicolumn{6}{@{}l}{\quad \quad \textbf{Image Models}} \\
\quad \quad \quad Nano-banana & 1.25\% & 27.50\% & 60.62\% & N/A & 0.00\% \\
\quad \quad \quad Nano-banana Pro & \hlyellow{0.00\%} & 22.50\% & 75.00\% & N/A & 5.00\% \\
\quad \quad \quad GPT-4o-image & 60.00\% & 7.50\% & 77.50\% & N/A & \hlyellow{\textbf{7.50\%}} \\
\quad \quad \quad GPT-image-1.5 & 2.50\% & \hlyellow{2.50\%} & \hlyellow{85.00\%} & N/A & 4.17\% \\
\quad \quad \quad Qwen-image & 2.50\% & 27.50\% & 38.33\% & N/A & 0.00\% \\

\bottomrule
\vspace{-4mm}
\end{tabular}
}
\end{adjustbox}
\end{table*}

\subsection{Evaluation Results}


\textbf{Key Finding: The Illusion of Competence and Physical Grounding.} Our analysis uncovers two critical disconnects in video generation reasoning. First, a \textbf{dichotomy of simulation}: \textbf{Veo-3 mimics the result}, generating direct solution paths via pattern matching but failing to respect impermeable boundaries. Conversely, \textbf{Sora-2 mimics the process}, performing visible ``reasoning'' (backtracking, hesitation) but losing logical coherence (hallucinating maze structures).  Second, human verification reveals an \textbf{evaluation gap}: Automated VLM metrics systematically overestimate model competence by missing transient ``physics violations.'' While Auto-Eval reports moderate success for Veo-3, human review reveals that the model effectively ``cheats'' by clipping through walls in fast motion—a failure mode invisible to current VLMs. Consequently, true adherence to physical constraints remains near zero, suggesting current models prioritize visual plausibility over logical validity.

\subsubsection{VLM-Based Evaluation}

\Cref{tab:maze_all_results} reveals distinct reasoning strategies across video and image generative models, exposing fundamental trade-offs between path planning, structural preservation, and physical constraint satisfaction. The final Maze row in \Cref{tab:main_results} is aggregated from this VLM-based overall metric; the pixel-based analysis below is a stricter deterministic diagnostic rather than the source of the main-table score.

\textbf{Video Models.} Among video generators, Veo-3 dominates with overall scores of 38.72\%--51.50\%, substantially outperforming Sora-2 (2.50\%--10.00\%) and Wan-2.2 (0.83\%--5.00\%). Veo-3 exhibits a ``Direct Execution'' reasoning style: its near-zero Action Reflection (0.00\%--2.50\%) confirms it generates a single, non-exploratory route without backtracking. While this enables high Target Achievement (50.78\%--61.50\%), the model's reliance on one-shot generation comes at the cost of physical precision, evidenced by Cross Wall rates of 15.62\%--25.64\%. This suggests Veo-3 solves mazes via pattern-matching rather than active search, often ``fudging'' collisions to maintain trajectory momentum. In contrast, Sora-2 exhibits ``Performative Reasoning'': it achieves high Action Reflection scores (40.00\%--77.50\%) by generating visible exploratory behaviors like backtracking. However, this exploration is fundamentally disconnected from valid state maintenance—the model hallucinates new maze structures (Maze Changed 45.00\%--67.50\%) while searching, and rarely achieves the target (10.00\%--25.00\%). Wan-2.2 demonstrates a failure of physical constraint satisfaction: its extreme Cross Wall rates (79.17\%--90.83\%) indicate it treats walls as visual suggestions rather than impermeable boundaries.

\textbf{Image Models.} The image model landscape reveals a striking dichotomy between structural preservation and constraint satisfaction. GPT-4o-image bypasses the reasoning task entirely, altering the maze structure (Maze Changed 60.00\%--95.00\%) to fabricate paths to the target—achieving high Target Achievement (72.50\%--90.00\%) but near-zero overall scores. GPT-image-1.5 represents a marked improvement: it maintains structural fidelity (Maze Changed 0.00\%--5.00\%), achieves competitive Target Achievement (80.83\%--90.00\%), and produces the best overall scores among image models on DFS mazes (3.33\%--13.33\%). The Nano-banana family preserves maze structure well (Maze Changed 0.00\%--10.50\%) but suffers from moderate Cross Wall rates (12.50\%--47.50\%), reflecting difficulty in precise path placement within narrow corridors. Interestingly, Qwen-image achieves the highest image model overall score (20.00\%) on Wilson's Easy mazes despite modest Target Achievement (67.50\%), suggesting it may be better suited to the more uniform passage widths characteristic of Wilson's algorithm.

\textbf{Cross-Cutting Observations.} Several patterns emerge across both modalities. First, high Target Achievement does not guarantee success: GPT-4o-image reaches targets in 72.50\%--90.00\% of cases but scores near zero overall due to structural violations. Second, difficulty scaling is asymmetric—video models maintain relatively stable performance across difficulty levels, while image models show steeper degradation from Easy to Hard. Third, the choice of maze generator matters: Wilson's algorithm produces more uniform mazes that some models (Qwen-image, Veo-3) handle better than the dead-end-heavy DFS mazes. These findings underscore that maze-solving requires the conjunction of structural preservation, physical constraint adherence, and goal achievement—capabilities that current generative models struggle to integrate.

\subsubsection{Limitations and Insights from VLM-Based Evaluation}

Our evaluation pipeline relies on a VLM-based automated evaluator (AutoEval), instantiated as Gemini-2.5-Pro. While AutoEval delivers consistent ratings for \textit{Easy} and \textit{Medium} tasks, its reliability degrades significantly at the \textit{Hard} difficulty level. In these complex scenarios, high-velocity agent movements challenge the VLM's temporal resolution. The evaluator frequently ``\textbf{drops frames},'' missing transient but critical violations—specifically \textbf{Cross Wall} and \textbf{Maze Changed} errors—that occur within single frames. Consequently, AutoEval can systematically overestimate performance on harder tasks. Beyond validation reliability, our analysis highlights the critical concept of \textbf{evaluability}. Models that explicitly visualize reasoning—such as Nano-banana, which renders its intended trajectory as a static blue path—fundamentally transform the evaluation task. This ``plan visualization'' converts a challenging temporal verification problem (tracking frame-by-frame collisions) into a straightforward spatial comparison (checking static path validity), thereby enhancing transparency and trustworthiness. However, this benefit relies on structural faithfulness. When prompted to generate similar trajectory visualizations, Veo-3 frequently produced erratic, hallucinated blue curves covering area the agent never visited. Far from aiding interpretation, these hallucinations obscured the model's actual reasoning and actively confused the Gemini-2.5-Pro evaluator, further degrading AutoEval accuracy. This underscores a key insight: while explicit visual reasoning can improve evaluability, it is only beneficial when the visualized artifacts remain grounded in the physical reality of the environment.

\subsubsection{Pixel-based Evaluation}

As a stricter supplementary diagnostic for limitations inherited from VLM-based evaluation, we propose a fully deterministic \textbf{pixel-based} evaluation pipeline that relies solely on classical computer vision operations, without invoking any vision–language or large language models. This evaluation directly compares generated outputs against ground-truth solutions at the pixel level, enabling fine-grained, reproducible, and cost-free assessment of maze-solving correctness.

\textbf{Image Generation.} We compare each generated image against its corresponding ground-truth solution using pure vision-based analysis. We first segment semantic components by color, including black walls, the green start marker, the red target marker, and the blue solution path. Based on these segments, we compute \textit{four} classes of metrics: (1) \textbf{Wall Crossing Detection} checks whether the generated blue path intersects maze walls, using erosion-based tolerance to account for anti-aliasing; crossings are flagged only when the intersection ratio exceeds a small threshold. (2) \textbf{Layout Matching} verifies structural consistency by computing IoU between wall regions and measuring centroid distances for start and target markers within a fixed pixel tolerance. (3) \textbf{Path Matching} evaluates route correctness by dilating thin generated paths to match cell width and computing IoU, precision, recall, and a cell-level \textbf{target achievement} score that measures whether the intended route is covered. (4) \textbf{Reflection/Exploration Detection} analyzes the skeletonized path to identify loops and excessive detours, capturing backtracking or exploratory behavior beyond the optimal solution. To align with VLM-based evaluation, each continuous metric is converted into a corresponding \textbf{gated metric} using predefined thresholds\footnote{Gated thresholds for image evaluation are defined as: maze match IoU $>0.8$, no wall crossing detected, target achievement $>0.95$, and reflection detected if the generated path contains loops or its length exceeds $1.3\times$ the solution path length.}. These binary indicators are then aggregated into a \textbf{gated overall score}, which equals 1 if and only if the layout is correct, no wall crossings are detected, and sufficient target achievement is reached (with no excessive reflection); otherwise, the overall score is 0. A perfect gated overall score therefore corresponds to a fully valid maze solution under strict pixel-level criteria.

\textbf{Video Generation.} We extend the same pixel-based principles to the temporal domain. We extract the maximum number of frames from each video (192 frames at 24 fps for Veo-3, 360 frames for Sora-2, and 120 frames for Wan-2.2) and track the green agent marker across frames to reconstruct the full traversal path. The accumulated path is then compared against the ground-truth solution image rather than intermediate video frames, ensuring a stable reference. Evaluation mirrors the image setting but incorporates temporal robustness: (1) \textbf{Wall Crossing Detection} is applied to the accumulated path with stronger erosion to avoid false positives near cell boundaries; (2) \textbf{Layout Matching} is computed for every frame, and the minimum score across all frames is reported to capture worst-case hallucinations or scene drift; (3) \textbf{Path Matching} compares the dilated accumulated path with the solution using IoU, precision, recall, and target achievement; and (4) \textbf{Reflection/Exploration Detection} identifies loops and excessive path length relative to the optimal route. As in the image setting, each continuous metric is mapped to a \textbf{gated metric} using strict thresholds\footnote{For video evaluation, gated thresholds follow the same criteria as images, except that layout matching is evaluated as the minimum IoU across all frames and wall crossing is flagged only if the crossing ratio exceeds $5\%$ to account for temporal accumulation noise.}. The \textbf{gated overall score} for a video equals 1 only when all required conditions are simultaneously satisfied—correct layout across all frames, absence of wall crossings, near-complete target coverage, and no excessive reflection—yielding a binary success indicator suitable for benchmarking. Any failure in these criteria results in an overall score of 0, ensuring that a perfect score reflects a temporally consistent and fully correct maze-solving trajectory.

\begin{figure*}[h]
    \centering
    \includegraphics[width=\textwidth, trim=15 10 15 0, clip]{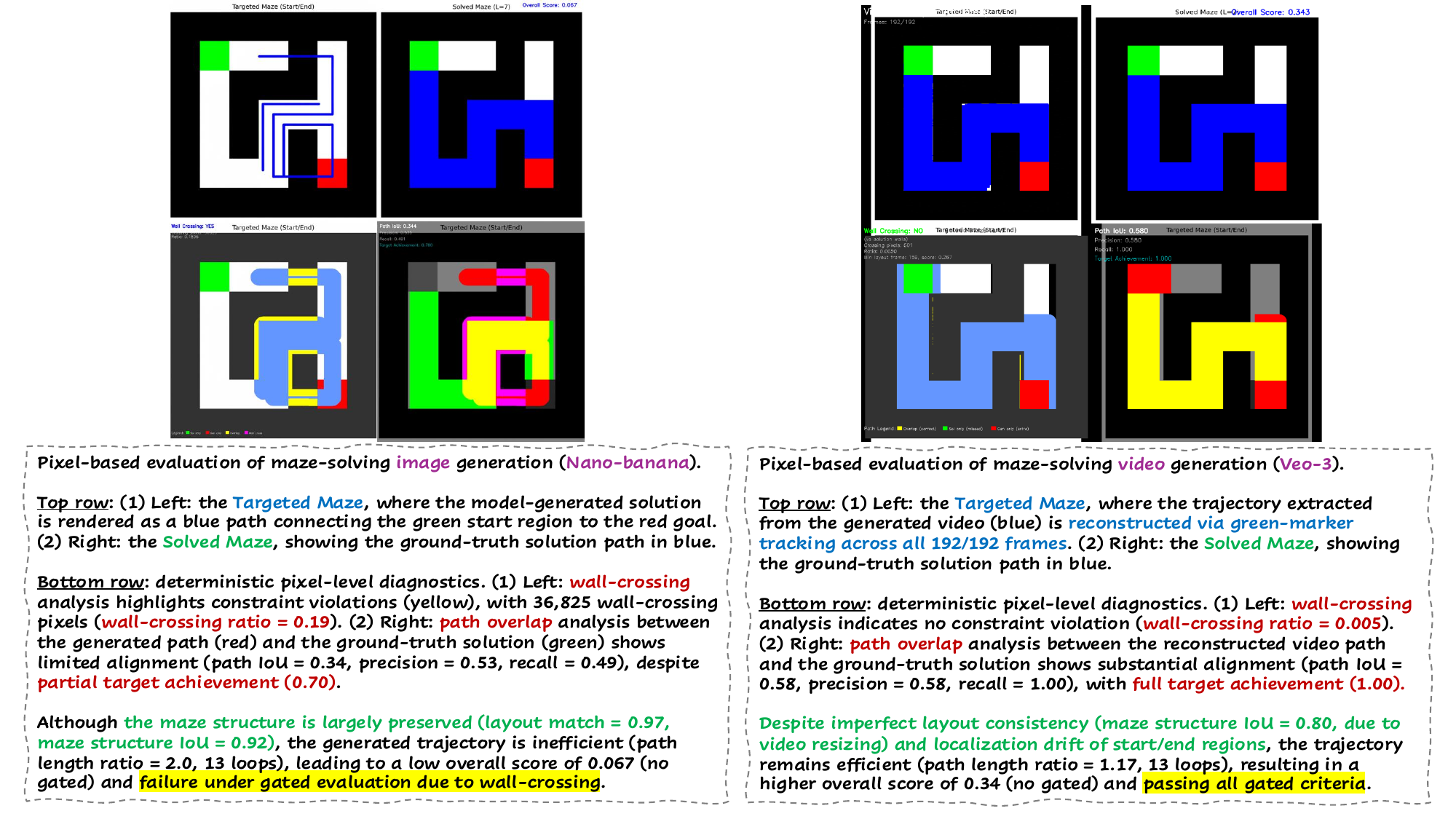}
    \caption{Pixel-based evaluation of maze-solving image generation (Nano-banana, left) and video generation (Veo-3, right).\looseness=-1}
    \vspace{+2mm}
    \label{fig:maze_pixel_evaluation}
\end{figure*}

\textbf{Qualitative Examples of Pixel-Based Evaluation.} \Cref{fig:maze_pixel_evaluation} presents two representative examples of pixel-based evaluation for maze-solving tasks, comparing image generation (Nano-banana, left) and video generation (Veo-3, right). For the \textbf{Nano-banana} model, the top row shows the targeted and solved mazes, while the bottom row reveals substantial wall-crossing violations and limited path alignment with the ground-truth solution, leading to \textit{failure} under gated evaluation despite high maze structure preservation. In contrast, for the \textbf{Veo-3} model, the top row displays the targeted and solved mazes with a trajectory reconstructed from the generated video, and the bottom row indicates negligible wall-crossing and strong path alignment with full target achievement, resulting in \textit{passing} all gated criteria despite imperfect layout consistency.

\begin{table*}[t!]
\centering
\small
\caption{Quantitative results for the \textbf{2D Maze} task using \textbf{pixel-based} evaluation method. We compare the performance of video models (Veo-3, Sora-2, and Wan-2.2) and image models (Nano-banana, Nano-banana Pro, GPT-4o-image, GPT-image-1.5, and Qwen-image) across two maze generation algorithms (DFS and Wilson) and three difficulty levels (Easy, Medium, and Hard).  Because image outputs do not support frame-by-frame reasoning, action-reflection–based metrics are omitted for image generative models (marked with ``N/A''). The best results among video models are highlighted with \colorbox{paleblue}{blue}, and the best results among image models are highlighted with \colorbox{paleyellow}{yellow}.}
\label{tab:maze_all_results_vis}
\begin{adjustbox}{max width=0.74\textwidth}
{
\begin{tabular}{@{}lccccc@{}}
\toprule
& \multicolumn{4}{c}{\textbf{Fine-grained Metrics}} & \multicolumn{1}{c}{\textbf{Primary Metric}} \\
\cmidrule(lr){2-5} \cmidrule(lr){6-6}

\textbf{Model} & \textbf{Maze Changed} $\downarrow$ & \textbf{Cross Wall} $\downarrow$ & \textbf{Target Achievement} $\uparrow$ & \textbf{Action Reflection} $\uparrow$ & \textbf{Overall} $\uparrow$ \\
\midrule

\multicolumn{6}{@{}l}{\textbf{Generator: Depth-First Search}} \\

\multicolumn{6}{@{}l}{\quad \textit{Level: Easy (3$\times$3–5$\times$5)}} \\
\multicolumn{6}{@{}l}{\quad \quad \textbf{Video Models}} \\
\quad \quad \quad Veo-3 & 71.00\% & \hlblue{76.00\%} & 33.50\% & 59.50\% & \hlblue{10.50\%} \\
\quad \quad \quad Sora-2 & 100.00\% & 97.50\% & \hlblue{45.00\%} & \hlblue{100.00\%} & 0.00\% \\
\quad \quad \quad Wan-2.2 & \hlblue{2.50\%} & 90.00\% & 25.00\% & 87.50\% & 0.00\% \\
\multicolumn{6}{@{}l}{\quad \quad \textbf{Image Models}} \\
\quad \quad \quad Nano-banana & \hlyellow{2.50\%} & 69.50\% & 44.00\% & N/A & \hlyellow{\textbf{14.00\%}} \\
\quad \quad \quad Nano-banana Pro & 7.50\% & \hlyellow{42.50\%} & 30.00\% & N/A & 12.50\% \\
\quad \quad \quad GPT-4o-image & 100.00\% & 77.50\% & 0.00\% & N/A & 0.00\% \\
\quad \quad \quad GPT-image-1.5 & 96.67\% & 87.50\% & 8.33\% & N/A & 1.67\% \\
\quad \quad \quad Qwen-image & 60.00\% & 89.17\% & \hlyellow{95.83\%} & N/A & 0.00\% \\

\multicolumn{6}{@{}l}{\quad \textit{Level: Medium (6$\times$6–9$\times$9)}} \\
\multicolumn{6}{@{}l}{\quad \quad \textbf{Video Models}} \\
\quad \quad \quad Veo-3 & 100.00\% & \hlblue{96.48\%} & 14.07\% & 43.22\% & 0.00\% \\
\quad \quad \quad Sora-2 & 100.00\% & 97.50\% & \hlblue{60.00\%} & \hlblue{100.00\%} & 0.00\% \\
\quad \quad \quad Wan-2.2 & \hlblue{7.50\%} & 100.00\% & 15.00\% & 82.50\% & 0.00\% \\
\multicolumn{6}{@{}l}{\quad \quad \textbf{Image Models}} \\
\quad \quad \quad Nano-banana & \hlyellow{0.00\%} & 78.50\% & 22.00\% & N/A & 2.00\% \\
\quad \quad \quad Nano-banana Pro & \hlyellow{0.00\%} & \hlyellow{30.00\%} & 37.50\% & N/A & \hlyellow{\textbf{30.00\%}} \\
\quad \quad \quad GPT-4o-image & 100.00\% & 77.50\% & 0.00\% & N/A & 0.00\% \\
\quad \quad \quad GPT-image-1.5 & 47.50\% & 88.33\% & 20.00\% & N/A & 3.33\% \\
\quad \quad \quad Qwen-image & 82.50\% & 84.17\% & \hlyellow{98.33\%} & N/A & 0.00\% \\

\multicolumn{6}{@{}l}{\quad \textit{Level: Hard (10$\times$10–13$\times$13)}} \\
\multicolumn{6}{@{}l}{\quad \quad \textbf{Video Models}} \\
\quad \quad \quad Veo-3 & 100.00\% & 98.50\% & 10.00\% & 22.50\% & 0.00\% \\
\quad \quad \quad Sora-2 & 100.00\% & \hlblue{95.00\%} & \hlblue{52.50\%} & \hlblue{60.00\%} & 0.00\% \\
\quad \quad \quad Wan-2.2 & \hlblue{0.00\%} & 97.50\% & 5.00\% & 67.50\% & 0.00\% \\
\multicolumn{6}{@{}l}{\quad \quad \textbf{Image Models}} \\
\quad \quad \quad Nano-banana & 15.00\% & 94.00\% & 10.00\% & N/A & \hlyellow{1.00\%} \\
\quad \quad \quad Nano-banana Pro & \hlyellow{0.00\%} & \hlyellow{50.00\%} & 7.50\% & N/A & 0.00\% \\
\quad \quad \quad GPT-4o-image & 100.00\% & 62.50\% & 0.00\% & N/A & 0.00\% \\
\quad \quad \quad GPT-image-1.5 & 1.67\% & 90.00\% & 5.83\% & N/A & 0.00\% \\
\quad \quad \quad Qwen-image & 98.33\% & 88.33\% & \hlyellow{100.00\%} & N/A & 0.00\% \\

\midrule
\multicolumn{6}{@{}l}{\textbf{Generator: Wilson's Algorithm}} \\

\multicolumn{6}{@{}l}{\quad \textit{Level: Easy (3$\times$3–5$\times$5)}} \\
\multicolumn{6}{@{}l}{\quad \quad \textbf{Video Models}} \\
\quad \quad \quad Veo-3 & 72.00\% & \hlblue{75.00\%} & 37.50\% & 79.00\% & \hlblue{\textbf{11.50\%}} \\
\quad \quad \quad Sora-2 & 100.00\% & 97.50\% & \hlblue{65.00\%} & \hlblue{100.00\%} & 0.00\% \\
\quad \quad \quad Wan-2.2 & \hlblue{0.00\%} & 82.50\% & 17.50\% & 90.00\% & 0.00\% \\
\multicolumn{6}{@{}l}{\quad \quad \textbf{Image Models}} \\
\quad \quad \quad Nano-banana & \hlyellow{0.00\%} & 83.50\% & 42.50\% & N/A & 5.00\% \\
\quad \quad \quad Nano-banana Pro & \hlyellow{0.00\%} & \hlyellow{65.00\%} & 30.00\% & N/A & \hlyellow{7.50\%} \\
\quad \quad \quad GPT-4o-image & 100.00\% & 92.50\% & 10.00\% & N/A & 0.00\% \\
\quad \quad \quad GPT-image-1.5 & 98.33\% & 94.17\% & 20.00\% & N/A & 0.83\% \\
\quad \quad \quad Qwen-image & 49.17\% & 90.83\% & \hlyellow{89.17\%} & N/A & 0.83\% \\

\multicolumn{6}{@{}l}{\quad \textit{Level: Medium (6$\times$6–9$\times$9)}} \\
\multicolumn{6}{@{}l}{\quad \quad \textbf{Video Models}} \\
\quad \quad \quad Veo-3 & 100.00\% & 97.50\% & 12.00\% & 56.00\% & 0.00\% \\
\quad \quad \quad Sora-2 & 100.00\% & \hlblue{95.00\%} & \hlblue{47.50\%} & \hlblue{100.00\%} & 0.00\% \\
\quad \quad \quad Wan-2.2 & \hlblue{5.00\%} & 100.00\% & 15.00\% & 95.00\% & 0.00\% \\
\multicolumn{6}{@{}l}{\quad \quad \textbf{Image Models}} \\
\quad \quad \quad Nano-banana & \hlyellow{0.00\%} & 93.50\% & 19.50\% & N/A & 0.00\% \\
\quad \quad \quad Nano-banana Pro & 2.50\% & \hlyellow{82.50\%} & 20.00\% & N/A & 0.00\% \\
\quad \quad \quad GPT-4o-image & 100.00\% & \hlyellow{82.50\%} & 0.00\% & N/A & 0.00\% \\
\quad \quad \quad GPT-image-1.5 & 45.00\% & 97.50\% & 14.17\% & N/A & 0.00\% \\
\quad \quad \quad Qwen-image & 80.83\% & 85.83\% & \hlyellow{99.17\%} & N/A & 0.00\% \\

\multicolumn{6}{@{}l}{\quad \textit{Level: Hard (10$\times$10–13$\times$13)}} \\
\multicolumn{6}{@{}l}{\quad \quad \textbf{Video Models}} \\
\quad \quad \quad Veo-3 & 100.00\% & 98.00\% & 6.00\% & 29.00\% & 0.00\% \\
\quad \quad \quad Sora-2 & 100.00\% & \hlblue{95.00\%} & \hlblue{55.00\%} & \hlblue{100.00\%} & 0.00\% \\
\quad \quad \quad Wan-2.2 & \hlblue{10.00\%} & 100.00\% & 0.00\% & 92.50\% & 0.00\% \\
\multicolumn{6}{@{}l}{\quad \quad \textbf{Image Models}} \\
\quad \quad \quad Nano-banana & 11.00\% & 99.50\% & \hlyellow{19.00\%} & N/A & 0.00\% \\
\quad \quad \quad Nano-banana Pro & \hlyellow{0.00\%} & \hlyellow{82.50\%} & 15.00\% & N/A & \hlyellow{\textbf{5.00\%}} \\
\quad \quad \quad GPT-4o-image & 100.00\% & 67.50\% & 0.00\% & N/A & 0.00\% \\
\quad \quad \quad GPT-image-1.5 & 0.00\% & 97.50\% & 5.00\% & N/A & 0.00\% \\
\quad \quad \quad Qwen-image & 100.00\% & 88.33\% & 100.00\% & N/A & 0.00\% \\

\bottomrule
\end{tabular}
}
\end{adjustbox}
\end{table*}

\textbf{Quantitative Evaluation Results.} \Cref{tab:maze_all_results_vis} presents the pixel-based evaluation results across all models and difficulty levels. The most striking observation is the \textbf{substantially lower overall scores} compared to VLM-based evaluation, confirming that pixel-level analysis imposes far stricter correctness criteria. Across all configurations, the best-performing model achieves only 30.00\% overall success (Nano-banana Pro on DFS Medium), with most models scoring at or near zero on Medium and Hard difficulty levels. Among \textbf{video models}, Veo-3 emerges as the only model capable of achieving non-zero overall scores, reaching 10.50\% on DFS Easy and 11.50\% on Wilson's Easy. However, its performance collapses entirely at Medium and Hard levels, where Cross Wall rates escalate to 96--99\% and Maze Changed rates reach 100\%. This confirms that Veo-3's apparent competence at easier levels does not generalize to more complex mazes. Sora-2 exhibits 100\% Maze Changed across all configurations, indicating complete structural hallucination that renders target achievement scores (45--65\%) meaningless—the model solves self-generated mazes rather than the input puzzles. Wan-2.2 demonstrates excellent layout preservation (0--10\% Maze Changed) but suffers from near-universal wall crossing violations (82--100\%), suggesting the model maintains visual fidelity while completely disregarding physical constraints. Among \textbf{image models}, Nano-banana Pro achieves the highest overall score at 30.00\% on DFS Medium, benefiting from relatively low Cross Wall rates (30--82.5\%) compared to other models. Nano-banana shows competitive performance on Easy tasks (5--14\%) but degrades rapidly at higher difficulties. Notably, Qwen-image achieves near-perfect Target Achievement scores (89--100\%) across all levels, yet scores near zero overall due to high Maze Changed (49--100\%) and Cross Wall (84--90\%) rates—demonstrating that reaching the target is trivial when the model ignores maze constraints. GPT-4o-image fails uniformly with 100\% Maze Changed and 0\% Target Achievement, confirming it cannot process input maze structures. In contrast, GPT-image-1.5 shows substantial improvement over GPT-4o-image: it achieves non-zero overall scores on DFS Easy (1.67\%) and DFS Medium (3.33\%), and demonstrates dramatically better layout preservation at Hard levels (0--1.67\% Maze Changed on DFS/Wilson Hard versus 100\% for GPT-4o-image). However, GPT-image-1.5 still suffers from high Cross Wall rates (87--97\%) and low Target Achievement (5--20\%), indicating that while it can maintain maze structure, it struggles with precise path placement within corridors. The results reveal a \textbf{clear difficulty scaling effect}: overall scores decrease monotonically from Easy to Hard for all models, with Cross Wall rates increasing correspondingly. This pattern suggests that current generative models lack genuine spatial reasoning capabilities—their limited success on Easy mazes reflects memorization or lucky pattern matching rather than algorithmic path planning. The pixel-based evaluation thus provides a stricter diagnostic assessment of model capabilities, exposing performance gaps that VLM-based metrics can underestimate.



\begin{figure*}[h]
    \centering
    \includegraphics[width=0.65\textwidth, trim=15 270 190 0, clip]{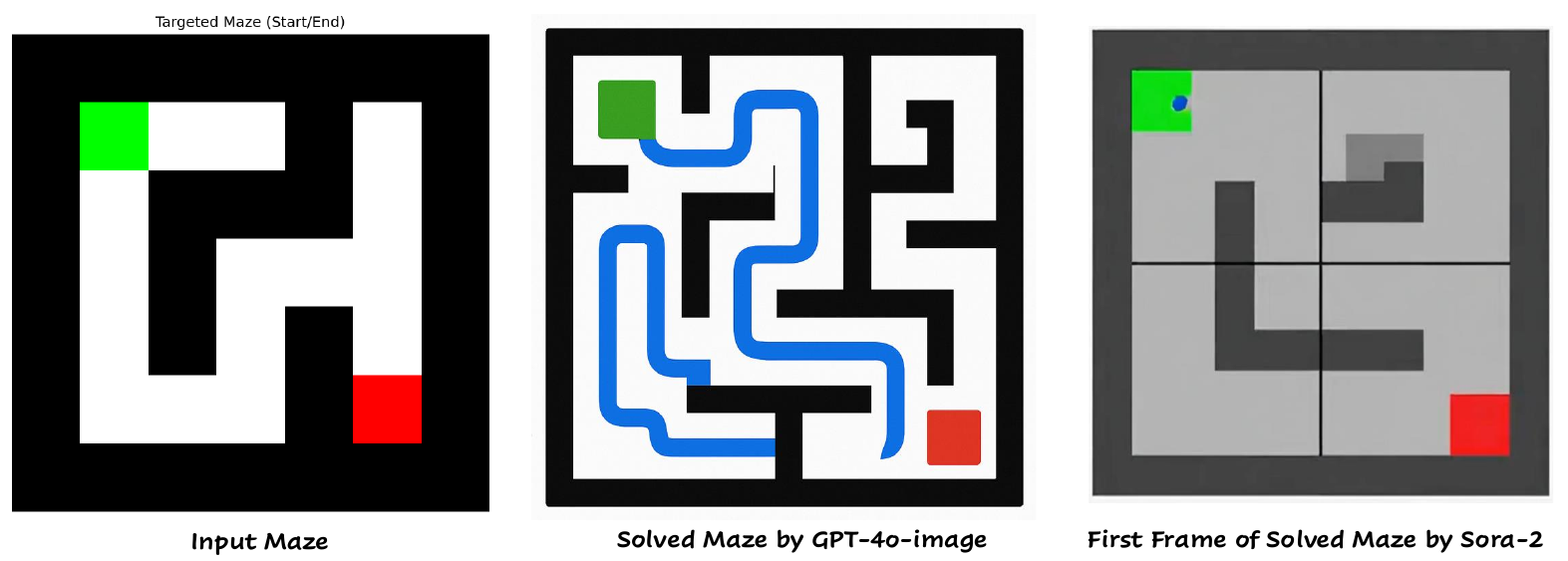}
    \caption{Input maze (left) and altered maze layouts generated by GPT-4o-image (middle) and Sora-2 (right).}
    \label{fig:maze_layout_mismatch}
\end{figure*}

\textbf{Failure Modes Analysis.} To investigate the root causes of model failure, we conduct an in-depth examination of two primary failure modes: \textbf{Maze Changing} and \textbf{Wall Crossing}. Manual review of OpenAI's video model Sora-2 and image model GPT-4o-image reveals systematic disregard for input maze structures, with generated outputs bearing little resemblance to the provided layouts. Quantitative analysis confirms this observation: both models exhibit a $100\%$ maze changing rate across all difficulty levels, accompanied by extremely low average maze structure IoU scores ($0.3210$ for GPT-4o-image and $0.1260$ for Sora-2), indicating complete structural hallucination rather than minor deviations (see \Cref{fig:maze_layout_mismatch} for representative examples). Notably, GPT-image-1.5 exhibits a markedly different failure profile: while it still struggles with Easy-level mazes (96--98\% Maze Changed), it demonstrates strong structural preservation at Hard levels (0--1.67\% Maze Changed), suggesting that the newer model has improved input conditioning capabilities. However, GPT-image-1.5's primary failure mode shifts to wall crossing (87--97\% across all levels), indicating that while it can faithfully reproduce maze layouts, it lacks the spatial reasoning to generate valid paths within those constraints. In contrast, Veo-3 demonstrates substantially better structural preservation with an average IoU of $0.6573$. However, its elevated maze changing rate stems from a fundamentally different failure mode—\textit{partial} layout deviations induced by wall crossing violations rather than wholesale structural rejection. To quantify this relationship, we analyze Veo-3 evaluation cases and observe a strong association between layout match failure and wall crossing action. The two conditions co-occur in $84.99\%$ of all cases, with highly directional conditional probabilities: $93.92\%$ of layout failures involve wall crossing, while $94.18\%$ of wall crossing events trigger layout failure. The Phi coefficient ($\phi = 0.3821$) indicates a statistically significant moderate positive correlation, confirming that wall crossing serves as a strong predictor of layout match failure. This finding validates our gating mechanism's design—when agents violate physical maze constraints, the penalty system correctly identifies and penalizes the error. The analysis reveals that for models like Veo-3 and GPT-image-1.5, structural failures are frequently downstream consequences of physical constraint violations, whereas for Sora-2 and GPT-4o-image, failures reflect fundamental inability to respect input spatial constraints.

\section{Sudoku}
\label{apx:sudoku}

We introduce the \textbf{Sudoku task} to evaluate a model's core abilities in \textbf{Constraint Satisfaction} and \textbf{Logical Reasoning}. This task directly probes a model's capacity for \textbf{logical inference}---its ability to derive valid conclusions under a structured set of logical rules. In Sudoku, the model must internalize the underlying constraints that govern valid solutions, ensuring that each symbol appears exactly once in every row, column, and subgrid. Moreover, the task explicitly tests \textbf{deductive reasoning}: the step-by-step application of these constraints to infer the only logically consistent values for each empty cell. Successful completion thus requires the model to integrate global structural understanding with local deductive consistency, producing a fully valid and complete grid \citep{seely2025sudoku}.

\begin{figure*}[h!]
    \centering
    \small
    \begin{tabular}{cccc}

        \subcaptionbox{4$\times$4 (Easy)\label{fig:4x4_easy}}{
            \includegraphics[width=0.2\textwidth]{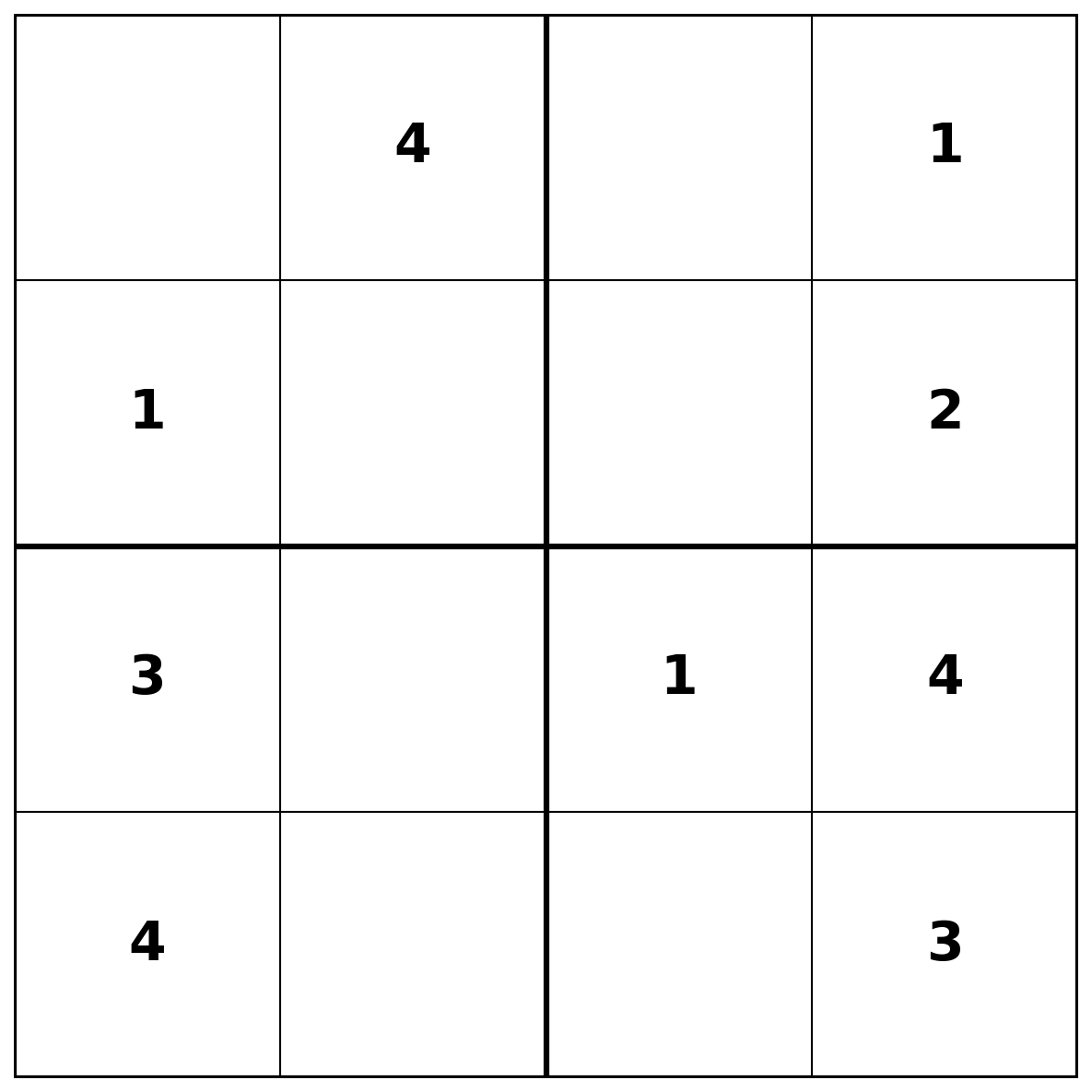}
        } &
        \subcaptionbox{4$\times$4 (Easy) Solution\label{fig:4x4_easy_solution}}{
            \includegraphics[width=0.2\textwidth]{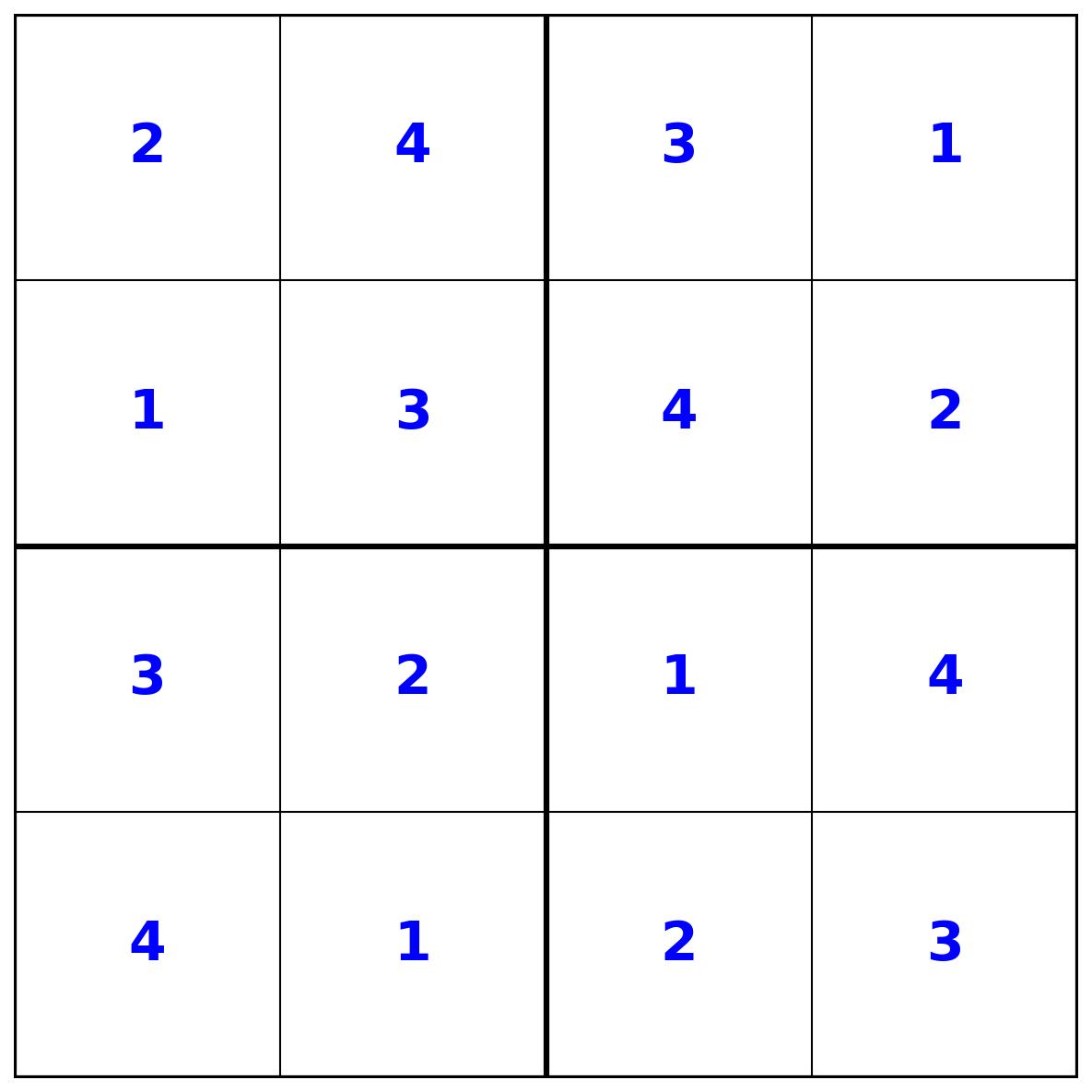}
        } &
        \subcaptionbox{9$\times$9 (Easy)\label{fig:9x9_easy}}{
            \includegraphics[width=0.2\textwidth]{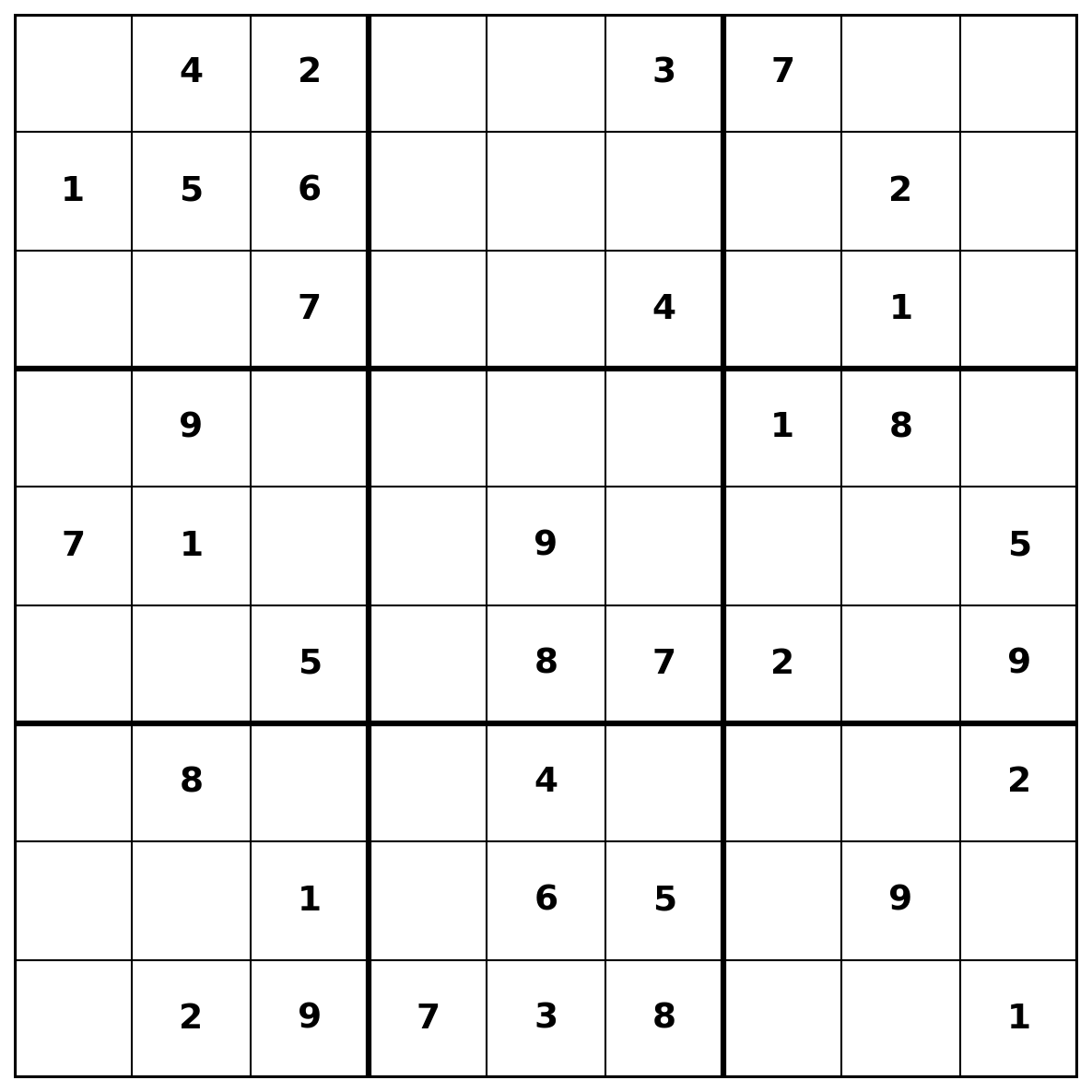}
        } &
        \subcaptionbox{9$\times$9 (Easy) Solution\label{fig:9x9_easy_solution}}{
            \includegraphics[width=0.2\textwidth]{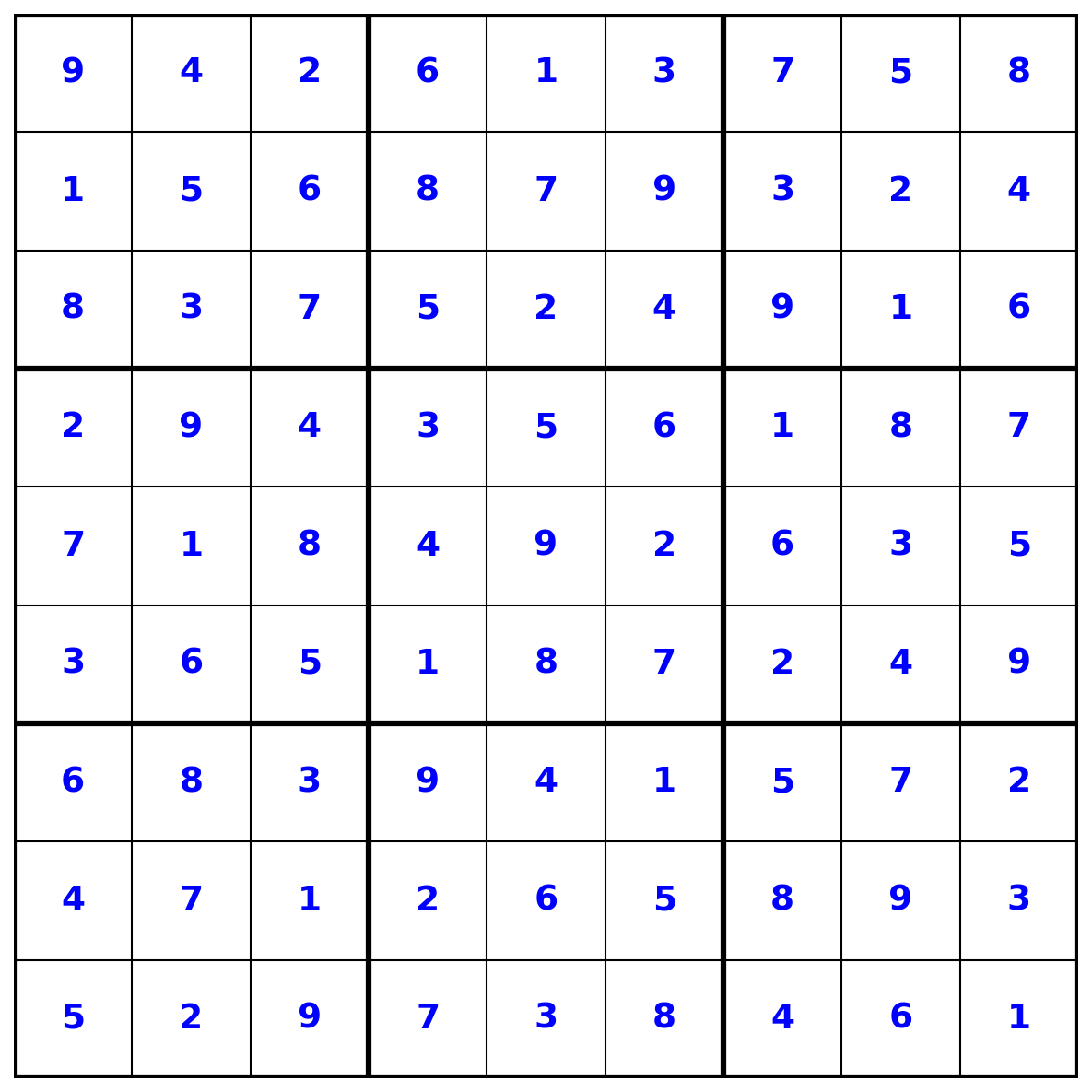}
        } \\[0.8em]
        \subcaptionbox{4$\times$4 (Medium)\label{fig:4x4_medium}}{
            \includegraphics[width=0.2\textwidth]{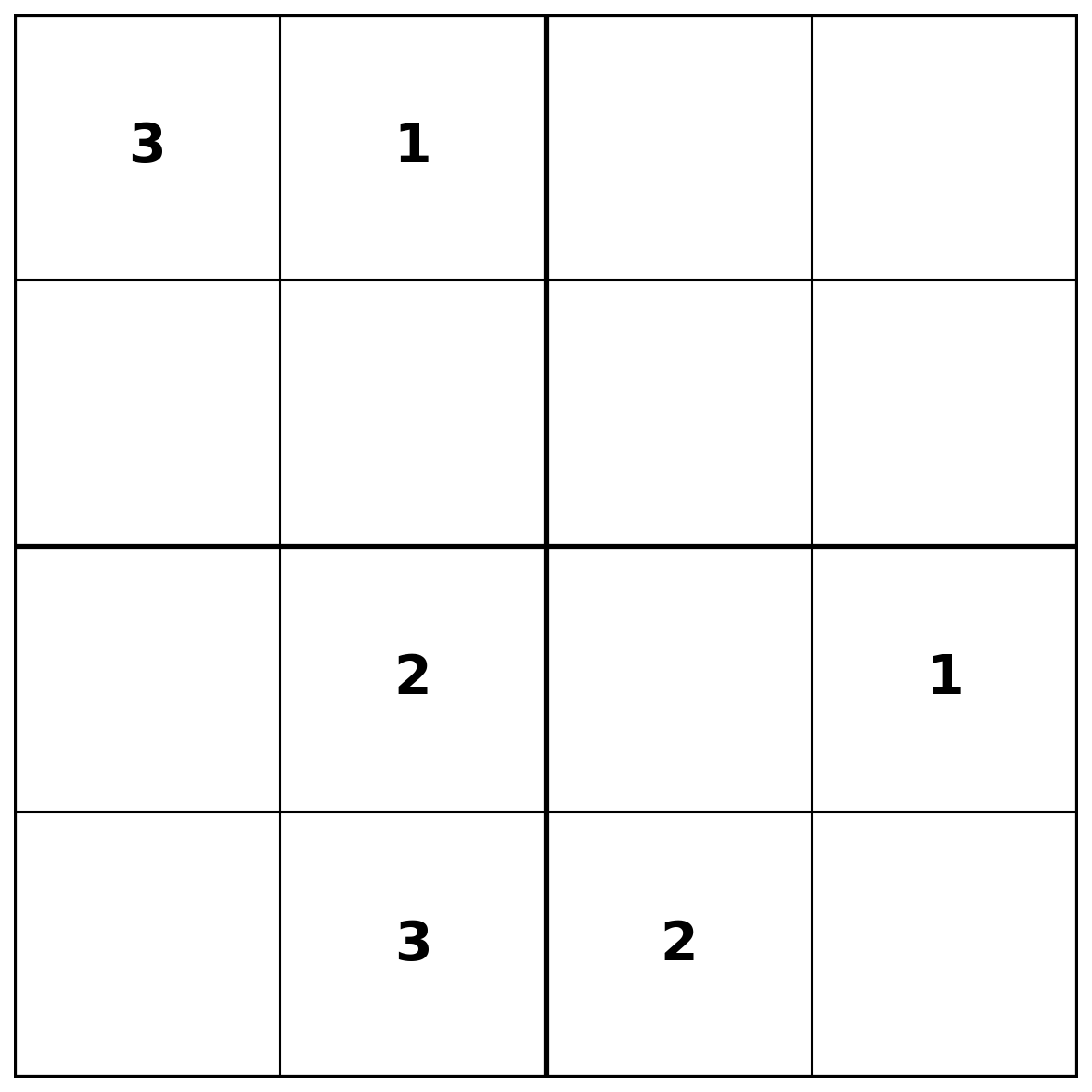}
        } &
        \subcaptionbox{4$\times$4 (Medium) Solution\label{fig:4x4_medium_solution}}{
            \includegraphics[width=0.2\textwidth]{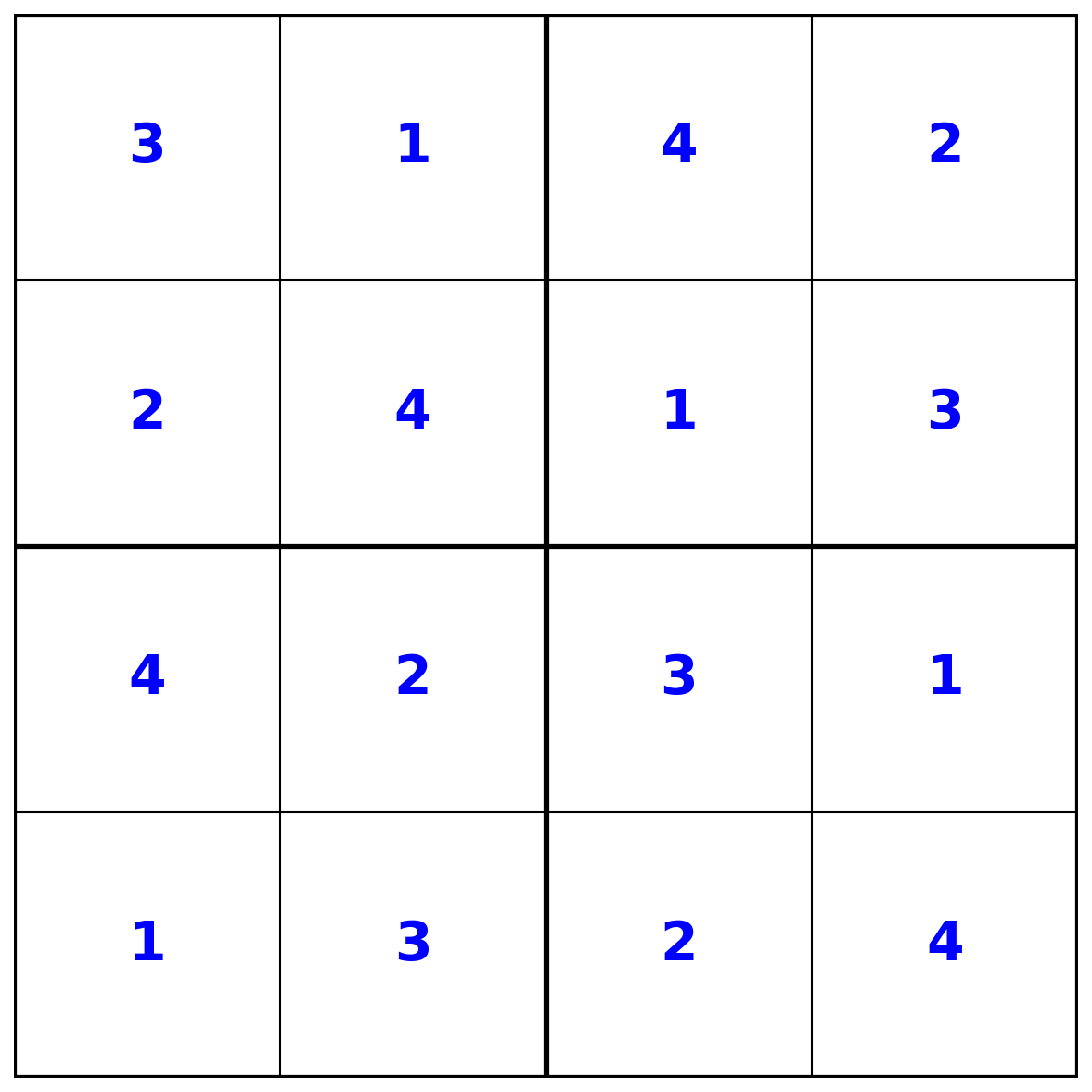}
        } &
        \subcaptionbox{9$\times$9 (Medium)\label{fig:9x9_medium}}{
            \includegraphics[width=0.2\textwidth]{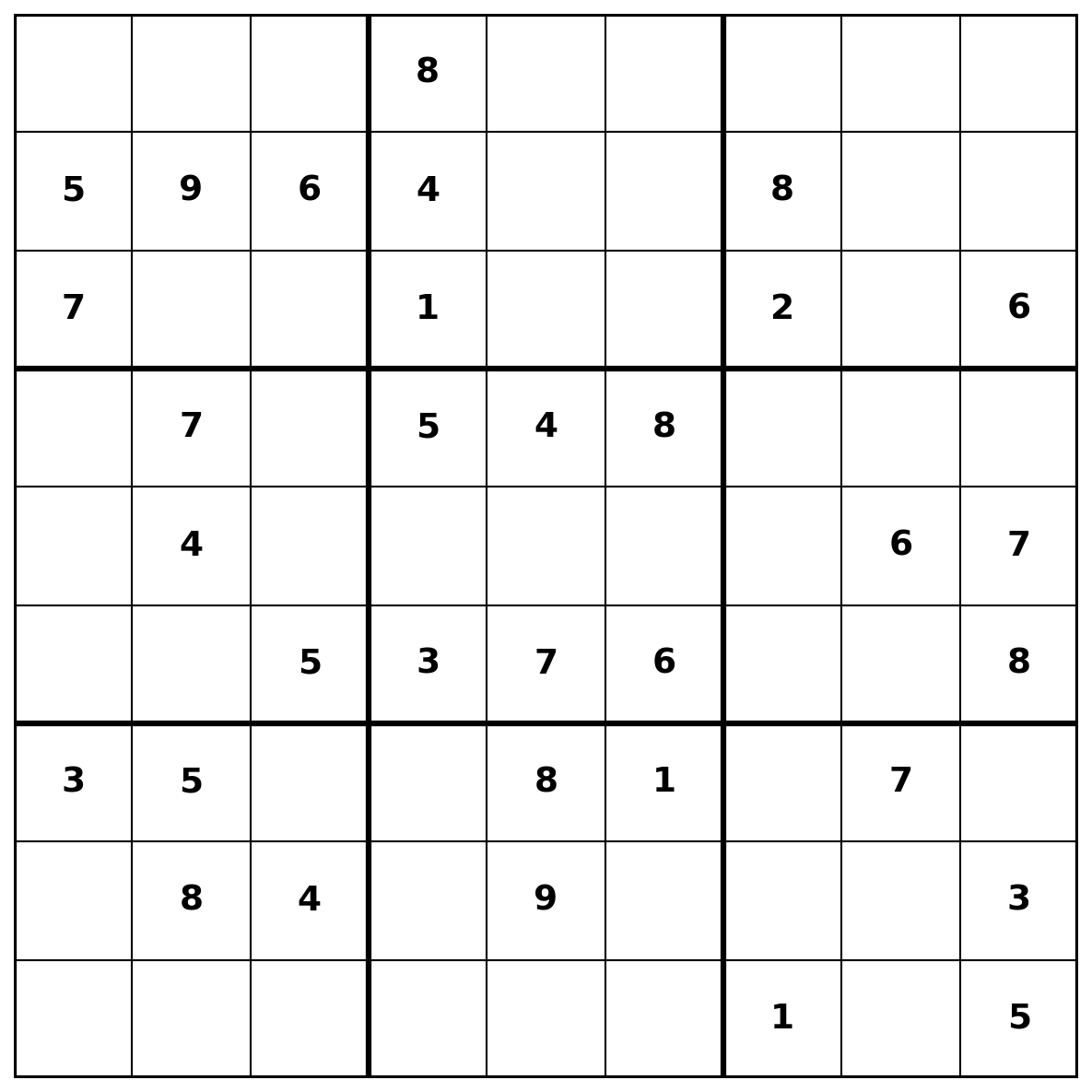}
        } &
        \subcaptionbox{9$\times$9 (Medium) Solution\label{fig:9x9_medium_solution}}{
            \includegraphics[width=0.2\textwidth]{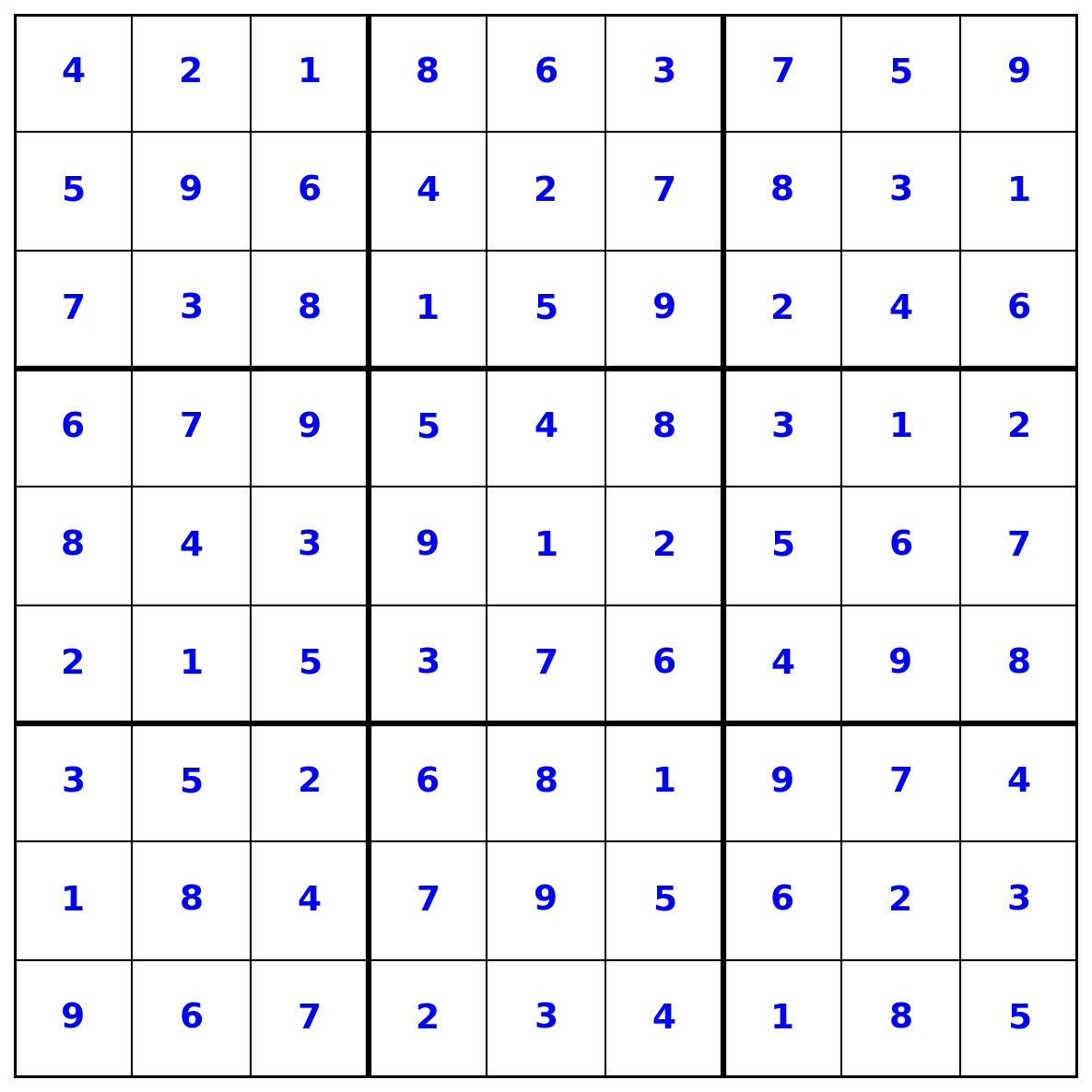}
        } \\[0.8em]
        \subcaptionbox{4$\times$4 (Hard)\label{fig:4x4_hard}}{
            \includegraphics[width=0.2\textwidth]{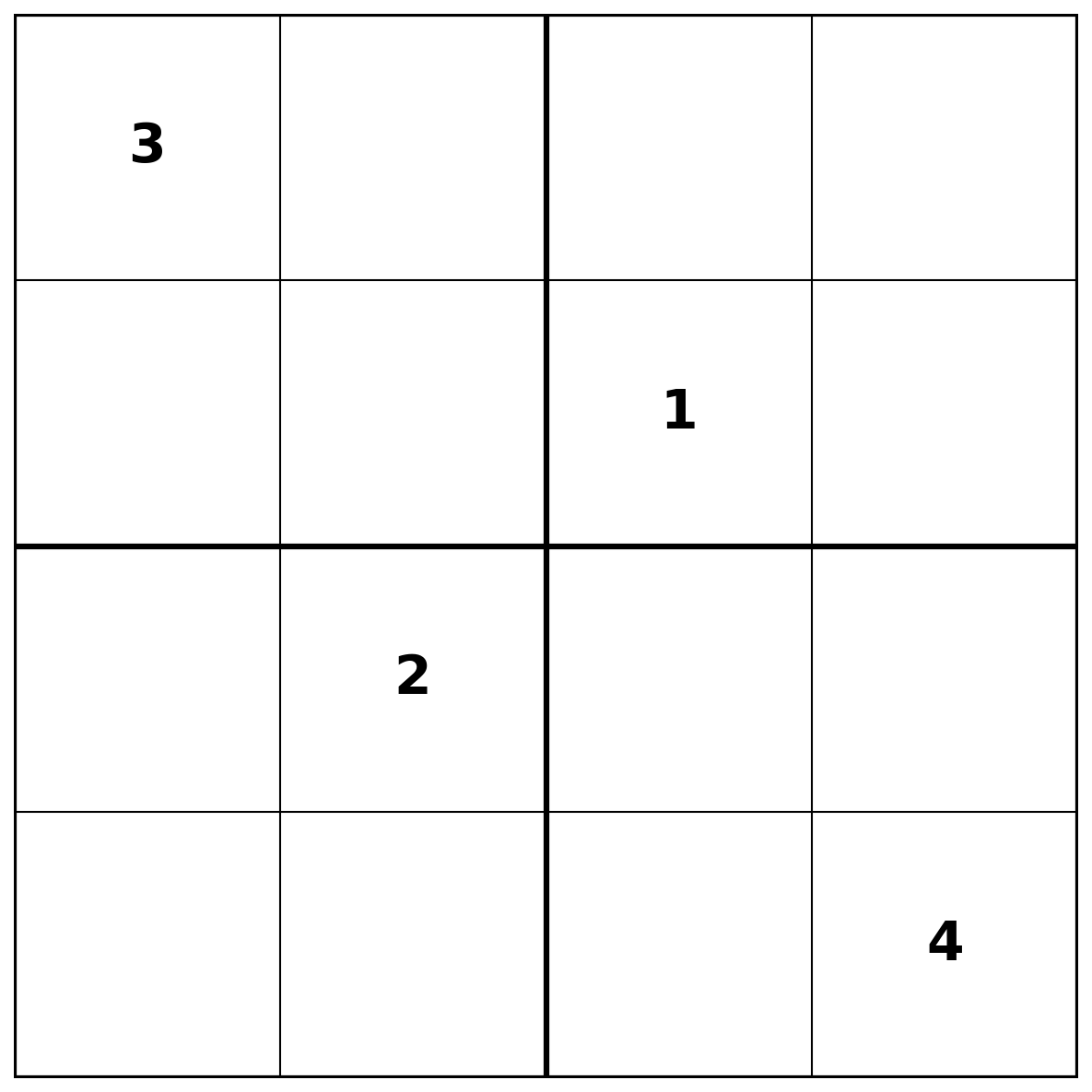}
        } &
        \subcaptionbox{4$\times$4 (Hard) Solution\label{fig:4x4_hard_solution}}{
            \includegraphics[width=0.2\textwidth]{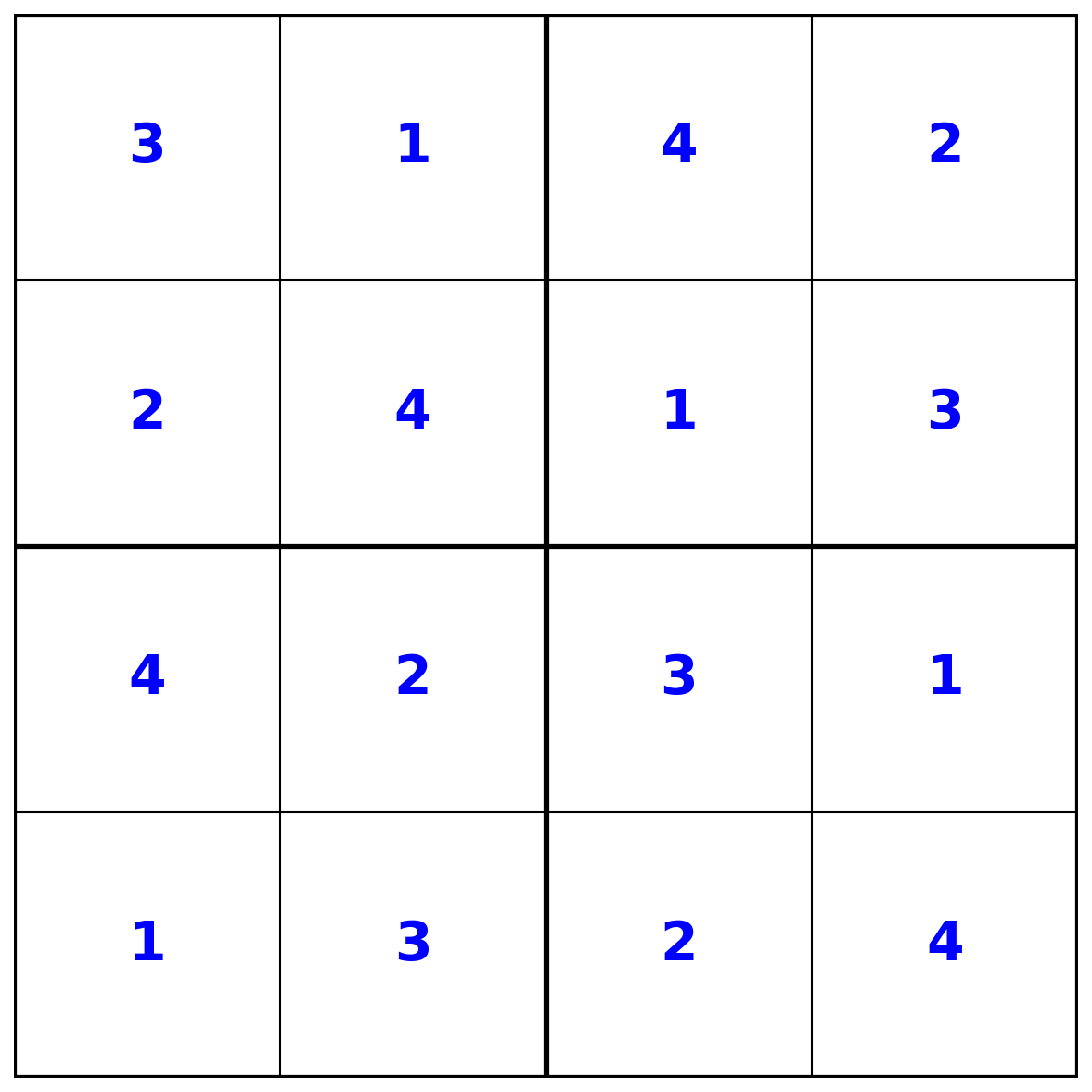}
        } &
        \subcaptionbox{9$\times$9 (Hard)\label{fig:9x9_hard}}{
            \includegraphics[width=0.2\textwidth]{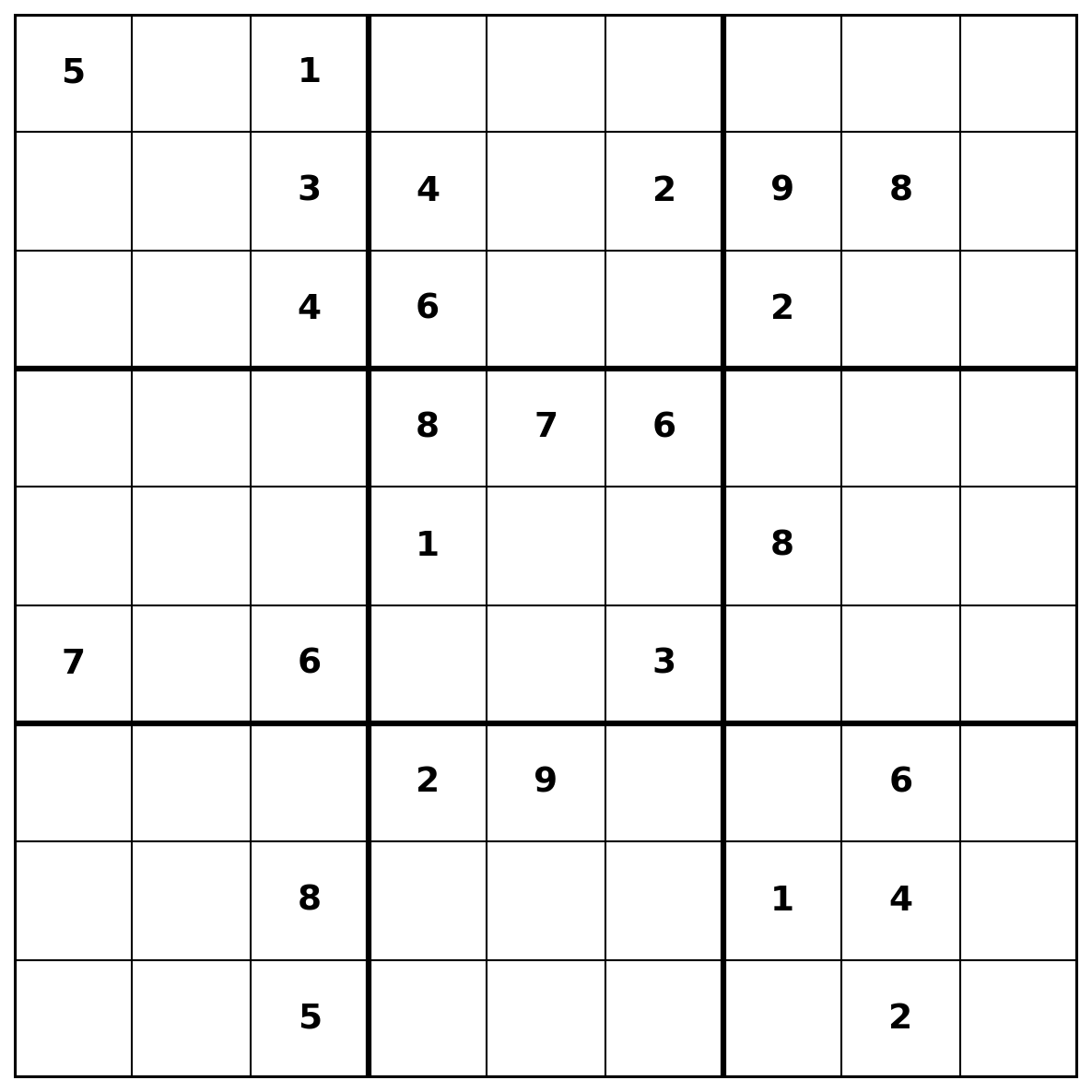}
        } &
        \subcaptionbox{9$\times$9 (Hard) Solution\label{fig:9x9_hard_solution}}{
            \includegraphics[width=0.2\textwidth]{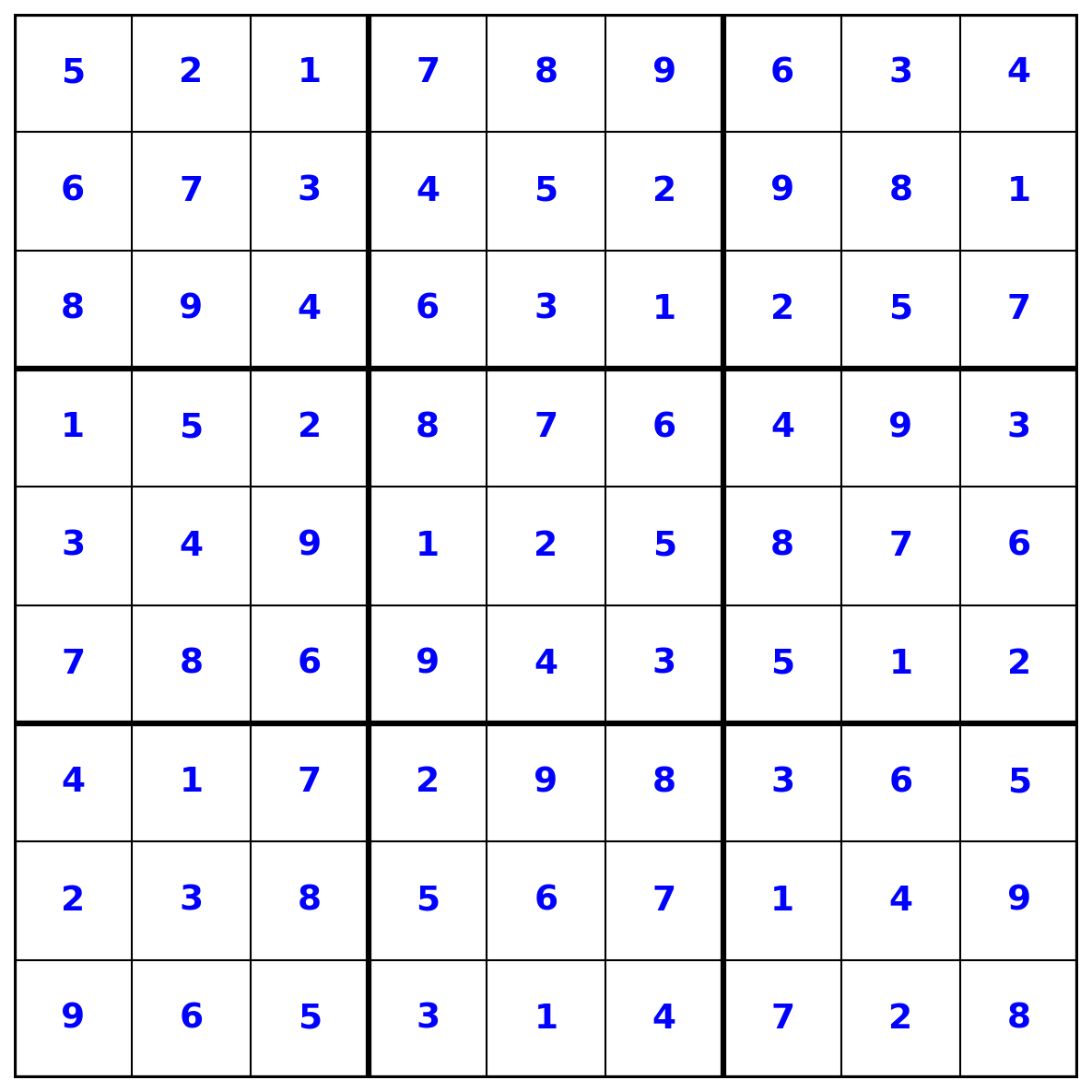}
        }
    \end{tabular}
    \caption{A 3$\times$4 grid showing Sudoku puzzles and their solutions across \textit{Easy}, \textit{Medium}, and \textit{Hard} difficulty levels for both 4$\times$4 and 9$\times$9 grid sizes, arranged so each puzzle is immediately followed by its solution.}
    \label{fig:sudoku_examples}
\end{figure*}

\subsection{Hard-Level Control}
\label{sec:appendix_sudoku_hardlevel}

To construct a diverse yet systematically controlled evaluation set, we programmatically generate puzzles that span a wide range of structural properties and reasoning demands. Task difficulty is explicitly controlled by varying parameters along \textit{two} primary axes, enabling fine-grained analysis of how model performance scales with increasing structural complexity and reasoning depth:
\vspace{-2mm}
\begin{itemize}
    \item \textbf{Grid Size (2 levels):} We generate puzzles in two standard configurations---4$\times$4 (digits 1--4) and 9$\times$9 (digits 1--9). The larger grid substantially increases combinatorial difficulty and requires deeper multi-step logical inference.
    \item \textbf{Puzzle Difficulty (3 levels):} For each grid size, we control difficulty by varying the number of initial \textit{clues} (pre-filled digits). The three levels—\textit{Easy} (many clues), \textit{Medium}, and \textit{Hard} (few clues while ensuring a unique solution)—modulate the search space and constraint-satisfaction complexity.
\end{itemize}

For every difficulty level (\textit{Easy}, \textit{Medium}, \textit{Hard}), we generate 100 puzzles: 50 for the 4$\times$4 grid and 50 for the 9$\times$9 grid. This results in a balanced evaluation set of \textbf{300} Sudoku puzzles spanning a controlled spectrum of structural sizes and reasoning challenges. \Cref{fig:sudoku_examples} provides representative puzzles and solutions for each difficulty level and grid size, with each puzzle displayed directly above its corresponding solution.

\subsection{Evaluation and Metrics}
\label{apx:sudoku_evaluation}

We evaluate both video and image outputs using a Vision-Language Model (VLM)–based evaluator, Gemini-2.5-Pro~\citep{comanici2025gemini}. The evaluator receives \textit{three} inputs: (i) the generated video or image, (ii) the ground-truth solved Sudoku grid, and (iii) a structured evaluation prompt (with modality-specific variants). Using these inputs, the VLM assesses whether the model produces a valid Sudoku solution and identifies failure modes related to rule consistency, clue preservation, and reasoning behavior.

From the evaluator's structured outputs, we derive \textbf{one} primary success metric and \textbf{four} fine-grained metrics:
\vspace{-2mm}
\begin{itemize}
    \item \textbf{Clues Changed (CC, Failure Mode)}: (i) Video: 1 if any original digits (``clues'') are modified, removed, or displaced in \textit{any} frame; 0 if all clues remain intact across the entire sequence. (ii) Image: 1 if any given clue differs from the original puzzle image; 0 otherwise.
    \item \textbf{Constraint Violation (CV, Failure Mode)}: The fraction of Sudoku constraints violated in the final output, including row, column, and subgrid constraints. A value of 0 indicates full rule compliance, while higher values indicate more severe violations.
    \item \textbf{Completion Accuracy (CA)}: The fraction of correctly filled \textit{originally empty} cells, computed by comparing the model’s final output against the ground-truth solution.
    \item \textbf{Action Reflection (AR)}: (i) Video: 1 if the video exhibits explicit \emph{reflection actions}, such as correcting earlier entries; 0 if updates appear simultaneously or occur in a non-reflective order. (ii) Image: Not applicable.
    \item \textbf{Overall Score:} 1 only if \textbf{Clues Changed=0} AND \textbf{Constraints Violation=0} AND \textbf{Completion Accuracy=1}; 0 otherwise.
\end{itemize}

\vspace{-3mm}

\begin{figure*}[h]
    \centering
    \includegraphics[width=0.9\textwidth, trim=10 170 0 0, clip]{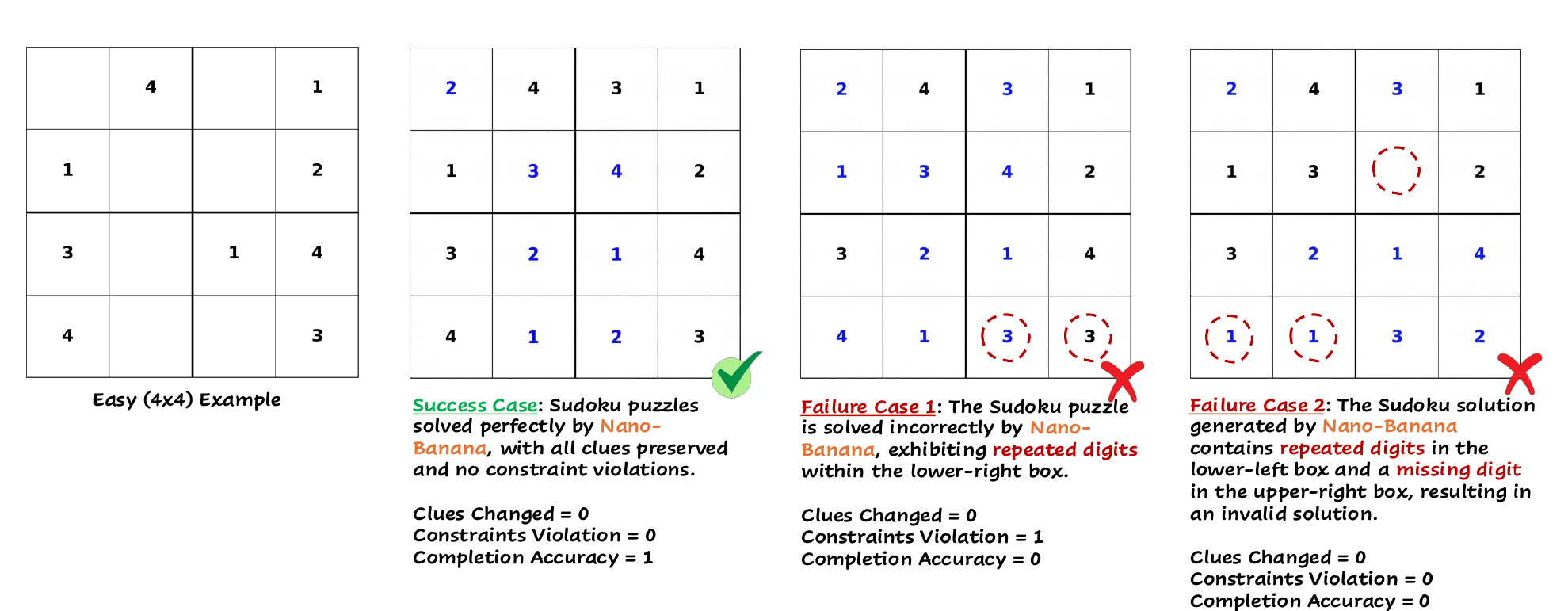}
    \caption{Case Study: Success and failure cases generated by \textbf{Nano-banana}.}
    \vspace{-3mm}
    \label{fig:sudoku_case_study_1}
\end{figure*}

\begin{figure*}[h]
    \centering
    \includegraphics[width=\textwidth, trim=15 0 15 0, clip]{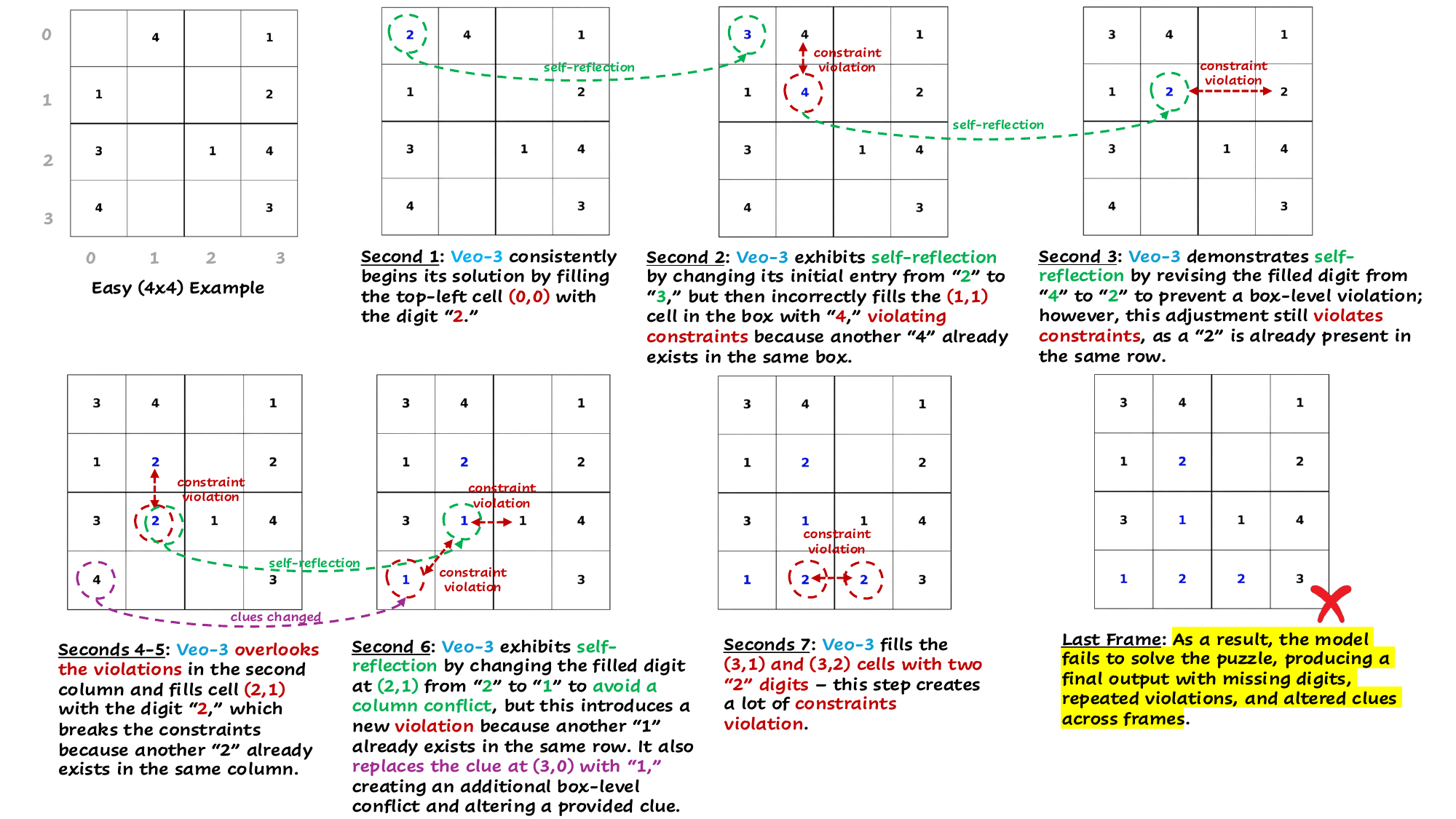}
    \caption{Case Study: Failure cases generated by \textbf{Veo-3} with frame-wise analysis highlighting \textit{three} key behaviors: positional bias, self-reflective edits, and temporal drift.}
    \vspace{-3mm}
    \label{fig:sudoku_case_study_2}
\end{figure*}

\subsection{Case Study}

\Cref{fig:sudoku_case_study_1} and \Cref{fig:sudoku_case_study_2} illustrate several solution trajectories generated by Nano-banana (image generative model) and Veo-3 (video generative model) for the same 4$\times$4 easy Sudoku puzzle. In the successful case, Nano-banana solves the puzzle perfectly—preserving all initial clues, following valid reasoning steps, and reaching the correct final configuration without any constraint violations. However, when the model fails, \textbf{the error patterns align closely with our diagnostic evaluation metrics: missing digits, repeated numbers within a row or box, constraint violations across frames, and changes to the original clues}. These failures often arise despite our detailed, instruction-focused generation prompts, suggesting that \textbf{Nano-banana still lacks the deeper abstract reasoning capacity required for consistent multi-step logical problem solving}. Although \textbf{Veo-3 performs worse overall} in producing fully correct solutions, its video outputs reveal a set of \textbf{systematic and interpretable reasoning behaviors}: \textbf{\underline{(1)}} Veo-3 demonstrates a strong \textcolor{blue}{\textbf{positional bias}}: it almost always initiates its reasoning from the top-left box of the grid. Humans, in contrast, typically adapt their solving order based on puzzle structure, difficulty, or available constraints—suggesting that \textbf{Veo-3 relies more on a fixed procedural heuristic than on dynamic reasoning}. \textbf{\underline{(2)}} Veo-3 frequently engages in \textcolor{blue}{\textbf{self-reflective edits}} during the generation process. These edits occasionally help the model avoid violations (\textit{e.g.}, at second 3, where it changes the (1,1) cell from ``4'' to ``2'' to resolve a conflict), but they can also be detrimental. At second 2, for example, Veo-3 overwrites a correct digit, changing the (0,0) entry from ``2'' to ``3,'' instantly invalidating the entire solution. This behavior illustrates a broader challenge in generative reasoning models: \textbf{self-reflection without grounded logical consistency can introduce more instability than benefit}. \textbf{\underline{(3)}} Veo-3 exhibits a characteristic \textcolor{blue}{\textbf{temporal drift}} pattern. Its early frames often remain stable and respect the given clues, but as generation progresses, the model increasingly modifies clues, introduces inconsistencies, or oscillates between alternative partial solutions. This suggests that the \textbf{model lacks a persistent internal representation of constraints, leading to reasoning degradation over time}, even when the prompt explicitly prohibits altering clues.

Overall, these case studies highlight the \textbf{gap between current generative models and true abstract reasoning ability}. Both Nano-banana and Veo-3 demonstrate surface-level competence—filling cells, correcting mistakes, or imitating stepwise processes—but neither maintains a robust, constraint-aware reasoning trajectory throughout the entire solution. These observations reinforce the importance of MMGR's diagnostic metrics: beyond assessing whether the final answer is correct, evaluating \emph{how} the model reasons is essential for understanding the strengths and limitations of generative reasoning models.

\begin{table*}[h!]
\centering
\small
\caption{Quantitative results for the \textbf{Sudoku} task using \textbf{VLM-based} evaluation method. We evaluate video models (Veo-3, Sora-2, and Wan-2.2) and image models (Nano-banana, Nano-banana Pro, GPT-4o-image, GPT-image-1.5, and Qwen-image) on two grid sizes (4$\times$4, 9$\times$9) and three difficulty levels (Easy, Medium, and Hard). Because image outputs do not support frame-by-frame reasoning, action-reflection–based metrics are omitted for image generative models (marked with ``N/A''). The best results among video models are highlighted with \colorbox{paleblue}{blue}, and the best results among image models are highlighted with \colorbox{paleyellow}{yellow}.}
\label{tab:sudoku_results}

\begin{adjustbox}{max width=0.82\textwidth}
{
    \begin{tabular}{@{}lccccc@{}}
    \toprule
    & \multicolumn{4}{c}{\textbf{Fine-grained Metrics}} & \multicolumn{1}{c}{\textbf{Primary Metric}} \\
    \cmidrule(lr){2-5} \cmidrule(lr){6-6}

    \textbf{Model} & \textbf{Clues Changed} $\downarrow$ & \textbf{Constraints Violation} $\downarrow$ & \textbf{Completion Accuracy} $\uparrow$ & \textbf{Action Reflection} $\uparrow$ & \textbf{Overall} $\uparrow$ \\
    \midrule

    \multicolumn{6}{@{}l}{\textbf{Grid Size: 4$\times$4}} \\

    \multicolumn{6}{@{}l}{\quad \textit{Level: Easy}} \\
    \multicolumn{6}{@{}l}{\quad \quad \textbf{Video Models}} \\
    \quad \quad \quad Veo-3 & \hlblue{72.40\%} & 98.40\% & \hlblue{37.22\%} & \hlblue{92.00\%} & \hlblue{\textbf{1.20\%}} \\
    \quad \quad \quad Sora-2 & 100.00\% & \hlblue{95.92\%} & 25.78\% & 32.65\% & 0.00\% \\
    \quad \quad \quad Wan-2.2 & 100.00\% & 100.00\% & 17.22\% & 4.67\% & 0.00\% \\
    \multicolumn{6}{@{}l}{\quad \quad \textbf{Image Models}} \\
    \quad \quad \quad Nano-banana & \hlyellow{18.61\%} & 17.20\% & 81.14\% & N/A & 0.00\% \\
    \quad \quad \quad Nano-banana Pro & 35.25\% & \hlyellow{1.33\%} & \hlyellow{97.05\%} & N/A & 0.00\% \\
    \quad \quad \quad GPT-4o-image & 21.55\% & 19.26\% & 74.99\% & N/A & 0.00\% \\
    \quad \quad \quad GPT-image-1.5 & 63.33\% & 32.06\% & 64.57\% & N/A & \hlyellow{\textbf{22.67\%}} \\
    \quad \quad \quad Qwen-image & 59.66\% & 45.33\% & 83.30\% & N/A & 0.00\% \\

    \multicolumn{6}{@{}l}{\quad \textit{Level: Medium}} \\
    \multicolumn{6}{@{}l}{\quad \quad \textbf{Video Models}} \\
    \quad \quad \quad Veo-3 & \hlblue{75.60\%} & 100.00\% & \hlblue{37.61\%} & \hlblue{95.60\%} & 0.00\% \\
    \quad \quad \quad Sora-2 & 100.00\% & 100.00\% & 21.71\% & 61.90\% & 0.00\% \\
    \quad \quad \quad Wan-2.2 & 100.00\% & 100.00\% & 14.36\% & 6.00\% & 0.00\% \\
    \multicolumn{6}{@{}l}{\quad \quad \textbf{Image Models}} \\
    \quad \quad \quad Nano-banana & \hlyellow{21.42\%} & 21.30\% & 67.45\% & N/A & 0.00\% \\
    \quad \quad \quad Nano-banana Pro & 34.81\% & \hlyellow{0.50\%} & \hlyellow{94.73\%} & N/A & 0.00\% \\
    \quad \quad \quad GPT-4o-image & 33.16\% & 22.45\% & 55.54\% & N/A & 0.00\% \\
    \quad \quad \quad GPT-image-1.5 & 60.67\% & 37.39\% & 52.99\% & N/A & \hlyellow{\textbf{10.67\%}} \\
    \quad \quad \quad Qwen-image & 37.82\% & 43.67\% & 82.83\% & N/A & 0.00\% \\

    \multicolumn{6}{@{}l}{\quad \textit{Level: Hard}} \\
    \multicolumn{6}{@{}l}{\quad \quad \textbf{Video Models}} \\
    \quad \quad \quad Veo-3 & \hlblue{70.00\%} & 100.00\% & \hlblue{30.38\%} & \hlblue{94.80\%} & 0.00\% \\
    \quad \quad \quad Sora-2 & 100.00\% & \hlblue{91.30\%} & 20.03\% & 60.87\% & 0.00\% \\
    \quad \quad \quad Wan-2.2 & 99.33\% & 100.00\% & 14.79\% & 6.00\% & 0.00\% \\
    \multicolumn{6}{@{}l}{\quad \quad \textbf{Image Models}} \\
    \quad \quad \quad Nano-banana & \hlyellow{25.84\%} & 25.80\% & 58.79\% & N/A & 0.00\% \\
    \quad \quad \quad Nano-banana Pro & 36.53\% & \hlyellow{0.83\%} & \hlyellow{91.08\%} & N/A & 0.00\% \\
    \quad \quad \quad GPT-4o-image & 35.73\% & 27.70\% & 47.49\% & N/A & 0.00\% \\
    \quad \quad \quad GPT-image-1.5 & 81.33\% & 39.11\% & 40.39\% & N/A & \hlyellow{\textbf{7.33\%}} \\
    \quad \quad \quad Qwen-image & 35.00\% & 47.50\% & 84.69\% & N/A & 0.00\% \\

    \midrule
    \multicolumn{6}{@{}l}{\textbf{Grid Size: 9$\times$9}} \\

    \multicolumn{6}{@{}l}{\quad \textit{Level: Easy}} \\
    \multicolumn{6}{@{}l}{\quad \quad \textbf{Video Models}} \\
    \quad \quad \quad Veo-3 & \hlblue{68.40\%} & 100.00\% & 15.47\% & \hlblue{70.00\%} & 0.00\% \\
    \quad \quad \quad Sora-2 & 95.74\% & \hlblue{95.74\%} & 8.47\% & 34.04\% & \hlblue{\textbf{4.26\%}} \\
    \quad \quad \quad Wan-2.2 & 87.33\% & 100.00\% & \hlblue{18.03\%} & 8.00\% & 0.00\% \\
    \multicolumn{6}{@{}l}{\quad \quad \textbf{Image Models}} \\
    \quad \quad \quad Nano-banana & \hlyellow{17.59\%} & 31.66\% & 58.73\% & N/A & 0.00\% \\
    \quad \quad \quad Nano-banana Pro & 19.09\% & 33.67\% & 51.92\% & N/A & 0.00\% \\
    \quad \quad \quad GPT-4o-image & 69.39\% & 59.57\% & 20.23\% & N/A & 0.00\% \\
    \quad \quad \quad GPT-image-1.5 & 100.00\% & 88.79\% & 13.85\% & N/A & 0.00\% \\
    \quad \quad \quad Qwen-image & 29.14\% & \hlyellow{5.78\%} & \hlyellow{72.02\%} & N/A & 0.00\% \\

    \multicolumn{6}{@{}l}{\quad \textit{Level: Medium}} \\
    \multicolumn{6}{@{}l}{\quad \quad \textbf{Video Models}} \\
    \quad \quad \quad Veo-3 & \hlblue{66.00\%} & \hlblue{99.20\%} & 13.66\% & \hlblue{72.00\%} & \hlblue{\textbf{0.40\%}} \\
    \quad \quad \quad Sora-2 & 100.00\% & 100.00\% & 9.43\% & 30.00\% & 0.00\% \\
    \quad \quad \quad Wan-2.2 & 85.33\% & 100.00\% & \hlblue{16.13\%} & 2.00\% & 0.00\% \\
    \multicolumn{6}{@{}l}{\quad \quad \textbf{Image Models}} \\
    \quad \quad \quad Nano-banana & 21.87\% & 35.54\% & 50.43\% & N/A & 0.00\% \\
    \quad \quad \quad Nano-banana Pro & \hlyellow{14.91\%} & 28.40\% & 46.47\% & N/A & 0.00\% \\
    \quad \quad \quad GPT-4o-image & 69.54\% & 59.69\% & 19.18\% & N/A & 0.00\% \\
    \quad \quad \quad GPT-image-1.5 & 100.00\% & 89.14\% & 14.62\% & N/A & 0.00\% \\
    \quad \quad \quad Qwen-image & 28.58\% & \hlyellow{7.41\%} & \hlyellow{73.74\%} & N/A & 0.00\% \\

    \multicolumn{6}{@{}l}{\quad \textit{Level: Hard}} \\
    \multicolumn{6}{@{}l}{\quad \quad \textbf{Video Models}} \\
    \quad \quad \quad Veo-3 & \hlblue{68.80\%} & 99.60\% & 13.45\% & \hlblue{70.40\%} & 0.00\% \\
    \quad \quad \quad Sora-2 & 92.86\% & \hlblue{92.86\%} & 8.65\% & 35.71\% & \hlblue{\textbf{7.14\%}} \\
    \quad \quad \quad Wan-2.2 & 93.00\% & 100.00\% & \hlblue{16.86\%} & 3.00\% & 0.00\% \\
    \multicolumn{6}{@{}l}{\quad \quad \textbf{Image Models}} \\
    \quad \quad \quad Nano-banana & 27.25\% & 38.16\% & 41.11\% & N/A & 0.00\% \\
    \quad \quad \quad Nano-banana Pro & \hlyellow{14.75\%} & 41.36\% & 40.11\% & N/A & 0.00\% \\
    \quad \quad \quad GPT-4o-image & 69.80\% & 57.66\% & 15.76\% & N/A & 0.00\% \\
    \quad \quad \quad GPT-image-1.5 & 100.00\% & 87.26\% & 12.81\% & N/A & 0.00\% \\
    \quad \quad \quad Qwen-image & 27.26\% & \hlyellow{10.44\%} & \hlyellow{73.59\%} & N/A & 0.00\% \\

    \bottomrule
    \end{tabular}
}
\end{adjustbox}
\end{table*}

\subsection{Evaluation Results}


\textbf{Key Finding: Image Models Outperform Video Models in Symbolic Reasoning.} \textbf{Image generation models significantly outperform video models} on Sudoku tasks, as video models suffer from severe temporal instability and a lack of global symbolic reasoning. While video models exhibit high \textbf{action reflection} (making step-by-step edits), they fail to maintain logical consistency, leading to frequent constraint violations and clue changes. Under the strict primary success criterion, Veo-3 remains at or near 0\% overall success across conditions, highlighting that current video models strictly fail at complex symbolic reasoning and that VLM-based auto-evaluation can overestimate model performance.

\subsubsection{VLM-Based Evaluation}

\Cref{tab:sudoku_results} reveals that \textbf{all generative models struggle with Sudoku under strict evaluation criteria}, though image models exhibit substantially better fine-grained metrics. The final Sudoku row in \Cref{tab:main_results} is aggregated from this VLM-based overall metric; the OCR-based analysis below is a stricter deterministic diagnostic rather than the source of the main-table score. \textbf{Image models (Nano-banana, Nano-banana Pro, GPT-4o-image, GPT-image-1.5, Qwen-image) consistently achieve higher completion accuracy and lower constraint violation rates}, reflecting stronger symbolic consistency when directly producing completed Sudoku grids. In contrast, \textbf{video models (Veo-3, Sora-2, Wan-2.2)} almost universally modify original clues and violate constraints, resulting in near-zero overall success.

\textbf{4$\times$4 Performance.} Even in the simplest setting, most models fail to produce fully correct solutions. Among video models, Veo-3 achieves the best overall score at only 1.20\% on Easy puzzles, while Sora-2 and Wan-2.2 completely fail (0\%). Video models exhibit severe clue modification (72\%–100\%) and near-universal constraint violations (96\%–100\%). GPT-image-1.5 emerges as the only model achieving meaningful overall success: 22.67\% on Easy, 10.67\% on Medium, and 7.33\% on Hard. Other image models achieve 0\% overall despite superior fine-grained metrics—Nano-banana Pro reaches 97\% completion accuracy with only 1.33\% constraint violations on Easy, but its 35\% clue-change rate prevents any valid solutions. Although Veo-3 maintains high action reflection (92\%–96\%), its completion accuracy remains low (30\%–38\%), and clue-change rates (70\%–76\%) indicate persistent temporal drift.\looseness=-1

\textbf{9$\times$9 Performance.} On full-scale Sudoku, where global structural consistency is essential, all models struggle severely. Paradoxically, video models achieve higher overall scores than image models: Sora-2 reaches 4.26\% on Easy and 7.14\% on Hard, while all image models achieve 0\% overall. However, this reflects VLM evaluation inconsistencies rather than genuine video model superiority—video models exhibit 66\%–100\% clue-change rates and near-universal constraint violations (92\%–100\%). Image models maintain better underlying metrics: Qwen-image achieves 72\%–74\% completion accuracy with only 5\%–10\% constraint violation rates, yet fails overall due to 27\%–29\% clue modifications. Nano-banana and Nano-banana Pro preserve clues better (15\%–27\% change rates) but suffer from higher constraint violations (28\%–41\%) at the 9$\times$9 scale.

The \textbf{stark contrast between fine-grained metrics and overall success} reveals the difficulty of Sudoku under strict evaluation: even models with high completion accuracy fail when any clue is modified or any constraint violated. Image models demonstrate stronger symbolic grounding through lower clue-change and constraint-violation rates, yet the combinatorial demands of perfect solutions expose limitations across all current generative architectures. Video models, despite producing visually coherent step-by-step animations with high action reflection (70\%–96\%), lack the constraint-aware reasoning needed to avoid cumulative errors—their sequential generation process introduces temporal instability that compounds across frames, making valid solutions nearly impossible.

\subsubsection{OCR-Based Evaluation}

\begin{table*}[h!]
\centering
\small
\caption{Quantitative results for the \textbf{Sudoku} task using \textbf{OCR-based} evaluation method. We evaluate video generative models (Veo-3, Sora-2, and Wan-2.2) and image generative models (Nano-banana, Nano-banana Pro, GPT-4o-image, GPT-image-1.5, and Qwen-image) on two grid sizes (4$\times$4, 9$\times$9) and three difficulty levels (Easy, Medium, and Hard). Because image outputs do not support frame-by-frame reasoning, action-reflection–based metrics are omitted for image generative models (marked with ``N/A''). The best results among video models are highlighted with \colorbox{paleblue}{blue}, and the best results among image models are highlighted with \colorbox{paleyellow}{yellow}.}
\label{tab:sudoku_results_vis}

\begin{adjustbox}{max width=0.83\textwidth}
{
    \begin{tabular}{@{}lccccc@{}}
    \toprule
    & \multicolumn{4}{c}{\textbf{Fine-grained Metrics}} & \multicolumn{1}{c}{\textbf{Primary Metric}} \\
    \cmidrule(lr){2-5} \cmidrule(lr){6-6}

    \textbf{Model} & \textbf{Clues Changed} $\downarrow$ & \textbf{Constraints Violation} $\downarrow$ & \textbf{Completion Accuracy} $\uparrow$ & \textbf{Action Reflection} $\uparrow$ & \textbf{Overall} $\uparrow$ \\
    \midrule

    \multicolumn{6}{@{}l}{\textbf{Grid Size: 4$\times$4}} \\

    \multicolumn{6}{@{}l}{\quad \textit{Level: Easy}} \\
    \multicolumn{6}{@{}l}{\quad \quad \textbf{Video Models}} \\
    \quad \quad \quad Veo-3 & \hlblue{80.40\%} & 75.17\% & \hlblue{28.49\%} & 23.20\% & 0.00\% \\
    \quad \quad \quad Sora-2 & 100.00\% & \hlblue{64.80\%} & 20.97\% & \hlblue{22.45\%} & 0.00\% \\
    \quad \quad \quad Wan-2.2 & 100.00\% & 91.67\% & 5.68\% & 18.00\% & 0.00\% \\
    \multicolumn{6}{@{}l}{\quad \quad \textbf{Image Models}} \\
    \quad \quad \quad Nano-banana & 0.00\% & \hlyellow{26.93\%} & 73.85\% &N/A& \hlyellow{\textbf{32.80\%}} \\
    \quad \quad \quad Nano-banana Pro & 0.00\% & 42.50\% & \hlyellow{87.35\%} &N/A& 14.00\% \\
    \quad \quad \quad GPT-4o-image & 0.00\% & 50.15\% & 57.63\% &N/A& 7.00\% \\
    \quad \quad \quad GPT-image-1.5 & 63.33\% & 32.06\% & 61.90\% &N/A& 22.67\% \\
    \quad \quad \quad Qwen-image & 97.33\% & 95.72\% & 3.84\% &N/A& 0.00\% \\

    \multicolumn{6}{@{}l}{\quad \textit{Level: Medium}} \\
    \multicolumn{6}{@{}l}{\quad \quad \textbf{Video Models}} \\
    \quad \quad \quad Veo-3 & \hlblue{86.00\%} & 90.50\% & \hlblue{25.47\%} & \hlblue{35.60\%} & 0.00\% \\
    \quad \quad \quad Sora-2 & 100.00\% & \hlblue{59.52\%} & 25.20\% & 33.33\% & 0.00\% \\
    \quad \quad \quad Wan-2.2 & 100.00\% & 93.17\% & 7.49\% & 10.00\% & 0.00\% \\
    \multicolumn{6}{@{}l}{\quad \quad \textbf{Image Models}} \\
    \quad \quad \quad Nano-banana & 0.00\% & \hlyellow{35.97\%} & 53.96\% &N/A& \hlyellow{\textbf{13.60\%}} \\
    \quad \quad \quad Nano-banana Pro & 0.00\% & 43.33\% & \hlyellow{85.62\%} &N/A& 10.00\% \\
    \quad \quad \quad GPT-4o-image & 0.00\% & 56.06\% & 41.47\% &N/A& 0.75\% \\
    \quad \quad \quad GPT-image-1.5 & 60.67\% & 37.39\% & 50.99\% &N/A& 10.67\% \\
    \quad \quad \quad Qwen-image & 92.67\% & 99.22\% & 2.53\% &N/A& 0.00\% \\

    \multicolumn{6}{@{}l}{\quad \textit{Level: Hard}} \\
    \multicolumn{6}{@{}l}{\quad \quad \textbf{Video Models}} \\
    \quad \quad \quad Veo-3 & \hlblue{90.80\%} & 91.53\% & \hlblue{26.02\%} & 31.20\% & 0.00\% \\
    \quad \quad \quad Sora-2 & 100.00\% & \hlblue{61.41\%} & 21.96\% & \hlblue{32.61\%} & 0.00\% \\
    \quad \quad \quad Wan-2.2 & 100.00\% & 94.83\% & 11.91\% & 4.00\% & 0.00\% \\
    \multicolumn{6}{@{}l}{\quad \quad \textbf{Image Models}} \\
    \quad \quad \quad Nano-banana & 0.00\% & 49.87\% & 42.92\% &N/A & 5.20\% \\
    \quad \quad \quad Nano-banana Pro & 0.00\% & \hlyellow{36.00\%} & \hlyellow{86.86\%} &N/A& \hlyellow{\textbf{18.00\%}} \\
    \quad \quad \quad GPT-4o-image & 0.00\% & 58.29\% & 36.80\% &N/A& 0.25\% \\
    \quad \quad \quad GPT-image-1.5 & 81.33\% & 39.11\% & 39.73\% &N/A& 7.33\% \\
    \quad \quad \quad Qwen-image & 91.33\% & 99.28\% & 2.92\% &N/A& 0.00\% \\

    \midrule
    \multicolumn{6}{@{}l}{\textbf{Grid Size: 9$\times$9}} \\

    \multicolumn{6}{@{}l}{\quad \textit{Level: Easy}} \\
    \multicolumn{6}{@{}l}{\quad \quad \textbf{Video Models}} \\
    \quad \quad \quad Veo-3 & \hlblue{88.00\%} & 99.91\% & 5.75\% & \hlblue{92.40\%} & 0.00\% \\
    \quad \quad \quad Sora-2 & 100.00\% & \hlblue{99.84\%} & \hlblue{6.63\%} & 55.32\% & 0.00\% \\
    \quad \quad \quad Wan-2.2 & 100.00\% & 100.00\% & 3.72\% & 96.00\% & 0.00\% \\
    \multicolumn{6}{@{}l}{\quad \quad \textbf{Image Models}} \\
    \quad \quad \quad Nano-banana & 0.00\% & 96.07\% & 15.70\% &N/A& 0.00\% \\
    \quad \quad \quad Nano-banana Pro & 0.00\% & \hlyellow{55.22\%} & \hlyellow{45.68\%} &N/A& 0.00\% \\
    \quad \quad \quad GPT-4o-image & 0.00\% & 87.25\% & 11.13\% &N/A& 0.00\% \\
    \quad \quad \quad GPT-image-1.5 & 100.00\% & 88.79\% & 13.18\% &N/A& 0.00\% \\
    \quad \quad \quad Qwen-image & 97.33\% & 100.00\% & 0.11\% &N/A& 0.00\% \\

    \multicolumn{6}{@{}l}{\quad \textit{Level: Medium}} \\
    \multicolumn{6}{@{}l}{\quad \quad \textbf{Video Models}} \\
    \quad \quad \quad Veo-3 & \hlblue{84.40\%} & 99.94\% & 6.70\% & \hlblue{94.00\%} & 0.00\% \\
    \quad \quad \quad Sora-2 & 100.00\% & 100.00\% & \hlblue{6.92\%} & 44.00\% & 0.00\% \\
    \quad \quad \quad Wan-2.2 & 100.00\% & 100.00\% & 3.03\% & 68.00\% & 0.00\% \\
    \multicolumn{6}{@{}l}{\quad \quad \textbf{Image Models}} \\
    \quad \quad \quad Nano-banana & 0.00\% & 96.37\% & 13.70\% &N/A& 0.00\% \\
    \quad \quad \quad Nano-banana Pro & 0.00\% & \hlyellow{53.58\%} & \hlyellow{35.92\%} &N/A& 0.00\% \\
    \quad \quad \quad GPT-4o-image & 0.00\% & 85.93\% & 10.47\% &N/A& 0.00\% \\
    \quad \quad \quad GPT-image-1.5 & 100.00\% & 89.14\% & 11.96\% &N/A& 0.00\% \\
    \quad \quad \quad Qwen-image & 92.67\% & 100.00\% & 0.11\% &N/A& 0.00\% \\

    \multicolumn{6}{@{}l}{\quad \textit{Level: Hard}} \\
    \multicolumn{6}{@{}l}{\quad \quad \textbf{Video Models}} \\
    \quad \quad \quad Veo-3 & \hlblue{81.60\%} & 99.99\% & \hlblue{6.93\%} & \hlblue{86.40\%} & 0.00\% \\
    \quad \quad \quad Sora-2 & 100.00\% & 100.00\% & 5.32\% & 50.00\% & 0.00\% \\
    \quad \quad \quad Wan-2.2 & 100.00\% & 100.00\% & 3.32\% & 50.00\% & 0.00\% \\
    \multicolumn{6}{@{}l}{\quad \quad \textbf{Image Models}} \\
    \quad \quad \quad Nano-banana & 0.00\% & 95.36\% & 13.12\% &N/A& 0.00\% \\
    \quad \quad \quad Nano-banana Pro & 0.00\% & \hlyellow{59.88\%} & \hlyellow{32.55\%} &N/A& 0.00\% \\
    \quad \quad \quad GPT-4o-image & 0.00\% & 86.25\% & 9.79\% &N/A& 0.00\% \\
    \quad \quad \quad GPT-image-1.5 & 100.00\% & 87.26\% & 12.81\% &N/A& 0.00\% \\
    \quad \quad \quad Qwen-image & 91.33\% & 100.00\% & 0.08\% &N/A& 0.00\% \\

    \bottomrule
    \end{tabular}
}
\end{adjustbox}
\end{table*}

As a stricter supplementary diagnostic for limitations of VLM-based evaluation, we introduce a fully deterministic \textbf{OCR-based} evaluation pipeline that relies solely on optical character recognition and classical computer vision operations, without invoking any vision–language or large language models. This evaluation directly extracts and compares digit placements against ground-truth solutions, enabling fine-grained, reproducible, and cost-free assessment of Sudoku-solving correctness.

\textbf{Image Generation.} We extract the complete Sudoku grid from each generated image using optical character recognition (OCR) with \texttt{PaddleOCR}. Each image is first preprocessed using adaptive thresholding and denoising, then segmented into equal-sized cells according to the grid resolution (4$\times$4 or 9$\times$9). OCR is applied independently to each cell, and recognized symbols are filtered to retain only valid digits (1--9). The reconstructed grid is evaluated against \textit{three} references: the original puzzle (to detect modifications to given clues), the ground-truth solution (to assess correctness), and standard Sudoku constraints (to verify validity). Based on this comparison, we compute \textit{three} core metrics: (1) \textbf{Clues Changed}, a binary indicator that flags whether any pre-filled digits from the original puzzle were altered (0 = no changes, 1 = at least one change); (2) \textbf{Constraint Violation}, measured as the fraction of row, column, or subgrid constraints that contain duplicate digits (0 = fully valid); and (3) \textbf{Completion Accuracy}, defined as the fraction of originally empty cells that are filled correctly with respect to the ground-truth solution. Since static images do not expose intermediate reasoning steps, action reflection is not applicable in this setting. We aggregate these metrics into an \textbf{overall score (perfect solution score)}, which equals 1 if and only if all correctness conditions are simultaneously satisfied: no clues are changed, no constraint violations occur, and completion accuracy is perfect (1.0). Any deviation from these requirements results in a score of 0. This strict binary score enables direct comparison with VLM-based judgments while preserving the determinism and interpretability of OCR-based evaluation.

\textbf{Video Generation.} We extend the OCR-based evaluation framework to the temporal domain by performing frame-by-frame analysis. For each video, we extract the maximum available frames (192 frames at 24 fps for Veo-3, 360 frames for Sora-2, and 120 frames for Wan-2.2) and apply the same OCR pipeline to each frame, yielding a sequence of intermediate Sudoku grid states that represent the solving process over time. Evaluation largely mirrors the image setting but incorporates temporal criteria. We compute: (1) \textbf{Clues Changed}, which flags whether any original clues are modified in \emph{any} frame; (2) \textbf{Constraint Violation}, evaluated on the final frame to ensure the completed grid satisfies all Sudoku rules; and (3) \textbf{Completion Accuracy}, measured on the final frame against the ground-truth solution. In addition, we introduce a video-specific metric, (4) \textbf{Action Reflection}, which captures step-by-step reasoning behavior. This metric verifies that digits are filled gradually rather than appearing all at once, by enforcing constraints on per-frame changes (average $\leq$3 cell updates per frame for 4$\times$4, $\leq$5 for 9$\times$9), prohibiting mass updates in any single frame (no more than 25\% of cells changing simultaneously), and requiring sustained activity (at least 20\% of frames exhibiting non-zero updates). The \textbf{overall score (perfect solution score)} for videos is defined identically to the image setting: it equals 1 if and only if no clues are changed, no constraint violations are present in the final frame, and completion accuracy is perfect; otherwise, the score is 0. While not required for overall correctness, the action reflection metric provides additional insight into the temporal reasoning dynamics of video generation. This evaluation framework remains fully deterministic, interpretable, and independent of language-based reasoning models, while uniquely capturing the procedural nature of video-based Sudoku solving.

\textbf{Quantitative Evaluation Results.} \Cref{tab:sudoku_results_vis} presents the OCR-based evaluation results across all models, grid sizes, and difficulty levels. The most striking finding is the \textbf{complete failure of all models on 9$\times$9 grids}: every model—both video and image—achieves 0\% overall score across all difficulty levels. This reveals that current generative models fundamentally lack the capacity for large-scale symbolic constraint satisfaction, regardless of generation modality.

Among \textbf{video models}, performance is uniformly poor across all settings. All three models (Veo-3, Sora-2, Wan-2.2) achieve 0\% overall score on every configuration, including the simplest 4$\times$4 Easy setting. The primary failure mode is \textbf{clue modification}: Sora-2 and Wan-2.2 alter original clues in 100\% of cases across all conditions, while Veo-3 shows slightly lower but still severe rates (80--91\%). Even when clues are partially preserved, constraint violation rates remain extremely high (59--100\%), and completion accuracy stays below 28\% for 4$\times$4 and below 7\% for 9$\times$9. Notably, video models exhibit high action reflection scores on 9$\times$9 grids (Veo-3: 86--94\%, Wan-2.2: 50--96\%), indicating that they produce visually coherent step-by-step animations—yet these animations are logically meaningless, as the underlying digit placements violate nearly all Sudoku constraints.

Among \textbf{image models}, Nano-banana and Nano-banana Pro demonstrate substantially stronger performance on 4$\times$4 puzzles. Both models achieve 0\% clue change rates across all conditions, indicating perfect preservation of input constraints. On 4$\times$4 Easy, Nano-banana achieves the highest overall score of 32.80\%, with Nano-banana Pro following at 14.00\%. Interestingly, Nano-banana Pro exhibits higher completion accuracy (85--87\%) but lower overall success, suggesting it fills more cells but with subtle constraint violations that disqualify solutions under strict evaluation. GPT-4o-image maintains clue integrity but suffers from higher constraint violation rates (50--58\%), limiting overall success to 0.25--7\%. GPT-image-1.5 presents a paradoxical case: despite high clue modification rates (60--81\% on 4$\times$4), it achieves meaningful overall success (22.67\% Easy, 10.67\% Medium, 7.33\% Hard)—the second-best performance among image models. However, on 9$\times$9 grids, GPT-image-1.5 fails completely with 100\% clue modification and 0\% overall, revealing that its 4$\times$4 success relies on strategies that do not scale. Qwen-image fails catastrophically due to near-universal clue modification (91--97\%), resulting in 0\% overall across all settings.

The \textbf{4$\times$4 to 9$\times$9 transition} exposes a fundamental scaling limitation. Even the best-performing image models (Nano-banana, Nano-banana Pro) collapse entirely on 9$\times$9 grids, with constraint violation rates jumping from 27--50\% to 53--96\%. Completion accuracy drops correspondingly from 42--87\% to 10--46\%. GPT-image-1.5 exhibits the most dramatic scaling failure: from 22.67\% overall success on 4$\times$4 Easy to 0\% on all 9$\times$9 configurations, with clue modification rates jumping from 63\% to 100\%. This precipitous decline suggests that models capable of local constraint satisfaction on small grids cannot generalize to the global coordination required for larger puzzles. The 9$\times$9 Sudoku demands simultaneous satisfaction of 27 constraints (9 rows, 9 columns, 9 boxes), each spanning 9 cells—a combinatorial challenge that exceeds current generative models' reasoning capacity.

Compared to VLM-based evaluation, the OCR-based pipeline reveals \textbf{substantially lower performance across all models}. The stricter, deterministic evaluation criteria—requiring exact digit recognition and perfect constraint satisfaction—expose failure modes that VLM-based assessment may overlook due to visual ambiguity or lenient judgment. These results reinforce our conclusion that current generative models, despite producing visually plausible outputs, lack genuine symbolic reasoning capabilities for complex constraint satisfaction tasks.


\begin{figure*}[h]
    \centering
    \includegraphics[width=\textwidth, trim=15 80 40 0, clip]{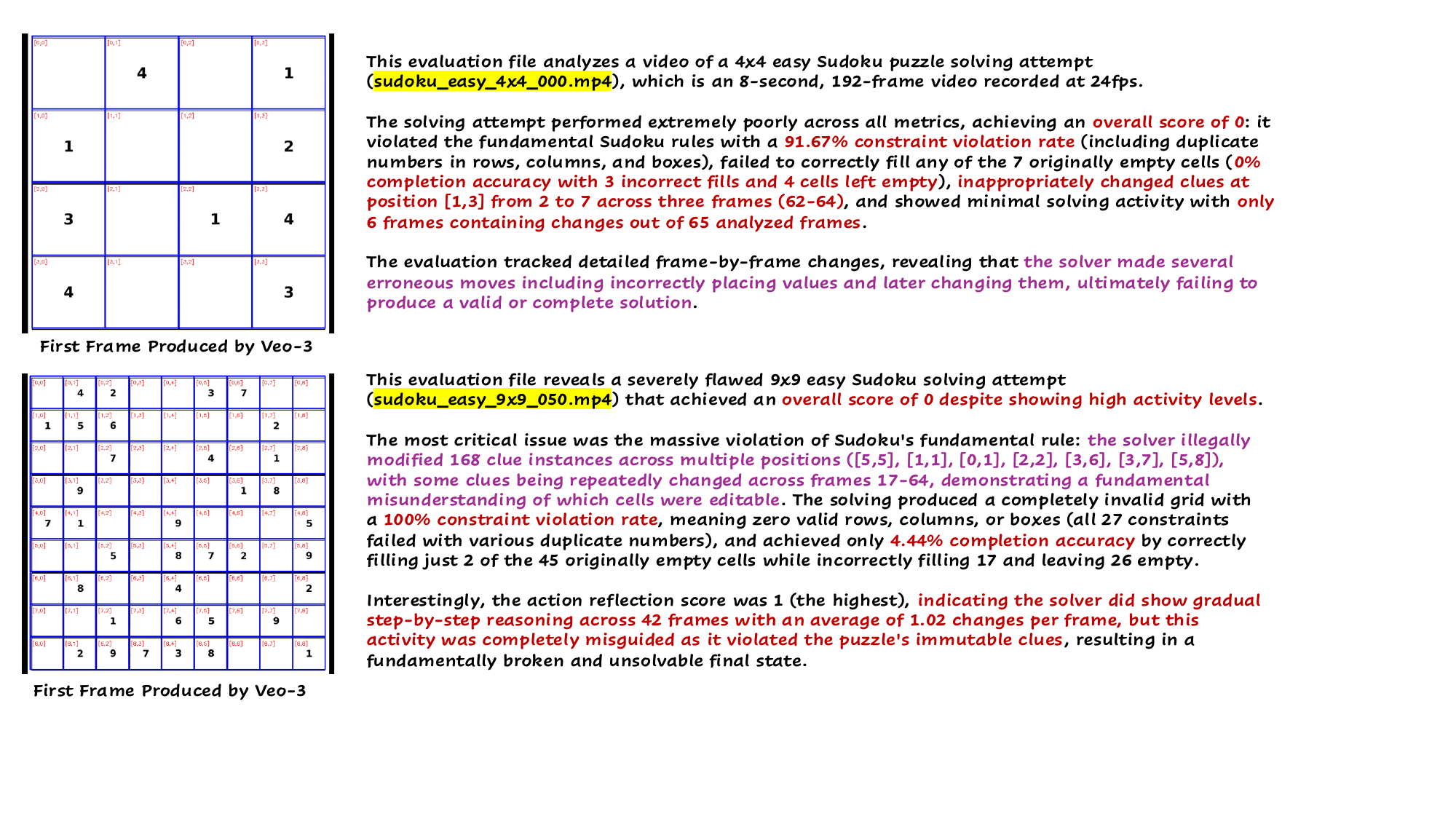}
    \caption{OCR-based evaluation of sudoku-solving video generation (Veo-3).}
    \vspace{-3mm}
    \label{fig:sudoku_ocr_evaluation}
\end{figure*}

\textbf{Failure Modes Analysis.} We analyze the OCR-based evaluation data to identify systematic failure patterns across all models. Our analysis reveals three primary failure modes: (1) Clue Modification, (2) Constraint Violation Cascades, and (3) Scaling Collapse. \textbf{Clue Modification.} Video models exhibit severe clue modification rates: Sora-2 and Wan-2.2 modify original clues in 100\% of all cases, while Veo-3 shows marginally better preservation with 85--91\% clue modification rates. Frame-by-frame analysis of Veo-3's 4$\times$4 outputs reveals an average of 91.3 clue changes per video affecting 34.5 frames—indicating that clue modifications accumulate progressively through generation rather than occurring as isolated errors. A single clue position may be modified repeatedly across multiple frames (\textit{e.g.}, position [1,2] altered in 28 consecutive frames), demonstrating temporal instability rather than deliberate correction. Among image models, Nano-banana, Nano-banana Pro, and GPT-4o-image achieve 100\% clue preservation across all conditions, demonstrating that static generation avoids the temporal drift inherent to video synthesis. GPT-image-1.5 presents an intermediate case: it modifies clues at high rates on 4$\times$4 grids (60--81\%) yet still achieves meaningful overall success, suggesting it may employ a different solving strategy that involves reinterpreting the input puzzle. On 9$\times$9 grids, GPT-image-1.5's clue modification reaches 100\%, matching the worst-performing models. Qwen-image exhibits catastrophic clue modification (91--97\% modification rate), with analysis showing 56--70\% of cells remain unfilled, suggesting fundamental generation failures. \textbf{Constraint Violation Cascades.} Even when clues are preserved, models fail to satisfy Sudoku's constraint structure. On 4$\times$4 grids, Nano-banana achieves 38.8\% perfect constraint rate on Easy puzzles, dropping to 26.4\% on Medium and 12.0\% on Hard—a +22.9 percentage point increase in constraint violations as difficulty increases. GPT-4o-image shows a similar pattern: constraint violation rates rise from 50.1\% (Easy) to 58.3\% (Hard). Video models exhibit even more severe violations: Veo-3 averages 85.7\% constraint violation on 4$\times$4, Sora-2 reaches 62.5\%, and Wan-2.2 reaches 93.2\%. Detailed violation analysis across 2,513 constraint failures in Nano-banana's 4$\times$4 outputs reveals the breakdown: column violations (40.6\%) occur most frequently, followed by box violations (33.4\%) and row violations (25.9\%)—suggesting that vertical dependencies are harder to maintain than horizontal ones. On 9$\times$9 grids, no model achieves a single instance of perfect constraint satisfaction: violation rates reach 99.9--100\% across all video models and 96--100\% for image models. Notably, Veo-3 generates out-of-range digits in 16.8\% of 4$\times$4 videos (185 out of 3,028 incorrect cells contain digits $>$4 or $<$1), revealing that constraint internalization fails at the most fundamental level. \textbf{Scaling Collapse.} The transition from 4$\times$4 to 9$\times$9 reveals a fundamental scaling limitation. Nano-banana's success rate drops from 32.8\% (4$\times$4 Easy) to 0\% (9$\times$9 Easy); GPT-4o-image falls from 7.0\% to 0\%; GPT-image-1.5 collapses from 22.67\% to 0\%; and video models, already at 0\% on 4$\times$4, remain at 0\%. Completion accuracy exhibits similar collapse across all models: Nano-banana drops from 73.9\% to 15.7\% (4.7$\times$ degradation), Nano-banana Pro from 87.3\% to 45.7\% (1.9$\times$), GPT-4o-image from 57.6\% to 11.1\% (5.2$\times$), GPT-image-1.5 from 61.9\% to 13.2\% (4.7$\times$), and Qwen-image from 3.8\% to 0.1\% (38$\times$). Video models show comparable patterns: Veo-3 drops from 26.7\% to 6.5\% completion accuracy, Sora-2 from 22.1\% to 6.6\%, and Wan-2.2 from 8.4\% to 3.4\%. This universal degradation indicates that models capable of local pattern matching on small grids cannot generalize to the global coordination required for larger puzzles. The 9$\times$9 Sudoku demands simultaneous satisfaction of 27 constraints spanning 81 cells—a combinatorial complexity that exceeds current generative models' reasoning capacity, regardless of generation modality.\looseness=-1

\section{Abstraction and Reasoning Corpus (ARC)}
\label{apx:arc}

We evaluate models on the \textbf{ARC task}, a benchmark designed to measure \textbf{Abstract Reasoning}, \textbf{Pattern Recognition}, and \textbf{Rule Induction}~\citep{chollet2019measure}. ARC probes a model’s ability to infer latent transformation rules from a small set of input–output examples and to apply those rules to unseen test cases—an ability central to \textit{fluid intelligence}. Unlike tasks with explicit instructions or predefined rule sets, ARC demands that the model autonomously identify and generalize the underlying visual and structural principles governing each puzzle. As a result, the task jointly tests \textit{2D spatial reasoning} (interpreting grid-based patterns) and \textit{logical reasoning} (deducing and executing abstract transformations).

\subsection{Hard-Level Control}
\label{sec:arc_hardlevel}

To construct a diverse and controllable evaluation suite, we build on the open-source ARC benchmark~\citep{chollet2019measure}. Our final dataset comprises \textbf{456} tasks drawn from two benchmark versions:
\vspace{-2mm}
\begin{itemize}
\item \textbf{ARC v1 (381 tasks):} The publicly released training set from the original benchmark, encompassing a wide variety of pattern-transformation and abstract reasoning problems.
\item \textbf{ARC v2 (75 tasks):} Newly added tasks that introduce novel pattern families and higher structural complexity, expanding the benchmark’s coverage.
\end{itemize}

This combined set provides broad coverage of transformation types while introducing sufficient novelty and difficulty for rigorous model evaluation. All tasks undergo manual curation to ensure clean pattern design, consistent formatting, and unambiguous transformation rules. Collectively, the benchmark spans a wide spectrum of transformation categories—including symmetry, rotation, scaling, color manipulation, and object-level reasoning—capturing the core abstractions and reasoning skills fundamental to ARC-style tasks.\footnote{Additional examples are available in the official repositories: \url{https://github.com/fchollet/ARC}, \url{https://github.com/michaelhodel/re-arc}, and the automated generation toolkit: \url{https://github.com/google/ARC-GEN}.}

\subsubsection{Two-Level Classification System}
To enable fine-grained analysis of model capabilities, we classify all cases along two dimensions:

\textbf{Level 1: Shape Consistency Classification.} We categorize each case based on whether the input and output grids maintain the same spatial structure:
\vspace{-2mm}
\begin{itemize}
    \item \textbf{Match (316 cases)}: The output grid has the same dimensions as the input grid. Transformations occur ``in-place'' through color changes, pattern filling, or local modifications. Examples include color replacement, symmetry completion, and pattern extension within fixed boundaries.
    \item \textbf{Mismatch (140 cases)}: The output grid dimensions differ from the input. These cases require spatial restructuring, such as cropping, extraction of sub-patterns, grid concatenation, or shape reconstruction. These are generally more challenging as they demand explicit understanding of spatial relationships beyond simple transformations.
\end{itemize}

\textbf{Level 2: Quantitative Difficulty Classification.} Within each shape-consistency category, we further assign each case to one of three difficulty levels—\textit{Easy}, \textit{Medium}, or \textit{Hard}—based on five quantitative grid-level features:
\vspace{-2mm}
\begin{enumerate}
    \item \textbf{Grid Size} (minimum side length): $<8$ cells $\rightarrow$ score 0; $8$-$15$ cells $\rightarrow$ score 1; $\geq 16$ cells $\rightarrow$ score 2.
    \item \textbf{Color Count} (distinct colors in input): $\leq 3$ $\rightarrow$ score 0; $4$-$6$ $\rightarrow$ score 1; $\geq 7$ $\rightarrow$ score 2.
    \item \textbf{Object Count} (distinct connected components): $\leq 4$ $\rightarrow$ score 0; $5$-$10$ $\rightarrow$ score 1; $>10$ $\rightarrow$ score 2.
    \item \textbf{Occupancy Ratio} (non-background cells / total cells): $\leq 0.25$ $\rightarrow$ score 0; $(0.25, 0.55]$ $\rightarrow$ score 1; $> 0.55$ $\rightarrow$ score 2.
    \item \textbf{$\Delta_{IO}$} (grid change ratio, for Match cases only): $\leq 0.2$ $\rightarrow$ score 0; $(0.2, 0.5]$ $\rightarrow$ score 1; $> 0.5$ $\rightarrow$ score 2. For Mismatch cases, this feature is not applicable.
\end{enumerate}

\begin{figure*}[t!]
    \centering
    \includegraphics[width=\textwidth,clip,trim=0 230 120 0]{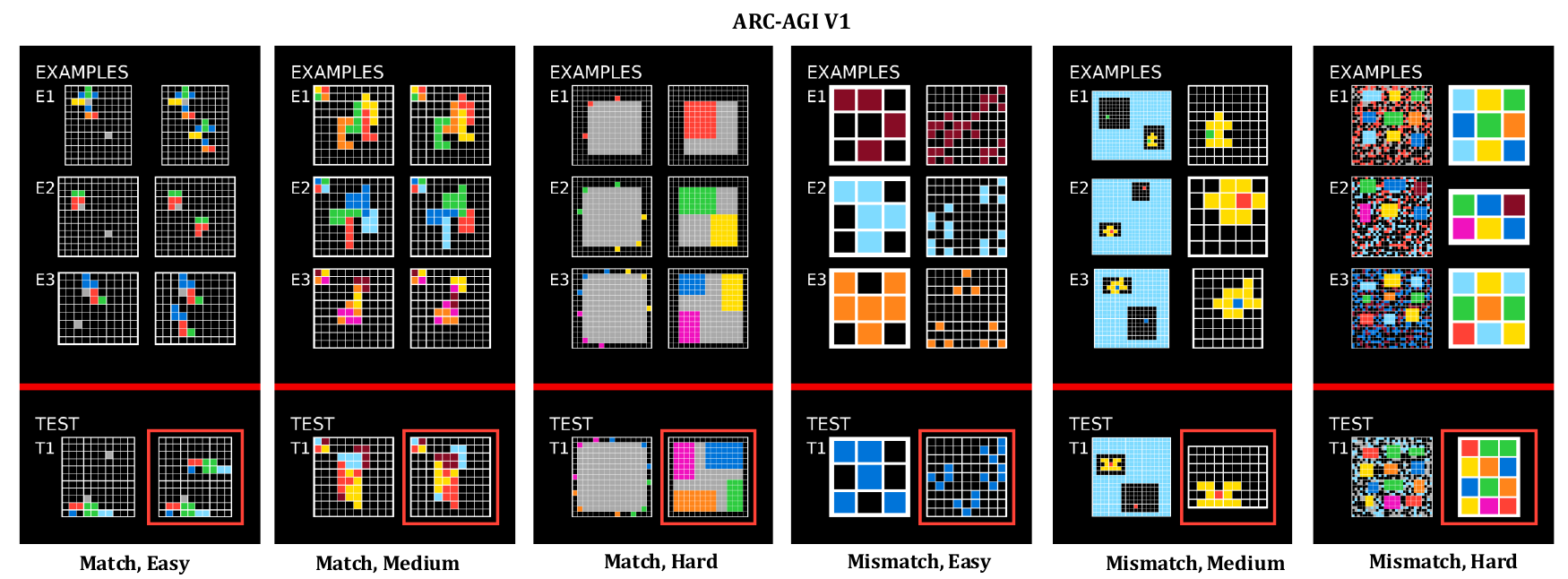}
    \vspace{-5mm}
    \caption{Examples from \textbf{ARC v1}, illustrating both Match and Mismatch tasks across three difficulty levels: Easy, Medium, and Hard.}
    \vspace{-3mm}
    \label{fig:arc_examples_v1}
\end{figure*}

\begin{figure*}[t!]
    \centering
    \includegraphics[width=\textwidth,clip,trim=40 230 70 0]{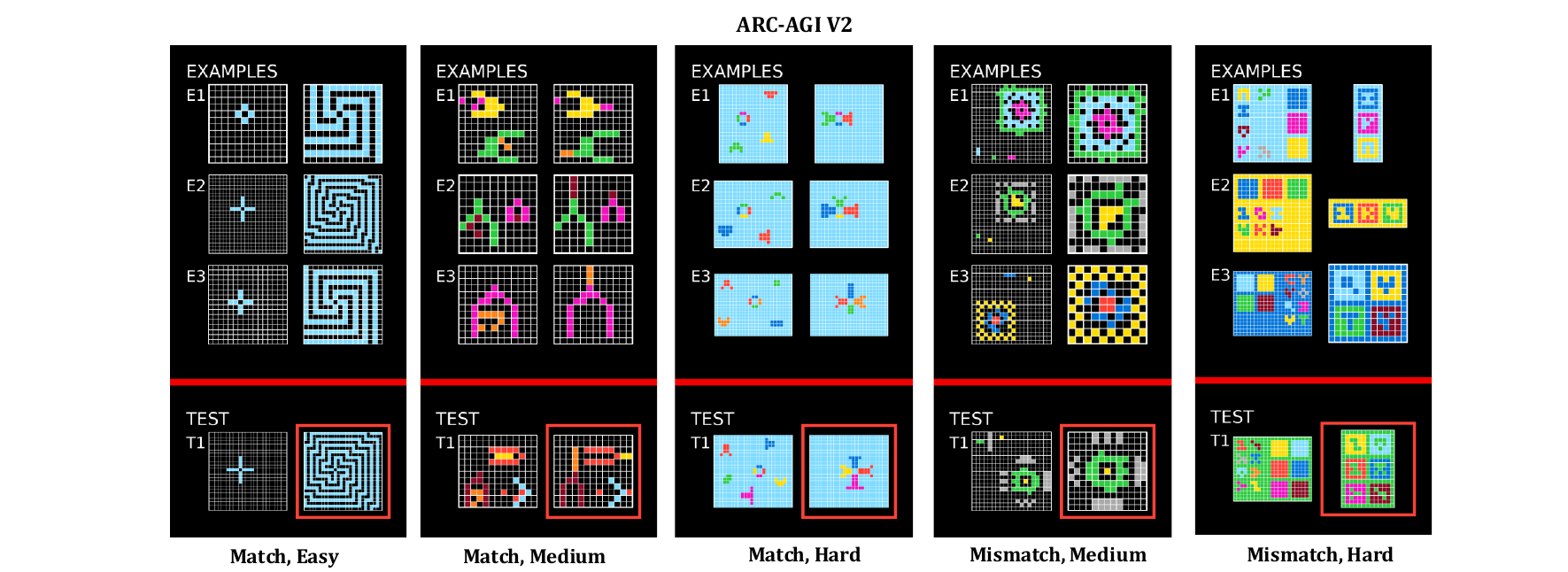}
    \vspace{-5mm}
    \caption{Examples from \textbf{ARC v2}, illustrating both Match and Mismatch tasks across three difficulty levels: Easy, Medium, and Hard.}
    \vspace{-3mm}
    \label{fig:arc_examples_v2}
\end{figure*}

A case's overall difficulty is determined by summing the applicable feature scores. For \textbf{Match} cases (5 features): \quad Easy $\leq 3$, Medium $4$–$6$, Hard $\geq 7$. For \textbf{Mismatch} cases (4 features): \quad Easy $\leq 2$, Medium $3$–$4$, Hard $\geq 5$. \Cref{tab:arc_distribution} reports the full distribution across categories. \Cref{fig:arc_examples_v1} and \Cref{fig:arc_examples_v2} present some selected examples.

\begin{table*}[h!]
\centering
\caption{Distribution of 456 ARC cases across shape consistency and difficulty levels, separated by v1 and v2. Percentages indicate the proportion within each shape consistency group for the corresponding benchmark version.\looseness=-1}
\label{tab:arc_distribution}
\small
\begin{adjustbox}{max width=0.6\textwidth}
{
\begin{tabular}{@{}llcccc@{}}
\toprule
\textbf{Version} & \textbf{Shape Consistency} & \textbf{Easy} & \textbf{Medium} & \textbf{Hard} & \textbf{Total} \\
\midrule
\multirow{2}{*}{\textbf{V1}}
& Match
& 102 (38.8\%)
& 124 (47.1\%)
& 37 (14.1\%)
& 263 \\
& Mismatch
& 34 (28.8\%)
& 57 (48.3\%)
& 27 (22.9\%)
& 118 \\
\addlinespace[2pt]
& \textbf{Total}
& \textbf{136}
& \textbf{181}
& \textbf{64}
& \textbf{381} \\
\midrule
\multirow{2}{*}{\textbf{V2}}
& Match
& 1 (1.9\%)
& 27 (50.9\%)
& 25 (47.2\%)
& 53 \\
& Mismatch
& /
& 7 (31.8\%)
& 15 (68.2\%)
& 22 \\
\addlinespace[2pt]
& \textbf{Total}
& \textbf{1}
& \textbf{34}
& \textbf{40}
& \textbf{75} \\
\midrule
\textbf{Overall}
& Match
& 103 (31.0\%)
& 151 (45.4\%)
& 62 (23.6\%)
& 316 \\
& Mismatch
& 34 (25.0\%)
& 64 (45.7\%)
& 42 (29.3\%)
& 140 \\
\addlinespace[2pt]
& \textbf{Total}
& \textbf{137 (30.0\%)}
& \textbf{215 (47.1\%)}
& \textbf{104 (22.8\%)}
& \textbf{456} \\
\bottomrule
\end{tabular}
}
\end{adjustbox}
\end{table*}

\vspace{-2mm}

\subsection{Evaluation and Metrics}
\label{apx:arc_evaluation}

We evaluate both video and image outputs using a Vision-Language Model (VLM)–based evaluator, Gemini-2.5 Pro~\citep{comanici2025gemini}. The evaluator takes \textit{four} inputs: (i) the demonstration examples, (ii) the test input, (iii) the ground-truth output, and (iv) the generated video or image. Using these inputs, the VLM determines whether the predicted transformation is correct and identifies errors in pattern recognition, structural consistency, and rule application.

From the evaluator’s structured responses, we compute \textbf{one} primary metric and \textbf{three} fine-grained diagnostic metrics:

\begin{itemize}
\item \textbf{Pattern Recognition:} 1 if the evaluator confirms that the model successfully identifies the transformation pattern from the demonstrations; 0 otherwise.
\item \textbf{Grid Integrity:} 1 if the generated output preserves the correct grid dimensions and structural layout; 0 if the grid is distorted or misaligned.
\item \textbf{Color Accuracy:} 1 if all colors are applied correctly according to the transformation rule; 0 otherwise.
\item \textbf{Valid Solution (Primary Metric):} 1 only if the generated output \textit{exactly} matches the ground-truth solution; 0 otherwise.
\end{itemize}

\subsection{Case Study}

\begin{figure*}[h!]
\centering
    \begin{subfigure}{0.85\textwidth}
        \centering
        \includegraphics[width=\textwidth, trim=0 0 0 0, clip]{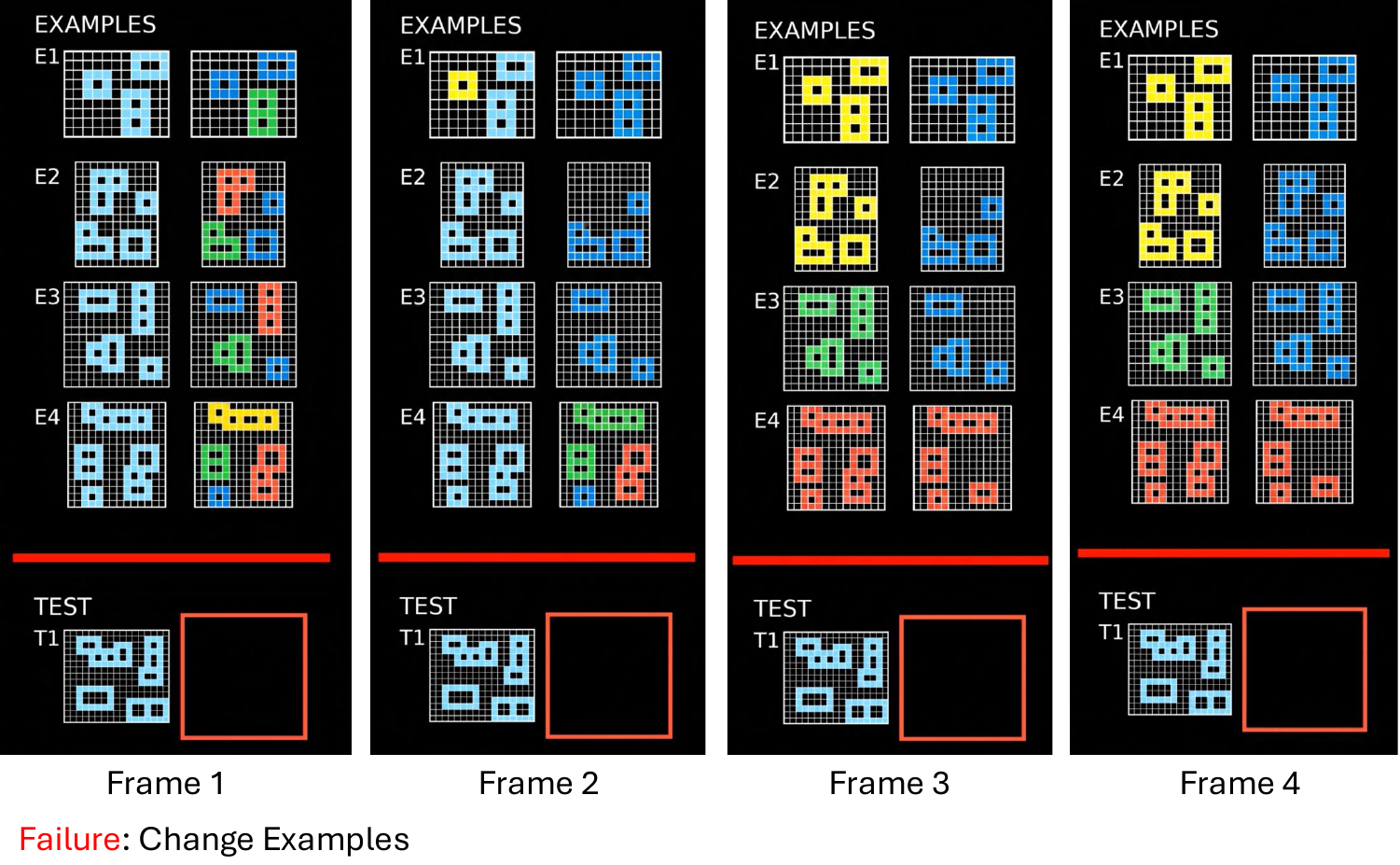}
        \caption{Inconsistent demonstration examples across video frames. The model fails to maintain the static demonstration inputs (E1–E4), with colors and patterns changing between frames. This behavior reflects a critical failure to preserve the given problem context.}
        \label{fig:arc_case_study_1}
    \end{subfigure}

    \vspace{1mm} 

    \begin{subfigure}{0.85\textwidth}
        \centering
        \includegraphics[width=\textwidth, trim=0 0 0 0, clip]{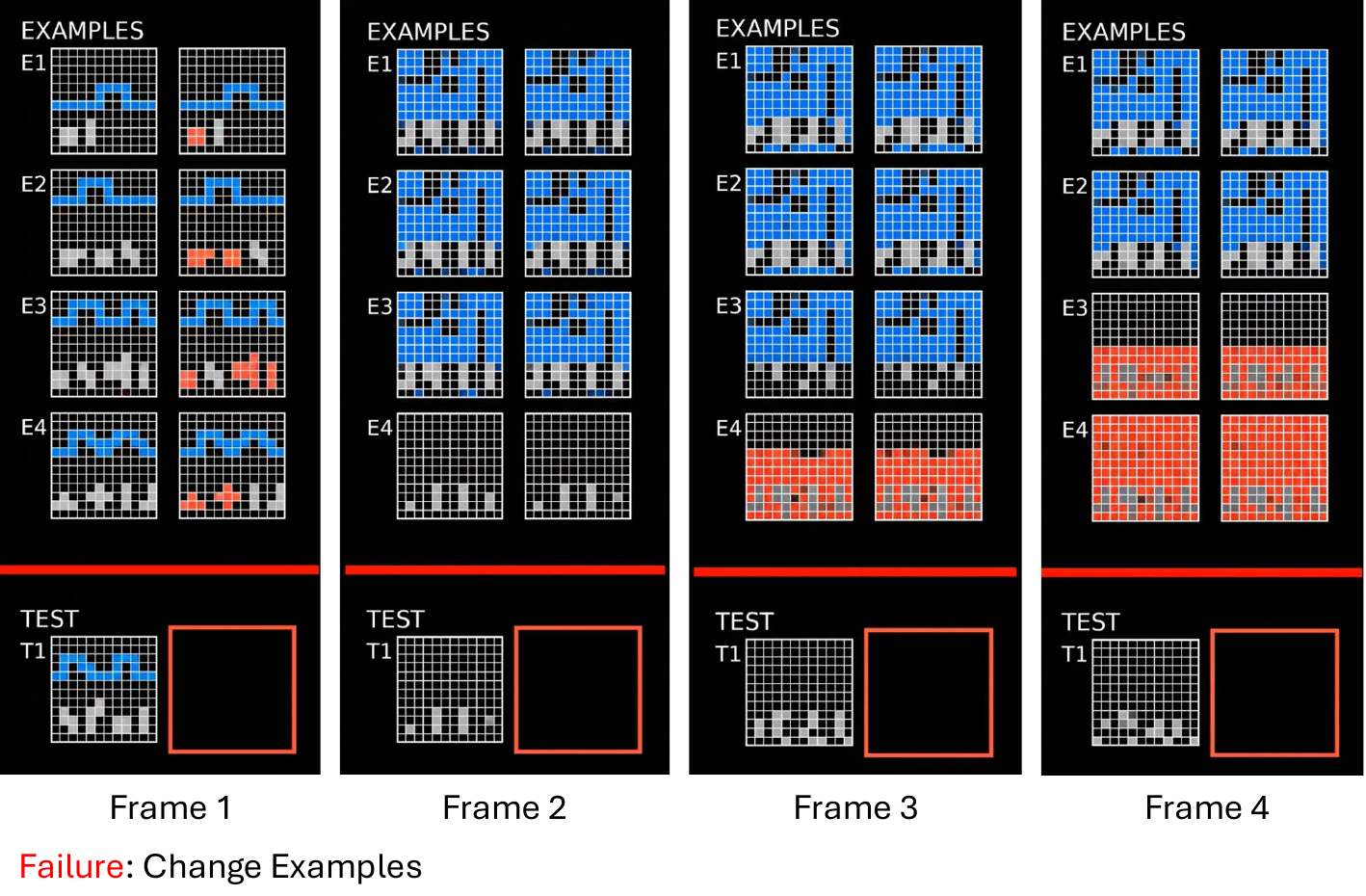}
        \caption{Progressive transformation of example demonstrations across frames. The examples (E1–E4) undergo unintended color and pattern evolution from Frame 1 to Frame 4, indicating a lack of temporal consistency in preserving the demonstration context during solution generation.}
        \label{fig:arc_case_study_2}
    \end{subfigure}

    \caption{Failure cases generated by \textbf{Veo-3} on \textbf{ARC} showing violation of static demonstration preservation. In both examples, the model fails to keep the input demonstrations unchanged through time, illustrating a key breakdown in temporal consistency for reasoning-based video generation.}
    \label{fig:arc_case_combined}
\end{figure*}

The case studies presented in \Cref{fig:arc_case_study_1} and \Cref{fig:arc_case_study_2} highlight a significant failure mode in video generation models applied to abstract reasoning tasks: \textbf{the inability to maintain temporal consistency for static information}. In both instances, the model fails to distinguish between the invariant problem context (the ``EXAMPLES'' E1-E4) and the dynamic solution generation (the ``TEST'' T1). As the video progresses from Frame 1 to Frame 4, the demonstration examples—which should remain strictly fixed to define the logic of the task—suffer from severe hallucinations, including unintended color shifts, pattern deformations, and progressive structural degradation (such as the complete erasure of grid contents in \Cref{fig:arc_case_study_2}). This ``drift'' suggests that the model treats the entire visual field as a mutable video sequence rather than respecting the logical constraints of the prompt, ultimately undermining the reasoning process by destabilizing the very ground truth required to solve the puzzle.

\subsection{Evaluation Results}


\textbf{Key Finding: The Abstract Reasoning Divide and Temporal Instability.} Our evaluation reveals two critical insights about abstract visual reasoning. First, a \textbf{fundamental modality gap}: in the final task-level score, \textbf{Nano-banana Pro} leads at 28.07\%, followed by \textbf{GPT-image-1.5} at 12.72\%, while the strongest video model, \textbf{Sora-2}, reaches 11.67\%. Second, a \textbf{temporal consistency failure}: video models struggle to maintain static demonstration examples, leading to ``context drift'' where the problem definition itself becomes corrupted during generation. These failures reveal that current video models cannot reliably execute abstract transformations despite occasionally producing visually plausible outputs.

\begin{table*}[t!]
\centering
\small
\caption{Final task-level results for the \textbf{ARC} task. Scores match the ARC row in \Cref{tab:main_results}; subsequent v1/v2 tables provide diagnostic split analyses rather than the final aggregation.}
\label{tab:arc_final_scores}
\begin{adjustbox}{max width=0.72\textwidth}
\begin{tabular}{@{}llc@{}}
\toprule
\textbf{Model} & \textbf{Type} & \textbf{Final score} $\uparrow$ \\
\midrule
Veo-3 & Video & 4.80\% \\
Sora-2 & Video & 11.67\% \\
Wan-2.2 & Video & 0.15\% \\
\midrule
Nano-banana & Image & 8.15\% \\
Nano-banana Pro & Image & \textbf{28.07\%} \\
GPT-4o-image & Image & 0.00\% \\
GPT-image-1.5 & Image & 12.72\% \\
Qwen-image & Image & 2.15\% \\
\bottomrule
\end{tabular}
\end{adjustbox}
\end{table*}

\begin{table*}[t!]
\centering
\small
\caption{Diagnostic quantitative results for the \textbf{ARC v1} split (381 cases). This table reports the original fine-grained v1 breakdown and is not the final task-level aggregation in \Cref{tab:arc_final_scores}.}
\label{tab:arc_v1_overall}
\begin{adjustbox}{max width=0.75\textwidth}
{
\begin{tabular}{@{}lcccc@{}}
\toprule
& \multicolumn{3}{c}{\textbf{Fine-grained Metrics}} & \textbf{Primary Metric} \\
\cmidrule(lr){2-4}
\textbf{Model} & \textbf{Pattern Recog.} $\uparrow$ & \textbf{Grid Integrity} $\uparrow$ & \textbf{Color Accuracy} $\uparrow$ & \textbf{Overall} $\uparrow$ \\
\midrule
\multicolumn{5}{@{}l}{\textbf{Video Models}} \\
\quad Veo-3 & 17.32\% & 32.98\% & 8.22\% & 5.16\% \\
\quad Sora-2 & 71.99\% & 94.58\% & 36.75\% & \textbf{20.18\%} \\
\quad Wan-2.2 & 0.61\% & 13.04\% & 0.17\% & 0.17\% \\
\multicolumn{5}{@{}l}{\textbf{Image Models}} \\
\quad Nano-banana & 28.42\% & 55.79\% & 12.63\% & 9.21\% \\
\quad Nano-banana Pro & 61.98\% & 84.73\% & 40.42\% & \textbf{30.54\%} \\
\quad GPT-4o-image & 1.05\% & 10.24\% & 0.52\% & 0.00\%\\
\quad Qwen-image & 1.31\% & 4.46\% & 0.52\% & 0.52\% \\
\bottomrule
\end{tabular}
}
\end{adjustbox}
\end{table*}

\begin{table*}[h!]
\centering
\small
\caption{Diagnostic quantitative results for the \textbf{ARC v2} split (75 cases). This table reports the original fine-grained v2 breakdown and is not the final task-level aggregation in \Cref{tab:arc_final_scores}.}
\label{tab:arc_v2_overall}
\begin{adjustbox}{max width=0.75\textwidth}
{
\begin{tabular}{@{}lcccc@{}}
\toprule
& \multicolumn{3}{c}{\textbf{Fine-grained Metrics}} & \textbf{Primary Metric} \\
\cmidrule(lr){2-4}
\textbf{Model} & \textbf{Pattern Recog.} $\uparrow$ & \textbf{Grid Integrity} $\uparrow$ & \textbf{Color Accuracy} $\uparrow$ & \textbf{Overall} $\uparrow$ \\
\midrule
\multicolumn{5}{@{}l}{\textbf{Video Models}} \\
\quad Veo-3 & 17.78\% & 31.11\% & 6.22\% & \textbf{4.00\%} \\
\quad Sora-2 & 4.00\% & 16.00\% & 1.33\% & 1.33\% \\
\quad Wan-2.2 & 0.00\% & 5.78\% & 0.00\% & 0.00\% \\
\multicolumn{5}{@{}l}{\textbf{Image Models}} \\
\quad Nano-banana & 18.67\% & 42.67\% & 8.00\% & 2.67\% \\
\quad Nano-banana Pro & 62.50\% & 83.93\% & 44.64\% & \textbf{30.36\%} \\
\quad GPT-4o-image & 1.33\% & 2.67\% & 1.33\% & 0.00\% \\
\quad Qwen-image & 1.33\% & 5.33\% & 1.33\% & 1.33\% \\
\bottomrule
\end{tabular}
}
\end{adjustbox}
\end{table*}

\subsubsection{VLM-Based Evaluation}

\textbf{Overall Performance Patterns.} The final task-level scores in \Cref{tab:arc_final_scores} establish the main hierarchy across ARC: \textbf{Nano-banana Pro} leads at 28.07\%, \textbf{GPT-image-1.5} reaches 12.72\%, and \textbf{Sora-2} is the strongest video model at 11.67\%. \Cref{tab:arc_v1_overall} and \Cref{tab:arc_v2_overall} should be read as diagnostic split analyses from the original fine-grained run. They show that Nano-banana Pro remains stable across v1 and v2, while Sora-2's strong v1 behavior does not transfer to v2, exposing brittleness under distribution shift.

\begin{table*}[h!]
\centering
\small
\caption{Diagnostic quantitative breakdown results (Match and Mismatch) for the \textbf{ARC v1} task (381 cases). We compare the performance of video generative models (Veo-3, Sora-2, and Wan-2.2) and image generative models (Nano-banana, Nano-banana Pro, GPT-4o-image, and Qwen-image).}
\label{tab:arc_v1_match_mismatch}
\begin{adjustbox}{max width=0.7\textwidth}
{
\begin{tabular}{@{}lcccc@{}}
\toprule
& \multicolumn{3}{c}{\textbf{Fine-grained Metrics}} & \textbf{Primary Metric} \\
\cmidrule(lr){2-4}
\textbf{Model / Category} & \textbf{Pattern Recog.} $\uparrow$ & \textbf{Grid Integrity} $\uparrow$ & \textbf{Color Accuracy} $\uparrow$ & \textbf{Overall} $\uparrow$ \\
\midrule
\multicolumn{5}{@{}l}{\textbf{Video Models}} \\
\multicolumn{5}{@{}l}{\quad \textit{Veo-3}} \\
\quad \quad Match & 17.74\% & 35.74\% & 9.51\% & 5.70\% \\
\quad \quad Mismatch & 16.38\% & 26.84\% & 5.37\% & 3.95\% \\
\multicolumn{5}{@{}l}{\quad \textit{Sora-2}} \\
\quad \quad Match & 67.29\% & 93.46\% & 35.05\% & \textbf{17.76\%} \\
\quad \quad Mismatch & 80.51\% & 96.61\% & 39.83\% & \textbf{24.58\%} \\
\multicolumn{5}{@{}l}{\quad \textit{Wan-2.2}} \\
\quad \quad Match & 0.51\% & 13.31\% & 0.00\% & 0.00\% \\
\quad \quad Mismatch & 0.85\% & 12.43\% & 0.56\% & 0.56\% \\
\midrule
\multicolumn{5}{@{}l}{\textbf{Image Models}} \\
\multicolumn{5}{@{}l}{\quad \textit{Nano-banana}} \\
\quad \quad Match & 24.05\% & 53.82\% & 11.07\% & 8.40\% \\
\quad \quad Mismatch & 38.14\% & 60.17\% & 16.10\% & 11.02\% \\
\multicolumn{5}{@{}l}{\quad \textit{Nano-banana Pro}} \\
\quad \quad Match & 62.24\% & 89.63\% & 42.74\% & \textbf{31.54\%} \\
\quad \quad Mismatch & 61.29\% & 72.04\% & 34.41\% & \textbf{27.96\%} \\
\multicolumn{5}{@{}l}{\quad \textit{GPT-4o-image}} \\
\quad \quad Match & 0.38\% & 9.51\% & 0.76\% & 0.00\% \\
\quad \quad Mismatch & 2.54\% & 11.86\% & 0.00\% & 0.00\% \\
\multicolumn{5}{@{}l}{\quad \textit{Qwen-image}} \\
\quad \quad Match & 0.76\% & 4.56\% & 0.76\% & 0.76\% \\
\quad \quad Mismatch & 2.54\% & 4.24\% & 0.00\% & 0.00\% \\
\bottomrule
\end{tabular}
}
\end{adjustbox}
\end{table*}

\begin{table*}[h!]
\centering
\small
\caption{Diagnostic quantitative breakdown results for the \textbf{ARC v1} task (381 cases) across different difficulty levels (Easy, Medium, and Hard). We compare the performance of video generative models (Veo-3, Sora-2, and Wan-2.2) and image generative models (Nano-banana, Nano-banana Pro, GPT-4o-image, and Qwen-image).}
\label{tab:arc_v1_difficulty}
\begin{adjustbox}{max width=0.65\textwidth}
{
\begin{tabular}{@{}lcccc@{}}
\toprule
& \multicolumn{3}{c}{\textbf{Fine-grained Metrics}} & \textbf{Primary Metric} \\
\cmidrule(lr){2-4}
\textbf{Model / Difficulty} & \textbf{Pattern Recog.} $\uparrow$ & \textbf{Grid Integrity} $\uparrow$ & \textbf{Color Accuracy} $\uparrow$ & \textbf{Overall} $\uparrow$ \\
\midrule
\multicolumn{5}{@{}l}{\textbf{Veo-3: Match}} \\
\quad Easy & 18.95\% & 40.52\% & 11.76\% & 5.88\% \\
\quad Medium & 17.74\% & 35.48\% & 8.06\% & 5.65\% \\
\quad Hard & 14.41\% & 23.42\% & 8.11\% & 5.41\% \\
\midrule
\multicolumn{5}{@{}l}{\textbf{Veo-3: Mismatch}} \\
\quad Easy & 19.61\% & 37.25\% & 6.86\% & 4.90\% \\
\quad Medium & 14.04\% & 23.39\% & 5.26\% & 3.51\% \\
\quad Hard & 17.28\% & 20.99\% & 3.70\% & 3.70\% \\
\midrule
\multicolumn{5}{@{}l}{\textbf{Sora-2: Match}} \\
\quad Easy & 73.17\% & 90.24\% & 42.68\% & 19.51\% \\
\quad Medium & 63.81\% & 95.24\% & 33.33\% & 19.05\% \\
\quad Hard & 62.96\% & 96.30\% & 18.52\% & 7.41\% \\
\midrule
\multicolumn{5}{@{}l}{\textbf{Sora-2: Mismatch}} \\
\quad Easy & 85.29\% & 91.18\% & 41.18\% & 29.41\% \\
\quad Medium & 80.70\% & 100.00\% & 36.84\% & 19.30\% \\
\quad Hard & 74.07\% & 96.30\% & 44.44\% & 29.63\% \\
\midrule
\multicolumn{5}{@{}l}{\textbf{Wan-2.2: Match}} \\
\quad Easy & 0.00\% & 16.34\% & 0.00\% & 0.00\% \\
\quad Medium & 0.54\% & 12.37\% & 0.00\% & 0.00\% \\
\quad Hard & 1.80\% & 8.11\% & 0.00\% & 0.00\% \\
\midrule
\multicolumn{5}{@{}l}{\textbf{Wan-2.2: Mismatch}} \\
\quad Easy & 0.00\% & 17.65\% & 0.00\% & 0.00\% \\
\quad Medium & 1.75\% & 9.36\% & 1.17\% & 1.17\% \\
\quad Hard & 0.00\% & 12.35\% & 0.00\% & 0.00\% \\
\midrule
\multicolumn{5}{@{}l}{\textbf{Nano-banana: Match}} \\
\quad Easy & 27.45\% & 56.86\% & 13.73\% & 8.82\% \\
\quad Medium & 20.33\% & 50.41\% & 8.13\% & 7.32\% \\
\quad Hard & 27.03\% & 56.76\% & 13.51\% & 10.81\% \\
\midrule
\multicolumn{5}{@{}l}{\textbf{Nano-banana: Mismatch}} \\
\quad Easy & 47.06\% & 85.29\% & 11.76\% & 11.76\% \\
\quad Medium & 35.09\% & 52.63\% & 19.30\% & 10.53\% \\
\quad Hard & 33.33\% & 44.44\% & 14.81\% & 11.11\% \\
\midrule
\multicolumn{5}{@{}l}{\textbf{Nano-banana Pro: Match}} \\
\quad Easy & 62.77\% & 90.43\% & 45.74\% & \textbf{31.91\%} \\
\quad Medium & 60.71\% & 89.29\% & 37.50\% & \textbf{29.46\%} \\
\quad Hard & 65.71\% & 88.57\% & 51.43\% & \textbf{37.14\%} \\
\midrule
\multicolumn{5}{@{}l}{\textbf{Nano-banana Pro: Mismatch}} \\
\quad Easy & 60.71\% & 71.43\% & 35.71\% & \textbf{25.00\%} \\
\quad Medium & 65.12\% & 74.42\% & 37.21\% & \textbf{30.23\%} \\
\quad Hard & 54.55\% & 68.18\% & 27.27\% & \textbf{27.27\%} \\
\midrule
\multicolumn{5}{@{}l}{\textbf{GPT-4o-image: Match}} \\
\quad Easy & 0.98\% & 10.78\% & 1.96\% & 0.00\% \\
\quad Medium & 0.00\% & 11.29\% & 0.00\% & 0.00\% \\
\quad Hard & 0.00\% & 0.00\% & 0.00\% & 0.00\% \\
\midrule
\multicolumn{5}{@{}l}{\textbf{GPT-4o-image: Mismatch}} \\
\quad Easy & 0.00\% & 11.76\% & 0.00\% & 0.00\% \\
\quad Medium & 1.75\% & 10.53\% & 0.00\% & 0.00\% \\
\quad Hard & 7.41\% & 14.81\% & 0.00\% & 0.00\% \\
\midrule
\multicolumn{5}{@{}l}{\textbf{Qwen-image: Match}} \\
\quad Easy & 0.98\% & 5.88\% & 0.98\% & 0.98\% \\
\quad Medium & 0.00\% & 3.23\% & 0.00\% & 0.00\% \\
\quad Hard & 2.70\% & 5.41\% & 2.70\% & 2.70\% \\
\midrule
\multicolumn{5}{@{}l}{\textbf{Qwen-image: Mismatch}} \\
\quad Easy & 2.94\% & 2.94\% & 0.00\% & 0.00\% \\
\quad Medium & 3.51\% & 7.02\% & 0.00\% & 0.00\% \\
\quad Hard & 0.00\% & 0.00\% & 0.00\% & 0.00\% \\
\bottomrule
\end{tabular}
}
\end{adjustbox}
\end{table*}

\begin{table*}[h!]
\centering
\small
\caption{Diagnostic quantitative breakdown results (Match and Mismatch) for the \textbf{ARC v2} task (75 cases). We compare the performance of video generative models (Veo-3, Sora-2, and Wan-2.2) and image generative models (Nano-banana, Nano-banana Pro, GPT-4o-image, and Qwen-image).}
\label{tab:arc_v2_match_mismatch}
\begin{adjustbox}{max width=0.7\textwidth}
{
\begin{tabular}{@{}lcccc@{}}
\toprule
& \multicolumn{3}{c}{\textbf{Fine-grained Metrics}} & \textbf{Primary Metric} \\
\cmidrule(lr){2-4}
\textbf{Model / Category} & \textbf{Pattern Recog.} $\uparrow$ & \textbf{Grid Integrity} $\uparrow$ & \textbf{Color Accuracy} $\uparrow$ & \textbf{Overall} $\uparrow$ \\
\midrule
\multicolumn{5}{@{}l}{\textbf{Video Models}} \\
\multicolumn{5}{@{}l}{\quad \textit{Veo-3}} \\
\quad \quad Match & 17.61\% & 32.08\% & 6.92\% & 3.77\% \\
\quad \quad Mismatch & 18.18\% & 28.79\% & 4.55\% & 4.55\% \\
\multicolumn{5}{@{}l}{\quad \textit{Sora-2}} \\
\quad \quad Match & 5.66\% & 18.87\% & 1.89\% & 1.89\% \\
\quad \quad Mismatch & 0.00\% & 9.09\% & 0.00\% & 0.00\% \\
\multicolumn{5}{@{}l}{\quad \textit{Wan-2.2}} \\
\quad \quad Match & 0.00\% & 4.40\% & 0.00\% & 0.00\% \\
\quad \quad Mismatch & 0.00\% & 9.09\% & 0.00\% & 0.00\% \\
\midrule
\multicolumn{5}{@{}l}{\textbf{Image Models}} \\
\multicolumn{5}{@{}l}{\quad \textit{Nano-banana}} \\
\quad \quad Match & 18.87\% & 45.28\% & 11.32\% & 3.77\% \\
\quad \quad Mismatch & 18.18\% & 36.36\% & 0.00\% & 0.00\% \\
\multicolumn{5}{@{}l}{\quad \textit{Nano-banana Pro}} \\
\quad \quad Match & 64.29\% & 88.10\% & 50.00\% & \textbf{33.33\%} \\
\quad \quad Mismatch & 57.14\% & 71.43\% & 28.57\% & \textbf{21.43\%} \\
\multicolumn{5}{@{}l}{\quad \textit{GPT-4o-image}} \\
\quad \quad Match & 0.00\% & 1.89\% & 0.00\% & 0.00\% \\
\quad \quad Mismatch & 4.55\% & 4.55\% & 4.55\% & 0.00\% \\
\multicolumn{5}{@{}l}{\quad \textit{Qwen-image}} \\
\quad \quad Match & 1.89\% & 5.66\% & 1.89\% & 1.89\% \\
\quad \quad Mismatch & 0.00\% & 4.55\% & 0.00\% & 0.00\% \\
\bottomrule
\end{tabular}
}
\end{adjustbox}
\end{table*}

\begin{table*}[h!]
\centering
\small
\caption{Diagnostic quantitative breakdown results for the \textbf{ARC v2} task (75 cases) across different difficulty levels (Easy, Medium, and Hard). We compare the performance of video generative models (Veo-3, Sora-2, and Wan-2.2) and image generative models (Nano-banana, Nano-banana Pro, GPT-4o-image, and Qwen-image).
\textbf{*} The Easy level of ARC v2 contains only one evaluation case; therefore, a model solving this case correctly achieves a 100\% score.
}
\label{tab:arc_v2_difficulty}
\begin{adjustbox}{max width=0.65\textwidth}
{
\begin{tabular}{@{}lcccc@{}}
\toprule
& \multicolumn{3}{c}{\textbf{Fine-grained Metrics}} & \textbf{Primary Metric} \\
\cmidrule(lr){2-4}
\textbf{Model / Difficulty} & \textbf{Pattern Recog.} $\uparrow$ & \textbf{Grid Integrity} $\uparrow$ & \textbf{Color Accuracy} $\uparrow$ & \textbf{Overall} $\uparrow$ \\
\midrule
\multicolumn{5}{@{}l}{\textbf{Veo-3: Match}} \\
\quad Easy & 0.00\% & 33.33\% & 0.00\% & 0.00\% \\
\quad Medium & 20.99\% & 34.57\% & 8.64\% & 6.17\% \\
\quad Hard & 14.67\% & 29.33\% & 5.33\% & 1.33\% \\
\midrule
\multicolumn{5}{@{}l}{\textbf{Veo-3: Mismatch}} \\
\quad Medium & 23.81\% & 42.86\% & 4.76\% & 4.76\% \\
\quad Hard & 15.56\% & 22.22\% & 4.44\% & 4.44\% \\
\midrule
\multicolumn{5}{@{}l}{\textbf{Sora-2: Match}} \\
\quad Easy & 0.00\% & 0.00\% & 0.00\% & 0.00\% \\
\quad Medium & 7.41\% & 22.22\% & 3.70\% & 3.70\% \\
\quad Hard & 4.00\% & 16.00\% & 0.00\% & 0.00\% \\
\midrule
\multicolumn{5}{@{}l}{\textbf{Sora-2: Mismatch}} \\
\quad Medium & 0.00\% & 28.57\% & 0.00\% & 0.00\% \\
\quad Hard & 0.00\% & 0.00\% & 0.00\% & 0.00\% \\
\midrule
\multicolumn{5}{@{}l}{\textbf{Wan-2.2: Match}} \\
\quad Easy & 0.00\% & 33.33\% & 0.00\% & 0.00\% \\
\quad Medium & 0.00\% & 4.94\% & 0.00\% & 0.00\% \\
\quad Hard & 0.00\% & 2.67\% & 0.00\% & 0.00\% \\
\midrule
\multicolumn{5}{@{}l}{\textbf{Wan-2.2: Mismatch}} \\
\quad Medium & 0.00\% & 9.52\% & 0.00\% & 0.00\% \\
\quad Hard & 0.00\% & 8.89\% & 0.00\% & 0.00\% \\
\midrule
\multicolumn{5}{@{}l}{\textbf{Nano-banana: Match}} \\
\quad Easy & 100.00\% & 100.00\% & 100.00\% & 100.00\%\textbf{*} \\
\quad Medium & 18.52\% & 55.56\% & 7.41\% & 0.00\% \\
\quad Hard & 16.00\% & 32.00\% & 12.00\% & 4.00\% \\
\midrule
\multicolumn{5}{@{}l}{\textbf{Nano-banana: Mismatch}} \\
\quad Medium & 42.86\% & 57.14\% & 0.00\% & 0.00\% \\
\quad Hard & 6.67\% & 26.67\% & 0.00\% & 0.00\% \\
\midrule
\multicolumn{5}{@{}l}{\textbf{Nano-banana Pro: Match}} \\
\quad Easy & 100.00\% & 100.00\% & 100.00\% & 100.00\%\textbf{*} \\
\quad Medium & 66.67\% & 95.24\% & 61.90\% & 38.10\% \\
\quad Hard & 60.00\% & 80.00\% & 35.00\% & 25.00\% \\
\midrule
\multicolumn{5}{@{}l}{\textbf{Nano-banana Pro: Mismatch}} \\
\quad Medium & 80.00\% & 100.00\% & 40.00\% & 20.00\% \\
\quad Hard & 44.44\% & 55.56\% & 22.22\% & 22.22\% \\
\midrule
\multicolumn{5}{@{}l}{\textbf{GPT-4o-image: Match}} \\
\quad Easy & 0.00\% & 0.00\% & 0.00\% & 0.00\% \\
\quad Medium & 0.00\% & 3.70\% & 0.00\% & 0.00\% \\
\quad Hard & 0.00\% & 0.00\% & 0.00\% & 0.00\% \\
\midrule
\multicolumn{5}{@{}l}{\textbf{GPT-4o-image: Mismatch}} \\
\quad Medium & 0.00\% & 0.00\% & 0.00\% & 0.00\% \\
\quad Hard & 6.67\% & 6.67\% & 6.67\% & 0.00\% \\
\midrule
\multicolumn{5}{@{}l}{\textbf{Qwen-image: Match}} \\
\quad Easy & 0.00\% & 0.00\% & 0.00\% & 0.00\% \\
\quad Medium & 3.70\% & 7.41\% & 3.70\% & 3.70\% \\
\quad Hard & 0.00\% & 4.00\% & 0.00\% & 0.00\% \\
\midrule
\multicolumn{5}{@{}l}{\textbf{Qwen-image: Mismatch}} \\
\quad Medium & 0.00\% & 0.00\% & 0.00\% & 0.00\% \\
\quad Hard & 0.00\% & 6.67\% & 0.00\% & 0.00\% \\
\bottomrule
\end{tabular}
}
\end{adjustbox}
\end{table*}

\textbf{The Modality Gap.} The results reveal a fundamental divide between image and video generation paradigms (\Cref{tab:arc_final_scores,tab:arc_v1_overall,tab:arc_v2_overall}). In final task-level scoring, the best image model outperforms the best video model by more than $2\times$ (28.07\% vs.\ 11.67\%). The diagnostic v1 split also reveals an important anomaly: Sora-2 achieves very high Pattern Recognition and Grid Integrity, yet still trails Nano-banana Pro on overall correctness. This suggests that video models can perceive patterns and maintain structure but fail at the final integration step: translating partial understanding into exact grid solutions. The gap amplifies on v2, where Nano-banana Pro remains stable while Sora-2 collapses from 20.18\% to 1.33\%, revealing limited generalization beyond familiar transformation patterns.

\textbf{Dataset Difficulty Progression.} Performance stratification by difficulty level (shown in \Cref{tab:arc_v1_difficulty,tab:arc_v2_difficulty}) reveals distinct scaling behaviors across model types:
\vspace{-2mm}
\begin{itemize}
    \item \textbf{Nano-banana Pro (Robust Stability):} Maintains remarkable consistency across Easy (30.89\%), Medium (30.39\%), and Hard (30.23\%) cases. Grid Integrity shows only modest degradation (86.18\% $\to$ 86.74\% $\to$ 77.91\%), confirming that this model has internalized abstract transformation rules rather than relying on surface-level pattern matching.
    \item \textbf{Sora-2 (Difficulty-Sensitive Collapse):} Exhibits pronounced degradation from Easy (22.22\%) to Medium (16.33\%) to Hard (10.64\%). Notably, Pattern Recognition drops from 76.07\% to 40.43\%, indicating that video models lose the ability to identify transformation rules as complexity increases.
    \item \textbf{Veo-3 (Consistent Underperformance):} Remains stagnant around 5\% across all difficulty levels, with Grid Integrity showing steady decline (39.66\% $\to$ 32.40\% $\to$ 24.04\%). This suggests that Veo-3's failures are systematic rather than difficulty-dependent.
\end{itemize}

\textbf{Shape Consistency Analysis.} Comparing Match cases (same input/output dimensions) against Mismatch cases (different dimensions) across ARC v1 and v2 reveals distinct model behaviors (shown in \Cref{tab:arc_v1_match_mismatch} and \Cref{tab:arc_v2_match_mismatch}):
\vspace{-2mm}
\begin{itemize}
    \item \textbf{The Inverse Pattern (v1 vs. v2):} In v1, both Sora-2 and the base Nano-banana model perform better on Mismatch cases than Match cases (Sora-2: 24.58\% vs 17.76\%; Nano-banana: 11.02\% vs 8.40\%). This suggests these models may leverage visual restructuring effectively when dimensions shift. However, this advantage disappears in the more complex v2, where Sora-2 collapses to 0.00\% on Mismatch cases.
    \item \textbf{Nano-banana Pro Stability:} Demonstrates the most robust handling of transformation types. In v1, it maintains high stability (31.54\% Match vs 27.96\% Mismatch). In v2, while it remains the top performer, it exhibits a clearer preference for Match cases (33.33\%) over Mismatch (21.43\%), indicating that increased task complexity reintroduces the expected difficulty penalty for dimension changes.
    \item \textbf{Model-Specific Difficulty Curves:} Other models show varied sensitivity to grid consistency. Veo-3 displays mixed behavior—dropping slightly in v1 (5.70\% to 3.95\%) but surprisingly improving on Mismatch in v2 (3.77\% to 4.55\%), potentially due to low sample variance. Conversely,  image models like Qwen-image consistently struggle with Mismatch tasks, collapsing to near-zero performance across both datasets.
\end{itemize}

\textbf{Dimensionality-Difficulty Interaction.} A granular intersectional analysis of difficulty and grid consistency (shown in \Cref{tab:arc_v1_difficulty} and \Cref{tab:arc_v2_difficulty}) exposes a paradox in video model behavior distinct from standard scaling laws:
\vspace{-2mm}
\begin{itemize}
    \item \textbf{The Sora-2 Anomaly (Hard Case Inversion):} On v1 Hard cases, Sora-2 displays a highly anomalous inverted profile, achieving nearly 4$\times$ higher accuracy on Mismatch tasks (29.63\%) compared to Match tasks (7.41\%). Fine-grained metrics reveal this is not a structural failure—\textbf{Grid Integrity} remains identical (96.30\%) across both categories. Instead, the collapse is driven by \textbf{Color Accuracy}, which plummets to 18.52\% on Match compared to 44.44\% on Mismatch. This implies video models struggle to execute complex pixel-level color transformations when constrained to a fixed canvas, but recover capability when generating new spatial structures.
    \item \textbf{Nano-banana Pro (Logical Scaling):} In contrast, the leading image model exhibits expected difficulty scaling. It consistently performs better on Match (fixed-dimension) tasks than Mismatch (dynamic-dimension) tasks across all difficulty levels (\textit{e.g.}, v1 Hard Match 37.14\% vs. Hard Mismatch 27.27\%), confirming a stable, dimension-agnostic reasoning core.
    \item \textbf{Brittleness of the Generative Advantage:} Sora-2's ``generative advantage'' on Mismatch tasks proves brittle. On the v2 benchmark, the anomaly vanishes entirely: Sora-2 collapses to 0.00\% on Hard Mismatch cases, while Nano-banana Pro maintains robust capability (22.22\%). This reinforces that while video models may occasionally exploit spatial restructuring, they lack the robust generalization of their image-based counterparts.
\end{itemize}

\textbf{Metric Cascade and Bottlenecks.} The four evaluation metrics form a consistent funnel where success becomes progressively more restrictive. For Nano-banana on v1: Grid Integrity (55.79\%) $\to$ Pattern Recognition (28.42\%) $\to$ Color Accuracy (12.63\%) $\to$ Overall (9.21\%). This cascade reveals that approximately 51\% of structurally correct outputs achieve pattern recognition, 44\% of those achieve color accuracy, and 73\% of color-accurate outputs reach full correctness. \textbf{Color Accuracy emerges as the critical bottleneck}, representing only 20--25\% of Grid Integrity performance across all models. This bottleneck is most severe for GPT-4o-image: despite achieving 10.24\% Grid Integrity, it manages only 0.52\% Color Accuracy, resulting in 0.00\% overall success. The cascade pattern suggests that models can learn structural rules but struggle with precise color-based execution.

\textbf{Version Comparison.} The transition from v1 to v2 exposes model robustness to novel pattern families:
\vspace{-2mm}
\begin{itemize}
    \item \textbf{Dramatic Video Model Collapse:} Sora-2 drops from 20.18\% to 1.33\% (93\% decline), while Nano-banana drops from 9.21\% to 2.67\% (71\% decline). Nano-banana Pro remains stable (30.54\% $\to$ 30.36\%).
    \item \textbf{Dataset Composition Effects:} v2 skews harder, containing only 1 Easy Match case versus 102 in v1. The difficulty distribution shift disproportionately penalizes models that rely on memorized patterns.
    \item \textbf{Grid Integrity Preservation:} Interestingly, Veo-3 maintains similar Grid Integrity across versions (32.98\% v1, 31.11\% v2) despite failing on the primary metric, suggesting it can preserve structure without understanding transformation semantics.
\end{itemize}

\subsubsection{Key Findings and Implications}

Three critical insights emerge from the ARC evaluation, revealing fundamental limitations in how current generative models approach abstract visual reasoning.

\textbf{1. The Temporal Consistency Barrier.} A defining failure mode of video generation on ARC is the \textbf{inability to maintain static context}. As illustrated in \Cref{fig:arc_case_study_1} and \Cref{fig:arc_case_study_2}, video models fail to distinguish between invariant problem context (the demonstration examples E1--E4) and the dynamic solution generation (test output T1). Demonstration examples—which should remain strictly fixed to define the transformation logic—undergo progressive hallucination: color shifts, pattern deformations, and structural degradation across frames. This ``context drift'' effectively corrupts the problem definition itself, undermining any chance of correct solution generation. The implication is clear: current video architectures treat the entire visual field as a mutable sequence rather than respecting logical invariants, making them fundamentally unsuited for reasoning tasks that require stable reference points.

\textbf{2. The Perception-Execution Gap.} The metric cascade analysis reveals a systematic disconnect between \textbf{pattern perception and precise execution}. Models frequently achieve reasonable Pattern Recognition (up to 71.99\% for Sora-2) and Grid Integrity (up to 94.58\%) scores, yet fail catastrophically on Overall accuracy. This suggests that current generative models can \textit{see} the transformation pattern but cannot \textit{execute} it correctly. The bottleneck consistently occurs at Color Accuracy, where performance drops to 20--25\% of Grid Integrity levels. This perception-execution gap implies that models may learn to recognize visual patterns without internalizing the precise symbolic rules governing color assignment—a fundamental limitation for tasks requiring exact constraint satisfaction.

\textbf{3. The Robustness-Memorization Trade-off.} The diagnostic collapse from v1 to v2—particularly Sora-2's 93\% relative decline—exposes a critical reliance on \textbf{memorized patterns rather than generalizable reasoning}. In contrast, Nano-banana Pro's stability across the split (30.54\% $\to$ 30.36\%) demonstrates that image-based generation can achieve more robust abstract reasoning. This dichotomy suggests that video models may be overfitting to temporal motion patterns in training data rather than learning the underlying transformation logic. For ARC specifically, where novel pattern families are explicitly designed to defeat memorization, this limitation is fatal. The finding has broader implications: video generation architectures may require fundamentally different training objectives or inductive biases to support genuine rule-based reasoning rather than sophisticated pattern matching.

\section{Math}
\label{apx:math}

We construct the \textbf{Visual Math} task to evaluate \textbf{Logical Reasoning}, \textbf{Logical Deduction}, and \textbf{2D Spatial Reasoning}. This task probes a model's ability to understand complex mathematical problems presented visually (\textit{e.g.}, geometry, diagrams) and challenges its \textbf{Temporal Reasoning} by requiring it to generate a video animating the step-by-step deduction from premises to solution \citep{huang2025vision}.

\subsection{Hard-Level Control}
\label{sec:math_datasets}

\begin{figure*}[t!]
    \centering
    \includegraphics[width=0.95\textwidth,clip,trim=0 140 220 0]{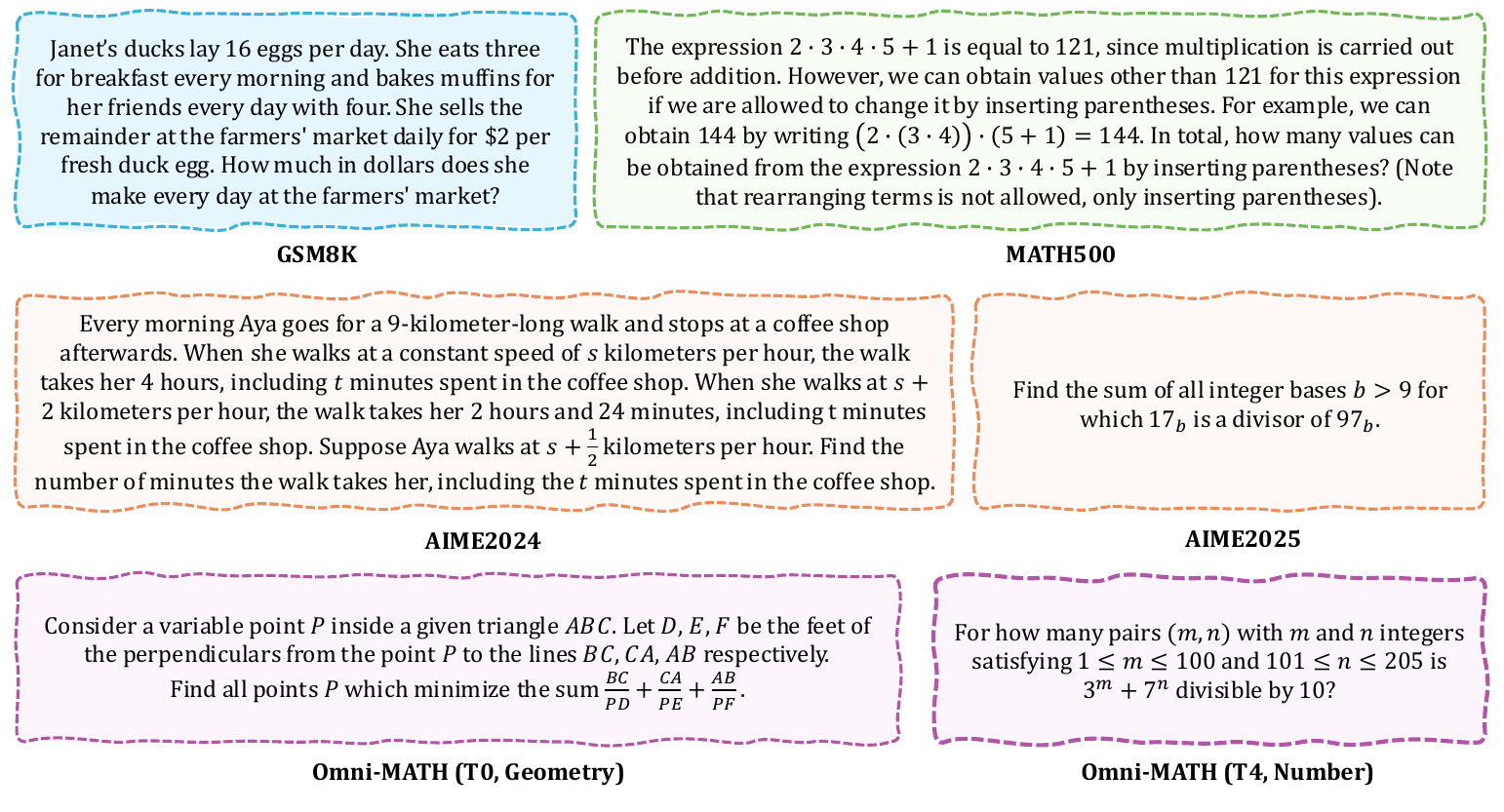}
    \vspace{-5mm}
    \caption{Examples from our selected five mathematical reasoning benchmarks.}
    \label{fig:math_examples}
\end{figure*}

We evaluate models across \textbf{five} benchmarks that span a wide spectrum of difficulty: \textbf{GSM8K}~\citep{cobbe2021gsm8k} (grade school), \textbf{MATH}~\citep{hendrycks2021math} (high school competition), the \textbf{AIME 2024}~\citep{aime2024} and \textbf{AIME 2025}~\citep{aime2025} invitational examinations, and \textbf{Omni-MATH}~\citep{omnimath2024} (Olympiad-level). For Omni-MATH, we further classify problems by \textit{difficulty} (five levels: T0--T4) and \textit{category} (eight types: Algebra, Applied Math, Calculus, Discrete Math, Geometry, Precalculus, Number Theory, and Other). \Cref{tab:math_datasets_summary} summarizes the evaluation benchmarks and sample counts, while \Cref{tab:omnimath_distribution} details the Omni-MATH sample distribution across difficulty levels and categories. \Cref{fig:math_examples} shows some selected examples.

\begin{table}[h]
\centering
\caption{Summary of mathematical reasoning benchmarks used for evaluation. GSM8K contains grade school math word problems. MATH covers high school competition mathematics across multiple subjects. AIME (American Invitational Mathematics Examination) represents advanced competition-level problems. Omni-MATH provides fine-grained categorization across difficulty levels and subject areas.\looseness=-1}
\label{tab:math_datasets_summary}
\begin{adjustbox}{max width=0.45\textwidth}{
\begin{tabular}{lcc}
\toprule
\textbf{Dataset} & \textbf{Level} & \textbf{Sample Count} \\
\midrule
\textbf{GSM8K} & Grade School & 50 \\
\textbf{MATH500} & High School Competition & 50 \\
\textbf{AIME 2024} & Invitational Competition & 30 \\
\textbf{AIME 2025} & Invitational Competition & 30 \\
\textbf{Omni-MATH} & Multi-level (T0-T4) & 167 \\
\midrule
\textbf{Total} & -- & \textbf{327} \\
\bottomrule
\end{tabular}
}
\end{adjustbox}
\end{table}

  \begin{table*}[h!]
  \begin{threeparttable}
  \centering
  \caption{Sample distribution of Omni-MATH dataset across difficulty levels and mathematical categories. \textbf{Note:} Omni-MATH provides a fine-grained ontology for mathematical problem classification. Difficulty levels range from T0 (easiest, suitable for middle school) to T4 (hardest, Olympiad-level problems). Categories cover the major branches of mathematics commonly tested in competitions and standardized assessments.}
  \label{tab:omnimath_distribution}
  \small
  \setlength{\tabcolsep}{5pt}
  \begin{tabular}{lcccccccc}
  \toprule
  \textbf{Difficulty} & \textbf{Algebra} & \textbf{Applied Math} & \textbf{Calculus} & \textbf{Discrete Math} & \textbf{Geometry} & \textbf{Precalculus} & \textbf{Number} & \textbf{Other} \\
  \midrule
  T0 (Easiest) & 4 & 5 & 5 & 5 & 5 & 5 & 4 & 1 \\
  T1 & 5 & 5 & 4 & 5 & 4 & 5 & 5 & 0 \\
  T2 & 5 & 5 & 5 & 5 & 5 & 5 & 5 & 1 \\
  T3 & 5 & 5 & 5 & 5 & 5 & 4 & 5 & 0 \\
  T4 (Hardest) & 5 & 5 & 1 & 5 & 5 & 4 & 5 & 0 \\
  \midrule
  \textbf{Total} & \textbf{24} & \textbf{25} & \textbf{20} & \textbf{25} & \textbf{24} & \textbf{23} & \textbf{24} & \textbf{2} \\
  \bottomrule
  \end{tabular}
  \begin{tablenotes}
  \small
  \item \textbf{Note:} Omni-MATH provides a fine-grained ontology for mathematical problem classification. Difficulty levels range from T0 (easiest, suitable for middle school) to T4
  (hardest, Olympiad-level problems). Categories cover the major branches of mathematics commonly tested in competitions and standardized assessments.
  \end{tablenotes}
  \end{threeparttable}
  \end{table*}

\subsection{Evaluation and Metrics}
\label{apx:math_evaluation}

We use Gemini-2.5-Pro~\citep{comanici2025gemini} as our evaluator to assess the correctness and quality of generated solutions. The evaluator analyzes intermediate reasoning steps, final answers, and reflective behavior (for video generations). We define the following metrics:
\vspace{-2mm}
\begin{itemize}
    \item \textbf{Final Correctness (FC)}: 1 if the final solution matches the ground truth answer, 0 otherwise. This metric verifies whether the model arrives at the correct conclusion.
    \item \textbf{Intermediate Correctness (IC)}: 1 if all reasoning steps leading to the solution are logically valid and mathematically sound, 0 otherwise. This evaluates the quality of the step-by-step deduction process.
    \item \textbf{Action Reflection (AR)} (videos only): 1 if the generated video exhibits self-correction behavior (\textit{e.g.}, revising incorrect steps, reconsidering approaches), 0 if the solution proceeds without reflection. This metric is not applicable to static image generations.
    \item \textbf{Overall Score}: 1 if and only if both \textbf{Final Correctness=1} AND \textbf{Intermediate Correctness=1}, 0 otherwise. This is the primary metric representing complete solution correctness.
\end{itemize}

\begin{figure*}[h!]
    \centering
    \small
    \setlength{\tabcolsep}{2pt}
    \begin{tabular}{cc}
        \subcaptionbox{t=0s\label{fig:math_case_0013_t0}}{
            \includegraphics[width=0.48\textwidth]{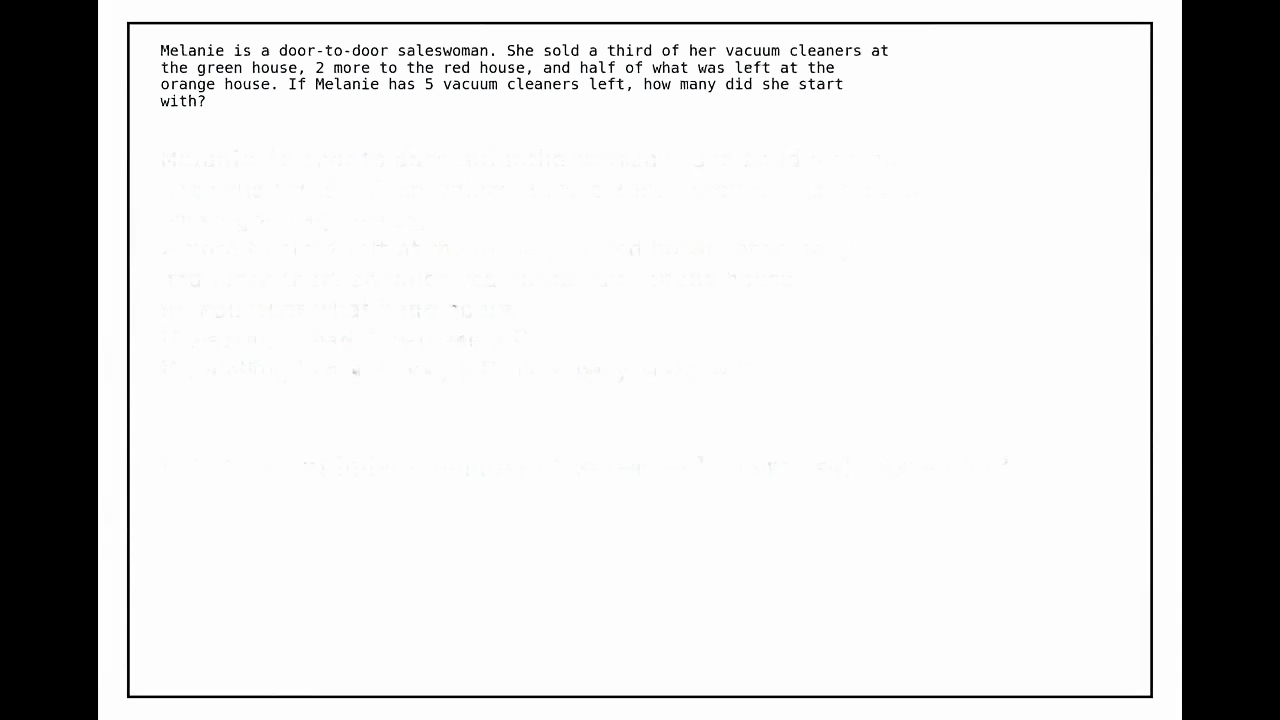}
        } &
        \subcaptionbox{t=2s\label{fig:math_case_0013_t2}}{
            \includegraphics[width=0.48\textwidth]{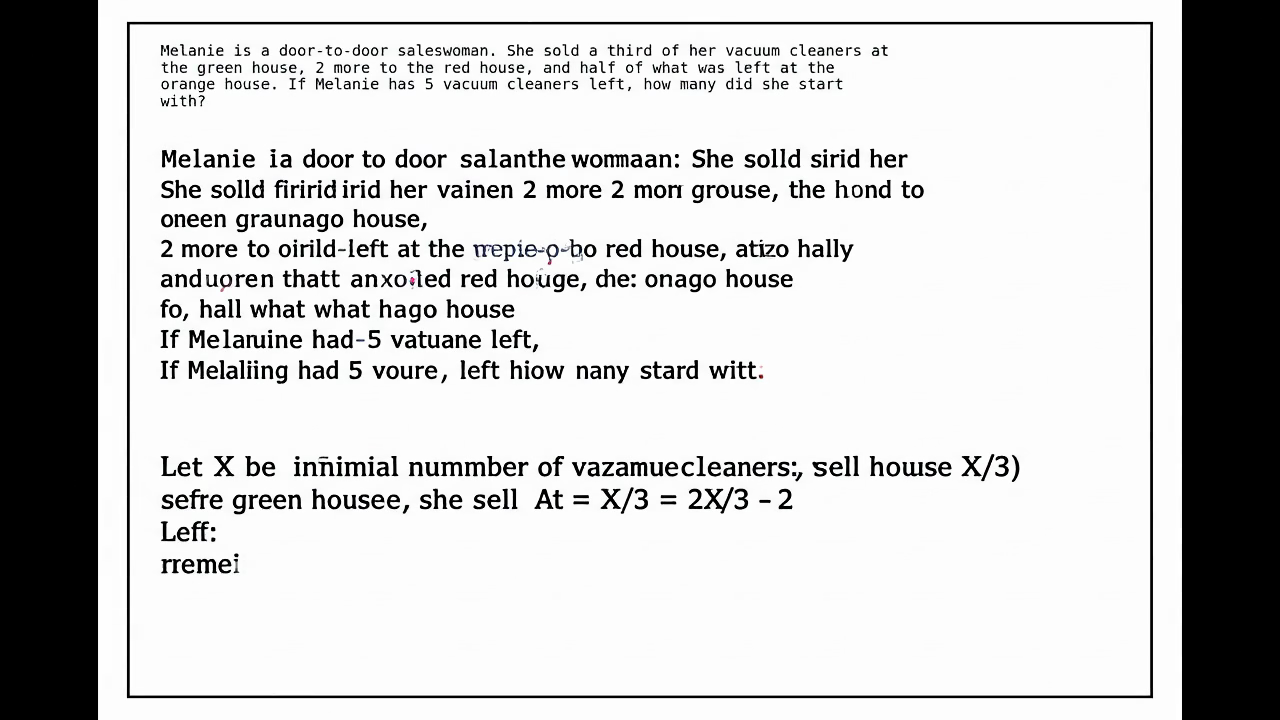}
        } \\[0.5em]
        \subcaptionbox{t=4s\label{fig:math_case_0013_t4}}{
            \includegraphics[width=0.48\textwidth]{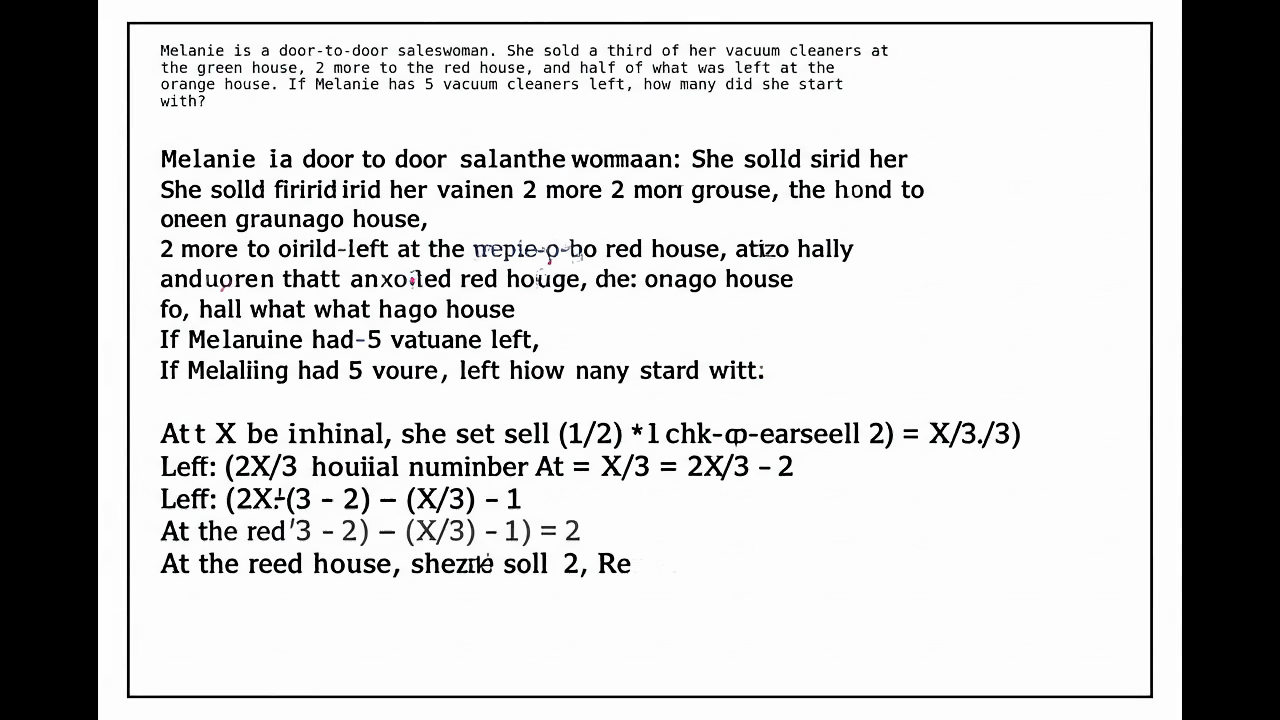}
        } &
        \subcaptionbox{t=6s\label{fig:math_case_0013_t6}}{
            \includegraphics[width=0.48\textwidth]{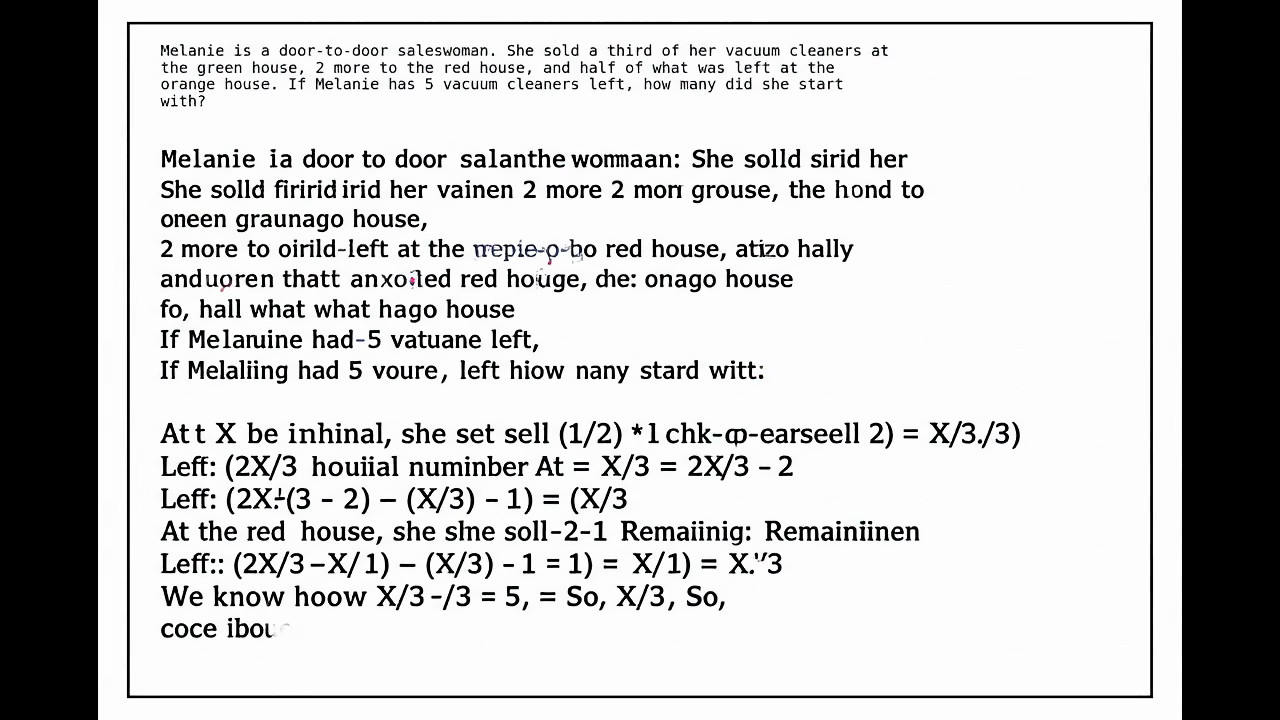}
        }
    \end{tabular}
    \vspace{-2mm}
    \caption{Temporal evolution of math problem solving at t=0s, 2s, 4s, 6s. Generated by Veo-3.}
    \label{fig:math_case_study_problem_0013}
\end{figure*}

\subsection{Evaluation Results}


\textbf{Key Finding: The Reasoning-outcome Disconnect.} Final task-level scores show that \textbf{Nano-banana Pro} (72.69\%) is competitive with the Gemini text baselines (71.38--74.14\%), while video models remain far behind (0--10.61\%). Diagnostic fine-grained results indicate a severe ``reasoning disconnect'' in video generation: outputs can reach plausible final answers while failing to preserve valid intermediate logic.

\subsubsection{Overall Performance Patterns}

\begin{table*}[t!]
\centering
\small
\caption{Final task-level results for the \textbf{Math} task. Scores match the Math row in \Cref{tab:main_results}; the detailed benchmark breakdowns that follow are diagnostic analyses from the fine-grained evaluation runs.}
\label{tab:math_final_scores}
\begin{adjustbox}{max width=0.72\textwidth}
\begin{tabular}{@{}llc@{}}
\toprule
\textbf{Model} & \textbf{Type} & \textbf{Final score} $\uparrow$ \\
\midrule
Veo-3 & Video & 10.61\% \\
Sora-2 & Video & 10.56\% \\
Wan-2.2 & Video & 0.00\% \\
\midrule
Nano-banana & Image & 11.47\% \\
Nano-banana Pro & Image & 72.69\% \\
GPT-4o-image & Image & 26.24\% \\
GPT-image-1.5 & Image & 25.52\% \\
Qwen-image & Image & 11.22\% \\
\midrule
Gemini-3-Flash & Text baseline & 71.38\% \\
Gemini-3-Pro & Text baseline & \textbf{74.14\%} \\
\bottomrule
\end{tabular}
\end{adjustbox}
\end{table*}

\begin{table*}[h!]
\small
\centering
\caption{Diagnostic fine-grained results for the \textbf{Math} task across the original benchmark splits. These per-dataset metrics explain process/outcome behavior for the model subset available in the fine-grained run; the final task-level scores are reported in \Cref{tab:math_final_scores}. ``--'' denotes a missing per-dataset result.}
\label{tab:math_results_grouped}
\begin{adjustbox}{max width=\textwidth}
\begin{tabular}{@{}lcccc@{}}
\toprule
& \multicolumn{3}{c}{\textbf{Fine-grained Metrics}} & \textbf{Primary Metric} \\
\cmidrule(lr){2-4}
\textbf{Model} & \textbf{Process Success Rate} $\uparrow$ & \textbf{Outcome Success Rate} $\uparrow$ & \textbf{Action Reflection} $\uparrow$ & \textbf{Overall Success Rate} $\uparrow$ \\
\midrule
\multicolumn{5}{@{}l}{\textbf{Dataset: \textit{GSM8K}}} \\
\multicolumn{5}{@{}l}{\quad \textbf{Video Models}} \\
\quad \quad Veo-3 & 12.00\% & 74.00\% & 12.00\% & 12.00\% \\
\quad \quad Sora-2 & 38.00\% & 64.00\% & 16.00\% & 30.00\% \\
\quad \quad Wan-2.2 & 2.00\% & 2.00\% & 0.00\% & 2.00\% \\
\multicolumn{5}{@{}l}{\quad \textbf{Image Models}} \\
\quad \quad Nano-banana & 44.00\% & 88.00\% & N/A & 42.00\% \\
\quad \quad Nano-banana Pro & 97.83\% & 97.83\% & N/A & 97.83\% \\
\quad \quad GPT-4o-image & 80.00\% & 83.48\% & N/A & 75.65\% \\
\quad \quad Qwen-image & 44.00\% & 44.00\% & N/A & 44.00\% \\

\multicolumn{5}{@{}l}{\textbf{Dataset: \textit{MATH500}}} \\
\multicolumn{5}{@{}l}{\quad \textbf{Video Models}} \\
\quad \quad Veo-3 & 20.00\% & 52.00\% & 14.00\% & 18.00\% \\
\quad \quad Sora-2 & 34.04\% & 59.57\% & 10.64\% & 31.91\% \\
\quad \quad Wan-2.2 & 3.33\% & 6.00\% & 0.00\% & 3.33\% \\
\multicolumn{5}{@{}l}{\quad \textbf{Image Models}} \\
\quad \quad Nano-banana & 16.00\% & 74.00\% & N/A & 16.00\% \\
\quad \quad Nano-banana Pro & 91.84\% & 91.84\% & N/A & 91.84\% \\
\quad \quad GPT-4o-image & 29.14\% & 42.45\% & N/A & 27.34\% \\
\quad \quad Qwen-image & -- & -- & N/A & -- \\

\multicolumn{5}{@{}l}{\textbf{Dataset: \textit{AIME24}}} \\
\multicolumn{5}{@{}l}{\quad \textbf{Video Models}} \\
\quad \quad Veo-3 & 8.33\% & 8.33\% & 5.00\% & 1.67\% \\
\quad \quad Sora-2 & 8.70\% & 13.04\% & 4.35\% & 4.35\% \\
\quad \quad Wan-2.2 & 15.56\% & 22.22\% & 11.11\% & 15.56\% \\
\multicolumn{5}{@{}l}{\quad \textbf{Image Models}} \\
\quad \quad Nano-banana & 5.00\% & 15.00\% & N/A & 0.00\% \\
\quad \quad Nano-banana Pro & 63.64\% & 36.36\% & N/A & 31.82\% \\
\quad \quad GPT-4o-image & 3.33\% & 10.00\% & N/A & 0.00\% \\
\quad \quad Qwen-image & 1.12\% & 1.12\% & N/A & 1.12\% \\

\multicolumn{5}{@{}l}{\textbf{Dataset: \textit{AIME25}}} \\
\multicolumn{5}{@{}l}{\quad \textbf{Video Models}} \\
\quad \quad Veo-3 & 3.33\% & 11.67\% & 1.67\% & 3.33\% \\
\quad \quad Sora-2 & 8.70\% & 21.74\% & 4.35\% & 0.00\% \\
\quad \quad Wan-2.2 & 5.56\% & 10.00\% & 3.33\% & 5.56\% \\
\multicolumn{5}{@{}l}{\quad \textbf{Image Models}} \\
\quad \quad Nano-banana & 1.75\% & 33.33\% & N/A & 1.75\% \\
\quad \quad Nano-banana Pro & 71.43\% & 90.48\% & N/A & 66.67\% \\
\quad \quad GPT-4o-image & 0.00\% & 3.33\% & N/A & 0.00\% \\
\quad \quad Qwen-image & 4.00\% & 4.00\% & N/A & 4.00\% \\

\multicolumn{5}{@{}l}{\textbf{Dataset: \textit{Omni-MATH}}} \\
\multicolumn{5}{@{}l}{\quad \textbf{Video Models}} \\
\quad \quad Veo-3 & 4.79\% & 15.57\% & 5.09\% & 3.89\% \\
\quad \quad Sora-2 & 0.62\% & 1.88\% & 6.88\% & 0.62\% \\
\quad \quad Wan-2.2 & 0.41\% & 3.46\% & 0.61\% & 0.41\% \\
\multicolumn{5}{@{}l}{\quad \textbf{Image Models}} \\
\quad \quad Nano-banana & 3.90\% & 39.94\% & N/A & 3.90\% \\
\quad \quad Nano-banana Pro & 65.77\% & 85.59\% & N/A & 63.06\% \\
\quad \quad GPT-4o-image & 4.79\% & 41.92\% & N/A & 4.79\% \\
\quad \quad Qwen-image & 6.67\% & 6.67\% & N/A & 6.67\% \\

\bottomrule
\end{tabular}
\end{adjustbox}
\end{table*}

\textbf{Final-Score Hierarchy.} \Cref{tab:math_final_scores} gives the task-level scores used in \Cref{tab:main_results}. Gemini-3-Pro is highest at 74.14\%, followed closely by Nano-banana Pro at 72.69\% and Gemini-3-Flash at 71.38\%. All video models remain far behind: Veo-3 and Sora-2 are near 10\%, while Wan-2.2 receives 0.00\% under the final scoring protocol.

\textbf{The Modality Gap.} The diagnostic quantitative results (\Cref{tab:math_results_grouped}) explain where this gap comes from. On the entry-level GSM8K benchmark, \textbf{Nano-banana Pro} achieves a near-perfect \textit{Overall Success Rate} of 97.83\%, whereas the best-performing video model, Sora-2, reaches only 30.00\%—a greater than $3\times$ performance disparity. This gap widens on competition-level mathematics; on AIME 25, Nano-banana Pro maintains a robust 66.67\%, while Sora-2 collapses to 0.00\%.

\textbf{The Illusion of Reasoning in Video.} A critical insight from the fine-grained metrics is the divergence between \textit{Outcome Success Rate} and \textit{Process Success Rate} in video models. On GSM8K, Veo-3 achieves a relatively high Outcome Success Rate of 74.00\% but a low Process Success Rate of 12.00\%. This substantial $\sim$62\% delta indicates that video models frequently arrive at the correct final solution through ``hallucinated'' or logically invalid visual transitions, effectively guessing the answer without successfully modeling the mathematical steps. In contrast, image models like Nano-banana Pro show near-perfect alignment (97.83\% for both metrics), demonstrating that their outputs rely on consistent step-by-step deduction.

\textbf{Action Reflection and Error Correction.} Video-specific \textit{Action Reflection} scores remain critically low across the board (0--16\%), implying a limited capacity for self-correction during temporal generation.
\vspace{-2mm}
\begin{itemize}
    \item Inverse Scaling: Interestingly, different video models struggle in different regimes. Sora-2 leads video models on easier tasks (30.00\% on GSM8K) but degrades on hard tasks.
    \item The Wan-2.2 Anomaly: Conversely, Wan-2.2 performs poorly on simple tasks (2.00\% on GSM8K) but unexpectedly outperforms other video models on the hardest benchmarks (15.56\% on AIME 24), suggesting it may possess latent reasoning capabilities that are not triggered by simpler prompts, or that its training distribution favors complex visual proofs over simple arithmetic.
\end{itemize}

\subsubsection{Dataset Difficulty Progression}

Performance degradation across datasets (\Cref{tab:math_results_grouped}) reveals three distinct scaling behaviors: robust retention (Nano-banana Pro), rapid collapse (GPT-4o-image), and the ``middle-peak'' stability of video models (Veo-3).

\textbf{GSM8K (Grade School): Hallucination Gap.} On this baseline benchmark, image models demonstrate mastery, with Nano-banana Pro hitting a ceiling of 97.8\%. In contrast, Veo-3 exhibits a defining characteristic of current video generation: the ``illusion of competence.'' While it achieves a high Outcome Success Rate of 74.00\%, its Process Success Rate is only 12.00\% (and Overall Success matches at 12.00\%). This massive disparity suggests that on simple word problems, Veo-3 retrieves correct answers from its training data without generating the corresponding visual reasoning to support them. Sora-2 performs better here (30.00\% Overall), but still trails image models by a wide margin.

\textbf{MATH500 (High School Competition): Video Stability Phenomenon.} As problem complexity increases, a counter-intuitive trend emerges for video models. While image models like GPT-4o-image suffer a 64\% relative decline (75.65\% $\to$ 27.34\%), Veo-3 actually improves, rising from 12.00\% on GSM8K to 18.00\% on MATH500. Similarly, Sora-2 remains stable (31.91\%). This suggests that the structured, formal nature of competition mathematics—which often involves distinct, sequential steps—may align better with video generation temporal priors than the varying linguistic structures of grade-school word problems. Meanwhile, Nano-banana Pro continues to dominate, maintaining a 91.84\% success rate.

\textbf{AIME \& Omni-MATH: Hard-Reasoning Frontier.} On the most challenging benchmarks, model behaviors diverge sharply:\looseness=-1
\vspace{-2mm}
\begin{itemize}
    \item \textbf{The Image SOTA:} Nano-banana Pro defies the difficulty curve, achieving 66.67\% on AIME25 and 63.06\% on Omni-MATH, proving that visual generation models can sustain complex reasoning chains.
    \item \textbf{The Video Shuffle:} The hierarchy among video models inverts at this level. Wan-2.2, which failed on easy tasks, spikes to 15.56\% on AIME24. Furthermore, on Omni-MATH, Veo-3 emerges as the most robust video model (3.89\%), significantly outperforming both Sora-2 (0.62\%) and Wan-2.2 (0.41\%). This indicates that while Veo-3 struggles with the ``exactness'' of simple arithmetic (GSM8K), it possesses a superior capacity for generalized, albeit imperfect, reasoning on novel, high-complexity tasks compared to its peers.
\end{itemize}

\subsubsection{Omni-MATH Deep Dive}

The fine-grained analysis of Omni-MATH reveals that difficulty and domain do not impact video and image models uniformly (\Cref{tab:omni_math_level} and \Cref{tab:omni_math_category}).

\textbf{Difficulty Analysis: The ``Inverse Scaling'' Anomaly.} Standard benchmarks usually show linear degradation as difficulty increases (T0 $\to$ T4). However, both model types exhibit counter-intuitive behavior here:
\vspace{-2mm}
\begin{itemize}
    \item \textbf{Video Non-Monotonicity:} Veo-3 displays a ``U-shaped'' performance curve. It starts at 6.06\% (T0), dips to a near-zero 1.47\% on T3, but surprisingly recovers to 5.00\% on T4 (the hardest tier). This suggests that T4 problems, while mathematically harder, may possess canonical structures (standard Olympiad templates) that video models can memorize and reproduce more effectively than the ambiguous, semi-structured problems found in intermediate tiers (T1--T3).
    \item \textbf{Image Inverse Scaling:} Nano-banana Pro exhibits arguably the most shocking result in the dataset: it performs better on the hardest problems than the easiest ones. Its Overall Success Rate skyrockets from 26.67\% on T1 to a commanding 82.76\% on T4. This implies the model is highly optimized for formal, high-complexity mathematical proofs rather than simpler, variable-heavy word problems.
\end{itemize}

\textbf{Category Analysis: The Visual vs. Abstract Divide.} Domain breakdowns expose the specific cognitive limitations of video generation.
\vspace{-2mm}
\begin{itemize}
    \item \textbf{Geometry:} Video models perform best in Geometry (Veo-3: 8.33\%), outperforming their own averages in other domains. This confirms that video generation logic aligns well with geometric construction, where reasoning steps (\textit{e.g.}, drawing auxiliary lines) correspond directly to visual frame transitions.
    \item \textbf{The Abstract Collapse:} Conversely, video models face catastrophic failure in domains requiring abstract symbolic manipulation. In \textbf{Calculus}, Discrete Math, Applied Math, and Precalculus, video models collapse to an Overall Success Rate of 0.00\%. Notably, in Precalculus, Veo-3 achieves a 21.74\% \textit{Outcome Success Rate} but 0.00\% overall, illustrating extreme ``hallucination''—guessing the right number through invalid visual morphing.
    \item \textbf{Image Robustness:} Unlike video models, Nano-banana Pro maintains high process correctness across all domains, including those where video fails (\textit{e.g.}, 53.33\% in Calculus; 68.75\% in Applied Math). This confirms that SOTA image models have successfully internalized the symbolic logic required for abstract math, whereas video models remain tethered to visual pattern matching.
\end{itemize}

\subsubsection{Key Findings and Implications}

Three critical insights emerge from the evaluation, pointing to fundamental divergences in how image and video architectures process mathematical logic.

\textbf{1. The Temporal Penalties of Reasoning.} The results establish that video generation currently penalizes mathematical reasoning rather than enhancing it. While image-based models like \textbf{Nano-banana Pro} have achieved near-mastery of complex logic (66\%+ on AIME), video models face a ``temporal tax.'' The requirement to maintain frame-to-frame coherence appears to compete with logical consistency, leading to a massive performance gap—often exceeding 4--6$\times$ on hard benchmarks. This suggests that current video architectures treat mathematical derivation as a \textit{visual texture} to be morphed, rather than a \textit{semantic chain} to be constructed, necessitating new architectures that decouple reasoning states from visual rendering.

\textbf{2. The ``Hallucination'' of Competence in Video.} A defining failure mode of current video models is \textbf{solution-answer dissociation}. Models like Veo-3 frequently achieve high Outcome Success Rates (\textbf{e.g.}, 74\% on GSM8K) while failing Process Success (12\%), indicating they reach correct answers through invalid or hallucinated visual transitions. This contrasts sharply with SOTA image models, where process and outcome metrics are tightly coupled ($>99\%$ correlation for Nano-banana Pro). This dissociation undermines the utility of video for educational or explanatory tasks, as the visual "proof" provided by the video is often mathematically illusory despite the final answer being correct.

\textbf{3. Domain-Specific ``Islands of Aptitude.''} Visual mathematical reasoning is not a monolithic capability but highly domain-dependent.

\begin{itemize}
    \item \textbf{The Geometry Bias:} Video models show a clear inductive bias for Geometry (8.33\% success), where the reasoning process (construction, transformation) is inherently spatial-temporal.
    \item \textbf{The Abstract Barrier:} Conversely, video models collapse to near-zero performance on abstract domains like Calculus and Discrete Math.
    \item \textbf{The ``Inverse Scaling'' Paradox:} Unexpectedly, models like Wan-2.2 (Video) and Nano-banana Pro (Image) perform significantly better on the hardest benchmarks (AIME, Omni-MATH T4) than on intermediate ones. This implies that high-difficulty problems often follow rigid, canonical templates that models can memorize, whereas ``simpler'' problems involving variable linguistic structures or non-standard arithmetic expose the fragility of their generalized reasoning.
\end{itemize}

\subsubsection{Case Study Analysis: Veo-3 Failure Modes}

Visual case studies of Veo-3's generation traces reveal three structural pathologies that explain the ``Reasoning-Outcome Disconnect'' observed in the quantitative results.

\textbf{1. Temporal Drift and Visual Hallucination.} The most pervasive failure mode is \textit{Temporal Drift}, where the logical integrity of the solution degrades as the video progresses, even if the initial setup is correct. This is the primary driver of the massive gap between Outcome Success (74.00\%) and Process Success (12.00\%) on GSM8K. In 62\% of cases, Veo-3 ``teleports'' to the correct final answer via visually fluid but mathematically invalid transitions—morphing numbers arbitrarily or skipping essential derivation steps. This suggests the model minimizes visual prediction error (pixel consistency) at the expense of logical semantic error, prioritizing a smooth-looking video over a mathematically valid proof.

\textbf{2. The "Correction Paradox" (Toxic Reflection).} Counter-intuitively, the model's self-correction mechanisms often act as noise injection rather than error mitigation. While Veo-3 exhibits an Action Reflection rate of 14.00\% on MATH500, these edits rarely salvage the solution. Qualitative analysis shows that when the model attempts to ``backtrack'' and rewrite a frame, it frequently introduces continuity errors or hallucinates new, irrelevant constraints. Rather than exhibiting genuine metacognition (detecting logical flaws), the reflection appears to be a stochastic process—randomly modifying parts of the visual field—which disrupts the linear chain of reasoning required for multi-step problems.

\textbf{3. Spatial Attention Collapse (Positional Bias).} Veo-3 displays a distinct ``foveal bias,'' disproportionately attending to the center of the visual field while neglecting constraints or auxiliary figures located at the periphery. This bias is particularly detrimental in Geometry tasks (where Veo-3 otherwise performs well, relative to other domains). In failed instances, the model successfully renders the central geometric construction but fails to incorporate numerical values or variable definitions positioned in the upper or lower margins. This spatial neglect results in ``correctly solved'' wrong problems—logic that is internally consistent but divorced from the specific boundary conditions of the prompt.

\begin{table*}[h!]
\small
\centering
\caption{Quantitative breakdown results for the \textbf{Omni-MATH} task. We evaluate video generative models (Veo-3, Sora-2, and Wan-2.2) and image generative models (Nano-banana, Nano-banana Pro, GPT-4o-image, and Qwen-image) on five \textbf{difficulty} levels (T0--T4).}
\label{tab:omni_math_level}
\begin{adjustbox}{max width=0.9\textwidth}{ 
\begin{tabular}{@{}lcccc@{}}
\toprule
& \multicolumn{3}{c}{\textbf{Fine-grained Metrics}} & \textbf{Primary Metric} \\
\cmidrule(lr){2-4}
\textbf{Model} & \textbf{Process Success Rate} $\uparrow$ & \textbf{Outcome Success Rate} $\uparrow$ & \textbf{Action Reflection} $\uparrow$ & \textbf{Overall Success Rate} $\uparrow$ \\
\midrule

\multicolumn{5}{@{}l}{\textbf{T0}} \\
\multicolumn{5}{@{}l}{\quad \textbf{Video Models}} \\
\quad \quad Veo-3 & 6.06\% & 12.12\% & 3.03\% & \textbf{6.06\%} \\
\quad \quad Sora-2 & 3.03\% & 6.06\% & 15.15\% & 3.03\% \\
\quad \quad Wan-2.2 & 0.98\% & 4.90\% & 2.94\% & 0.98\% \\
\multicolumn{5}{@{}l}{\quad \textbf{Image Models}} \\
\quad \quad Nano-banana & 0.00\% & 33.82\% & N/A & 0.00\% \\
\quad \quad Nano-banana Pro & 53.85\% & 84.62\% & N/A & \textbf{53.85\%} \\
\quad \quad GPT-4o-image & 0.00\% & 0.00\% & N/A & 0.00\% \\
\quad \quad Qwen-image & 22.00\% & 22.00\% & N/A & 22.00\% \\
\addlinespace 

\multicolumn{5}{@{}l}{\textbf{T1}} \\
\multicolumn{5}{@{}l}{\quad \textbf{Video Models}} \\
\quad \quad Veo-3 & 4.55\% & 4.55\% & 1.52\% & \textbf{3.03\%} \\
\quad \quad Sora-2 & 0.00\% & 3.12\% & 3.12\% & 0.00\% \\
\quad \quad Wan-2.2 & 0.00\% & 6.06\% & 0.00\% & 0.00\% \\
\multicolumn{5}{@{}l}{\quad \textbf{Image Models}} \\
\quad \quad Nano-banana & 0.00\% & 19.70\% & N/A & 0.00\% \\
\quad \quad Nano-banana Pro & 26.67\% & 60.00\% & N/A & \textbf{26.67\%} \\
\quad \quad GPT-4o-image & 0.00\% & 0.00\% & N/A & 0.00\% \\
\quad \quad Qwen-image & 8.42\% & 13.68\% & N/A & 8.42\% \\
\addlinespace

\multicolumn{5}{@{}l}{\textbf{T2}} \\
\multicolumn{5}{@{}l}{\quad \textbf{Video Models}} \\
\quad \quad Veo-3 & 5.56\% & 16.67\% & 2.78\% & \textbf{5.56\%} \\
\quad \quad Sora-2 & 0.00\% & 0.00\% & 2.86\% & 0.00\% \\
\quad \quad Wan-2.2 & 0.93\% & 3.70\% & 0.00\% & 0.93\% \\
\multicolumn{5}{@{}l}{\quad \textbf{Image Models}} \\
\quad \quad Nano-banana & 4.17\% & 47.22\% & N/A & 4.17\% \\
\quad \quad Nano-banana Pro & 70.00\% & 96.67\% & N/A & \textbf{66.67\%} \\
\quad \quad GPT-4o-image & 2.78\% & 2.78\% & N/A & 2.78\% \\
\quad \quad Qwen-image & 7.92\% & 8.91\% & N/A & 7.92\% \\
\addlinespace

\multicolumn{5}{@{}l}{\textbf{T3}} \\
\multicolumn{5}{@{}l}{\quad \textbf{Video Models}} \\
\quad \quad Veo-3 & 1.96\% & 13.24\% & 1.47\% & \textbf{1.47\%} \\
\quad \quad Sora-2 & 0.00\% & 0.00\% & 9.68\% & 0.00\% \\
\quad \quad Wan-2.2 & 0.00\% & 1.96\% & 0.00\% & 0.00\% \\
\multicolumn{5}{@{}l}{\quad \textbf{Image Models}} \\
\quad \quad Nano-banana & 5.88\% & 45.59\% & N/A & 5.88\% \\
\quad \quad Nano-banana Pro & 70.83\% & 79.17\% & N/A & \textbf{62.50\%} \\
\quad \quad GPT-4o-image & 0.00\% & 2.86\% & N/A & 0.00\% \\
\quad \quad Qwen-image & 16.67\% & 12.22\% & N/A & 12.22\% \\
\addlinespace

\multicolumn{5}{@{}l}{\textbf{T4}} \\
\multicolumn{5}{@{}l}{\quad \textbf{Video Models}} \\
\quad \quad Veo-3 & 5.17\% & 36.67\% & 15.00\% & \textbf{5.00\%} \\
\quad \quad Sora-2 & 0.00\% & 0.00\% & 3.45\% & 0.00\% \\
\quad \quad Wan-2.2 & 0.00\% & 0.00\% & 0.00\% & 0.00\% \\
\multicolumn{5}{@{}l}{\quad \textbf{Image Models}} \\
\quad \quad Nano-banana & 9.52\% & 53.97\% & N/A & 9.52\% \\
\quad \quad Nano-banana Pro & 82.76\% & 93.10\% & N/A & \textbf{82.76\%} \\
\quad \quad GPT-4o-image & 0.00\% & 6.67\% & N/A & 0.00\% \\
\quad \quad Qwen-image & 25.68\% & 22.97\% & N/A & 22.97\% \\

\bottomrule
\end{tabular}
}
\end{adjustbox}
\end{table*}

\begin{table*}[p]
\small
\centering
\caption{Quantitative breakdown results for the \textbf{Omni-MATH} task. We evaluate video generative models (Veo-3, Sora-2, and Wan-2.2) and image generative models (Nano-banana, Nano-banana Pro, GPT-4o-image, and Qwen-image) on eight \textbf{categories}. Because image outputs do not support frame-by-frame reasoning, action-reflection–based metrics are omitted for image generative models (marked with ``N/A'').}
\label{tab:omni_math_category}
\begin{adjustbox}{max width=0.78\textwidth}{
\begin{tabular}{@{}lcccc@{}}
\toprule
& \multicolumn{3}{c}{\textbf{Fine-grained Metrics}} & \textbf{Primary Metric} \\
\cmidrule(lr){2-4}
\textbf{Model} & \textbf{Process Success Rate} $\uparrow$ & \textbf{Outcome Success Rate} $\uparrow$ & \textbf{Action Reflection} $\uparrow$ & \textbf{Overall Success Rate} $\uparrow$ \\
\midrule
\multicolumn{5}{@{}l}{\textbf{Algebra}} \\
\multicolumn{5}{@{}l}{\quad \textbf{Video Models}} \\
\quad \quad Veo-3 & 4.17\% & 12.50\% & 4.17\% & 4.17\% \\
\quad \quad Sora-2 & 4.35\% & 8.70\% & 13.04\% & \textbf{4.35\%} \\
\quad \quad Wan-2.2 & 0.00\% & 5.56\% & 0.00\% & 0.00\% \\
\multicolumn{5}{@{}l}{\quad \textbf{Image Models}} \\
\quad \quad Nano-banana & 4.17\% & 45.83\% & N/A & 4.17\% \\
\quad \quad Nano-banana Pro & 61.11\% & 83.33\% & N/A & \textbf{55.56\%} \\
\quad \quad GPT-4o-image & 0.00\% & 4.00\% & N/A & 0.00\% \\
\quad \quad Qwen-image & 19.70\% & 24.24\% & N/A & 19.70\% \\

\multicolumn{5}{@{}l}{\textbf{Applied Math}} \\
\multicolumn{5}{@{}l}{\quad \textbf{Video Models}} \\
\quad \quad Veo-3 & 0.00\% & 20.00\% & 8.00\% & 0.00\% \\
\quad \quad Sora-2 & 0.00\% & 0.00\% & 8.00\% & 0.00\% \\
\quad \quad Wan-2.2 & 0.00\% & 5.33\% & 0.00\% & 0.00\% \\
\multicolumn{5}{@{}l}{\quad \textbf{Image Models}} \\
\quad \quad Nano-banana & 16.00\% & 32.00\% & N/A & 16.00\% \\
\quad \quad Nano-banana Pro & 68.75\% & 87.50\% & N/A & \textbf{68.75\%} \\
\quad \quad GPT-4o-image & 4.00\% & 4.00\% & N/A & 4.00\% \\
\quad \quad Qwen-image & 14.49\% & 14.49\% & N/A & 14.49\% \\

\multicolumn{5}{@{}l}{\textbf{Calculus}} \\
\multicolumn{5}{@{}l}{\quad \textbf{Video Models}} \\
\quad \quad Veo-3 & 5.00\% & 10.00\% & 5.00\% & 0.00\% \\
\quad \quad Sora-2 & 0.00\% & 0.00\% & 10.00\% & 0.00\% \\
\quad \quad Wan-2.2 & 0.00\% & 1.67\% & 0.00\% & 0.00\% \\
\multicolumn{5}{@{}l}{\quad \textbf{Image Models}} \\
\quad \quad Nano-banana & 0.00\% & 35.00\% & N/A & 0.00\% \\
\quad \quad Nano-banana Pro & 66.67\% & 80.00\% & N/A & \textbf{53.33\%} \\
\quad \quad GPT-4o-image & 0.00\% & 0.00\% & N/A & 0.00\% \\
\quad \quad Qwen-image & 15.79\% & 15.79\% & N/A & 15.79\% \\

\multicolumn{5}{@{}l}{\textbf{Discrete Math}} \\
\multicolumn{5}{@{}l}{\quad \textbf{Video Models}} \\
\quad \quad Veo-3 & 0.00\% & 12.00\% & 4.00\% & 0.00\% \\
\quad \quad Sora-2 & 0.00\% & 0.00\% & 4.17\% & 0.00\% \\
\quad \quad Wan-2.2 & 0.00\% & 5.33\% & 0.00\% & 0.00\% \\
\multicolumn{5}{@{}l}{\quad \textbf{Image Models}} \\
\quad \quad Nano-banana & 0.00\% & 24.00\% & N/A & 0.00\% \\
\quad \quad Nano-banana Pro & 54.55\% & 72.73\% & N/A & \textbf{54.55\%} \\
\quad \quad GPT-4o-image & 0.00\% & 0.00\% & N/A & 0.00\% \\
\quad \quad Qwen-image & 15.94\% & 17.39\% & N/A & 15.94\% \\

\multicolumn{5}{@{}l}{\textbf{Geometry}} \\
\multicolumn{5}{@{}l}{\quad \textbf{Video Models}} \\
\quad \quad Veo-3 & 8.33\% & 25.00\% & 4.17\% & \textbf{8.33\%} \\
\quad \quad Sora-2 & 0.00\% & 0.00\% & 0.00\% & 0.00\% \\
\quad \quad Wan-2.2 & 0.00\% & 1.39\% & 4.17\% & 0.00\% \\
\multicolumn{5}{@{}l}{\quad \textbf{Image Models}} \\
\quad \quad Nano-banana & 0.00\% & 45.83\% & N/A & 0.00\% \\
\quad \quad Nano-banana Pro & 78.57\% & 92.86\% & N/A & \textbf{78.57\%} \\
\quad \quad GPT-4o-image & 0.00\% & 0.00\% & N/A & 0.00\% \\
\quad \quad Qwen-image & 23.64\% & 20.00\% & N/A & 18.18\% \\

\multicolumn{5}{@{}l}{\textbf{Precalculus}} \\
\multicolumn{5}{@{}l}{\quad \textbf{Video Models}} \\
\quad \quad Veo-3 & 4.35\% & 21.74\% & 8.70\% & 0.00\% \\
\quad \quad Sora-2 & 0.00\% & 0.00\% & 0.00\% & 0.00\% \\
\quad \quad Wan-2.2 & 0.00\% & 1.45\% & 0.00\% & 0.00\% \\
\multicolumn{5}{@{}l}{\quad \textbf{Image Models}} \\
\quad \quad Nano-banana & 13.04\% & 65.22\% & N/A & 13.04\% \\
\quad \quad Nano-banana Pro & 68.18\% & 86.36\% & N/A & \textbf{68.18\%} \\
\quad \quad GPT-4o-image & 0.00\% & 4.17\% & N/A & 0.00\% \\
\quad \quad Qwen-image & 20.90\% & 16.42\% & N/A & 16.42\% \\

\multicolumn{5}{@{}l}{\textbf{Number}} \\
\multicolumn{5}{@{}l}{\quad \textbf{Video Models}} \\
\quad \quad Veo-3 & 4.17\% & 8.33\% & 12.50\% & \textbf{4.17\%} \\
\quad \quad Sora-2 & 0.00\% & 0.00\% & 0.00\% & 0.00\% \\
\quad \quad Wan-2.2 & 1.39\% & 1.39\% & 0.00\% & 1.39\% \\
\multicolumn{5}{@{}l}{\quad \textbf{Image Models}} \\
\quad \quad Nano-banana & 0.00\% & 50.00\% & N/A & 0.00\% \\
\quad \quad Nano-banana Pro & 64.29\% & 92.86\% & N/A & \textbf{64.29\%} \\
\quad \quad GPT-4o-image & 0.00\% & 4.00\% & N/A & 0.00\% \\
\quad \quad Qwen-image & 2.82\% & 4.23\% & N/A & 2.82\% \\

\bottomrule
\end{tabular}
}
\end{adjustbox}
\end{table*}

\section{Embodied Navigation}
\label{sec:embodied}

We develop the \textbf{Embodied Navigation} task to evaluate \textbf{3D Spatial Understanding}, \textbf{2D Spatial Grounding}, \textbf{Temporal Reasoning}, and \textbf{Physical Commonsense}. This task probes a model’s ability to reason about ego-centric environments, interpreting scene geometry, anticipating future states, and respecting physical constraints, and challenges its multi-step planning ability by requiring it to generate a video or image that depicts the agent’s trajectory as it navigates toward the goal.

\subsection{Task Definition}
\label{sec:embodied_task-definition}

\begin{figure*}[htbp]
    \centering
    \includegraphics[width=\textwidth]{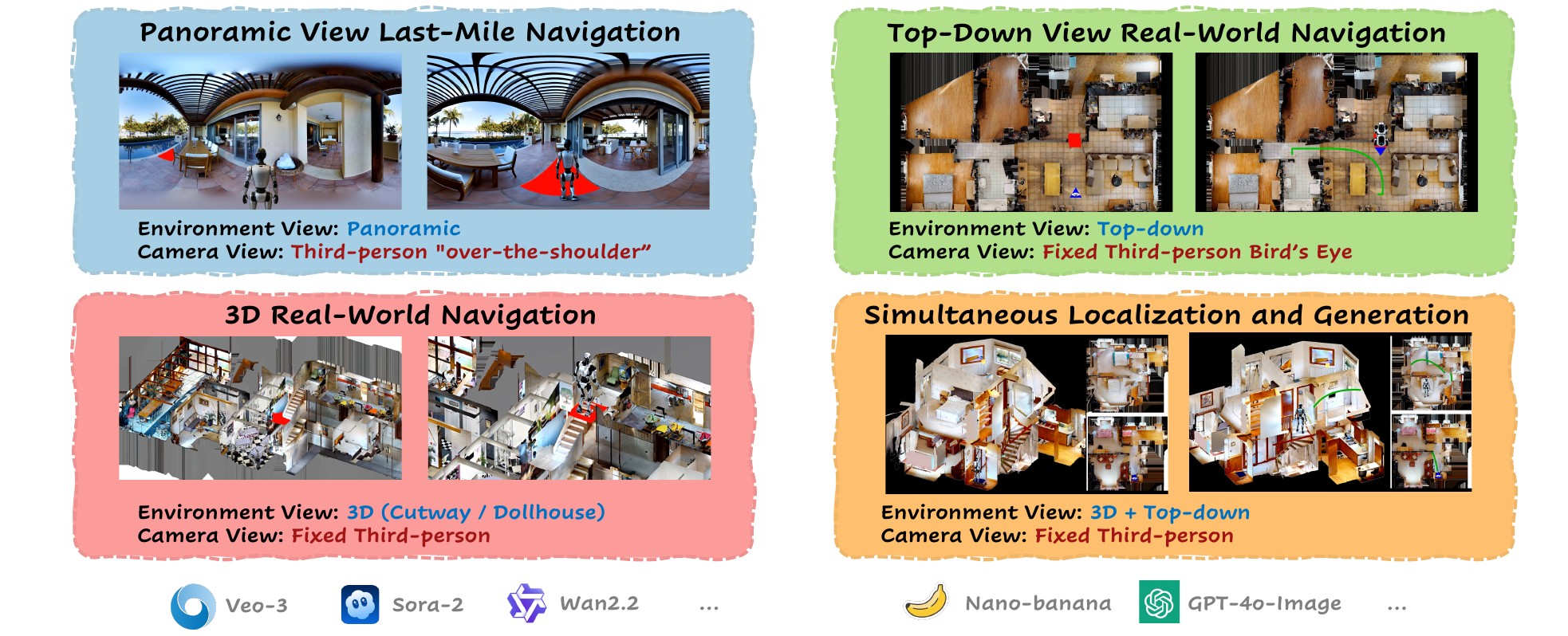}
    \caption{Visual illustration of the four Embodied Navigation task settings: Panoramic View Last-Mile Navigation, Top-down View Real-World Navigation, 3D Real-World Navigation, and Simultaneous Localization and Generation (SLAG).}
    \label{fig:embodied_tasks}
\end{figure*}

\Cref{fig:embodied_tasks} illustrates the four Embodied Navigation task settings evaluated in our benchmark:
\vspace{-2mm}
\begin{itemize}
    \item \textbf{Panoramic View Last-Mile Navigation (Last-Mile Nav.)} (\Cref{apx:last_mile_navigation}) presents a $360^\circ$ panoramic environment from a third-person ``over-the-shoulder'' perspective, requiring models to reason over wide-field visual context for short-range navigation.
    \item \textbf{Top-down View Real-World Navigation (Top-down View Nav.)} (\Cref{apx:topdown}) uses a fixed bird’s-eye camera, emphasizing global spatial planning and long-horizon path prediction.
    \item \textbf{3D Real-World Navigation (3D R.-W. Nav.)} (\Cref{apx:3dnav}) adopts cutaway/dollhouse-style renderings to expose full 3D structure, challenging models to ground navigation decisions in multi-room, multi-level layouts from a fixed third-person view.
    \item \textbf{Simultaneous Localization and Generation (SLAG)} (\Cref{apx:slag}) combines both 3D and top-down environment views, requiring models to jointly localize the agent and generate the surrounding scene layout. Together, these settings form a comprehensive testbed for evaluating spatial reasoning, geometric understanding, and scene-generation capabilities of modern video and image generative models.
\end{itemize}

\subsection{Hard-Level Control}
\label{sec:embodied_hard-level-control}

\begin{table*}[t]
\centering
\caption{Distribution of evaluation samples across the 24 hard-level configurations defined by environmental complexity, view fidelity, trajectory distance, and destination specification. Counts are reported for the four embodied navigation tasks: Panoramic View Last-Mile Navigation (Last-Mile Nav.), Top-down View Real-World Navigation (Top-down View Nav.), 3D Real-World Navigation (3D R.-W. Nav.), and Simultaneous Localization and Generation (SLAG).}
\label{tab:distribution_embodied_hard-level_samples}
\small
\resizebox{\linewidth}{!}{
\begin{tabular}{llllcccc}
\toprule
\textbf{Env. Complexity} & \textbf{View Fidelity} & \textbf{Distance Level} & \textbf{Destination Type} & \textbf{Last-Mile Nav.} & \textbf{Top-down View Nav.} & \textbf{3D R.-W. Nav.} & \textbf{SLAG} \\
\midrule

\multirow{12}{*}{1 Floor}
    & \multirow{4}{*}{quality03}
        & \multirow{2}{*}{short}
            & color mark & 5 & 5 & 5 & 5 \\
        &   &   & location description & 5 & 5 & 5 & 5 \\
        &   & \multirow{2}{*}{long}
            & color mark & 5 & 5 & 5 & 5 \\
        &   &   & location description & 5 & 5 & 5 & 5 \\
\cmidrule(lr){2-8}

    & \multirow{4}{*}{quality04}
        & \multirow{2}{*}{short}
            & color mark & 5 & 5 & 5 & 5 \\
        &   &   & location description & 5 & 5 & 5 & 5 \\
        &   & \multirow{2}{*}{long}
            & color mark & 5 & 5 & 5 & 5 \\
        &   &   & location description & 5 & 5 & 5 & 5 \\
\cmidrule(lr){2-8}

    & \multirow{4}{*}{quality05}
        & \multirow{2}{*}{short}
            & color mark & 5 & 5 & 5 & 5 \\
        &   &   & location description & 5 & 5 & 5 & 5 \\
        &   & \multirow{2}{*}{long}
            & color mark & 5 & 5 & 5 & 5 \\
        &   &   & location description & 5 & 5 & 5 & 5 \\

\midrule

\multirow{12}{*}{2 Plus Floors}
    & \multirow{4}{*}{quality03}
        & \multirow{2}{*}{short}
            & color mark & 5 & 5 & 5 & 5 \\
        &   &   & location description & 5 & 5 & 5 & 5 \\
        &   & \multirow{2}{*}{long}
            & color mark & 5 & 5 & 5 & 5 \\
        &   &   & location description & 5 & 5 & 5 & 5 \\
\cmidrule(lr){2-8}

    & \multirow{4}{*}{quality04}
        & \multirow{2}{*}{short}
            & color mark & 5 & 5 & 5 & 5 \\
        &   &   & location description & 5 & 5 & 5 & 5 \\
        &   & \multirow{2}{*}{long}
            & color mark & 5 & 5 & 5 & 5 \\
        &   &   & location description & 5 & 5 & 5 & 5 \\
\cmidrule(lr){2-8}

    & \multirow{4}{*}{quality05}
        & \multirow{2}{*}{short}
            & color mark & 5 & 5 & 5 & 5 \\
        &   &   & location description & 5 & 5 & 5 & 5 \\
        &   & \multirow{2}{*}{long}
            & color mark & 5 & 5 & 5 & 5 \\
        &   &   & location description & 5 & 5 & 5 & 5 \\

\midrule
\textbf{Total} & \multicolumn{3}{c}{All 24 Configurations} & \textbf{120} & \textbf{120} & \textbf{120} & \textbf{120} \\
\bottomrule
\end{tabular}
}
\end{table*}

To build a consistent and controllable evaluation suite across the four embodied navigation tasks—Panoramic View Last-Mile Navigation, Top-down View Real-World Navigation, 3D Real-World Navigation, and Simultaneous Localization and Generation (SLAG)—we structure the dataset along four hard-level axes: \textbf{Environmental Complexity}, \textbf{View Fidelity}, \textbf{Trajectory Distance}, and \textbf{Destination Specification}. Each axis is defined through a unified set of principles shared across tasks, while accommodating modality-specific differences in sensing, spatial representation, and scene geometry. This organization ensures cross-task comparability while preserving the unique challenges intrinsic to each navigation setting.

\begin{figure*}[t]
    \centering
    \begin{subfigure}[b]{0.48\linewidth}
        \includegraphics[width=\linewidth]{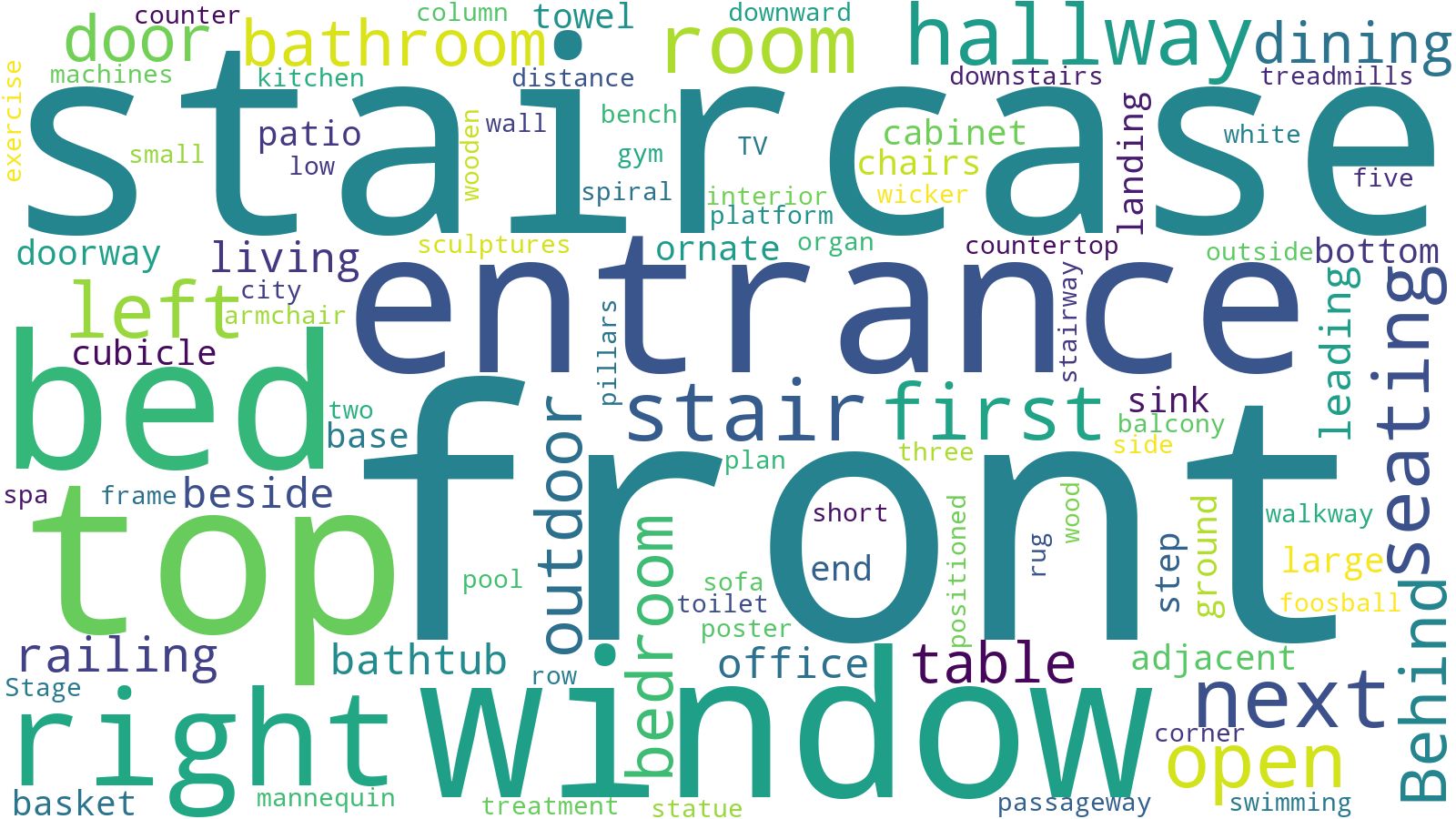}
        \caption{Sample Vocabulary Distribution of  Panoramic View Last-Mile Navigation.}
        \label{subfig:cloud_a}
    \end{subfigure}
    \hfill
    \begin{subfigure}[b]{0.48\linewidth}
        \includegraphics[width=\linewidth]{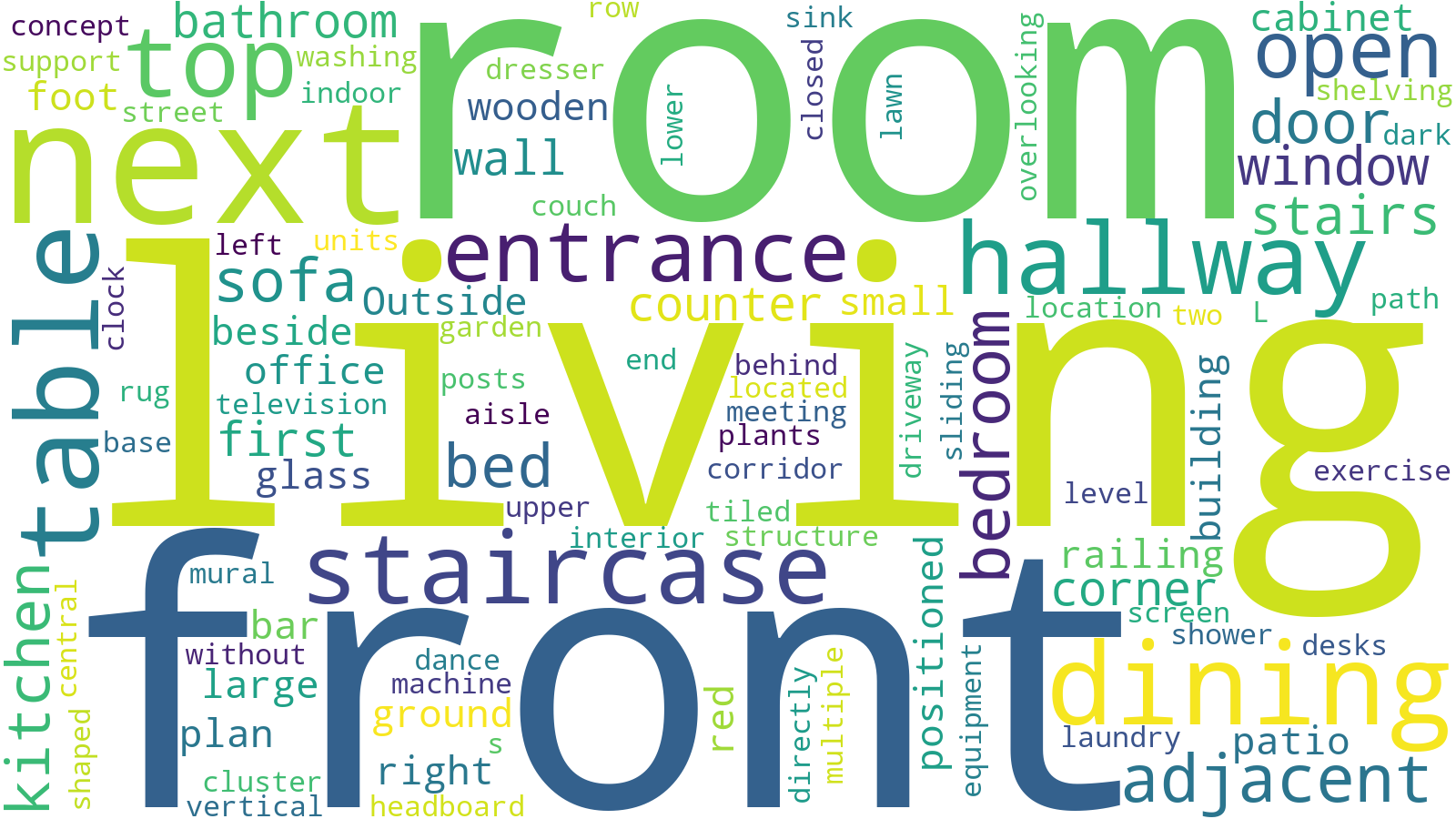}
        \caption{Sample Vocabulary Distribution of 3D Real-World Navigation.}
        \label{subfig:cloud_b}
    \end{subfigure}

    \vspace{-1mm}
    \caption{Word Clouds of Semantic Target Descriptions for Destination Specification (Hard Level). It visualizes the frequency of terms used by annotators to describe target locations using natural language.
    The dominance of spatial prepositions (\textit{e.g.}, ``front,'' ``top,'' ``next,'' ``right'') and structural landmarks (\textit{e.g.}, ``staircase,'' ``window,'' ``entrance,'' ``living [room]'') indicates that annotators heavily rely on relative positioning and salient architectural features. This validates the hard-level protocol, where ambiguity in complex environments is resolved by adding spatially anchored landmarks and floor identifiers to uniquely isolate the intended area.}
    \label{fig:dest-spec-wordcloud}
    \vspace{-2mm}
\end{figure*}

\subsubsection{Environmental Complexity}

We source scenes from photo-realistic indoor scans in Matterport3D~\citep{Matterport3D} and HM3D~\citep{HM3D}, as well as rendered environments in Habitat~\citep{savva2019habitat}. Environmental complexity varies by the structural layout visible to the agent. In Panoramic View Last-Mile Navigation, complexity is determined by the spatial arrangement captured within a single panorama: \textit{floor01} scenes correspond to single-floor homes with no vertical transitions, whereas \textit{floor02plus} scenes include multi-level structures with either implicit stairs not fully visible in the panorama or explicit staircases enabling vertical navigation. For the remaining three tasks, which rely on rendered 3D views or top-down maps, multi-level environments present fully connected floors and additional branching regions. These multi-floor layouts may include basements, attics, outdoor pools, and gardens. By varying structural scale while keeping other factors fixed, the benchmark ensures controlled yet diverse navigation challenges.

\subsubsection{View Fidelity}

Because the four navigation tasks use distinct rendering modalities, we define view fidelity in a task-specific yet cross-comparable manner. For Panoramic View Last-Mile Navigation, fidelity captures how much of the environment is visually accessible. Human raters evaluate the extent and spatial distribution of occlusions from foreground objects—\textit{i.e.}, how much of the room layout is obstructed and how many landmarks remain visible. Scores range from 3 to 5, corresponding to \textit{quality03} through \textit{quality05}. For Top-down View Real-World Navigation, 3D Real-World Navigation, and SLAG, fidelity reflects a more holistic assessment of scene realism and navigability. Raters consider factors such as the presence of holes or cracks, furnishing quality, door openness, and the plausibility of interaction with the environment. These scenes are likewise scored on the 3–5 scale aligned with \textit{quality03} to \textit{quality05}.

\subsubsection{Trajectory Distance}

Trajectory distance is defined as the geodesic separation between the agent’s starting point and the target location. We categorize trajectories into two types. \textbf{Short} trajectories involve relatively direct motion: they require no major turns and may include vertical movement (\textit{e.g.}, reaching an upper floor) without substantial directional changes. \textbf{Long} trajectories, in contrast, include at least one significant turn. To maintain comparability across the four tasks, long trajectories are selected so that they share partial path structure with the corresponding short cases at the same hard level.

\subsubsection{Destination Specification}

Targets are specified either through direct visual annotation or through a natural-language description of the same location. For \textbf{color-based targets} (\textit{color mark}), annotators highlight the target region in the input image using a pure red overlay (\#ff0000), ensuring that it is visually distinguishable from its surroundings. For \textbf{semantic targets} (\textit{location description}), annotators describe the corresponding region using natural language. Although both specifications are intended to reference the same location, ambiguity may arise when multiple similar regions exist. In such cases, annotators include additional disambiguating details—such as floor identifiers and spatially anchored landmarks—to ensure the description uniquely identifies the intended target area.\looseness=-1

These four axes allow us to systematically control difficulty across all 24 configuration slots summarized in the table, while keeping the evaluation consistent across tasks that vary in sensing modality, scene rendering, and navigation objective.
\Cref{tab:distribution_embodied_hard-level_samples}
provides the detailed distribution of the Embodied Navigation samples.

\subsection{Evaluation and Metrics}\label{sec:embodied_eval_metrics}

\begin{figure*}[htbp]
    \centering
    \includegraphics[width=0.9\textwidth]{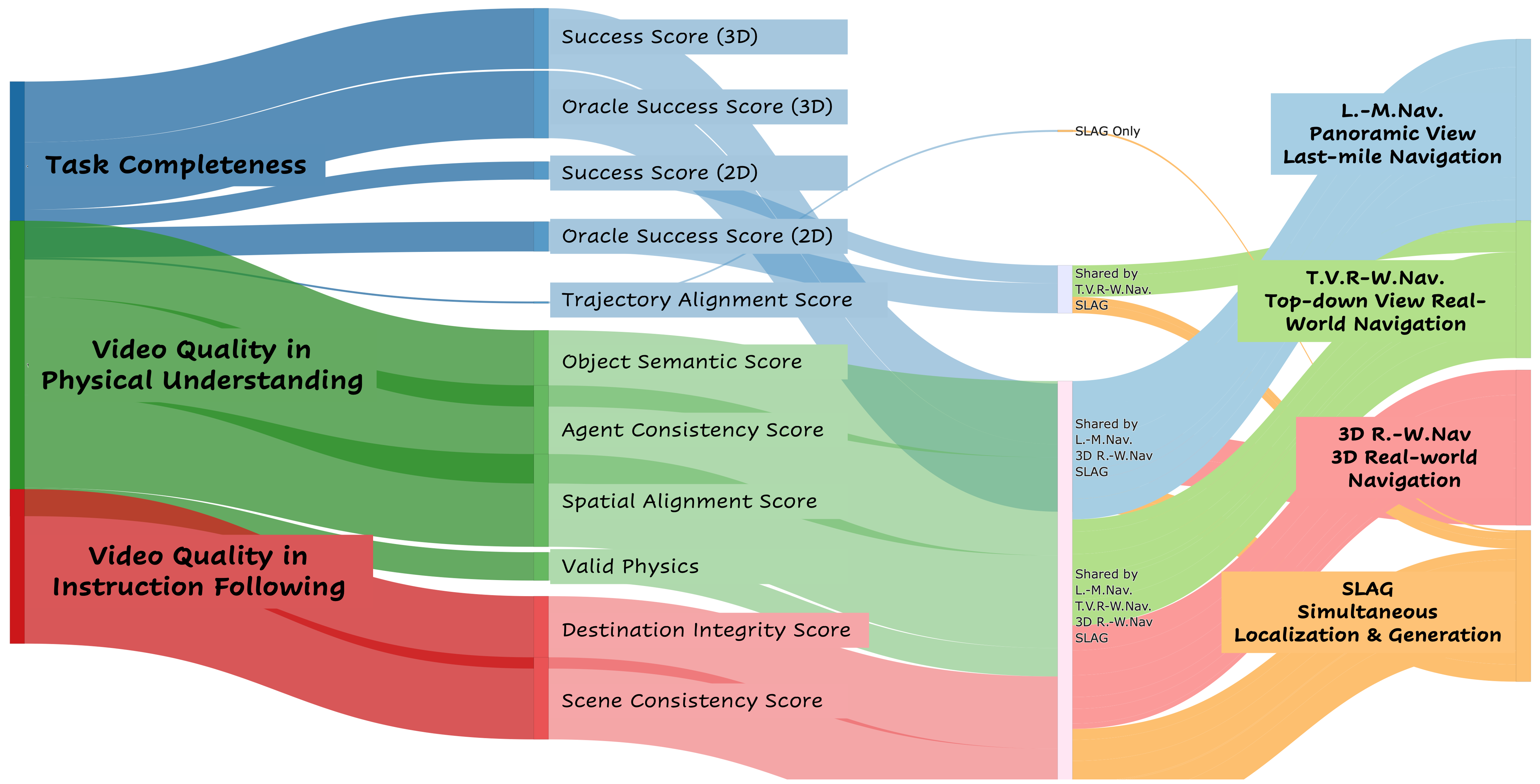}
    \caption{Evaluation metrics flow. Decomposition of the three main metrics \textbf{Task Completeness}, \textbf{Video Quality in Physical Understanding}, and \textbf{Video Quality in Instruction Following} into fine-grained components (\textit{e.g.}, Consistency, Physical Plausibility, Instruction Alignment) and their mapping to the corresponding navigation tasks.}
    \label{fig:embodied_metrics_flow}
\end{figure*}

Evaluating navigation-conditioned video or image generation requires metrics that jointly capture geometric fidelity and visual-semantic reasoning. Our evaluation focuses on whether the agent moves plausibly through the environment, correctly interprets the navigation instruction, adheres to the physical layout of the scene, and preserves the identity and spatial integrity of the target destination. All metrics are binary and are computed directly from the agent’s execution trace and the generated video frames. The definitions below consolidate the evaluation protocol used across all navigation tasks, drawing from the criteria specified in the evaluation prompt templates and supplementary rules.
As illustrated in \Cref{fig:embodied_metrics_flow}, this framework is adaptive: while several fine-grained metrics are shared across multiple tasks, others are task-specific to address unique navigation modalities.

\subsubsection{Task Completeness Metrics}

These metrics assess \textbf{whether the agent reaches or meaningfully approaches the correct destination based solely on geometric information}. They intentionally ignore visual fidelity, semantic correctness, and physical plausibility, which are evaluated by separate metrics.
\vspace{-2mm}
\begin{itemize}
    \item \textbf{Success Score (S.S. 2D).} Measures whether the agent’s final position lies within the highlighted or textually specified goal region in the 2D overhead map. The score is $1$ if the final coordinates fall entirely inside the goal footprint; otherwise $0$.
    \item \textbf{Oracle Success Score (O.S. 2D).} Provides partial credit when the agent comes sufficiently close to the 2D goal during navigation. The score is $1$ if the agent’s path ever intersects or touches the goal region, even if it does not stop there; otherwise $0$.
    \item \textbf{Success Score (S.S. 3D).} Checks whether the agent ends inside the correct destination volume in the 3D navigation sequence. This metric is purely geometric and independent of any visual discrepancies at the destination. The score is $1$ if the final 3D position is within the target volume; otherwise $0$.
    \item \textbf{Oracle Success Score (O.S. 3D).} Grants credit when the agent enters the vicinity of the correct 3D destination at any point during its rollout. The score is $1$ if the trajectory ever crosses the predefined proximity threshold around the target; otherwise $0$.
    \item \textbf{Trajectory Alignment Score (Traj. Ali.).} Evaluates whether the agent’s 2D projected route is consistent with its 3D motion path, focusing on major turns and spatial transitions. A score of $1$ indicates strong correspondence between the two trajectories; otherwise $0$.
\end{itemize}

\subsubsection{Video Quality in Physical Understanding}

These metrics assess \textbf{whether the agent’s motion obeys basic physical principles} and remains consistent with the underlying scene geometry. They focus on physical plausibility, continuity, and spatial coherence rather than destination correctness.
\vspace{-2mm}
\begin{itemize}
    \item \textbf{Object Semantic Score (Obj. Sem.).} Evaluates whether the agent interacts with the environment in a physically valid way. The agent must not collide with, pass through, or visually intersect solid structures such as walls, furniture, or appliances. Score is $1$ if no collision or penetration is observed; otherwise $0$.
    \item \textbf{Agent Consistency Score (Agent Con.).} Measures temporal continuity and identity preservation of the navigating agent. For image generation tasks: (1) The agent’s trajectory must remain continuous across frames, and (2) exactly one agent should appear throughout the navigation sequence. Score is $1$ if the same agent moves smoothly and consistently across all frames; otherwise $0$.
    \item \textbf{Spatial Alignment Score (Spa. Ali.).} Checks whether the agent’s heading, motion direction, and elevation changes remain coherent with the expected physical layout. For image generation tasks: (1) The initial position must be visually identifiable when provided, and (2) the agent’s initial facing direction must align with its first movement. Score is $1$ if heading, transitions, and movement direction are physically and visually consistent; otherwise $0$.
\end{itemize}

\subsubsection{Video Quality in Instruction Following}

Because a video generation model can ``cheat'' in embodied navigation—by fabricating a visually plausible destination, altering the environment, or painting a new target beneath the agent—we impose strict constraints to ensure faithful instruction following. These metrics evaluate \textbf{whether the generated video preserves the intended destination and maintains a static, coherent scene}.

\begin{itemize}
    \item \textbf{Destination Integrity Score (Des. Inte.).} Assesses whether the destination region is preserved and correctly interpreted by the model. According to the supplementary rules: (1) The red-marked target region must remain unchanged in size, position, texture, and overall appearance; (2) The agent must not rely on hallucinated alternatives (\textit{e.g.}, newly created look-alike objects or fabricated goal markers). The score is $1$ if the original destination remains intact and the agent stops within that region; otherwise $0$.
    \item \textbf{Scene Consistency Score (Scene Con.).} Evaluates whether the environment remains static throughout the video. No objects, lighting, geometry, or layout elements may appear, disappear, deform, or shift in a way that violates the static-scene assumption. The score is $1$ if the scene stays unchanged across all frames; otherwise $0$.
\end{itemize}

Across all four navigation tasks, these video-quality metrics employ a unified binary (pass/fail) definition to ensure consistent evaluation.

\subsubsection{Holistic Performance}

Across all embodied navigation tasks, we define an \textbf{Overall} metric that measures end-to-end success under a strict, holistic criterion: a sample is considered correct only if \textbf{all fine-grained evaluation metrics simultaneously achieve a score of $1$}. This requirement highlights a key observation about current generative models: strong performance on isolated metrics does not necessarily translate into coherent, successful task execution. The Overall score thus captures the true difficulty of producing videos that are simultaneously geometrically correct, physically plausible, visually consistent, and instruction-faithful.

We run automatic VLM-based evaluation for the Embodied Navigation task. The automatic setup uses Gemini-2.5-Pro \citep{comanici2025gemini} with a structured evaluation prompt; it receives (i) the model-generated video or image and (ii) task-specific context, and returns binary metric scores and brief justifications.

\subsubsection{Geo-Align}\label{sec:embodied_eval_method_geo_align}
\textbf{Embodied Navigation.} To evaluate the physical faithfulness of the Panoramic View Last-Mile Navigation, we employ \textit{Geo-Align}, a deterministic geometric alignment pipeline that recovers agent trajectories from generated videos for rigorous spatial verification. For the ground truth (GT), shortest-path coordinates are recorded via manual control in Habitat; for generated sequences, we utilize VGGT~\cite{wang2025vggt} to reconstruct a 3D point cloud and sample predicted poses. The synchronization process involves converting panoramic frames into cubemaps, retaining only four lateral faces to align with the front-view bias of VGGT, and performing scale normalization by matching predicted start-to-end Euclidean displacement with the GT. By centering start coordinates and estimating an optimal XZ-plane rotation to align headings, we enable the computation of navigation efficiency metrics, like Success weighted by Path Length (SPL)~\cite{zhang2024visionandlanguage}.


To provide a multidimensional assessment of the reconstructed trajectories, we evaluate performance using the following suite of metrics:

\begin{itemize}
    \item \textbf{Task Success}: We report the binary Success Rate (\textbf{SR})~\cite{anderson2018vision}, defined as reaching the goal within a specified distance threshold, and Success weighted by Path Length (\textbf{SPL})~\cite{zhao_mind_2023}, which accounts for the efficiency of the agent's route relative to the shortest path.
    \item \textbf{Path Fidelity}: We employ normalized Dynamic Time Warping (\textbf{nDTW})~\cite{ilharco2019general} and its success-weighted variant (\textbf{sDTW})~\cite{ilharco2019general} to measure the spatio-temporal similarity between the predicted and ground-truth trajectories, capturing how well the model adheres to the intended geometric path.
    \item \textbf{Trajectory Error}: Accuracy is quantified through Absolute Trajectory Error (\textbf{ATE}), which measures the global translation drift, and Relative Pose Error (\textbf{RPE}), which evaluates local frame-to-frame drift and rotational consistency.
    \item \textbf{Path Length}: We compare the total distance of the \textbf{Predicted} (\textbf{Pred}) trajectory against the \textbf{Ground Truth} (\textbf{GT}) to identify behaviors such as excessive looping (over-generation) or premature stopping.
    \item \textbf{Navigation Error}: To assess terminal precision, we measure the final distance discrepancy (\textbf{Nav}) and the final heading misalignment (\textbf{Dir}) between the predicted agent's pose and the target destination.
\end{itemize}

\subsubsection{View-Syn}\label{sec:embodied_eval_method_view_syn}
For tasks with static observers, such as Top-down View, 3D Dollhouse View, and SLAG, we employ \textit{View-Syn}, a distributional evaluator that measures the alignment between synthesized and GT visual sequences.
Given that fixed perspectives constrain explicit pose extraction, we utilize Fréchet Video Distance (FVD)~\cite{unterthiner2019fvd} for video-generating models to capture temporal and spatial dynamics. For image-generating models, we apply Fréchet Inception Distance (FID)~\cite{heusel2017gans} by extracting the final frame from each GT navigation video to serve as the reference image, comparing it against the generated outputs.
To ensure consistency across the evaluation suite, we normalize the raw FVD and FID values into the unit interval $[0, 1]$. We define the normalized score $S$ such that a lower distributional distance corresponds to a higher alignment score:
\begin{equation}
S = \max \left( 0, \min \left( 1, 1 - \frac{d - d_{\min}}{d_{\max} - d_{\min}} \right) \right)
\end{equation}
, where $d \in \{\text{FID}, \text{FVD}\}$ represents the raw distributional metric, while $d_{\min}$ and $d_{\max}$ denote the empirical bounds of the metric's distribution. Thus, the higher the normalized FVD or FID, the better the quality of generated videos or images.
GT sequences are curated by recording an agent’s shortest-path navigation from synchronized fixed-camera viewpoints within the simulator. This methodology provides a high-level proxy for temporal coherence and semantic goal-verification, evaluating the model's ability to synthesize transitions within a global reference frame where traditional geometric alignment is inapplicable.

To comprehensively evaluate the performance of view synthesis, we utilize five distinct metrics that capture both perceptual quality and distributional similarity:

\begin{itemize}
    \item \textbf{Fréchet Video Distance (FVD)~\cite{unterthiner2019fvd}:} Measures the temporal and spatial distribution alignment between synthesized and ground-truth video sequences.
    \item \textbf{Fréchet Inception Distance (FID)~\cite{heusel2017gans}:} Evaluates the visual quality and diversity of image generation by comparing feature distributions of final synthesized frames against reference images.
    \item \textbf{Structural Similarity Index Measure (SSIM)~\cite{zhou2004ssim}:} Assesses the degradation of structural information, focusing on luminance, contrast, and pixel-wise spatial correlation.
    \item \textbf{Learned Perceptual Image Patch Similarity (LPIPS)~\cite{zhang2018perceptual}:} Utilizes deep features to measure the perceptual distance between frames, aligning more closely with human visual judgment than traditional heuristics.
    \item \textbf{Peak Signal-to-Noise Ratio (PSNR):} Quantifies the ratio between the maximum possible power of a signal and the power of corrupting noise, serving as a standard measure for reconstruction quality.
\end{itemize}

\subsection{Overall Evaluation Results}

We evaluate embodied navigation with one final score per task, matching the four embodied rows in \Cref{tab:main_results}. The comparative final-score summary is reported in \Cref{tab:embodied_final_scores}; later fine-grained tables are diagnostic breakdowns for the model subsets available in the original detailed runs.

\subsubsection{Comparative Analysis of Generators}

As detailed in \Cref{tab:embodied_final_scores}, \textbf{Nano-banana Pro} leads all four embodied tasks: Last-Mile Navigation (75.83\%), Top-down View Navigation (33.05\%), 3D Real-World Navigation (85.00\%), and SLAG (37.29\%). The base Nano-banana model remains strong on Last-Mile and 3D navigation, while GPT-image-1.5 is the most competitive non-Nano image model. Among video models, \textbf{Veo-3} is competitive on Last-Mile Navigation (60.00\%) but falls sharply on Top-down View and SLAG, showing that temporal generation does not reliably preserve goal state, geometry, and cross-view alignment.

Performance varies significantly by task complexity:

\begin{itemize}
    \item \textbf{Last-Mile Navigation:} Nano-banana Pro (75.83\%) and Nano-banana (74.17\%) lead, while Veo-3 is the strongest video model at 60.00\%.
    \item \textbf{Top-down View Navigation:} This task is the hardest embodied row overall. Nano-banana Pro leads at 33.05\%, with GPT-image-1.5 second at 26.27\% and Veo-3 third at 19.49\%.
    \item \textbf{3D Real-World Navigation:} Nano-banana Pro reaches 85.00\%, followed by Nano-banana (79.17\%) and GPT-image-1.5 (77.92\%). Video models remain below 25\%.
    \item \textbf{SLAG:} Nano-banana Pro leads at 37.29\%, followed by GPT-image-1.5 (31.36\%) and Nano-banana (28.79\%). The low video scores highlight persistent cross-view alignment failures.
\end{itemize}

\begin{table*}[t!]
\centering
\small
\setlength{\tabcolsep}{4pt}
\caption{Final task-level scores for \textbf{Embodied Navigation}. Scores match the four embodied rows in \Cref{tab:main_results}; subsequent fine-grained tables report diagnostic subset metrics for the model sets available in the original detailed runs.}
\label{tab:embodied_final_scores}
\begin{adjustbox}{max width=\textwidth}
\begin{tabular}{@{}lcccccccc@{}}
\toprule
\textbf{Task} & \textbf{Veo-3} & \textbf{Sora-2} & \textbf{Wan-2.2} & \textbf{Nano-banana} & \textbf{Nano-banana Pro} & \textbf{GPT-4o-image} & \textbf{GPT-image-1.5} & \textbf{Qwen-image} \\
\midrule
Last-Mile Nav. & 60.00\% & 0.00\% & 14.17\% & 74.17\% & \textbf{75.83\%} & 0.00\% & 55.84\% & 16.67\% \\
Top-down View Nav. & 19.49\% & 3.39\% & 5.09\% & 11.11\% & \textbf{33.05\%} & 3.39\% & 26.27\% & 5.08\% \\
3D R.-W. Nav. & 22.50\% & 0.00\% & 24.17\% & 79.17\% & \textbf{85.00\%} & 13.33\% & 77.92\% & 38.33\% \\
SLAG & 11.02\% & 12.50\% & 0.85\% & 28.79\% & \textbf{37.29\%} & 16.67\% & 31.36\% & 6.78\% \\
\bottomrule
\end{tabular}
\end{adjustbox}
\end{table*}

\subsection{Key Observations}

In this section, we distill the key insights that emerge from evaluating state-of-the-art video and image generation models across the four embodied navigation tasks. These findings offer a high-level view of current strengths and limitations, setting the stage for deeper investigation. The task-specific tables below are diagnostic fine-grained breakdowns from the original subset runs, so their model coverage can be narrower than the final-score summary in \Cref{tab:embodied_final_scores}. In the following sections—\Cref{sec:embodied_pano_analysis,sec:embodied_2d_analysis,sec:embodied_3d_analysis,sec:embodied_slag_analysis}—we introduce each subtask in detail and provide comprehensive quantitative and qualitative analyses.\looseness=-1

\paragraph{Finding 1: Strong Capability in Ego-Centric Navigation Video Generation.} Our results indicate that video generation models possess a surprisingly robust capability for third-person, over-the-shoulder perspective navigation generation. Beyond simple frame coherence, these models demonstrate abilities that have been a long-pursuing goal in embodied navigation~\citep{batra2020objectnav, anderson2018vision}, such as the understanding of \textbf{scene layouts}~\citep{li2023layout, fuentes2015visual, henriques2018mapnet} and \textbf{semantic meanings}~\citep{cartillier2021semantic, chaplot2020object, zhou2025navgpt2, irshad2021sasra, wani2020multion, blanco2008toward, konolige2011navigation, gomez2020hybrid, an2022bevbert}, effectively recognizing goal objects and maintaining contextual consistency even when the agent's body is visible within the frame. Notably, they exhibit strong capabilities in \textbf{imagining spatial relationships}~\citep{koh2021pathdreamer, qin2025navigatediff, sridhar2024nomad, bar2025navigation, shah2025foresightnav}, allowing them to consistently model the geometric interaction between the agent and the environment. This confirms that video diffusion models implicitly learn powerful geometric priors and environment continuity, which are critical for perceiving the immediate surroundings from an embodied perspective. However, such spatial imagination does not translate into functional navigation except in the \textbf{Panoramic View Last-Mile Navigation} task. Only when the goal is already visible, and the required trajectory is short and locally constrained, do models like Veo-3 achieve meaningful success rates. For longer-horizon or multi-view settings, their ability to reason over distance, occlusion, or multi-step spatial transformations rapidly deteriorates, exposing the gap between ego-centric perception and actionable planning.

\paragraph{Finding 2: Free-form Generation is Insufficient for Navigation.} Tasks requiring controlled movement, like Top-down View Real-World Navigation, SLAG, and Instruction-based Target Search, reveal a consistent failure pattern: \textbf{video models excel at producing aesthetically coherent frames but fail to follow navigation constraints}. Models such as Sora-2 often generate \emph{plausible but irrelevant} motion, drift off the intended path, or remain static despite instructions. Even with perfect semantic grounding (\textit{e.g.}, Sora-2 achieving 87.35\% Object Semantic Score), Success Scores remain near zero, underscoring that generative quality does not equal navigational utility. The inability to reliably execute directed motion suggests that current video models lack mechanisms for consistent temporal control, trajectory commitment, or adherence to spatial rules imposed by the prompt.\looseness=-1

\paragraph{Finding 3: Cross-View Spatial Alignment Exists, But Must Be Explicitly Activated.} Our SLAG experiments reveal an unexpected finding: while trajectory alignment remains highly challenging, \textbf{scene consistency improves markedly when models are evaluated in a cross-view alignment framework}. In SLAG, aligning generated motion with a 2D top-down map forces the model to maintain geometric coherence across viewpoints, and this pressure appears to \emph{activate} latent spatial priors that remain unused in other tasks. Veo-3, in particular, maintains high Scene Consistency even when its navigation fails, suggesting robust internal world modeling. However, this ability does not emerge in free-form video generation or text-only prompting. It requires structured multimodal conditioning, indicating that cross-view alignment is a learnable but currently underutilized capability in video models.

\paragraph{Additional Profile of Strengths and Weaknesses.}
Our results suggest a preliminary ``navigation capability profile'' for video generation models:
\begin{itemize}
    \item \textbf{Strengths:} strong semantic grounding, accurate object identity preservation, coherent ego-centric spatial reasoning, and latent ability for cross-view geometric alignment.
    \item \textbf{Weaknesses:} poor rule-following under long-horizon instructions, inability to maintain goal-directed motion, hallucination under abstract destination descriptions (\textit{e.g.}, Veo-3 generating an entirely new scene matching the semantic description), and brittleness to environmental complexity.
\end{itemize}

\paragraph{Summary.} Taken together, these findings suggest that \textbf{current video generation models possess strong ego-centric spatial perception, semantic grounding, and latent cross-view alignment capability, but they fail to reliably execute goal-driven navigation under free-form generation}. Their cross-view understanding must be explicitly elicited through structured multimodal constraints such as SLAG, while long-horizon trajectory control and rule following remain largely unsolved.
In the following sections, we provide detailed analyses for each navigation task, including fine-grained quantitative and qualitative breakdowns.


\section{Panoramic View Last-Mile Navigation (Last-Mile Nav.)} \label{apx:last_mile_navigation}

\subsection{Task Definition}

\begin{figure*}[htbp]
    \centering
    \includegraphics[width=\linewidth]{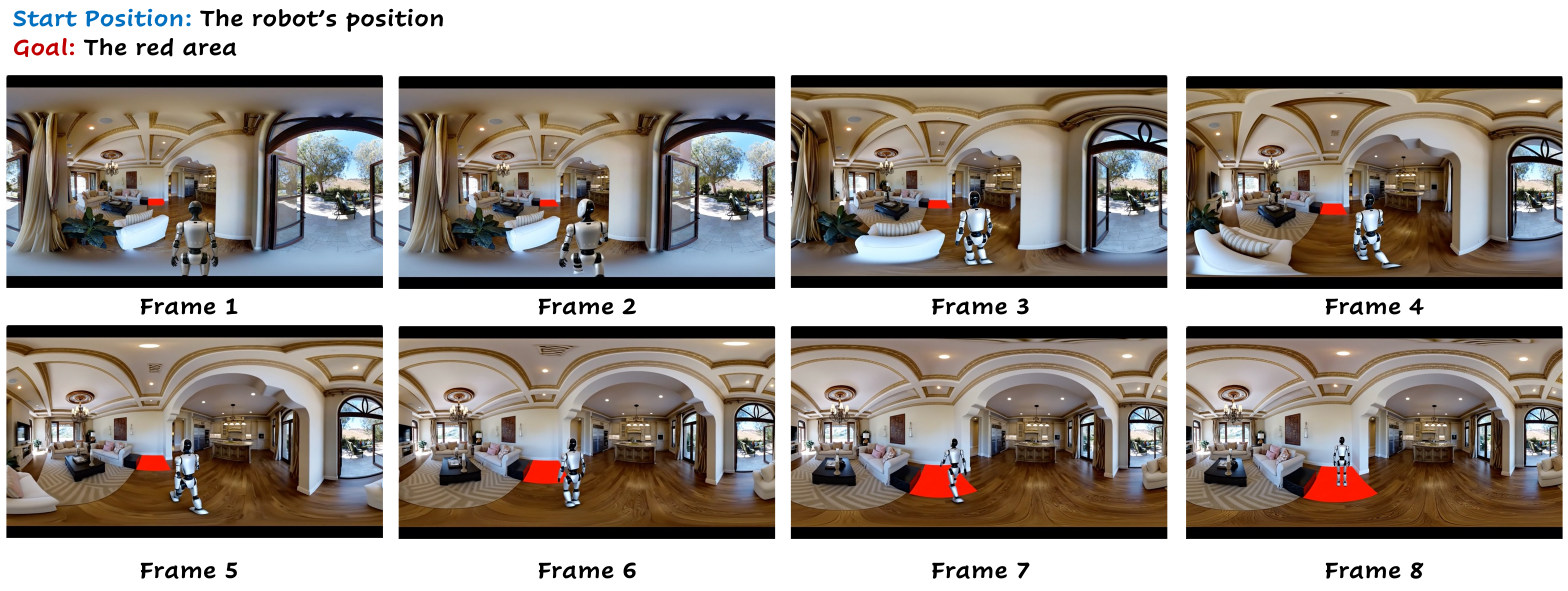}
    \vspace{-5mm}
    \caption{A successful case completed by \textbf{Veo-3} on the \textbf{Panoramic View Last-Mile Navigation} task. The model navigates the final segment of the route from a first-person panoramic viewpoint, demonstrating coherent mental mapping, spatial awareness, and accurate localization needed to reach the precise target destination.}
    \label{fig:lastmile-nav}
    \vspace{-2mm}
\end{figure*}

The Panoramic View Last-Mile Navigation task evaluates fine-grained, embodied decision-making when a target is already within the agent’s visual field. Distinct from large-scale route planning, last-mile navigation isolates the critical final phase of movement: \textbf{precisely localizing a visible goal and generating an optimal short-horizon trajectory toward it}. In this setting, the model receives a single panoramic RGB observation and must infer a feasible motion plan leading directly to the destination. Targets are either explicitly defined by a red marker or implicitly specified via object-class descriptions, necessitating a synthesis of geometric reasoning and semantic recognition. A successful execution by Veo-3 is illustrated in \Cref{fig:lastmile-nav}. Ultimately, this task probes a model’s capacity to interpret egocentric $360^\circ$ spatial layouts, estimate relative pose and depth, and execute precise actions to bridge perception and control in the final meters of navigation.

\subsection{Evaluation and Metrics}

Evaluating navigation-conditioned video generation presents a dual challenge: quantifying geometric precision while ensuring visual-semantic fidelity. Our framework explicitly assesses whether the agent executes plausible motion, maintains structural consistency, and preserves the semantic integrity of the destination. All metrics are binary (pass/fail) and come from the rollout traces and generated frames.

We adopt the shared embodied evaluation protocol in Section~\ref{sec:embodied_eval_metrics}: all metrics are binary and labeled via the automatic VLM/human rules on the rollout traces and generated frames. The Physical Understanding and Instruction Following checks (Object Semantic, Agent Consistency, Spatial Alignment, Destination Integrity, Scene Consistency) are unchanged; here we only note how task completeness metrics are used for Panoramic Last-Mile Navigation and how the gated scores are formed.


\subsubsection{Task Completeness Metrics (Geometry Only)}

These metrics isolate navigational accuracy, evaluating whether the agent reaches the destination based purely on \textbf{geometric coordinates}. Visual fidelity and physical consistency are excluded here and assessed by subsequent metrics.

\begin{itemize}
    \item \textbf{Success Score 3D (S.S. 3D).} Measures whether the final state of the agent coincides with the target location. This metric is strictly geometric; it is satisfied if the agent terminates its trajectory within the defined volume of the destination, regardless of visual preservation.
    \item \textbf{Oracle Success Score 3D (O.S. 3D).} A relaxed variation of Success Score 3D that credits the agent for reaching the target at any point during the trajectory. The metric passes if the agent enters the destination's proximity threshold at any timestep, irrespective of where it ultimately stops.
\end{itemize}

\subsubsection{Gate Metrics and Holistic Performance}

To ensure a rigorous assessment, we introduce composite ``Gate Metrics'' that intersect geometric success with semantic and physical validity. These metrics serve as strict filters, disqualifying trajectories that achieve goal conditions through generative artifacts such as hallucination or geometry warping.

\begin{itemize}
    \item \textbf{Success (3D) with Original Destination.} This metric imposes a strict visual-geometric consistency check. It counters the tendency of generative models to exploit scene manipulation—such as fabricating a target at the agent's feet or warping the room layout—to satisfy stopping conditions artificially. A trajectory is successful under this metric only if it satisfies the \textbf{Success Score 3D (S.S. 3D)} while simultaneously passing the \textbf{Destination Integrity (Des. Inte.)} and \textbf{Scene Consistency (Scene Con.)} checks.
    \item \textbf{Physics Validity.} A holistic measure of physical plausibility, ensuring that movement is grounded in reality rather than dream-like transitions. Geometric success is disregarded if the agent violates fundamental physical laws. This metric requires the simultaneous satisfaction of three conditions: (1) \textbf{Object Semantic Score (Obj. Sem.)} (no collisions); (2) \textbf{Agent Consistency Score (Agent Con.)} (identity persistence); and (3) \textbf{Spatial Alignment Score (Spa. Ali.)} (pose coherence).
    \item \textbf{Overall Success} = Success Score 3D $\land$ Oracle Success Score 3D $\land$ Object Semantic $\land$ Agent Consistency $\land$ Spatial Alignment $\land$ Destination Integrity $\land$ Scene Consistency; a sample passes only when all seven binary checks are $1$. This strict criterion highlights that strong performance on individual metrics does not guarantee successful, plausible end-to-end task completion.

\end{itemize}

\subsection{Evaluation Results}
\label{sec:embodied_pano_analysis}


\textbf{Key Finding: Model Capabilities \& The Evaluation Gap.}
\begin{itemize}
    \item \textbf{Performance Stratification:} In the selected diagnostic subset, results are sharply stratified: \textbf{Veo-3} and \textbf{Nano-banana} function, while Sora-2 and GPT-4o fail completely. Surprisingly, the image-based Nano-banana often outperforms Veo-3 in complex environments and instruction following.
    \item \textbf{Auto-Metric Failure:} A massive reliability gap exists between evaluation methods. Auto-metrics rate Veo-3 at \textbf{73.33\%} success, while humans rate it at only \textbf{25.00\%}. Auto-evaluators successfully track camera motion but fail to penalize physical hallucinations and logic violations.
\end{itemize}

\subsubsection{VLM-Based Evaluation}

The diagnostic subset results in \Cref{tab:embodied_task01_models_results} reveal a distinct stratification in model capabilities for the Panoramic View Last-Mile Navigation task. Within this selected model set, the performance landscape is strictly binary: \textbf{Veo-3} and \textbf{Nano-banana} demonstrate functional navigation capabilities, while \textbf{Sora-2} and \textbf{GPT-4o-image} fail to complete the task entirely (0.00\% Overall Success).

\paragraph{Top-Tier Performance (Veo-3 vs. Nano-banana).} While Veo-3 is the primary video model of interest, Nano-banana serves as a surprisingly robust baseline, frequently outperforming the video model in complex scenarios.
\begin{itemize}
    \item \textbf{Environmental Robustness:} In simple environments (\textit{floor01}), both models achieve parity with a 73.33\% Overall Success rate. However, as environmental complexity increases (\textit{floor02plus}), Veo-3's performance degrades sharply to 46.67\%, whereas Nano-banana maintains a high success rate of 75.00\%.
    \item \textbf{Visual Fidelity Scaling:} A notable divergence occurs in how the models handle visual quality. Nano-banana exhibits positive scaling with fidelity, improving from 60.00\% success at \textit{quality03} to 82.50\% at \textit{quality05}. In contrast, Veo-3 plateaus, peaking at only 62.50\% regardless of increased visual fidelity.
\end{itemize}

\paragraph{Instruction Following Capabilities.} A critical differentiator found in the results is the response to destination specifications.
\begin{itemize}
    \item \textbf{Visual vs. Textual Guidance:} When the destination is specified by a visual marker (\textit{color mark}), both models perform comparably ($\approx$70\%).
    \item \textbf{Descriptive Navigation:} When the destination is defined by text instructions (\textit{location description}), Veo-3 struggles, dropping to 50.00\% success. Conversely, Nano-banana excels, achieving its highest success rate of 78.33\%. This suggests that while Veo-3 offers superior temporal consistency (\textit{Agent Consistency} >93\% across most settings), Nano-banana possesses superior multimodal understanding, allowing it to map textual descriptions to visual goals more effectively.
\end{itemize}

\paragraph{Navigation Failures (Sora-2 and GPT-4o-image).} The failing models illustrate two distinct modes of error.
\begin{itemize}
    \item \textbf{Static vs. Dynamic Failure:} GPT-4o-image achieves high Object Semantic Scores (up to 100\%) but fails to generate the temporal trajectory required for navigation, resulting in a 0.00\% success rate.
    \item \textbf{Steerability Failure:} Sora-2 presents a ``steerability'' problem. While it maintains decent Physical Understanding (\textit{e.g.}, 88.33\% Object Semantic Score in \textit{floor01}) and generates coherent video, it lacks \textit{Destination Integrity} (0.00\%). The model hallucinates plausible scenes but cannot be constrained to a specific coordinate or target path.
\end{itemize}

\paragraph{Impact of Trajectory Distance.} As anticipated, performance for the leading video model, Veo-3, degrades with trajectory length, dropping from 66.67\% (Short) to 53.33\% (Long). Nano-banana remains more resilient to distance, maintaining a 66.67\% success rate even on long trajectories, further validating the robustness of image-based stepwise generation over holistic video generation for this specific navigation benchmark.

\begin{table*}[htbp]
\centering
\small
\caption{Quantitative results for the \textbf{Panoramic View Last-Mile Navigation} benchmark. We compare performance across two video generative models (Veo-3 and Sora-2) and two image generative models (Nano-banana, and GPT-4o-image).}
\label{tab:embodied_task01_models_results}
\resizebox{0.95\linewidth}{!}{
\begin{tabular}{@{}lcccccccccc@{}}
\toprule
 & \multicolumn{2}{c}{\textbf{Task Completeness}} & \multicolumn{3}{c}{\textbf{Physical Understanding}} & \multicolumn{2}{c}{\textbf{Instruction Following}} & \multicolumn{2}{c}{\textbf{Gate Metric}} & \multicolumn{1}{c}{\textbf{Holistic Metric}} \\
\cmidrule(lr){2-3} \cmidrule(lr){4-6} \cmidrule(lr){7-8} \cmidrule(lr){9-10} \cmidrule(lr){11-11}
\textbf{Model} & \begin{tabular}[c]{@{}c@{}}\textbf{Success} \\ \textbf{Score} \\\textbf{(3D)}\end{tabular} & \begin{tabular}[c]{@{}c@{}}\textbf{Oracle} \\ \textbf{Success}\\ \textbf{Score (3D)}\end{tabular} & \begin{tabular}[c]{@{}c@{}}\textbf{Object} \\ \textbf{Semantic} \\ \textbf{Score}\end{tabular} & \begin{tabular}[c]{@{}c@{}}\textbf{Agent} \\ \textbf{Consistency} \\ \textbf{Score}\end{tabular} & \begin{tabular}[c]{@{}c@{}}\textbf{Spatial} \\ \textbf{Alignment} \\ \textbf{Score}\end{tabular} & \begin{tabular}[c]{@{}c@{}}\textbf{Destination} \\ \textbf{Integrity} \\ \textbf{Score}\end{tabular} & \begin{tabular}[c]{@{}c@{}}\textbf{Scene} \\ \textbf{Consistency} \\ \textbf{Score}\end{tabular} & \begin{tabular}[c]{@{}c@{}}\textbf{Success (3D)} \\ \textbf{Original} \\ \textbf{Destination}\end{tabular} & \begin{tabular}[c]{@{}c@{}}\textbf{Physics} \\ \textbf{Validness}\end{tabular} & \begin{tabular}[c]{@{}c@{}}\textbf{Overall} \\ \textbf{Success}\end{tabular} \\
\midrule
\multicolumn{11}{@{}l}{\textbf{Environmental Complexity}} \\
\multicolumn{11}{@{}l}{\quad \textit{Level: floor01}} \\
\multicolumn{11}{@{}l}{\quad \textbf{Video Models}} \\
\quad \quad Veo-3 & 90.00\% & 90.00\% & 93.33\% & 93.33\% & 93.33\% & 90.00\% & 98.33\% & 90.00\% & 81.67\% & \textbf{73.33\%} \\
\quad \quad Sora-2 & 0.00\% & 1.67\% & 93.33\% & 88.33\% & 93.33\% & 0.00\% & 0.00\% & 0.00\% & 81.67\% & 0.00\% \\
\multicolumn{11}{@{}l}{\quad \textbf{Image Models}} \\
\quad \quad Nano-banana & 76.67\% & 80.00\% & 96.67\% & 88.33\% & 88.33\% & 80.00\% & 91.67\% & 75.00\% & 88.33\% & \textbf{73.33\%} \\
\quad \quad GPT-4o-image & 0.00\% & 0.00\% & 100.00\% & 1.67\% & 0.00\% & 0.00\% & 0.00\% & 0.00\% & 0.00\% & 0.00\% \\
\multicolumn{11}{@{}l}{\quad \textit{Level: floor02plus}} \\
\multicolumn{11}{@{}l}{\quad \textbf{Video Models}} \\
\quad \quad Veo-3 & 58.33\% & 66.67\% & 95.00\% & 93.33\% & 91.67\% & 55.00\% & 91.67\% & 55.00\% & 83.33\% & \textbf{46.67\%} \\
\quad \quad Sora-2 & 0.00\% & 0.00\% & 81.36\% & 77.97\% & 76.27\% & 0.00\% & 1.69\% & 0.00\% & 64.41\% & 0.00\% \\
\multicolumn{11}{@{}l}{\quad \textbf{Image Models}} \\
\quad \quad Nano-banana & 80.00\% & 80.00\% & 96.67\% & 90.00\% & 86.67\% & 80.00\% & 90.00\% & 78.33\% & 86.67\% & \textbf{75.00\%} \\
\quad \quad GPT-4o-image & 0.00\% & 0.00\% & 91.67\% & 0.00\% & 0.00\% & 0.00\% & 0.00\% & 0.00\% & 0.00\% & 0.00\% \\
\multicolumn{11}{@{}l}{\textbf{View Fidelity}} \\
\multicolumn{11}{@{}l}{\quad \textit{Level: quality03}} \\
\multicolumn{11}{@{}l}{\quad \textbf{Video Models}} \\
\quad \quad Veo-3 & 72.50\% & 80.00\% & 92.50\% & 92.50\% & 90.00\% & 70.00\% & 92.50\% & 70.00\% & 77.50\% & \textbf{55.00\%} \\
\quad \quad Sora-2 & 0.00\% & 2.56\% & 79.49\% & 82.05\% & 76.92\% & 0.00\% & 2.56\% & 0.00\% & 69.23\% & 0.00\% \\
\multicolumn{11}{@{}l}{\quad \textbf{Image Models}} \\
\quad \quad Nano-banana & 62.50\% & 62.50\% & 92.50\% & 87.50\% & 82.50\% & 62.50\% & 80.00\% & 62.50\% & 82.50\% & \textbf{60.00\%} \\
\quad \quad GPT-4o-image & 0.00\% & 0.00\% & 90.00\% & 0.00\% & 0.00\% & 0.00\% & 0.00\% & 0.00\% & 0.00\% & 0.00\% \\
\multicolumn{11}{@{}l}{\quad \textit{Level: quality04}} \\
\multicolumn{11}{@{}l}{\quad \textbf{Video Models}} \\
\quad \quad Veo-3 & 75.00\% & 75.00\% & 95.00\% & 95.00\% & 90.00\% & 75.00\% & 97.50\% & 75.00\% & 82.50\% & \textbf{62.50\%} \\
\quad \quad Sora-2 & 0.00\% & 0.00\% & 85.00\% & 80.00\% & 87.50\% & 0.00\% & 0.00\% & 0.00\% & 67.50\% & 0.00\% \\
\multicolumn{11}{@{}l}{\quad \textbf{Image Models}} \\
\quad \quad Nano-banana & 85.00\% & 90.00\% & 100.00\% & 87.50\% & 87.50\% & 90.00\% & 92.50\% & 80.00\% & 87.50\% & \textbf{80.00\%} \\
\quad \quad GPT-4o-image & 0.00\% & 0.00\% & 100.00\% & 0.00\% & 0.00\% & 0.00\% & 0.00\% & 0.00\% & 0.00\% & 0.00\% \\
\multicolumn{11}{@{}l}{\quad \textit{Level: quality05}} \\
\multicolumn{11}{@{}l}{\quad \textbf{Video Models}} \\
\quad \quad Veo-3 & 75.00\% & 80.00\% & 95.00\% & 92.50\% & 97.50\% & 72.50\% & 95.00\% & 72.50\% & 87.50\% & \textbf{62.50\%} \\
\quad \quad Sora-2 & 0.00\% & 0.00\% & 97.50\% & 87.50\% & 90.00\% & 0.00\% & 0.00\% & 0.00\% & 82.50\% & 0.00\% \\
\multicolumn{11}{@{}l}{\quad \textbf{Image Models}} \\
\quad \quad Nano-banana & 87.50\% & 87.50\% & 97.50\% & 92.50\% & 92.50\% & 87.50\% & 100.00\% & 87.50\% & 92.50\% & \textbf{82.50\%} \\
\quad \quad GPT-4o-image & 0.00\% & 0.00\% & 97.50\% & 2.50\% & 0.00\% & 0.00\% & 0.00\% & 0.00\% & 0.00\% & 0.00\% \\
\multicolumn{11}{@{}l}{\textbf{Trajectory Distance}} \\
\multicolumn{11}{@{}l}{\quad \textit{Level: short}} \\
\multicolumn{11}{@{}l}{\quad \textbf{Video Models}} \\
\quad \quad Veo-3 & 81.67\% & 83.33\% & 96.67\% & 93.33\% & 90.00\% & 80.00\% & 98.33\% & 80.00\% & 83.33\% & \textbf{66.67\%} \\
\quad \quad Sora-2 & 0.00\% & 0.00\% & 88.33\% & 83.33\% & 85.00\% & 0.00\% & 1.67\% & 0.00\% & 76.67\% & 0.00\% \\
\multicolumn{11}{@{}l}{\quad \textbf{Image Models}} \\
\quad \quad Nano-banana & 86.67\% & 86.67\% & 98.33\% & 93.33\% & 91.67\% & 86.67\% & 95.00\% & 85.00\% & 91.67\% & \textbf{81.67\%} \\
\quad \quad GPT-4o-image & 0.00\% & 0.00\% & 95.00\% & 1.67\% & 0.00\% & 0.00\% & 0.00\% & 0.00\% & 0.00\% & 0.00\% \\
\multicolumn{11}{@{}l}{\quad \textit{Level: long}} \\
\multicolumn{11}{@{}l}{\quad \textbf{Video Models}} \\
\quad \quad Veo-3 & 66.67\% & 73.33\% & 91.67\% & 93.33\% & 95.00\% & 65.00\% & 91.67\% & 65.00\% & 81.67\% & \textbf{53.33\%} \\
\quad \quad Sora-2 & 0.00\% & 1.69\% & 86.44\% & 83.05\% & 84.75\% & 0.00\% & 0.00\% & 0.00\% & 69.49\% & 0.00\% \\
\multicolumn{11}{@{}l}{\quad \textbf{Image Models}} \\
\quad \quad Nano-banana & 70.00\% & 73.33\% & 95.00\% & 85.00\% & 83.33\% & 73.33\% & 86.67\% & 68.33\% & 83.33\% & \textbf{66.67\%} \\
\quad \quad GPT-4o-image & 0.00\% & 0.00\% & 96.67\% & 0.00\% & 0.00\% & 0.00\% & 0.00\% & 0.00\% & 0.00\% & 0.00\% \\
\multicolumn{11}{@{}l}{\textbf{Destination Specification}} \\
\multicolumn{11}{@{}l}{\quad \textit{Level: color mark}} \\
\multicolumn{11}{@{}l}{\quad \textbf{Video Models}} \\
\quad \quad Veo-3 & 83.33\% & 88.33\% & 96.67\% & 93.33\% & 93.33\% & 80.00\% & 93.33\% & 80.00\% & 85.00\% & \textbf{70.00\%} \\
\quad \quad Sora-2 & 0.00\% & 1.69\% & 86.44\% & 77.97\% & 81.36\% & 0.00\% & 0.00\% & 0.00\% & 67.80\% & 0.00\% \\
\multicolumn{11}{@{}l}{\quad \textbf{Image Models}} \\
\quad \quad Nano-banana & 76.67\% & 80.00\% & 95.00\% & 85.00\% & 81.67\% & 80.00\% & 86.67\% & 73.33\% & 81.67\% & \textbf{70.00\%} \\
\quad \quad GPT-4o-image & 0.00\% & 0.00\% & 93.33\% & 0.00\% & 0.00\% & 0.00\% & 0.00\% & 0.00\% & 0.00\% & 0.00\% \\
\multicolumn{11}{@{}l}{\quad \textit{Level: location description}} \\
\multicolumn{11}{@{}l}{\quad \textbf{Video Models}} \\
\quad \quad Veo-3 & 65.00\% & 68.33\% & 91.67\% & 93.33\% & 91.67\% & 65.00\% & 96.67\% & 65.00\% & 80.00\% & \textbf{50.00\%} \\
\quad \quad Sora-2 & 0.00\% & 0.00\% & 88.33\% & 88.33\% & 88.33\% & 0.00\% & 1.67\% & 0.00\% & 78.33\% & 0.00\% \\
\multicolumn{11}{@{}l}{\quad \textbf{Image Models}} \\
\quad \quad Nano-banana & 80.00\% & 80.00\% & 98.33\% & 93.33\% & 93.33\% & 80.00\% & 95.00\% & 80.00\% & 93.33\% & \textbf{78.33\%} \\
\quad \quad GPT-4o-image & 0.00\% & 0.00\% & 98.33\% & 1.67\% & 0.00\% & 0.00\% & 0.00\% & 0.00\% & 0.00\% & 0.00\% \\
\bottomrule
\end{tabular} }
\end{table*}

\subsubsection{Geo-Align Evaluation}
\begin{table*}[htbp]
\centering
\small
\caption{The quantitative results reveal a distinct stratification in model capabilities for the \textbf{Panoramic View Last-Mile Navigation} task. The performance landscape is strictly competitive: \textbf{Veo-3} demonstrates superior navigation fidelity across all environments, while \textbf{Sora-2} and \textbf{Wan-2.2} show significant degradation in success rates and path efficiency (\textbf{SPL}) as environmental complexity increases.}
\label{tab:vggt_task01_results}
\resizebox{0.8\linewidth}{!}{
\begin{tabular}{@{}lcccccccccc@{}}
\toprule
 & \multicolumn{2}{c}{\textbf{Task Success}} & \multicolumn{2}{c}{\textbf{Path Fidelity}} & \multicolumn{2}{c}{\textbf{Trajectory Error}} & \multicolumn{2}{c}{\textbf{Path Length}} & \multicolumn{2}{c}{\textbf{Navigation Error}} \\
\cmidrule(lr){2-3} \cmidrule(lr){4-5} \cmidrule(lr){6-7} \cmidrule(lr){8-9} \cmidrule(lr){10-11}
\textbf{Model} & \textbf{SR} & \textbf{SPL} & \textbf{nDTW} & \textbf{sDTW} & \textbf{ATE} & \textbf{RPE} & \textbf{Pred} & \textbf{GT} & \textbf{Nav} & \textbf{Dir} \\
\midrule
\multicolumn{11}{@{}l}{\textbf{Environmental Complexity}} \\
\multicolumn{11}{@{}l}{\quad \textit{Level: floor01}} \\
\quad \quad Veo-3 & $83.3\%$ & $50.6\%$ & $53.9\%$ & $48.7\%$ & $0.91$ & $0.37$ & $6.80$ & $3.56$ & $0.66$ & $11.18$ \\
\quad \quad Sora-2 & $55.0\%$ & $26.5\%$ & $41.7\%$ & $29.0\%$ & $1.70$ & $0.60$ & $11.43$ & $3.56$ & $1.40$ & $14.02$ \\
\quad \quad Wan-2.2 & $53.3\%$ & $31.7\%$ & $44.5\%$ & $32.6\%$ & $2.65$ & $0.77$ & $16.48$ & $3.56$ & $1.46$ & $24.06$ \\
\multicolumn{11}{@{}l}{\quad \textit{Level: floor02plus}} \\
\quad \quad Veo-3 & $50.0\%$ & $32.3\%$ & $41.1\%$ & $28.7\%$ & $1.46$ & $0.48$ & $8.61$ & $3.75$ & $1.60$ & $19.47$ \\
\quad \quad Sora-2 & $25.0\%$ & $9.9\%$ & $32.0\%$ & $13.7\%$ & $1.85$ & $0.59$ & $10.90$ & $3.75$ & $2.00$ & $17.13$ \\
\quad \quad Wan-2.2 & $33.9\%$ & $14.9\%$ & $34.5\%$ & $17.7\%$ & $1.74$ & $0.60$ & $11.00$ & $3.75$ & $1.61$ & $16.61$ \\
\midrule
\multicolumn{11}{@{}l}{\textbf{View Fidelity}} \\
\multicolumn{11}{@{}l}{\quad \textit{Level: quality03}} \\
\quad \quad Veo-3 & $62.5\%$ & $42.5\%$ & $46.0\%$ & $35.3\%$ & $1.20$ & $0.37$ & $7.98$ & $4.43$ & $1.10$ & $12.39$ \\
\quad \quad Sora-2 & $22.5\%$ & $10.7\%$ & $32.2\%$ & $11.8\%$ & $2.09$ & $0.57$ & $12.60$ & $4.43$ & $2.14$ & $16.71$ \\
\quad \quad Wan-2.2 & $37.5\%$ & $22.5\%$ & $38.3\%$ & $21.5\%$ & $3.41$ & $0.78$ & $19.82$ & $4.43$ & $1.68$ & $17.28$ \\
\multicolumn{11}{@{}l}{\quad \textit{Level: quality05}} \\
\quad \quad Veo-3 & $77.5\%$ & $46.8\%$ & $53.2\%$ & $47.2\%$ & $0.93$ & $0.39$ & $6.53$ & $3.08$ & $0.83$ & $14.28$ \\
\quad \quad Sora-2 & $50.0\%$ & $23.0\%$ & $40.8\%$ & $28.1\%$ & $1.64$ & $0.63$ & $10.98$ & $3.08$ & $1.59$ & $14.80$ \\
\quad \quad Wan-2.2 & $52.5\%$ & $29.2\%$ & $45.0\%$ & $31.8\%$ & $1.41$ & $0.53$ & $9.01$ & $3.08$ & $1.27$ & $19.79$ \\
\midrule
\multicolumn{11}{@{}l}{\textbf{Trajectory Complexity}} \\
\multicolumn{11}{@{}l}{\quad \textit{Level: noturn}} \\
\quad \quad Veo-3 & $67.2\%$ & $37.5\%$ & $49.0\%$ & $39.4\%$ & $1.06$ & $0.46$ & $6.83$ & $2.84$ & $0.91$ & $14.98$ \\
\quad \quad Sora-2 & $48.3\%$ & $21.4\%$ & $40.8\%$ & $26.9\%$ & $1.50$ & $0.57$ & $9.06$ & $2.84$ & $1.48$ & $16.13$ \\
\quad \quad Wan-2.2 & $53.4\%$ & $28.2\%$ & $43.9\%$ & $31.4\%$ & $1.31$ & $0.57$ & $8.54$ & $2.84$ & $1.25$ & $18.55$ \\
\multicolumn{11}{@{}l}{\quad \textit{Level: oneturn}} \\
\quad \quad Veo-3 & $67.2\%$ & $46.1\%$ & $46.4\%$ & $38.8\%$ & $1.29$ & $0.38$ & $8.52$ & $4.47$ & $1.32$ & $15.38$ \\
\quad \quad Sora-2 & $32.8\%$ & $15.6\%$ & $33.3\%$ & $16.4\%$ & $2.04$ & $0.61$ & $13.28$ & $4.47$ & $1.91$ & $14.92$ \\
\quad \quad Wan-2.2 & $34.5\%$ & $18.9\%$ & $35.5\%$ & $19.4\%$ & $3.11$ & $0.81$ & $19.13$ & $4.47$ & $1.81$ & $22.38$ \\
\midrule
\multicolumn{11}{@{}l}{\textbf{Destination Specification}} \\
\multicolumn{11}{@{}l}{\quad \textit{Level: color}} \\
\quad \quad Veo-3 & $63.8\%$ & $39.3\%$ & $47.9\%$ & $37.5\%$ & $1.08$ & $0.41$ & $7.40$ & $3.65$ & $1.11$ & $16.17$ \\
\quad \quad Sora-2 & $31.0\%$ & $13.9\%$ & $34.1\%$ & $17.1\%$ & $1.89$ & $0.66$ & $11.93$ & $3.65$ & $1.89$ & $18.92$ \\
\quad \quad Wan-2.2 & $37.9\%$ & $19.3\%$ & $36.4\%$ & $21.8\%$ & $2.93$ & $0.85$ & $17.64$ & $3.65$ & $1.84$ & $26.49$ \\
\multicolumn{11}{@{}l}{\quad \textit{Level: object}} \\
\quad \quad Veo-3 & $70.7\%$ & $44.3\%$ & $47.4\%$ & $40.7\%$ & $1.26$ & $0.43$ & $7.95$ & $3.65$ & $1.12$ & $14.19$ \\
\quad \quad Sora-2 & $50.0\%$ & $23.1\%$ & $39.9\%$ & $26.2\%$ & $1.65$ & $0.52$ & $10.42$ & $3.65$ & $1.50$ & $12.12$ \\
\quad \quad Wan-2.2 & $50.0\%$ & $27.9\%$ & $43.0\%$ & $29.0\%$ & $1.49$ & $0.53$ & $10.02$ & $3.65$ & $1.22$ & $14.44$ \\
\bottomrule
\end{tabular}
}
\end{table*}

\begin{figure*}[htbp]
    \centering
    \includegraphics[width=0.8\textwidth]{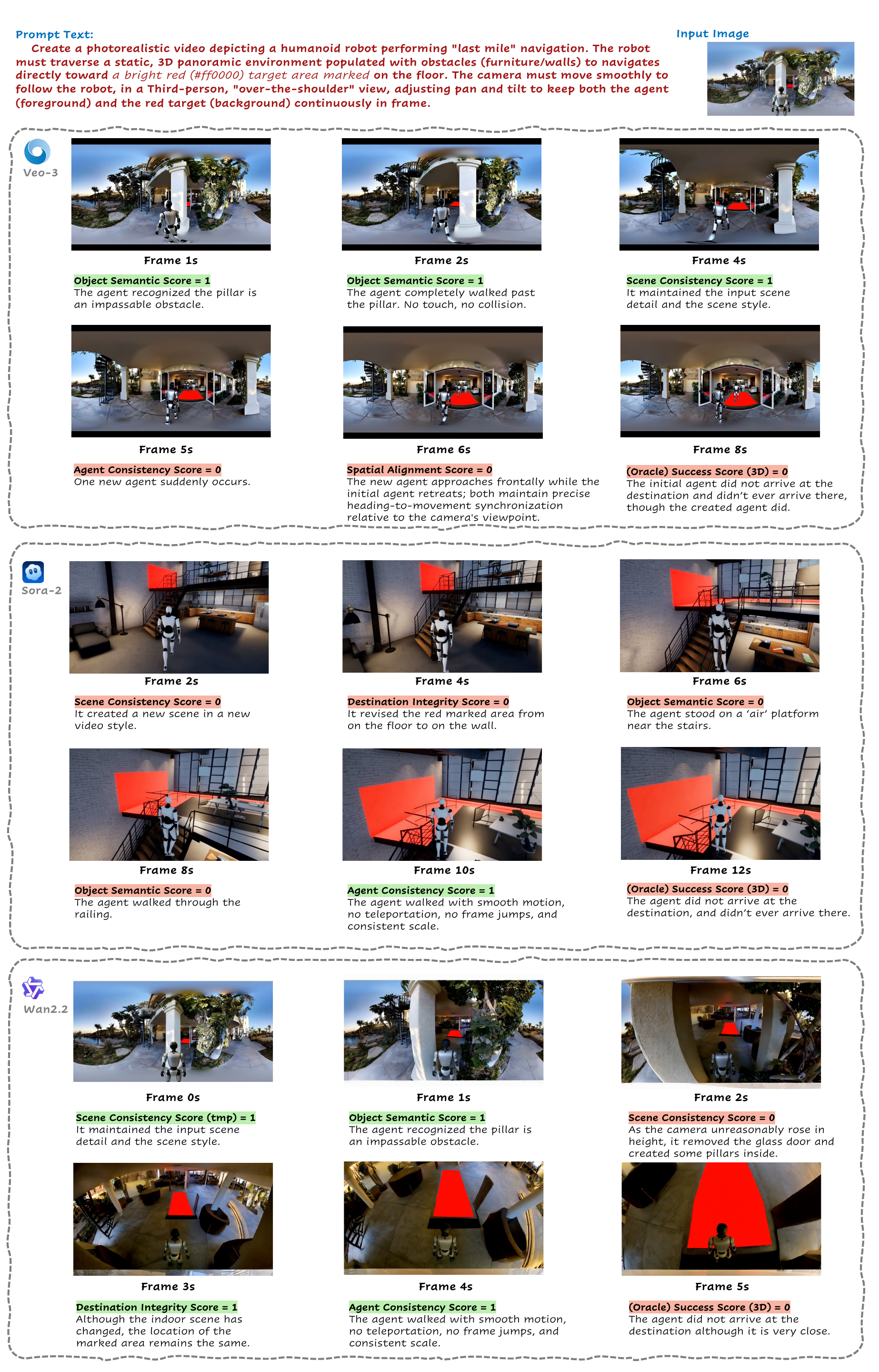}
    \vspace{-2mm}
    \caption{\textbf{Qualitative Comparison on Panoramic View Last-Mile Navigation.}
    We analyze three models (Veo-3, Sora-2, and Wan-2.2) navigating toward a red target.  The figure highlights distinct failure modes: \textit{Veo-3} suffers from agent inconsistency (hallucinating a second agent), \textit{Sora-2} exhibits severe geometric and physical violations (style shift, clipping through railings), and \textit{Wan-2.2} struggles with scene stability (altering structural elements like doors and pillars) despite smooth agent motion.}
    \label{fig:embodied_last_mile_navigation}
    \vspace{-3mm}
\end{figure*}

\subsection{Case Study: Qualitative Failure Modes} \label{sec:embodied_case_study}

This case study, visualized in \Cref{fig:embodied_last_mile_navigation}, evaluates the ``world modeling'' capabilities of video generation models during a last-mile navigation task. In this scenario, a humanoid robot must traverse a 3D environment to reach a specific target marked on the floor. The analysis reveals that while models can generate plausible motion, they suffer from distinct breakdowns in object permanence, physical laws, and instruction grounding.

\paragraph{Veo-3 (Temporal Identity Collapse).} Initially, Veo-3 demonstrates strong spatial reasoning. The agent successfully perceives a pillar as a solid obstacle and navigates around it, exhibiting valid collision avoidance. However, the generation fails critically in \textit{Agent Consistency} at the 5-second mark. The model suffers from a ``doppelgänger'' hallucination, where a second agent spontaneously manifests and approaches the camera while the initial agent retreats. This loss of agent identity results in an \textit{Oracle Success Score} of 0, as the intended actor never completes the trajectory.

\paragraph{Sora-2 (Semantic and Physical Drift).} Sora-2 struggles to maintain the fundamental constraints of the scene. By Frame 2s, it exhibits severe semantic drift, abruptly shifting the visual style and ungrounding the instruction: the model incorrectly remaps the red target marker from the floor to a vertical wall. Furthermore, the physics engine collapses; the agent is observed floating on a non-existent ``air platform'' and subsequently clipping through a solid railing. This confirms that Sora-2 prioritizes visual fluidity over collision geometry or logical consistency.\looseness=-1

\paragraph{Wan-2.2 (Environmental Instability).} In contrast to the others, Wan-2.2 maintains high \textit{Agent Consistency}, producing smooth, continuous motion without teleportation or scaling artifacts. However, it fails to maintain the ``stage'' around the actor. As the camera angle shifts (Frame 2s), the model suffers from background object permanence failures, inexplicably erasing a glass door and hallucinating new pillars within the room. While it preserves \textit{Destination Integrity} better than Sora-2 (keeping the target relative to the agent), the volatility of the environment ultimately prevents successful navigation.

\section{Top-down View Real-World Navigation (Top-down View Nav.)}
\label{apx:topdown}

\subsection{Task Definition}

The Top-down Real-World Navigation task benchmarks a model's ability to synthesize advanced spatial reasoning with semantic understanding in complex, human-centric environments. This task requires models to interpret diverse and partially occluded top-down layouts---such as floor plans---to generate optimal trajectories. Simultaneously, it tests robust semantic grounding by demanding goal identification based on textual descriptions of varying abstraction.

\begin{figure*}[t!]
    \centering
    \includegraphics[width=\linewidth]{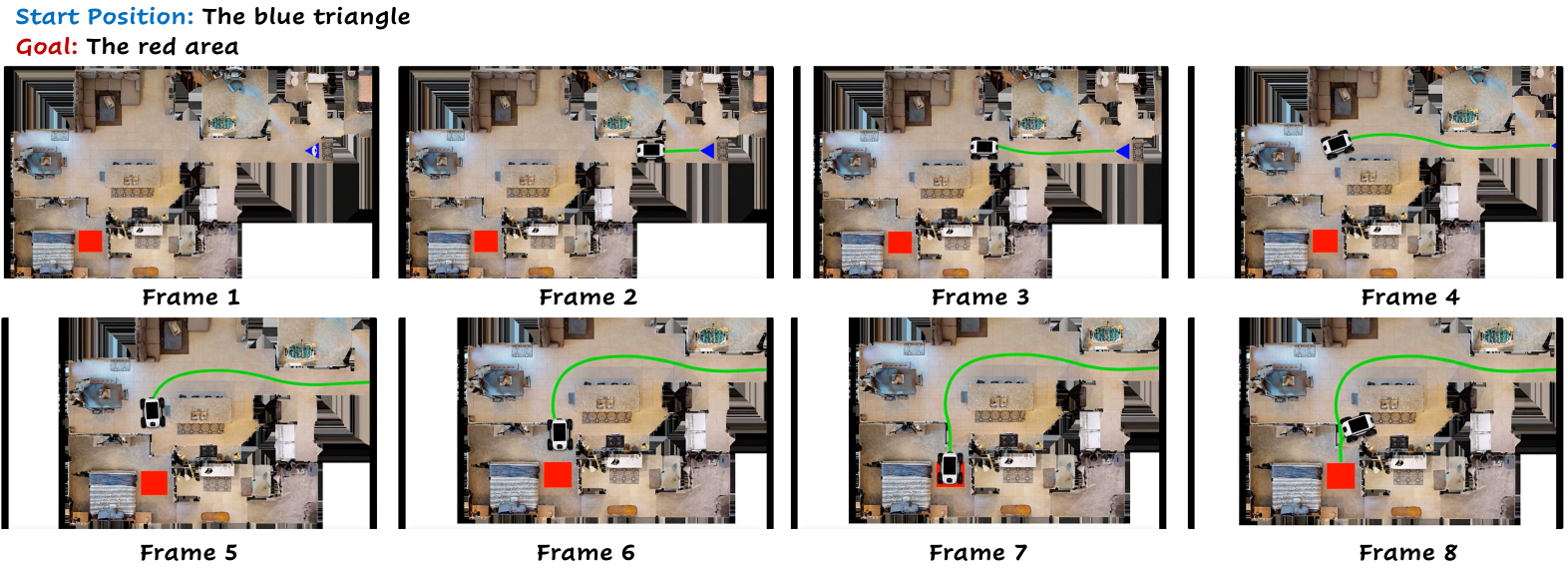}

    \caption{\textbf{Top-down View Real-World Navigation} task showing bird's-eye view navigation. The model must interpret the abstract 2D map representation and plan a path from start to goal, demonstrating 2D spatial reasoning and temporal reasoning from a global perspective. A successful case completed by Veo-3 on the Top-down View Real-World Navigation task.}

    \label{fig:topdown-nav}
\end{figure*}

\subsection{Evaluation and Metrics}

We adopt the shared embodied evaluation protocol in Section~\ref{sec:embodied_eval_metrics}: Physical Understanding and Instruction Following checks (Object Semantic, Agent Consistency, Spatial Alignment, Destination Integrity, Scene Consistency) are unchanged. Here we only note how task completeness metrics are used for Top-down View Real-World Navigation and how the gated scores are formed.


\subsubsection{Task Completeness Metrics (Geometry Only)}

These metrics isolate navigational accuracy, evaluating whether the agent reaches the destination based purely on \textbf{geometric coordinates}. Visual fidelity and physical consistency are excluded here and assessed by subsequent metrics.

\begin{itemize}
    \item \textbf{Success Score 2D (S.S. 2D).} This metric evaluates whether the agent terminates its trajectory strictly within the designated goal region in the top-down 2D view. Success is achieved if and only if the agent's final position is contained entirely within the goal footprint.
    \item \textbf{Oracle Success Score 2D (O.S. 2D).} This metric determines if the agent encounters the goal region at any point during the rollout, regardless of the stopping location. Credit is awarded if the agent's trajectory intersects with or passes adjacent to the 2D goal boundary at any timestep.
\end{itemize}

\subsubsection{Gate Metrics and Holistic Performance}


As in Panoramic Last-Mile Navigation, gates guard against generative shortcuts, but here the challenge comes from the 2D top-down view: success must hold on a static map, and Physical Validness is stricter because object recognition and agent orientation are blurrier from overhead.

\begin{itemize}
    \item \textbf{Success (2D) Original Destination.} This composite metric mitigates the tendency of video generation models to hallucinate solution shortcuts. It is satisfied if and only if the Success Score 2D, Destination Integrity Score, and Scene Consistency Score are met simultaneously. This ensures that successful arrival is valid and not the result of the model altering the scene layout or destination coordinates to simplify the task.
    \item \textbf{Physical Validity.} To enforce physical realism, this composite metric acts as a strict consistency check. It is satisfied only when the Object Semantic Score, Agent Consistency Score, and Spatial Alignment Score are simultaneously met. This precludes physical anomalies such as object clipping, teleportation, or unnatural drift.
    \item \textbf{Overall Success} = Success Score 2D $\land$ Oracle Success Score 2D $\land$ Object Semantic $\land$ Agent Consistency $\land$ Spatial Alignment $\land$ Destination Integrity $\land$ Scene Consistency; a sample passes only when all seven binary checks are $1$.

\end{itemize}

Finally, across all embodied navigation tasks, we introduce the \textbf{Overall} metric to evaluate holistic performance. A sample is considered successful only if all fine-grained evaluation metrics are satisfied. This strict criterion demonstrates that strong performance on individual sub-metrics does not guarantee successful end-to-end task completion.

\subsection{Evaluation Results}
\label{sec:embodied_2d_analysis}


\textbf{Key Finding: The Reality \& Modality Gap.}
\begin{itemize}
    \item \textbf{Physics Hallucination:} Video models like Veo-3 excel at path coherence but fail at physical realism. Automatic metrics miss this, inflating success rates (37.14\%) compared to human judgment (10.34\%) by ignoring violations like wall-clipping.
    \item \textbf{The Semantic Cliff:} Performance drops by \textbf{>3$\times$} when instructions shift from visual markers to textual descriptions. Models struggle significantly to ground abstract linguistic commands into spatial actions.
\end{itemize}

The Top-down Real-World Navigation task evaluates a model's ability to synthesize spatial planning with semantic understanding. The following analysis dissects model performance across environmental complexities, instruction modalities, and visual fidelities, followed by a critical examination of automatic metric reliability against human judgment.

\subsubsection{VLM-Based Evaluation}

\paragraph{Video vs. Image-based Navigation.} In the diagnostic subset shown in \Cref{tab:embodied_task02_models_results}, \textbf{Veo-3} is the strongest listed model for navigation success and trajectory adherence; the full final-score ranking, including newer image models, is reported in \Cref{tab:embodied_final_scores}.
\begin{itemize}
    \item \textbf{The Temporal Advantage:} In the baseline setting (\textit{floor01}), Veo-3 achieves a \textbf{Success Score (2D)} of 65.71\%, significantly surpassing the image-based Nano-banana (41.67\%) and the video-based Sora-2 (22.81\%). This advantage stems from Veo-3's superior \textbf{Spatial Alignment Score} (91.43\%), indicating that video generation models with strong temporal attention can better maintain trajectory coherence than frame-by-frame image generators.
    \item \textbf{The Semantic Trade-off:} Interestingly, while Nano-banana lags in navigation, it remains highly competitive in static recognition. It achieves an \textbf{Object Semantic Score} of 88.89\% in simple environments—surpassing Veo-3 (77.14\%) and Sora-2 (82.46\%). This suggests that current image models excel at ``what'' is in the scene (grounding), while video models excel at ``how'' to move through it (dynamics).
\end{itemize}

\paragraph{Resilience to Complexity and Noise.}
\begin{itemize}
    \item \textbf{Environmental Complexity:} Performance degrades universally as environments become more intricate. Transitioning from \textit{floor01} to \textit{floor02plus}, Veo-3's \textbf{Overall Success} drops sharply from 37.14\% to 13.89\%. However, its \textbf{Oracle Success Score} remains robust (83.33\%), implying that the model often identifies the correct path but fails to execute it without violating physical constraints (\textit{e.g.}, collisions), a hypothesis supported by the low \textbf{Physics Validness} (33.33\%).
    \item \textbf{View Fidelity:} The results indicate a positive correlation between visual quality and navigation success for video models. Veo-3's Overall Success improves from 20.83\% in lower fidelity settings (\textit{quality03}) to 29.17\% in higher fidelity (\textit{quality04}), suggesting that clearer visual cues are critical for maintaining the ``driver'' capability in video generation.
\end{itemize}

\paragraph{Instruction Following: The Modality Gap.} A distinct performance gap exists between visual and textual grounding (see \textit{Destination Specification}).
\begin{itemize}
    \item \textbf{Visual Markers:} When targets are specified via a simple ``color mark,'' Veo-3 achieves a robust \textbf{Overall Success} of 38.89\%.
    \item \textbf{Textual Descriptions:} When targets require semantic parsing (``location description''), success rates collapse. Veo-3 falls to 11.43\%, and Nano-banana drops to 8.33\%. This confirms that mapping abstract linguistic descriptions to spatial layouts remains a significant hurdle compared to direct visual matching.
\end{itemize}

\begin{table*}[htbp]
\centering
\small
\caption{Quantitative results for the \textbf{2D Top-down Navigation} benchmark. We compare Sora-2, Veo-3, Nano-banana, and GPT-4o-image across various dimensions.}
\label{tab:embodied_task02_models_results}
\resizebox{0.9\linewidth}{!}{
\begin{tabular}{@{}lccccccccc@{}}
\toprule
 & \multicolumn{2}{c}{\textbf{Task Completeness}} & \multicolumn{3}{c}{\textbf{Physical Understanding}} & \multicolumn{2}{c}{\textbf{Instruction Following}} & \multicolumn{1}{c}{\textbf{Gate Metric}} & \multicolumn{1}{c}{\textbf{Holistic Metric}} \\
\cmidrule(lr){2-3} \cmidrule(lr){4-6} \cmidrule(lr){7-8} \cmidrule(lr){9-9} \cmidrule(lr){10-10}
\textbf{Model} & \begin{tabular}[c]{@{}c@{}}\textbf{Success} \\ \textbf{Score} \\\textbf{(2D)}\end{tabular} & \begin{tabular}[c]{@{}c@{}}\textbf{Oracle} \\ \textbf{Success}\\ \textbf{Score (2D)}\end{tabular} & \begin{tabular}[c]{@{}c@{}}\textbf{Object} \\ \textbf{Semantic} \\ \textbf{Score}\end{tabular} & \begin{tabular}[c]{@{}c@{}}\textbf{Agent} \\ \textbf{Consistency} \\ \textbf{Score}\end{tabular} & \begin{tabular}[c]{@{}c@{}}\textbf{Spatial} \\ \textbf{Alignment} \\ \textbf{Score}\end{tabular} & \begin{tabular}[c]{@{}c@{}}\textbf{Destination} \\ \textbf{Integrity} \\ \textbf{Score}\end{tabular} & \begin{tabular}[c]{@{}c@{}}\textbf{Scene} \\ \textbf{Consistency} \\ \textbf{Score}\end{tabular} & \begin{tabular}[c]{@{}c@{}}\textbf{Physics} \\ \textbf{Validness}\end{tabular} & \begin{tabular}[c]{@{}c@{}}\textbf{Overall} \\ \textbf{Success}\end{tabular} \\
\midrule
\multicolumn{10}{@{}l}{\textbf{Environmental Complexity}} \\
\multicolumn{10}{@{}l}{\quad \textit{Level: floor01}} \\
\multicolumn{10}{@{}l}{\quad \textbf{Video Models}} \\
\quad \quad Veo-3 & 65.71\% & 85.71\% & 77.14\% & 62.86\% & 91.43\% & 65.71\% & 45.71\% & 54.29\% & \textbf{37.14\%} \\
\quad \quad Sora-2 & 22.81\% & 47.37\% & 82.46\% & 68.42\% & 78.95\% & 7.02\% & 8.77\% & 52.63\% & 1.75\% \\
\multicolumn{10}{@{}l}{\quad \textbf{Image Models}} \\
\quad \quad Nano-banana & 41.67\% & 52.78\% & 88.89\% & 50.00\% & 77.78\% & 33.33\% & 38.89\% & 50.00\% & \textbf{5.56\%} \\
\quad \quad GPT-4o-image & 10.34\% & 15.52\% & 55.17\% & 6.90\% & 56.90\% & 1.72\% & 1.72\% & 6.90\% & 1.72\% \\
\multicolumn{10}{@{}l}{\quad \textit{Level: floor02plus}} \\
\multicolumn{10}{@{}l}{\quad \textbf{Video Models}} \\
\quad \quad Veo-3 & 25.00\% & 83.33\% & 61.11\% & 44.44\% & 80.56\% & 33.33\% & 45.71\% & 33.33\% & \textbf{13.89\%} \\
\quad \quad Sora-2 & 28.81\% & 55.93\% & 89.83\% & 54.24\% & 86.44\% & 15.25\% & 23.73\% & 49.15\% & 5.08\% \\
\multicolumn{10}{@{}l}{\quad \textbf{Image Models}} \\
\quad \quad Nano-banana & 38.89\% & 52.78\% & 86.11\% & 55.56\% & 86.11\% & 36.11\% & 66.67\% & 50.00\% & \textbf{16.67\%} \\
\quad \quad GPT-4o-image & 16.67\% & 23.33\% & 51.67\% & 11.67\% & 41.67\% & 5.00\% & 10.00\% & 10.00\% & 5.00\% \\
\multicolumn{10}{@{}l}{\textbf{View Fidelity}} \\
\multicolumn{10}{@{}l}{\quad \textit{Level: quality03}} \\
\multicolumn{10}{@{}l}{\quad \textbf{Video Models}} \\
\quad \quad Veo-3 & 45.83\% & 79.17\% & 66.67\% & 33.33\% & 75.00\% & 50.00\% & 43.48\% & 29.17\% & \textbf{20.83\%} \\
\quad \quad Sora-2 & 34.21\% & 52.63\% & 92.11\% & 63.16\% & 81.58\% & 18.42\% & 18.42\% & 57.89\% & 7.89\% \\
\multicolumn{10}{@{}l}{\quad \textbf{Image Models}} \\
\quad \quad Nano-banana & 41.67\% & 58.33\% & 83.33\% & 45.83\% & 75.00\% & 41.67\% & 58.33\% & 37.50\% & \textbf{12.50\%} \\
\quad \quad GPT-4o-image & 21.05\% & 23.68\% & 65.79\% & 13.16\% & 57.89\% & 5.26\% & 5.26\% & 10.53\% & 5.26\% \\
\multicolumn{10}{@{}l}{\quad \textit{Level: quality04}} \\
\multicolumn{10}{@{}l}{\quad \textbf{Video Models}} \\
\quad \quad Veo-3 & 37.50\% & 83.33\% & 70.83\% & 58.33\% & 95.83\% & 41.67\% & 45.83\% & 54.17\% & \textbf{29.17\%} \\
\quad \quad Sora-2 & 23.08\% & 46.15\% & 87.18\% & 66.67\% & 89.74\% & 7.69\% & 17.95\% & 51.28\% & 2.56\% \\
\multicolumn{10}{@{}l}{\quad \textbf{Image Models}} \\
\quad \quad Nano-banana & 37.50\% & 50.00\% & 87.50\% & 50.00\% & 83.33\% & 33.33\% & 37.50\% & 50.00\% & 4.17\% \\
\quad \quad GPT-4o-image & 12.50\% & 17.50\% & 52.50\% & 12.50\% & 50.00\% & 5.00\% & 10.00\% & 12.50\% & \textbf{5.00\%} \\
\multicolumn{10}{@{}l}{\quad \textit{Level: quality05}} \\
\multicolumn{10}{@{}l}{\quad \textbf{Video Models}} \\
\quad \quad Veo-3 & 52.17\% & 91.30\% & 69.57\% & 69.57\% & 86.96\% & 56.52\% & 47.83\% & 47.83\% & \textbf{26.09\%} \\
\quad \quad Sora-2 & 20.51\% & 56.41\% & 79.49\% & 53.85\% & 76.92\% & 7.69\% & 12.82\% & 43.59\% & 0.00\% \\
\multicolumn{10}{@{}l}{\quad \textbf{Image Models}} \\
\quad \quad Nano-banana & 41.67\% & 50.00\% & 91.67\% & 62.50\% & 87.50\% & 29.17\% & 62.50\% & 62.50\% & \textbf{16.67\%} \\
\quad \quad GPT-4o-image & 7.50\% & 17.50\% & 42.50\% & 2.50\% & 40.00\% & 0.00\% & 2.50\% & 2.50\% & 0.00\% \\
\multicolumn{10}{@{}l}{\textbf{Trajectory Distance}} \\
\multicolumn{10}{@{}l}{\quad \textit{Level: noturn}} \\
\multicolumn{10}{@{}l}{\quad \textbf{Video Models}} \\
\quad \quad Veo-3 & 36.11\% & 83.33\% & 61.11\% & 52.78\% & 83.33\% & 38.89\% & 54.29\% & 38.89\% & \textbf{25.00\%} \\
\quad \quad Sora-2 & 29.31\% & 60.34\% & 89.66\% & 62.07\% & 84.48\% & 15.52\% & 18.97\% & 48.28\% & 1.72\% \\
\multicolumn{10}{@{}l}{\quad \textbf{Image Models}} \\
\quad \quad Nano-banana & 41.67\% & 52.78\% & 88.89\% & 50.00\% & 83.33\% & 36.11\% & 52.78\% & 44.44\% & \textbf{11.11\%} \\
\quad \quad GPT-4o-image & 20.34\% & 23.73\% & 57.63\% & 10.17\% & 57.63\% & 5.08\% & 8.47\% & 10.17\% & 5.08\% \\
\multicolumn{10}{@{}l}{\quad \textit{Level: oneturn}} \\
\multicolumn{10}{@{}l}{\quad \textbf{Video Models}} \\
\quad \quad Veo-3 & 54.29\% & 85.71\% & 77.14\% & 54.29\% & 88.57\% & 60.00\% & 37.14\% & 48.57\% & \textbf{25.71\%} \\
\quad \quad Sora-2 & 22.41\% & 43.10\% & 82.76\% & 60.34\% & 81.03\% & 6.90\% & 13.79\% & 53.45\% & 5.17\% \\
\multicolumn{10}{@{}l}{\quad \textbf{Image Models}} \\
\quad \quad Nano-banana & 38.89\% & 52.78\% & 86.11\% & 55.56\% & 80.56\% & 33.33\% & 52.78\% & 55.56\% & \textbf{11.11\%} \\
\quad \quad GPT-4o-image & 6.78\% & 15.25\% & 49.15\% & 8.47\% & 40.68\% & 1.69\% & 3.39\% & 6.78\% & 1.69\% \\
\multicolumn{10}{@{}l}{\textbf{Destination Specification}} \\
\multicolumn{10}{@{}l}{\quad \textit{Level: color mark}} \\
\multicolumn{10}{@{}l}{\quad \textbf{Video Models}} \\
\quad \quad Veo-3 & 52.78\% & 91.67\% & 80.56\% & 69.44\% & 91.67\% & 58.33\% & 63.89\% & 61.11\% & \textbf{38.89\%} \\
\quad \quad Sora-2 & 17.24\% & 53.45\% & 77.59\% & 56.90\% & 77.59\% & 12.07\% & 24.14\% & 44.83\% & 1.72\% \\
\multicolumn{10}{@{}l}{\quad \textbf{Image Models}} \\
\quad \quad Nano-banana & 58.33\% & 77.78\% & 91.67\% & 44.44\% & 86.11\% & 55.56\% & 55.56\% & 41.67\% & \textbf{13.89\%} \\
\quad \quad GPT-4o-image & 10.34\% & 13.79\% & 39.66\% & 3.45\% & 34.48\% & 1.72\% & 3.45\% & 3.45\% & 1.72\% \\
\multicolumn{10}{@{}l}{\quad \textit{Level: location description}} \\
\multicolumn{10}{@{}l}{\quad \textbf{Video Models}} \\
\quad \quad Veo-3 & 37.14\% & 77.14\% & 57.14\% & 37.14\% & 80.00\% & 40.00\% & 26.47\% & 25.71\% & \textbf{11.43\%} \\
\quad \quad Sora-2 & 34.48\% & 50.00\% & 94.83\% & 65.52\% & 87.93\% & 10.34\% & 8.62\% & 56.90\% & 5.17\% \\
\multicolumn{10}{@{}l}{\quad \textbf{Image Models}} \\
\quad \quad Nano-banana & 22.22\% & 27.78\% & 83.33\% & 61.11\% & 77.78\% & 13.89\% & 50.00\% & 58.33\% & \textbf{8.33\%} \\
\quad \quad GPT-4o-image & 16.67\% & 25.00\% & 66.67\% & 15.00\% & 63.33\% & 5.00\% & 8.33\% & 13.33\% & 5.00\% \\
\bottomrule
\end{tabular}
}
\end{table*}

\subsubsection{View-Syn Evaluation}

\begin{table*}[h]
  \centering
  \caption{Top-down View Real-World Navigation — Video (FVD/SSIM/PSNR/LPIPS) and Image (FID normalized); higher is better except LPIPS}
  \begin{tabular}{lcccccc}
    \toprule
    Model & Modality & FVD Norm $\uparrow$ & FID Norm $\uparrow$ & SSIM $\uparrow$ & PSNR $\uparrow$ & LPIPS $\downarrow$ \\
    \midrule
    Sora-2          & Video & 0.7079 & --     & 0.4072 & 7.21 & 0.8345 \\
    Veo-3           & Video & 0.6731 & --     & 0.1471 & 6.65 & 0.8800 \\
    Wan-2.2         & Video & 0.6285 & --     & 0.4333 & 7.81 & 0.7955 \\
    GPT-image       & Image & --     & 0.6153 & --     & --   & --     \\
    Nano-banana     & Image & --     & 0.6153 & --     & --   & --     \\
    Qwen-image      & Image & --     & 0.6003 & --     & --   & --     \\
    GPT-image-1.5   & Image & --     & 0.5247 & --     & --   & --     \\
    Nano-banana-pro & Image & --     & 0.4890 & --     & --   & --     \\
    \bottomrule
  \end{tabular}
\end{table*}

\subsection{Case Study: Qualitative Failure Modes}

\begin{figure*}[htbp]

    \centering
    \includegraphics[width=0.87\textwidth]{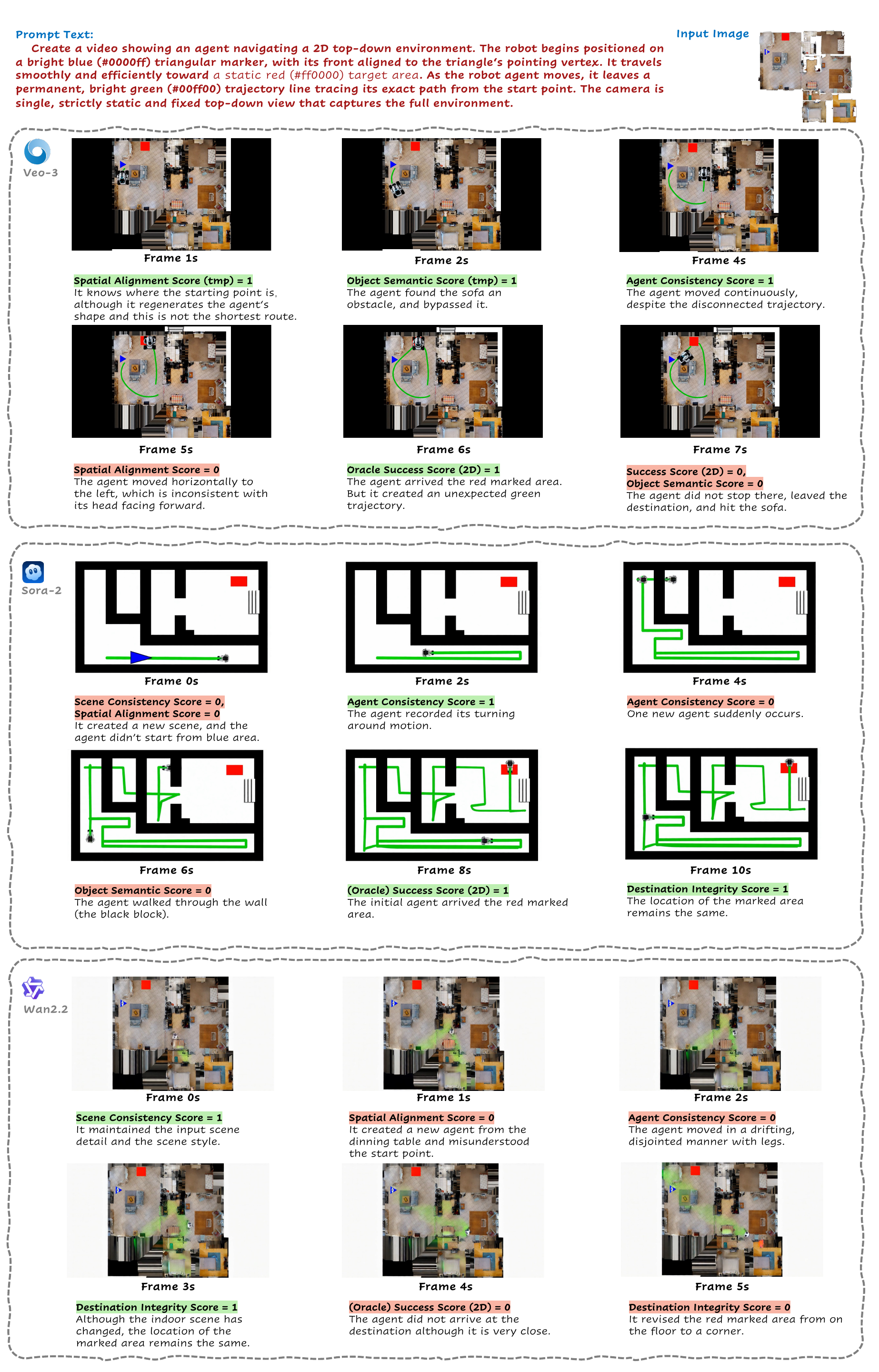}
    \caption{Qualitative Comparison on Top-down View Real-World Navigation. We analyze three models (Veo-3, Sora-2, and Wan-2.2) navigating toward a red target.}
    \label{fig:embodied_casestudy_2d}
    \vspace{-3mm}

\end{figure*}

This analysis, presented in \Cref{fig:embodied_casestudy_2d}, examines the performance of three video generation models—Veo-3, Sora-2, and Wan-2.2—on a top-down embodied navigation task. Each model is instructed to generate a video depicting an agent navigating from a blue triangular start position to a red target within a static environment, enabling a direct comparison of their spatial reasoning, trajectory consistency, and goal-directed behavior.

\textbf{Veo-3: Strong Spatial Grounding with Control and Termination Failures.} Veo-3 exhibits the strongest initial spatial grounding among the three models, correctly identifying the agent's starting position at the blue triangle and demonstrating partial semantic awareness of obstacles. In particular, the agent recognizes the sofa as a navigable obstruction and initially routes around it. However, the generation deteriorates due to failures in movement physics and action termination. The agent displays a critical orientation error, moving laterally to the left while facing forward, which results in a \textbf{Spatial Alignment Score} of 0 at Frame 5s. Although the agent briefly enters the red-marked target region (\textbf{Oracle Success Score} = 1), it fails to terminate upon arrival. Instead, the trajectory continues beyond the goal, deviates unpredictably, and ultimately collides with the sofa. This post-arrival drift causes both the \textbf{Success Score} and \textbf{Object Semantic Score} to drop to 0, revealing a failure to couple goal completion with action cessation.

\textbf{Sora-2: Severe Scene Hallucination and Physical Constraint Violations.} Sora-2 fails to preserve both the spatial ground truth and the physical constraints of the input environment. From the outset, the model reconstructs an entirely new scene in which the agent no longer originates from the blue start location. Temporal consistency further collapses as the generation introduces severe hallucinations, including the spontaneous appearance of a second agent at Frame 4s. The agent also violates fundamental physical boundaries by passing through solid obstacles (the black block), yielding an \textbf{Object Semantic Score} of 0. While an agent eventually reaches a red-marked region, the complete loss of scene integrity and environmental correspondence renders the navigation invalid with respect to the original prompt, highlighting a failure in maintaining world-state continuity.\looseness=-1

\textbf{Wan-2.2: Surface-Level Scene Consistency with Semantic Role Confusion.} Wan-2.2 preserves the visual style and low-level layout of the scene more effectively than Sora-2, achieving an initial \textbf{Scene Consistency Score} of 1. However, this apparent stability masks a deeper semantic failure. The model misidentifies the controllable agent, erroneously animating a dining table as the acting entity, which immediately induces spatial misalignment. The resulting motion is disjointed and drifting, lacking coherent locomotion or intentional control. Compounding this issue, the model fails to maintain destination integrity: the red target region shifts from its original floor location to a corner of the scene, causing the \textbf{Destination Integrity Score} to drop to 0. As a result, the agent never reaches the correct target, demonstrating that visual fidelity alone is insufficient for semantically grounded navigation.

\section{3D Real-World Navigation (3D R.-W. Nav.)}
\label{apx:3dnav}

\subsection{Task Definition}

This task evaluates an agent's foundational capabilities in \textbf{3D Spatial Reasoning} and \textbf{Visual-Semantic Grounding} within real-world scanned environments~\citep{Matterport3D,HM3D,savva2019habitat, anderson2018vision, zhu2017target}. It probes the agent's ability to \textbf{interpret egocentric visual streams} and \textbf{parse complex 3D environmental geometry} using datasets such as Matterport3D, HM3D, and Habitat~\citep{Matterport3D,HM3D, chaplot2020learning}. Furthermore, the task challenges the model's sequential decision-making by requiring the generation of valid action trajectories (\textit{e.g.}, move forward, turn left) and tests its semantic reasoning by demanding the identification of goals based on varied abstract or textual descriptions.

\begin{figure*}[htbp]
    \centering
    \includegraphics[width=\linewidth]{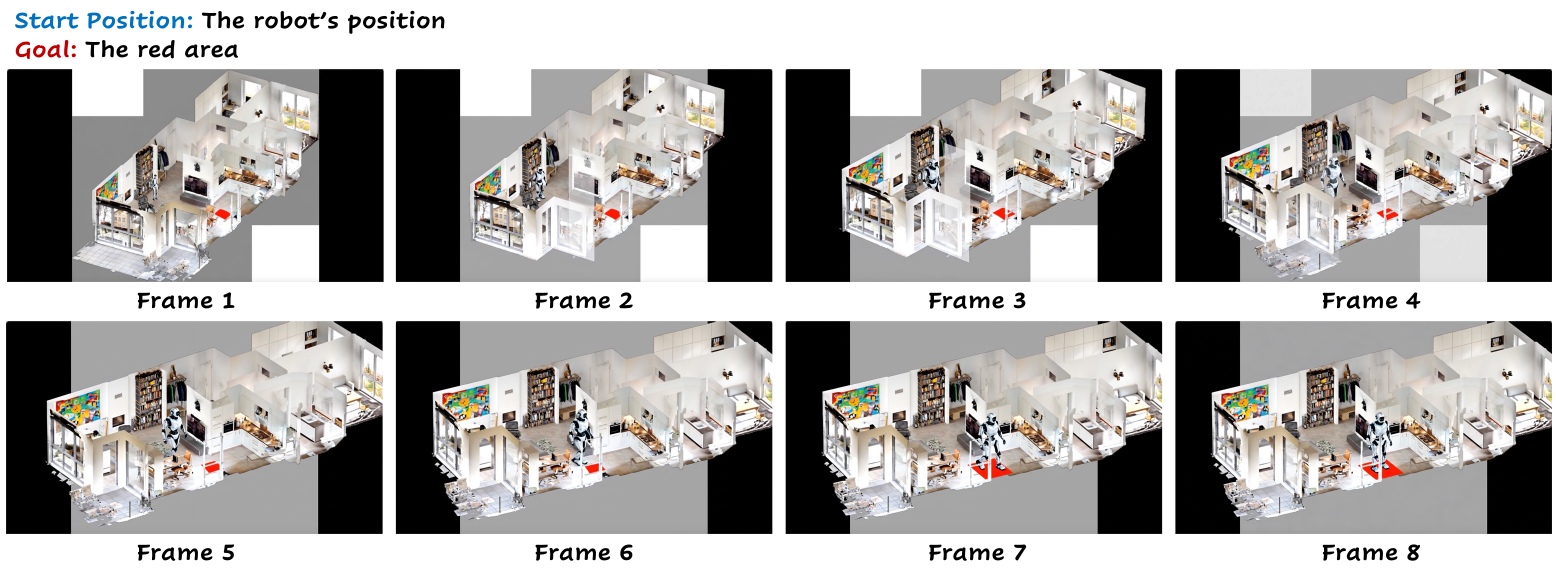}
    \vspace{-3mm}
    \caption{\textbf{3D Real-World Navigation} task showing navigation through a realistic 3D environment. The model must navigate from a starting position to a goal location using visual cues from the third-person perspective. A successful case completed by Veo-3 on the 3D Real-World Navigation task.}
    \vspace{-3mm}
    \label{fig:3d-nav}
\end{figure*}

\subsection{Evaluation and Metrics}



We adopt the shared embodied evaluation protocol in Section~\ref{sec:embodied_eval_metrics}: all metrics are binary and labeled via the automatic VLM/human rules on rollout traces and generated frames. Physical Understanding and Instruction Following checks (Object Semantic, Agent Consistency, Spatial Alignment, Destination Integrity, Scene Consistency) are unchanged; here we outline how they are applied in 3D Real-World Navigation and how the gates are formed.

\subsubsection{Task Completeness Metrics (Geometry Only)}


We report \textbf{Success Score 3D} and \textbf{Oracle Success Score 3D} as in Section~\ref{sec:embodied_eval_metrics}, testing whether the agent stops inside or ever enters the destination volume. These ignore visual fidelity and physical plausibility, which the gate metrics enforce.

\begin{itemize}
    \item \textbf{Success Score 3D (S.S. 3D).}
This checks whether the agent stops at the correct location in the 3D navigation sequence. The evaluation is geometric only and does not depend on whether the destination appearance is preserved. It is satisfied if the agent stops inside the intended destination volume.

    \item \textbf{Oracle Success Score 3D (O.S. 3D).}
This credits the agent if it ever comes within a proximity threshold of the correct 3D destination, regardless of its final stop. It is satisfied if the agent reaches the destination vicinity at any time during the video.
\end{itemize}

\subsubsection{Gate Metrics and Holistic Performance}

In 3D R.-W. Nav., the gates surface common failure modes: agents may “teleport” between floors, cut through occluded rooms, or alter the dollhouse layout to satisfy geometric hits. Requiring destination integrity and scene consistency prevents shortcutting via hallucinated staircases or duplicated goals, while the physics gate catches identity drift and implausible motion in the third-person view.

As with the Panoramic and Top-down tasks, strict conjunctions block generative shortcuts (\textit{e.g.}, hallucinating nearer goals or warping 3D structure), but the 3D setting stresses multi-floor occlusions and third-person viewpoint drift:

\begin{itemize}
    \item  \textbf{Success (3D) Original Destination.} This composite metric addresses the tendency of video generation models to hallucinate solution shortcuts. It is satisfied only if the Success Score 3D, Destination Integrity Score, and Scene Consistency Score are all met simultaneously. This metric ensures that a successful arrival is valid and not the result of the model altering the scene layout or destination location to simplify the task.
    \item \textbf{Physics Validness.} To provide a holistic view of physical understanding, this composite metric serves as a strict gate. It is satisfied only if the Object Semantic Score, Agent Consistency Score, and Spatial Alignment Score are all met simultaneously. This ensures the agent does not clip through objects, teleport, or drift unnaturally.
    \item \textbf{Overall Success} = Success Score 3D $\land$ Oracle Success Score 3D $\land$ Object Semantic $\land$ Agent Consistency $\land$ Spatial Alignment $\land$ Destination Integrity $\land$ Scene Consistency; a sample passes only when all seven binary checks are $1$.

\end{itemize}


\subsection{Evaluation Results}
\label{sec:embodied_3d_analysis}


\textbf{Key Finding: The ``Simulation Gap'' in Video Generation.} Current video generation models (Veo-3, Sora-2) exhibit a fundamental disconnect between \textbf{visual plausibility} and \textbf{physical causality}. While they often achieve high raw success rates in simple navigation tasks (85--89\%), they fail catastrophically on holistic metrics (0--3\%) due to severe scene hallucination and physics violations. In contrast, image-based agents (Nano-banana) demonstrate robust ``simulation'' capabilities, maintaining near-perfect scene consistency ($>$95\%) and physical validity.

\subsubsection{VLM-Based Evaluation}

\Cref{tab:embodied_task03_models_results} benchmarks Veo-3 against Sora-2, Nano-banana, and GPT-4o-image. The results highlight a distinct hierarchy in embodied navigation capabilities, separated by model modality and architectural stability.

\begin{table*}[htbp]
\centering
\small
\caption{Quantitative results for the \textbf{3D Real-world Navigation} benchmark. We compare Sora-2, Veo-3, Nano-banana, and GPT-4o-image across the same hard-level dimensions.}
\label{tab:embodied_task03_models_results}
\resizebox{\linewidth}{!}{
\begin{tabular}{@{}lcccccccccc@{}}
\toprule
 & \multicolumn{2}{c}{\textbf{Task Completeness}} & \multicolumn{3}{c}{\textbf{Physical Understanding}} & \multicolumn{2}{c}{\textbf{Instruction Following}} & \multicolumn{2}{c}{\textbf{Gate Metric}} & \multicolumn{1}{c}{\textbf{Holistic Metric}} \\
\cmidrule(lr){2-3} \cmidrule(lr){4-6} \cmidrule(lr){7-8} \cmidrule(lr){9-10} \cmidrule(lr){11-11}
\textbf{Evaluation} & \begin{tabular}[c]{@{}c@{}}\textbf{Success} \\ \textbf{Score} \\\textbf{(3D)}\end{tabular} & \begin{tabular}[c]{@{}c@{}}\textbf{Oracle} \\ \textbf{Success}\\ \textbf{Score (3D)}\end{tabular} & \begin{tabular}[c]{@{}c@{}}\textbf{Object} \\ \textbf{Semantic} \\ \textbf{Score}\end{tabular} & \begin{tabular}[c]{@{}c@{}}\textbf{Agent} \\ \textbf{Consistency} \\ \textbf{Score}\end{tabular} & \begin{tabular}[c]{@{}c@{}}\textbf{Spatial} \\ \textbf{Alignment} \\ \textbf{Score}\end{tabular} & \begin{tabular}[c]{@{}c@{}}\textbf{Destination} \\ \textbf{Integrity} \\ \textbf{Score}\end{tabular} & \begin{tabular}[c]{@{}c@{}}\textbf{Scene} \\ \textbf{Consistency} \\ \textbf{Score}\end{tabular} & \begin{tabular}[c]{@{}c@{}}\textbf{Success (3D)} \\ \textbf{Original} \\ \textbf{Destination}\end{tabular} & \begin{tabular}[c]{@{}c@{}}\textbf{Physics} \\ \textbf{Validness}\end{tabular} & \begin{tabular}[c]{@{}c@{}}\textbf{Overall} \\ \textbf{Success}\end{tabular} \\
\midrule
\multicolumn{11}{@{}l}{\textbf{Environmental Complexity}} \\
\multicolumn{11}{@{}l}{\quad \textit{Level: floor01}} \\
\multicolumn{11}{@{}l}{\quad \textbf{Video Models}} \\
\quad \quad Veo-3 & 85.00\% & 91.67\% & 81.67\% & 86.67\% & 63.33\% & 53.33\% & 15.00\% & 11.67\% & 40.00\% & \textbf{3.33\%} \\
\quad \quad Sora-2 & 88.89\% & 97.22\% & 72.22\% & 83.33\% & 97.22\% & 77.78\% & 44.44\% & 0.00\% & 55.00\% & 0.00\% \\
\multicolumn{11}{@{}l}{\quad \textbf{Image Models}} \\
\quad \quad Nano-banana & 75.00\% & 75.00\% & 100.00\% & 97.22\% & 97.22\% & 75.00\% & 86.11\% & 75.00\% & 97.22\% & \textbf{75.00\%} \\
\quad \quad GPT-4o-image & 13.33\% & 13.33\% & 88.33\% & 48.33\% & 58.33\% & 13.33\% & 21.67\% & 13.33\% & 35.00\% & 13.33\% \\
\multicolumn{11}{@{}l}{\quad \textit{Level: floor02plus}} \\
\multicolumn{11}{@{}l}{\quad \textbf{Video Models}} \\
\quad \quad Veo-3 & 68.33\% & 78.33\% & 75.00\% & 68.33\% & 73.33\% & 38.33\% & 18.33\% & 5.00\% & 33.33\% & \textbf{0.00\%} \\
\quad \quad Sora-2 & 72.22\% & 77.78\% & 69.44\% & 75.00\% & 83.33\% & 58.33\% & 36.11\% & 0.00\% & 60.00\% & \textbf{0.00\%} \\
\multicolumn{11}{@{}l}{\quad \textbf{Image Models}} \\
\quad \quad Nano-banana & 83.33\% & 86.11\% & 94.44\% & 97.22\% & 94.44\% & 86.11\% & 100.00\% & 83.33\% & 94.44\% & \textbf{83.33\%} \\
\quad \quad GPT-4o-image & 15.00\% & 16.67\% & 75.00\% & 56.67\% & 51.67\% & 15.00\% & 15.00\% & 13.33\% & 40.00\% & 13.33\% \\
\multicolumn{11}{@{}l}{\textbf{View Fidelity}} \\
\multicolumn{11}{@{}l}{\quad \textit{Level: quality03}} \\
\multicolumn{11}{@{}l}{\quad \textbf{Video Models}} \\
\quad \quad Veo-3 & 65.00\% & 77.50\% & 72.50\% & 75.00\% & 70.00\% & 40.00\% & 15.00\% & 5.00\% & 30.00\% & \textbf{0.00\%} \\
\quad \quad Sora-2 & 75.00\% & 83.33\% & 79.17\% & 83.33\% & 91.67\% & 66.67\% & 41.67\% & 0.00\% & 67.50\% & \textbf{0.00\%} \\
\multicolumn{11}{@{}l}{\quad \textbf{Image Models}} \\
\quad \quad Nano-banana & 75.00\% & 79.17\% & 95.83\% & 95.83\% & 91.67\% & 79.17\% & 91.67\% & 75.00\% & 91.67\% & \textbf{75.00\%} \\
\quad \quad GPT-4o-image & 7.50\% & 7.50\% & 80.00\% & 45.00\% & 52.50\% & 5.00\% & 7.50\% & 5.00\% & 30.00\% & 5.00\% \\
\multicolumn{11}{@{}l}{\quad \textit{Level: quality04}} \\
\multicolumn{11}{@{}l}{\quad \textbf{Video Models}} \\
\quad \quad Veo-3 & 82.50\% & 85.00\% & 75.00\% & 85.00\% & 65.00\% & 50.00\% & 20.00\% & 10.00\% & 42.50\% & \textbf{2.50\%} \\
\quad \quad Sora-2 & 91.67\% & 95.83\% & 54.17\% & 75.00\% & 83.33\% & 70.83\% & 41.67\% & 0.00\% & 45.00\% & 0.00\% \\
\multicolumn{11}{@{}l}{\quad \textbf{Image Models}} \\
\quad \quad Nano-banana & 87.50\% & 87.50\% & 100.00\% & 100.00\% & 100.00\% & 87.50\% & 91.67\% & 87.50\% & 100.00\% & \textbf{87.50\%} \\
\quad \quad GPT-4o-image & 17.50\% & 20.00\% & 80.00\% & 57.50\% & 52.50\% & 20.00\% & 27.50\% & 17.50\% & 40.00\% & 17.50\% \\
\multicolumn{11}{@{}l}{\quad \textit{Level: quality05}} \\
\multicolumn{11}{@{}l}{\quad \textbf{Video Models}} \\
\quad \quad Veo-3 & 82.50\% & 92.50\% & 87.50\% & 72.50\% & 70.00\% & 47.50\% & 15.00\% & 10.00\% & 37.50\% & \textbf{2.50\%} \\
\quad \quad Sora-2 & 75.00\% & 83.33\% & 79.17\% & 79.17\% & 95.83\% & 66.67\% & 37.50\% & 0.00\% & 60.00\% & 0.00\% \\
\multicolumn{11}{@{}l}{\quad \textbf{Image Models}} \\
\quad \quad Nano-banana & 75.00\% & 75.00\% & 95.83\% & 95.83\% & 95.83\% & 75.00\% & 95.83\% & 75.00\% & 95.83\% & \textbf{75.00\%} \\
\quad \quad GPT-4o-image & 17.50\% & 17.50\% & 85.00\% & 55.00\% & 60.00\% & 17.50\% & 20.00\% & 17.50\% & 42.50\% & 17.50\% \\
\multicolumn{11}{@{}l}{\textbf{Trajectory Distance}} \\
\multicolumn{11}{@{}l}{\quad \textit{Level: short}} \\
\multicolumn{11}{@{}l}{\quad \textbf{Video Models}} \\
\quad \quad Veo-3 & 73.33\% & 85.00\% & 81.67\% & 78.33\% & 68.33\% & 46.67\% & 16.67\% & 8.33\% & 38.33\% & \textbf{3.33\%} \\
\quad \quad Sora-2 & 72.22\% & 83.33\% & 80.56\% & 77.78\% & 86.11\% & 63.89\% & 41.67\% & 0.00\% & 55.00\% & 0.00\% \\
\multicolumn{11}{@{}l}{\quad \textbf{Image Models}} \\
\quad \quad Nano-banana & 77.78\% & 80.56\% & 97.22\% & 97.22\% & 94.44\% & 80.56\% & 88.89\% & 77.78\% & 94.44\% & \textbf{77.78\%} \\
\quad \quad GPT-4o-image & 13.33\% & 13.33\% & 85.00\% & 56.67\% & 58.33\% & 11.67\% & 16.67\% & 11.67\% & 45.00\% & 11.67\% \\
\multicolumn{11}{@{}l}{\quad \textit{Level: long}} \\
\multicolumn{11}{@{}l}{\quad \textbf{Video Models}} \\
\quad \quad Veo-3 & 80.00\% & 85.00\% & 75.00\% & 76.67\% & 68.33\% & 45.00\% & 16.67\% & 8.33\% & 35.00\% & \textbf{0.00\%} \\
\quad \quad Sora-2 & 88.89\% & 91.67\% & 61.11\% & 80.56\% & 94.44\% & 72.22\% & 38.89\% & 0.00\% & 60.00\% & \textbf{0.00\%} \\
\multicolumn{11}{@{}l}{\quad \textbf{Image Models}} \\
\quad \quad Nano-banana & 80.56\% & 80.56\% & 97.22\% & 97.22\% & 97.22\% & 80.56\% & 97.22\% & 80.56\% & 97.22\% & \textbf{80.56\%} \\
\quad \quad GPT-4o-image & 15.00\% & 16.67\% & 78.33\% & 48.33\% & 51.67\% & 16.67\% & 20.00\% & 15.00\% & 30.00\% & 15.00\% \\
\multicolumn{11}{@{}l}{\textbf{Destination Specification}} \\
\multicolumn{11}{@{}l}{\quad \textit{Level: color mark}} \\
\multicolumn{11}{@{}l}{\quad \textbf{Video Models}} \\
\quad \quad Veo-3 & 93.33\% & 96.67\% & 66.67\% & 73.33\% & 70.00\% & 31.67\% & 28.33\% & 15.00\% & 28.33\% & \textbf{3.33\%} \\
\quad \quad Sora-2 & 88.89\% & 91.67\% & 58.33\% & 75.00\% & 88.89\% & 69.44\% & 36.11\% & 0.00\% & 51.67\% & 0.00\% \\
\multicolumn{11}{@{}l}{\quad \textbf{Image Models}} \\
\quad \quad Nano-banana & 80.56\% & 80.56\% & 100.00\% & 100.00\% & 100.00\% & 80.56\% & 91.67\% & 80.56\% & 100.00\% & \textbf{80.56\%} \\
\quad \quad GPT-4o-image & 16.67\% & 18.33\% & 71.67\% & 46.67\% & 45.00\% & 18.33\% & 20.00\% & 16.67\% & 30.00\% & 16.67\% \\
\multicolumn{11}{@{}l}{\quad \textit{Level: location description}} \\
\multicolumn{11}{@{}l}{\quad \textbf{Video Models}} \\
\quad \quad Veo-3 & 60.00\% & 73.33\% & 90.00\% & 81.67\% & 66.67\% & 60.00\% & 5.00\% & 1.67\% & 45.00\% & \textbf{0.00\%} \\
\quad \quad Sora-2 & 72.22\% & 83.33\% & 83.33\% & 83.33\% & 91.67\% & 66.67\% & 44.44\% & 0.00\% & 63.33\% & \textbf{0.00\%} \\
\multicolumn{11}{@{}l}{\quad \textbf{Image Models}} \\
\quad \quad Nano-banana & 77.78\% & 80.56\% & 94.44\% & 94.44\% & 91.67\% & 80.56\% & 94.44\% & 77.78\% & 91.67\% & \textbf{77.78\%} \\
\quad \quad GPT-4o-image & 11.67\% & 11.67\% & 91.67\% & 58.33\% & 65.00\% & 10.00\% & 16.67\% & 10.00\% & 45.00\% & 10.00\% \\
\bottomrule
\end{tabular} }
\end{table*}

\paragraph{Image Models vs. Video Models (Nano-banana Dominance).} Within the selected diagnostic subset, \textbf{Nano-banana} emerges as the distinct leader, showcasing that image generation is currently far more reliable for this embodied task than video generation.
\begin{itemize}
    \item \textbf{Consistency as a Foundation:} Nano-banana maintains near-perfect scores in \textit{Scene Consistency} (86--100\%) and \textit{Physics Validness} (94--100\%) across all complexity levels. This stability allows it to achieve high Overall Success rates (75--87.5\%) that video models cannot approach.
    \item \textbf{Resilience to Complexity:} As environment complexity increases (from \textit{floor01} to \textit{floor02plus}), Nano-banana actually improves its overall success from 75\% to 83.33\%. In sharp contrast, video models collapse; Veo-3's holistic success drops to 0\% in complex multi-floor environments, unable to reconcile geometric consistency with longer navigation horizons.
\end{itemize}

\paragraph{Video Model Trade-offs (Veo-3 vs. Sora-2).} While both video models struggle with holistic success (mostly 0\%--3\%), they exhibit different failure modes:
\begin{itemize}
    \item \textbf{Sora-2 (Better Geometry, Worse Adherence):} Sora-2 generally outperforms Veo-3 in \textit{Spatial Alignment} (\textit{e.g.}, 97.22\% vs 63.33\% in \textit{floor01}) and \textit{Scene Consistency} (44.44\% vs 15.00\%). However, it suffers from a critical flaw in \textit{Gate Metric: Success (Original Destination)}, scoring 0.00\% across almost all categories. This indicates that while Sora-2 generates smooth, consistent video, it hallucinates the destination or drifts significantly from the prompt's specific target.
    \item \textbf{Veo-3 (Better Adherence, Worse Physics):} Veo-3 is more ``compliant'' with the task, scoring higher on reaching the original destination (11.67\% in \textit{floor01}). However, it achieves this by sacrificing physical laws, as evidenced by its abysmal \textit{Scene Consistency} scores (dropping to 5.00\% in location tasks) and frequent physics violations.
\end{itemize}

\paragraph{Baseline Comparison.} GPT-4o-image acts as a semantic baseline. While it demonstrates strong object recognition (\textit{Object Semantic Score} $\approx$ 80--90\%), it lacks the spatial reasoning to navigate, resulting in low success scores ($\approx$ 13--17\%). This confirms that embodied navigation requires more than just ``seeing'' the scene; it requires a consistent internal world model that GPT-4o-image lacks.

\subsubsection{View-Syn Evaluation}

\begin{table*}[h]
  \centering
  \caption{3D Real-World Navigation — Video (FVD/SSIM/PSNR/LPIPS) and Image (FID normalized); higher is better except LPIPS}
  \begin{tabular}{lcccccc}
    \toprule
    Model & Modality & FVD Norm $\uparrow$ & FID Norm $\uparrow$ & SSIM $\uparrow$ & PSNR $\uparrow$ & LPIPS $\downarrow$ \\
    \midrule
    Veo-3          & Video & 0.7708 & --     & 0.2987 & 7.35 & 0.8205 \\
    Wan-2.2        & Video & 0.7107 & --     & 0.4562 & 9.11 & 0.7652 \\
    Sora-2         & Video & 0.6870 & --     & 0.4471 & 9.19 & 0.8052 \\
    GPT-image-1.5  & Image & --     & 0.6747 & --     & --   & --     \\
    Nano-banana-pro& Image & --     & 0.6740 & --     & --   & --     \\
    GPT-image      & Image & --     & 0.5673 & --     & --   & --     \\
    Nano-banana    & Image & --     & 0.5673 & --     & --   & --     \\
    Qwen-image     & Image & --     & 0.5627 & --     & --   & --     \\
    \bottomrule
  \end{tabular}
\end{table*}

\subsection{Case Study: Qualitative Failure Modes}

\begin{figure*}[htbp]
    \centering
    \includegraphics[width=0.8\textwidth]{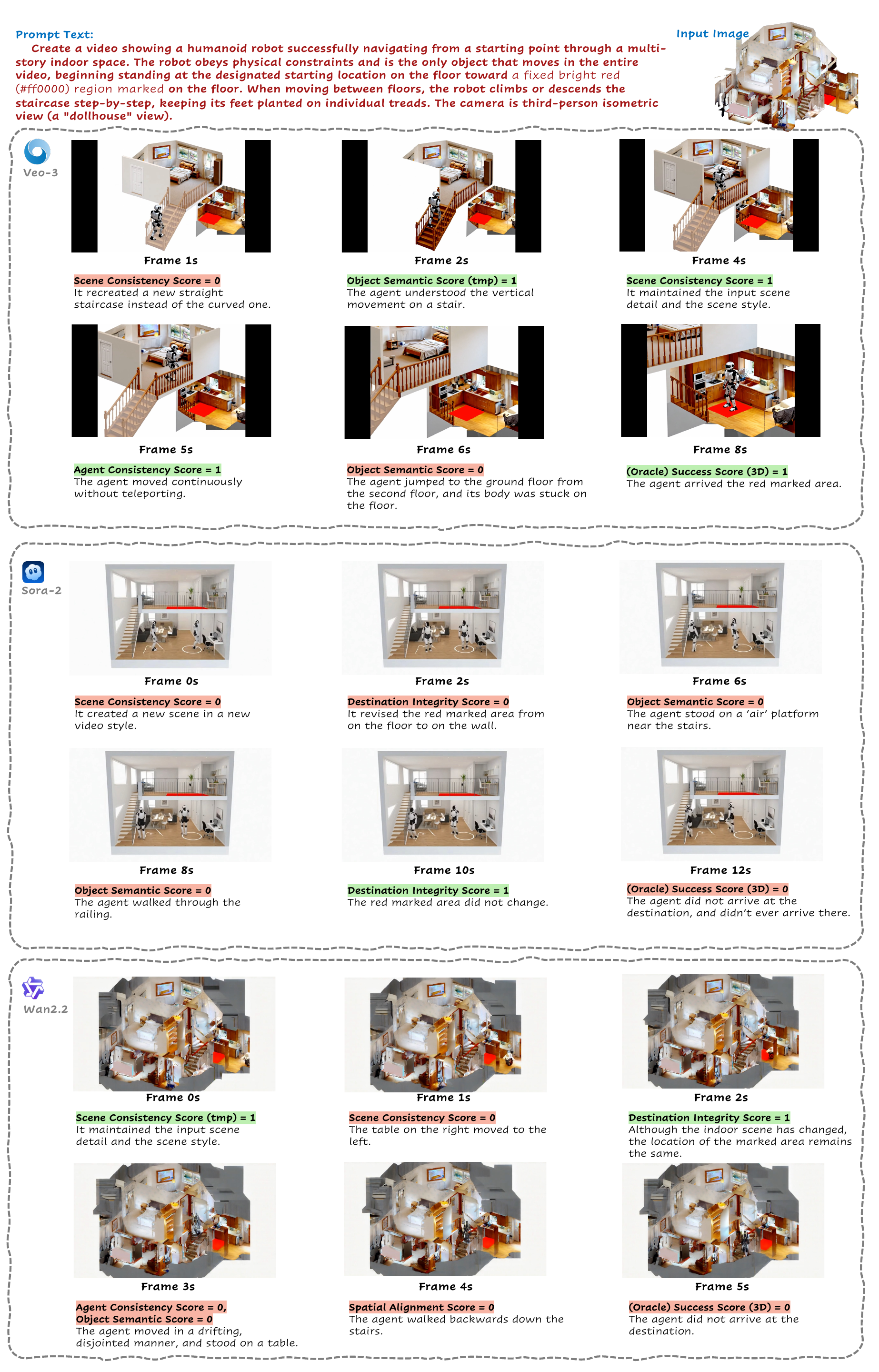}
    \caption{Case Study for 3D Real-World Navigation comparing Veo-3, Sora-2, and Wan-2.2. The visualization highlights model-specific failure modes: Veo-3 achieves the destination despite physics violations (floor jumping) and scene alterations (staircase geometry); Sora-2 suffers from severe scene hallucination and style drift; and Wan-2.2 exhibits temporal inconsistency with drifting agents and moving furniture.}
    \label{fig:embodied_casestudy_3d}
\end{figure*}

This case study in \Cref{fig:embodied_casestudy_3d} evaluates the ability of three video generation models—Veo-3, Sora-2, and Wan-2.2—to simulate a humanoid robot navigating a multi-story indoor space to a specific red target zone. The prompt requires strict adherence to physical constraints, step-by-step stair climbing, and a consistent ``dollhouse'' isometric view.

\textbf{Veo-3 Analysis: Task Success with Physical Compromises.} Veo-3 was the only model to achieve a successful arrival at the destination (Oracle Success Score = 1). However, this success came with significant hallucinations and physics violations:
\begin{itemize}
    \item \textbf{Scene Consistency:} The model failed to maintain the structural integrity of the input scene, notably hallucinating a straight staircase in place of the original curved one.
    \item \textbf{Physical Fidelity:} While the agent initially demonstrated understanding of vertical movement, it ultimately failed the physics constraint by jumping directly from the second floor to the ground floor, resulting in the body clipping into the floor. Despite these errors, the agent maintained consistency without teleporting earlier in the sequence.
\end{itemize}

\textbf{Sora-2 Analysis: Severe Hallucination and Physics Failures.} Sora-2 struggled significantly with both scene adherence and physical laws:
\begin{itemize}
    \item \textbf{Style and Destination Drift:} The model completely disregarded the input image, creating a new scene with a different video style. Crucially, it altered the task parameters by moving the red destination marker from the floor to the wall.
    \item \textbf{Physics Violations:} The agent displayed zero understanding of solid geometry (Object Semantic Score = 0), standing on ``air'' platforms near the stairs and walking directly through the railing. Consequently, the agent never arrived at the valid destination.
\end{itemize}

\textbf{Wan-2.2 Analysis: Temporal Instability and Agent Drifting.} Wan-2.2 initially maintained the scene details well but quickly devolved into temporal incoherence:
\begin{itemize}
    \item \textbf{Environmental Instability:} The static environment proved unstable; by Frame 1s, furniture (a table) shifted position from right to left.
    \item \textbf{Agent Control:} The navigation was disjointed. The agent drifted, stood on top of tables, and walked backwards down the stairs. This resulted in a failure to reach the destination.
\end{itemize}

While Veo-3 was the only model to ``complete'' the navigation task, all three models struggled with strict physical plausibility in a 3D space. Veo-3 prioritized path completion over geometry; Sora-2 hallucinated an entirely new reality; and Wan-2.2 failed to keep the static environment stationary.

\section{Simultaneous Localization and Generation (SLAG)}
\label{apx:slag}

\subsection{Task Definition}

The \textbf{Simultaneous Localization and Generation (SLAG)} task extends the paradigm of Simultaneous Localization and Mapping (SLAM)~\cite{durrant2006slam} by coupling 3D spatial navigation with real-time generative mapping. Unlike traditional methods that estimate position within a static or progressively built map, SLAG actively synthesizes a synchronized 2D top-down trajectory that corresponds to the robot's movement through a 3D environment. As a humanoid robot navigates complex indoor scenes—such as photorealistic apartments rendered in a ``dollhouse'' perspective—the system dynamically generates a 2D representation of its progress from start to goal. The core objective of SLAG is to achieve precise \textbf{spatiotemporal alignment} between \textbf{physical 3D navigation} and \textbf{generative 2D plotting}, demonstrating the synergy between real-time perception and generative modeling in spatial understanding.

\begin{figure*}[htbp]
    \centering
    \includegraphics[width=\linewidth]{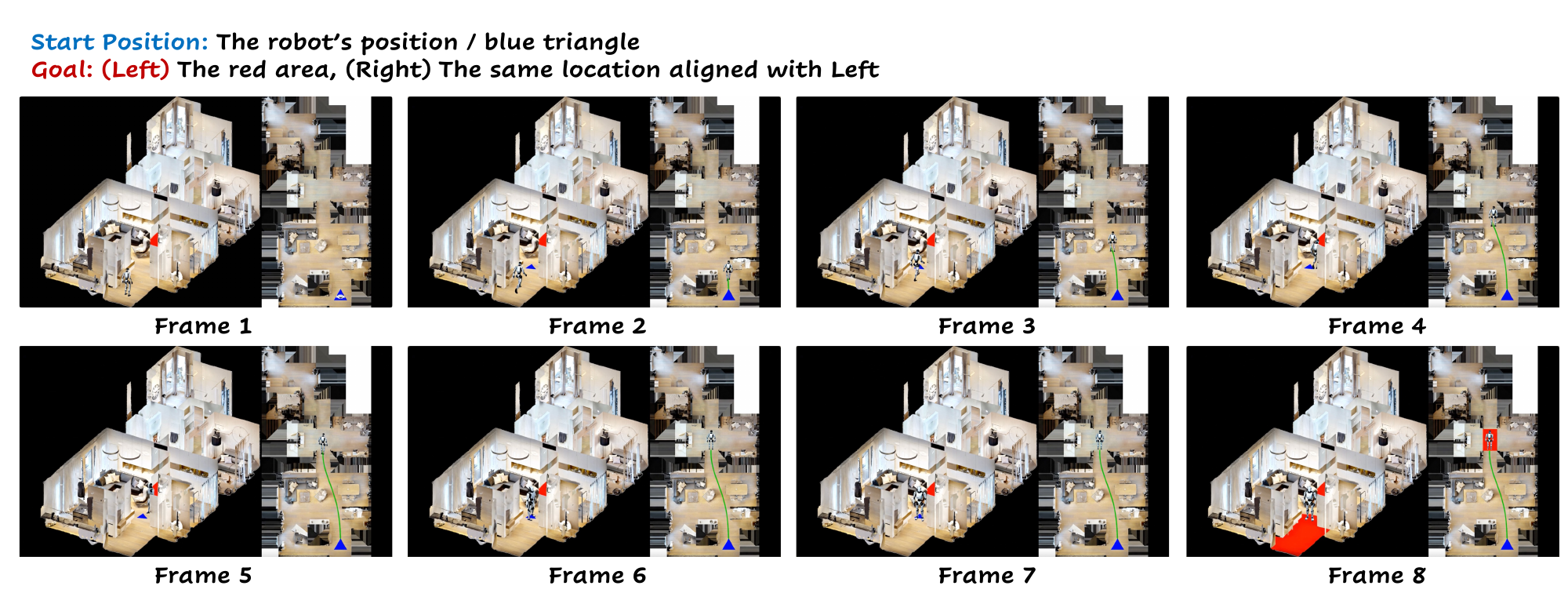}
    \caption{Simultaneous Localization and Goal-reaching (SLAG) task requiring the model to simultaneously maintain spatial awareness of its position while navigating toward a goal in an unknown environment.}
    \label{fig:slag-nav}
\end{figure*}

\subsection{Evaluation and Metrics}


SLAG differs from the previous tasks by requiring \textbf{simultaneous 3D navigation and 2D generative mapping}, making \textbf{trajectory alignment} the primary bottleneck.
Similarly, we adopt the shared embodied evaluation protocol in Section~\ref{sec:embodied_eval_metrics}. Physical Understanding and Instruction Following checks (Object Semantic, Agent Consistency, Spatial Alignment, Destination Integrity, Scene Consistency) are unchanged. Here we focus on the SLAG-specific task completeness set and gates.

\subsubsection{Task Completeness Metrics (Geometry Only)}

Five task completeness metrics are evaluated jointly: \textbf{Success Score 2D}, \textbf{Oracle Success Score 2D}, \textbf{Success Score 3D}, \textbf{Oracle Success Score 3D} (all from Section~\ref{sec:embodied_eval_metrics}), and \textbf{Trajectory Alignment} to ensure the projected 2D path matches the 3D motion. The 2D map is provided, but the destination is not; alignment hinges on correctly projecting the 3D navigation onto the given map, with 2D success conditioned on the same 3D destination.

\begin{itemize}
    \item \textbf{Success Score (S.S. 2D).} Measures whether the agent’s final position lies within the highlighted or textually specified goal region in the 2D overhead map. The score is $1$ if the final coordinates fall entirely inside the goal footprint; otherwise $0$.
    \item \textbf{Oracle Success Score (O.S. 2D).} Provides partial credit when the agent comes sufficiently close to the 2D goal during navigation. The score is $1$ if the agent’s path ever intersects or touches the goal region, even if it does not stop there; otherwise $0$.
    \item \textbf{Success Score (S.S. 3D).} Checks whether the agent ends inside the correct destination volume in the 3D navigation sequence. This metric is purely geometric and independent of any visual discrepancies at the destination. The score is $1$ if the final 3D position is within the target volume; otherwise $0$.
    \item \textbf{Oracle Success Score (O.S. 3D).} Grants credit when the agent enters the vicinity of the correct 3D destination at any point during its rollout. The score is $1$ if the trajectory ever crosses the predefined proximity threshold around the target; otherwise $0$.
    \item \textbf{Trajectory Alignment Score.} Evaluates whether the agent’s 2D projected route is consistent with its 3D motion path, focusing on major turns and spatial transitions. A score of $1$ indicates strong correspondence between the two trajectories; otherwise $0$.
\end{itemize}

\subsubsection{Gate Metrics and Holistic Performance}

As in the prior tasks, gates block generative shortcuts, but only a 3D destination gate is possible because the 2D map has no fixed ground-truth target (it is generated, not given):

In SLAG, the holistic gate exposes compounded failures: models often align one panel but not the other, hallucinate extra corridors in the 2D drawing, or “teleport” in 3D while the 2D trace stays smooth. Because the 2D goal is synthesized, the 3D destination gate plus trajectory alignment are key to preventing such mismatches.

\begin{itemize}
    \item \textbf{Success (3D) with Original Destination} = Success Score 3D $\land$ Destination Integrity $\land$ Scene Consistency.
    \item \textbf{Physics Validness} = Object Semantic $\land$ Agent Consistency $\land$ Spatial Alignment.
    \item \textbf{Overall Success} = Success Score 2D $\land$ Oracle Success Score 2D $\land$ Success Score 3D $\land$ Oracle Success Score 3D $\land$ Trajectory Alignment $\land$ Object Semantic $\land$ Agent Consistency $\land$ Spatial Alignment $\land$ Destination Integrity $\land$ Scene Consistency; a sample passes only when all ten binary checks are $1$.
\end{itemize}

\subsection{Evaluation Results}
\label{sec:embodied_slag_analysis}

The Simultaneous Localization and Generation (SLAG) task represents a significant leap in embodied AI, requiring models to maintain strict temporal and spatial alignment between 3D physical navigation and 2D generative mapping. The following diagnostic analysis dissects the performance of the selected model subset, revealing a critical trade-off between the visual fluidity of video models and the logical adherence of image-based models.

\textbf{Key Findings: Reality \& Modality Gap.}
\begin{itemize}[leftmargin=*]
    \item \textbf{Alignment bottleneck.} Trajectory Alignment remains the weakest link for video baselines (44.07\%--55.17\% vs. Nano-banana’s 88.89\%/76.67\%), and Destination Integrity stays low for them (20.34\%--44.83\%). Even with strong Scene Consistency (45.76\%--89.66\%), cross-view synchronization caps holistic success.
    \item \textbf{Semantic cliff.} Switching from color marks to textual location descriptions collapses Nano-banana’s 3D Success (81.25\% $\rightarrow$ 17.65\%) and Holistic score (50.00\% $\rightarrow$ 8.82\%); GPT-4o-image drops similarly (31.03\% $\rightarrow$ 1.67\%). Video models show the same gap (Veo-3: 17.24\% $\rightarrow$ 5.17\%, Sora-2: 19.64\% $\rightarrow$ 6.67\%), confirming spatial language grounding is the dominant failure mode.
    \item \textbf{Cross-view priors activate under pressure.} Forced 3D-to-2D alignment sustains Scene Consistency (Veo-3: 45.76\% $\rightarrow$ 57.89\%, Nano-banana: 100.00\% $\rightarrow$ 96.67\%, GPT-4o-image: 98.33\% $\rightarrow$ 94.83\%) even when trajectories misalign, suggesting latent spatial priors surface only with structured multimodal conditioning.
\end{itemize}


\subsubsection{VLM-Based Evaluation}


\Cref{tab:embodied_task04_models_results} highlights the performance disparities across Veo-3, Sora-2, Nano-banana, and GPT-4o-image. While video generation models often excel at visual fidelity, the requirement for synchronized spatial logic proves challenging.

\begin{table*}[htbp]
\centering
\small
\caption{Quantitative results for the \textbf{Simultaneous Localization and Generation} benchmark. We compare automatic evaluations across different models: Sora-2, Veo-3, Nano-banana, and GPT-4o-image.}
\label{tab:embodied_task04_models_results}
\resizebox{\linewidth}{!}{
\begin{tabular}{@{}lccccccccccccc@{}}
\toprule
 & \multicolumn{5}{c}{\textbf{Task Completeness}} & \multicolumn{3}{c}{\textbf{Physical Understanding}} & \multicolumn{2}{c}{\textbf{Instruction Following}} & \multicolumn{2}{c}{\textbf{Gate Metric}} & \multicolumn{1}{c}{\textbf{Holistic Metric}} \\
\cmidrule(lr){2-6} \cmidrule(lr){7-9} \cmidrule(lr){10-11} \cmidrule(lr){12-13} \cmidrule(lr){14-14}
\textbf{Evaluation} & \begin{tabular}[c]{@{}c@{}}\textbf{Success} \\ \textbf{Score} \\ \textbf{(3D)}\end{tabular} & \begin{tabular}[c]{@{}c@{}}\textbf{Success} \\ \textbf{Score} \\ \textbf{(2D)}\end{tabular} & \begin{tabular}[c]{@{}c@{}}\textbf{Oracle}\\ \textbf{Success} \\ \textbf{Score (3D)}\end{tabular} & \begin{tabular}[c]{@{}c@{}}\textbf{Oracle}\\ \textbf{Success} \\ \textbf{Score (2D)}\end{tabular} & \begin{tabular}[c]{@{}c@{}}\textbf{Trajectory} \\ \textbf{Alignment} \\ \textbf{Score}\end{tabular} & \begin{tabular}[c]{@{}c@{}}\textbf{Object} \\ \textbf{Semantic} \\ \textbf{Score}\end{tabular} & \begin{tabular}[c]{@{}c@{}}\textbf{Agent} \\ \textbf{Consistency} \\ \textbf{Score}\end{tabular} & \begin{tabular}[c]{@{}c@{}}\textbf{Spatial} \\ \textbf{Alignment} \\ \textbf{Score}\end{tabular} & \begin{tabular}[c]{@{}c@{}}\textbf{Destination} \\ \textbf{Integrity} \\ \textbf{Score} \end{tabular} & \begin{tabular}[c]{@{}c@{}}\textbf{Scene} \\  \textbf{Consistency} \\ \textbf{Score}\end{tabular} & \begin{tabular}[c]{@{}c@{}}\textbf{Success (3D)} \\ \textbf{Original} \\ \textbf{Destination} \end{tabular} & \begin{tabular}[c]{@{}c@{}}\textbf{Physics} \\ \textbf{Validness}\end{tabular} & \begin{tabular}[c]{@{}c@{}}\textbf{Overall} \\ \textbf{Success}\end{tabular} \\
\midrule
\multicolumn{14}{@{}l}{\textbf{Environmental Complexity}} \\
\multicolumn{14}{@{}l}{\quad \textit{Level: floor01}} \\
\multicolumn{14}{@{}l}{\quad \textbf{Video Models}} \\
\quad \quad Veo-3 & 52.54\% & 44.07\% & 59.32\% & 49.15\% & 44.07\% & 72.88\% & 52.54\% & 67.80\% & 20.34\% & 45.76\% & 18.64\% & 45.76\% & \textbf{11.86\%} \\
\quad \quad Sora-2 & 29.31\% & 27.59\% & 34.48\% & 37.93\% & 55.17\% & 84.48\% & 84.48\% & 72.41\% & 32.76\% & 79.31\% & 22.41\% & 65.52\% & 10.34\% \\
\multicolumn{14}{@{}l}{\quad \textbf{Image Models}} \\
\quad \quad Nano-banana & 55.56\% & 41.67\% & 55.56\% & 47.22\% & 88.89\% & 83.33\% & 83.33\% & 77.78\% & 50.00\% & 100.00\% & 38.89\% & 69.44\% & \textbf{27.78\%} \\
\quad \quad GPT-4o-image & 35.00\% & 28.33\% & 36.67\% & 31.67\% & 66.67\% & 81.67\% & 76.67\% & 61.67\% & 26.67\% & 98.33\% & 25.00\% & 58.33\% & 25.00\% \\
\multicolumn{14}{@{}l}{\quad \textit{Level: floor02plus}} \\
\multicolumn{14}{@{}l}{\quad \textbf{Video Models}} \\
\quad \quad Veo-3 & 21.05\% & 38.60\% & 26.32\% & 42.11\% & 33.33\% & 50.88\% & 42.11\% & 52.63\% & 31.58\% & 57.89\% & 15.79\% & 31.58\% & 10.53\% \\
\quad \quad Sora-2 & 29.31\% & 32.76\% & 31.03\% & 37.93\% & 48.28\% & 82.76\% & 67.24\% & 68.97\% & 44.83\% & 89.66\% & 20.69\% & 56.90\% & \textbf{15.52\%} \\
\multicolumn{14}{@{}l}{\quad \textbf{Image Models}} \\
\quad \quad Nano-banana & 40.00\% & 56.67\% & 40.00\% & 63.33\% & 76.67\% & 86.67\% & 86.67\% & 70.00\% & 50.00\% & 96.67\% & 36.67\% & 60.00\% & \textbf{30.00\%} \\
\quad \quad GPT-4o-image & 12.07\% & 17.24\% & 13.79\% & 20.69\% & 43.10\% & 68.97\% & 48.28\% & 37.93\% & 17.24\% & 94.83\% & 6.90\% & 34.48\% & 6.90\% \\
\multicolumn{14}{@{}l}{\textbf{View Fidelity}} \\
\multicolumn{14}{@{}l}{\quad \textit{Level: quality03}} \\
\multicolumn{14}{@{}l}{\quad \textbf{Video Models}} \\
\quad \quad Veo-3 & 35.00\% & 40.00\% & 35.00\% & 40.00\% & 35.00\% & 60.00\% & 45.00\% & 52.50\% & 22.50\% & 45.00\% & 15.00\% & 37.50\% & 10.00\% \\
\quad \quad Sora-2 & 27.50\% & 27.50\% & 27.50\% & 32.50\% & 47.50\% & 85.00\% & 77.50\% & 75.00\% & 37.50\% & 77.50\% & 22.50\% & 62.50\% & \textbf{17.50\%} \\
\multicolumn{14}{@{}l}{\quad \textbf{Image Models}} \\
\quad \quad Nano-banana & 50.00\% & 62.50\% & 50.00\% & 62.50\% & 91.67\% & 91.67\% & 95.83\% & 91.67\% & 58.33\% & 100.00\% & 41.67\% & 83.33\% & \textbf{41.67\%} \\
\quad \quad GPT-4o-image & 30.00\% & 25.00\% & 32.50\% & 32.50\% & 65.00\% & 75.00\% & 65.00\% & 52.50\% & 20.00\% & 97.50\% & 17.50\% & 52.50\% & 17.50\% \\
\multicolumn{14}{@{}l}{\quad \textit{Level: quality04}} \\
\multicolumn{14}{@{}l}{\quad \textbf{Video Models}} \\
\quad \quad Veo-3 & 46.15\% & 51.28\% & 51.28\% & 58.97\% & 43.59\% & 66.67\% & 46.15\% & 69.23\% & 33.33\% & 48.72\% & 25.64\% & 35.90\% & \textbf{15.38\%} \\
\quad \quad Sora-2 & 26.32\% & 34.21\% & 26.32\% & 42.11\% & 55.26\% & 84.21\% & 71.05\% & 71.05\% & 34.21\% & 84.21\% & 21.05\% & 60.53\% & 10.53\% \\
\multicolumn{14}{@{}l}{\quad \textbf{Image Models}} \\
\quad \quad Nano-banana & 45.83\% & 45.83\% & 45.83\% & 54.17\% & 70.83\% & 87.50\% & 79.17\% & 58.33\% & 50.00\% & 95.83\% & 37.50\% & 54.17\% & 20.83\% \\
\quad \quad GPT-4o-image & 25.00\% & 30.00\% & 25.00\% & 30.00\% & 57.50\% & 72.50\% & 62.50\% & 50.00\% & 32.50\% & 97.50\% & 22.50\% & 47.50\% & \textbf{22.50\%} \\
\multicolumn{14}{@{}l}{\quad \textit{Level: quality05}} \\
\multicolumn{14}{@{}l}{\quad \textbf{Video Models}} \\
\quad \quad Veo-3 & 29.73\% & 32.43\% & 43.24\% & 37.84\% & 37.84\% & 59.46\% & 51.35\% & 59.46\% & 21.62\% & 62.16\% & 10.81\% & 43.24\% & 8.11\% \\
\quad \quad Sora-2 & 34.21\% & 28.95\% & 44.74\% & 39.47\% & 52.63\% & 81.58\% & 78.95\% & 65.79\% & 44.74\% & 92.11\% & 21.05\% & 60.53\% & \textbf{10.53\%} \\
\multicolumn{14}{@{}l}{\quad \textbf{Image Models}} \\
\quad \quad Nano-banana & 50.00\% & 33.33\% & 50.00\% & 44.44\% & 88.89\% & 72.22\% & 77.78\% & 72.22\% & 38.89\% & 100.00\% & 33.33\% & 55.56\% & \textbf{22.22\%} \\
\quad \quad GPT-4o-image & 15.79\% & 13.16\% & 18.42\% & 15.79\% & 42.11\% & 78.95\% & 60.53\% & 47.37\% & 13.16\% & 94.74\% & 7.89\% & 39.47\% & 7.89\% \\
\multicolumn{14}{@{}l}{\textbf{Trajectory Distance}} \\
\multicolumn{14}{@{}l}{\quad \textit{Level: short}} \\
\multicolumn{14}{@{}l}{\quad \textbf{Video Models}} \\
\quad \quad Veo-3 & 38.98\% & 40.68\% & 45.76\% & 45.76\% & 44.07\% & 61.02\% & 54.24\% & 64.41\% & 28.81\% & 62.71\% & 20.34\% & 42.37\% & 15.25\% \\
\quad \quad Sora-2 & 40.35\% & 31.58\% & 43.86\% & 40.35\% & 52.63\% & 85.96\% & 82.46\% & 75.44\% & 47.37\% & 89.47\% & 29.82\% & 66.67\% & \textbf{15.79\%} \\
\multicolumn{14}{@{}l}{\quad \textbf{Image Models}} \\
\quad \quad Nano-banana & 54.55\% & 51.52\% & 54.55\% & 54.55\% & 87.88\% & 90.91\% & 90.91\% & 90.91\% & 57.58\% & 100.00\% & 42.42\% & 78.79\% & \textbf{36.36\%} \\
\quad \quad GPT-4o-image & 27.12\% & 20.34\% & 28.81\% & 25.42\% & 55.93\% & 77.97\% & 62.71\% & 50.85\% & 28.81\% & 96.61\% & 18.64\% & 45.76\% & 18.64\% \\
\multicolumn{14}{@{}l}{\quad \textit{Level: long}} \\
\multicolumn{14}{@{}l}{\quad \textbf{Video Models}} \\
\quad \quad Veo-3 & 35.09\% & 42.11\% & 40.35\% & 45.61\% & 33.33\% & 63.16\% & 40.35\% & 56.14\% & 22.81\% & 40.35\% & 14.04\% & 35.09\% & 7.02\% \\
\quad \quad Sora-2 & 18.64\% & 28.81\% & 22.03\% & 35.59\% & 50.85\% & 81.36\% & 69.49\% & 66.10\% & 30.51\% & 79.66\% & 13.56\% & 55.93\% & \textbf{10.17\%} \\
\multicolumn{14}{@{}l}{\quad \textbf{Image Models}} \\
\quad \quad Nano-banana & 42.42\% & 45.45\% & 42.42\% & 54.55\% & 78.79\% & 78.79\% & 78.79\% & 57.58\% & 42.42\% & 96.97\% & 33.33\% & 51.52\% & \textbf{21.21\%} \\
\quad \quad GPT-4o-image & 20.34\% & 25.42\% & 22.03\% & 27.12\% & 54.24\% & 72.88\% & 62.71\% & 49.15\% & 15.25\% & 96.61\% & 13.56\% & 47.46\% & 13.56\% \\
\multicolumn{14}{@{}l}{\textbf{Destination Specification}} \\
\multicolumn{14}{@{}l}{\quad \textit{Level: color mark}} \\
\multicolumn{14}{@{}l}{\quad \textbf{Video Models}} \\
\quad \quad Veo-3 & 53.45\% & 51.72\% & 60.34\% & 56.90\% & 43.10\% & 56.90\% & 46.55\% & 65.52\% & 32.76\% & 67.24\% & 25.86\% & 34.48\% & 17.24\% \\
\quad \quad Sora-2 & 44.64\% & 46.43\% & 51.79\% & 57.14\% & 48.21\% & 82.14\% & 73.21\% & 67.86\% & 62.50\% & 92.86\% & 33.93\% & 58.93\% & \textbf{19.64\%} \\
\multicolumn{14}{@{}l}{\quad \textbf{Image Models}} \\
\quad \quad Nano-banana & 81.25\% & 59.38\% & 81.25\% & 68.75\% & 87.50\% & 84.38\% & 87.50\% & 78.12\% & 71.88\% & 100.00\% & 62.50\% & 71.88\% & \textbf{50.00\%} \\
\quad \quad GPT-4o-image & 44.83\% & 36.21\% & 44.83\% & 41.38\% & 55.17\% & 70.69\% & 62.07\% & 51.72\% & 37.93\% & 96.55\% & 31.03\% & 46.55\% & 31.03\% \\
\multicolumn{14}{@{}l}{\quad \textit{Level: location description}} \\
\multicolumn{14}{@{}l}{\quad \textbf{Video Models}} \\
\quad \quad Veo-3 & 20.69\% & 31.03\% & 25.86\% & 34.48\% & 34.48\% & 67.24\% & 48.28\% & 55.17\% & 18.97\% & 36.21\% & 8.62\% & 43.10\% & 5.17\% \\
\quad \quad Sora-2 & 15.00\% & 15.00\% & 15.00\% & 20.00\% & 55.00\% & 85.00\% & 78.33\% & 73.33\% & 16.67\% & 76.67\% & 10.00\% & 63.33\% & \textbf{6.67\%} \\
\multicolumn{14}{@{}l}{\quad \textbf{Image Models}} \\
\quad \quad Nano-banana & 17.65\% & 38.24\% & 17.65\% & 41.18\% & 79.41\% & 85.29\% & 82.35\% & 70.59\% & 29.41\% & 97.06\% & 14.71\% & 58.82\% & \textbf{8.82\%} \\
\quad \quad GPT-4o-image & 3.33\% & 10.00\% & 6.67\% & 11.67\% & 55.00\% & 80.00\% & 63.33\% & 48.33\% & 6.67\% & 96.67\% & 1.67\% & 46.67\% & 1.67\% \\
\bottomrule
\end{tabular} }
\end{table*}

\paragraph{The Logic vs. Motion Trade-off.} Quantitative evaluation again ranks \textbf{Nano-banana} as the strongest SLAG performer: it posts the top Holistic Metric on \textit{floor01} (27.78\%), edging out GPT-4o-image (25.00\%), while Veo-3 and Sora-2 remain far lower (11.86\% and 10.34\%).
\begin{itemize}
    \item \textbf{Trajectory Alignment Dominance:} Nano-banana achieves a dominant Trajectory Alignment Score of 88.89\% on \textit{floor01}. Video baselines trail sharply (Veo-3: 44.07\%, Sora-2: 55.17\%), underscoring that smooth motion does not guarantee coordinate-level alignment with the map.
    \item \textbf{Gate-Induced Failures for Video Models:} Despite respectable physical validity (Physics Validness: Veo-3 at 45.76\%, Sora-2 at 65.52\%), the \textbf{Success (3D) Original Destination} gate collapses to 18.64\% and 22.41\%, respectively, versus Nano-banana’s 38.89\%. Low destination integrity (20.34\%--32.76\% for video models) is the primary culprit behind their depressed holistic scores.
\end{itemize}

\paragraph{The Grounding Gap: Visual vs. Linguistic Complexity.} Performance across instruction types shows that linguistic grounding is a steeper barrier than visual grounding.
\begin{itemize}
    \item \textbf{Drastic Drop on Text Prompts:} When the destination is specified by a \textit{Color Mark}, Nano-banana excels (81.25\% 3D Success, 50.00\% Holistic). Switching to a \textit{Location Description} collapses 3D Success to 17.65\% and Holistic to 8.82\%.
    \item \textbf{Model Consistency:} GPT-4o-image mirrors this trend, sliding from 44.83\% to 3.33\% (3D Success) and from 31.03\% to 1.67\% (Holistic). Video models show the same gap (Veo-3: 17.24\% $\rightarrow$ 5.17\% Holistic; Sora-2: 19.64\% $\rightarrow$ 6.67\%), confirming that textual spatial instructions remain the weakest link.
\end{itemize}

\paragraph{Environmental Scaling.} As complexity increases from \textit{floor01} to \textit{floor02plus}:
\begin{itemize}
    \item \textbf{Nano-banana} loses geometric reliability (3D Success 55.56\% $\rightarrow$ 40.00\%) yet nudges its Holistic Metric upward (27.78\% $\rightarrow$ 30.00\%) thanks to stable Scene Consistency (100.00\% $\rightarrow$ 96.67\%) and stronger 2D alignment.
    \item \textbf{Sora-2} remains limited (Holistic 10.34\% $\rightarrow$ 15.52\%) despite high physical priors (Physics Validness 65.52\% $\rightarrow$ 56.90\%), indicating that added floors do not resolve its instruction-following issues.
    \item \textbf{GPT-4o-image} degrades sharply (Holistic 25.00\% $\rightarrow$ 6.90\%), revealing brittleness of single-image rollouts once multi-floor reasoning is required.
\end{itemize}

\subsubsection{View-Syn Evaluation}

\begin{table*}[h]
  \centering
  \caption{Simultaneous Localization and Generation — Video (FVD/SSIM/PSNR/LPIPS) and Image (FID normalized); higher is better except LPIPS}
  \begin{tabular}{lcccccc}
    \toprule
    Model & Modality & FVD Norm $\uparrow$ & FID Norm $\uparrow$ & SSIM $\uparrow$ & PSNR $\uparrow$ & LPIPS $\downarrow$ \\
    \midrule
    Sora-2          & Video & 0.7775 & --     & 0.4061 & 7.87 & 0.8281 \\
    Veo-3           & Video & 0.7097 & --     & 0.1612 & 6.92 & 0.8501 \\
    Wan-2.2         & Video & 0.5847 & --     & 0.3285 & 7.54 & 0.8182 \\
    GPT-image       & Image & --     & 0.7037 & --     & --   & --     \\
    Nano-banana     & Image & --     & 0.7037 & --     & --   & --     \\
    Nano-banana-pro & Image & --     & 0.6907 & --     & --   & --     \\
    Qwen-image      & Image & --     & 0.5375 & --     & --   & --     \\
    GPT-image-1.5   & Image & --     & 0.4356 & --     & --   & --     \\
    \bottomrule
  \end{tabular}
\end{table*}

\subsection{Case Study: Qualitative Failure Modes}


\begin{figure*}[htbp]
    \centering
    \includegraphics[width=0.8\textwidth]{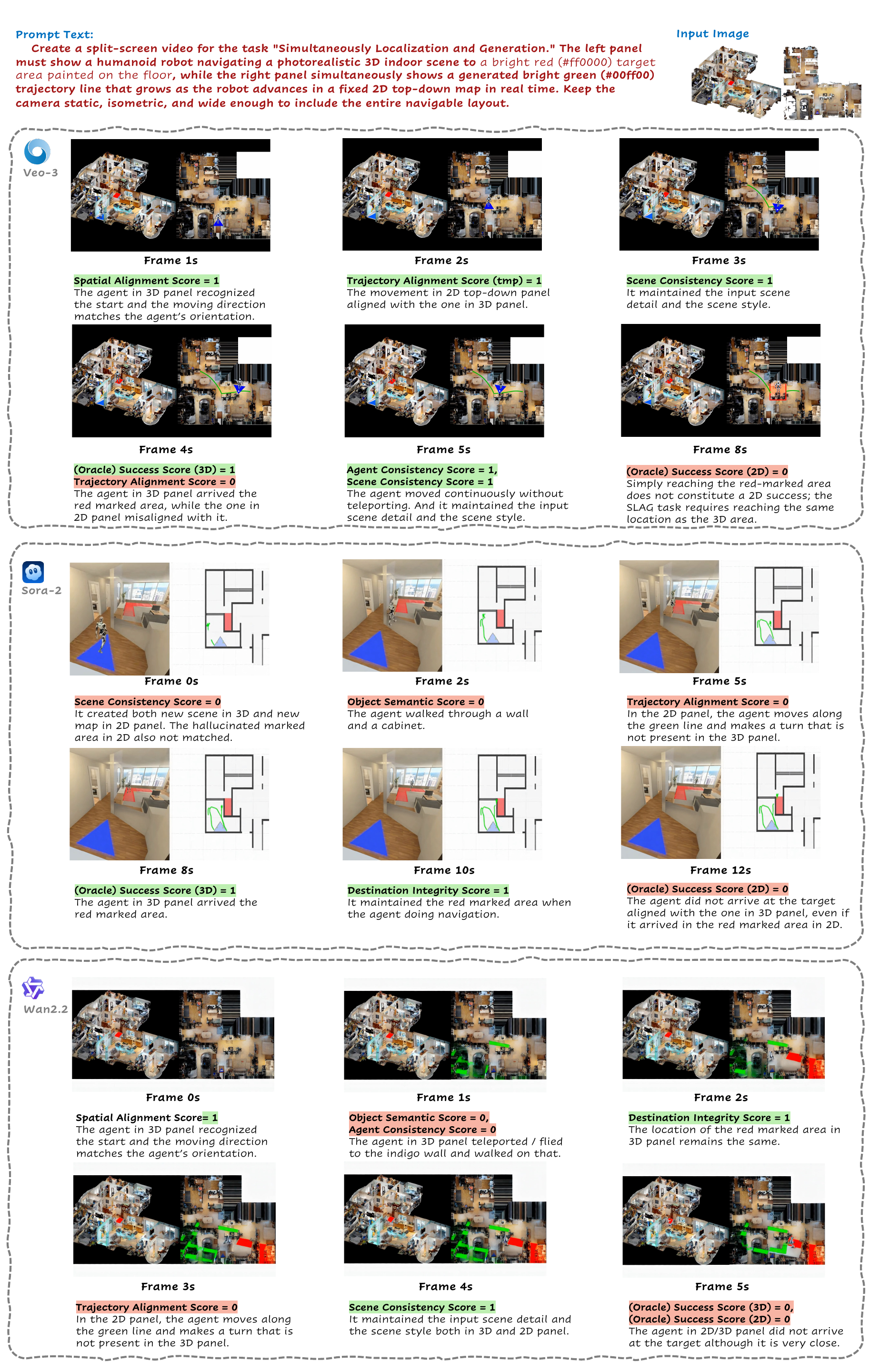}
    \caption{Case Study for \textbf{Simultaneous Localization and Generation}. The figure compares Veo-3, Sora-2, and Wan-2.2 on a split-screen navigation task requiring synchronization between a 3D view and a 2D map. Veo-3 achieves 3D target success but suffers from 2D trajectory misalignment. Sora-2 exhibits severe scene hallucination and physics violations (clipping through walls). Wan-2.2 demonstrates agent inconsistency (teleporting) and fails to reach the final target.}
    \label{fig:embodied_casestudy_slag}
\end{figure*}

The Simultaneous Localization and Generation task highlights a common failure mode across all three models in Figure \ref{fig:embodied_casestudy_slag}: \textbf{the inability to maintain semantic linkage between split-screen representations over time}.

\paragraph{Veo-3: High Fidelity, Low Logic.} Veo-3 offers the most stable visual experience, maintaining high scene consistency in this case study. The agent correctly identifies the start position and successfully reaches the red-marked goal in the 3D view. However, the \textbf{Trajectory Alignment} fails. As the 3D agent moves forward, the 2D dot on the map drifts independently, failing to mirror the agent's path. This exemplifies the ``disembodied'' nature of the generation—the model understands it needs to generate a map and a video, but treats them as separate artistic tasks rather than coupled data streams.

\paragraph{Sora-2: Severe Hallucination.} Sora-2 struggles to maintain the input reality. (1) \textbf{Scene Hallucination:} It immediately fails Scene Consistency ($Score=0$) by replacing the input floor plan with a hallucinated layout. (2) \textbf{Physics Violations:} Consistent with the low Physics Validness scores, the agent is observed clipping through a cabinet.(3) \textbf{Phantom Turns:} The 2D agent executes a turn that the 3D agent never makes, further emphasizing the lack of cross-modal attention in the model architecture.

\paragraph{Wan-2.2: The Teleportation Problem.} Wan-2.2 exhibits the most erratic behavior, leading to complete task failure. (1) \textbf{Agent Instability:} The model suffers from severe Agent Consistency failures ($Score=0$). At Frame 1s, the agent teleports—effectively ``flying''—onto an indigo wall, treating a vertical surface as a walkable floor. (2) \textbf{Navigation Failure:} Unlike Veo-3, Wan-2.2 fails to reach the destination in either view. The agent stops short, resulting in zero scores for both Success (2D) and Success (3D). This model demonstrates that without strong physical priors, video generation models devolve into surrealism rather than embodied simulation.

\section{Physical Commonsense}
\label{apx:physics}

We introduce the \textbf{Physical Commonsense} task to evaluate a model’s foundational understanding of ``intuitive physics'' \citep{battaglia2013simulation,yi2019clevrer,wu2015galileo}, a critical prerequisite for robust world modeling. This task probes \textbf{Physical Reasoning} by requiring models to generate videos that combine photorealism with physical plausibility \citep{bear2021physion, riochet2021intphys}. Beyond static visual fidelity, the task evaluates whether the model captures essential principles—such as gravity, momentum, collisions, and material properties (rigidity, fluid dynamics)—and correctly models causal dynamics \citep{bakhtin2019phyre, allen2020rapid}. The evaluation spans both \textbf{3D Spatial Reasoning} (spatial object interactions) and \textbf{Temporal Reasoning} (sequential cause-and-effect), organized along two complementary axes: (1) \textbf{Physical Concepts}, which tests fundamental principles using the VideoPhy ontology \citep{videophy2024}; and (2) \textbf{Sports Scenarios}, which probe compositional reasoning through dynamic, high-velocity human movements.

\subsection{Data Sources and Task Structure}

\textbf{Physical Concepts (Fundamental Interactions).} To systematically evaluate atomic physical principles, we adopt the structured ontology from VideoPhy~\citep{videophy2024} and VideoPhy-2~\citep{bansal2025videophy2}. We draw from a diverse pool of captioned interactions, including: \textbf{Solid–Solid} (\textit{e.g.}, rigid collisions, stacking), \textbf{Solid–Fluid} (\textit{e.g.}, splashing, buoyancy), and \textbf{Fluid–Fluid} (\textit{e.g.}, diffusion, mixing). These prompts isolate specific physical laws, allowing us to test statics, dynamics, and kinematics in controlled environments.

\textbf{Sports Scenarios (Compositional Reasoning).} To evaluate physical reasoning in complex, real-world contexts, we synthesize a complementary dataset of sports-oriented prompts. These scenarios naturally require the integration of multiple physical laws simultaneously. The dataset spans: \textbf{Precision \& Arts} (\textit{e.g.}, ballet pirouettes requiring angular momentum), \textbf{Winter Sports} (\textit{e.g.}, skiing moguls involving friction and gravity), \textbf{Aquatics} (\textit{e.g.}, diving and swimming involving fluid resistance), and \textbf{Athletics}. These prompts test the model's ability to maintain physical consistency during complex human-object interactions.

\subsection{Hard-Level Control and Evaluation Taxonomy}

To ensure a rigorous assessment, we curate a balanced evaluation set of 50 samples (\Cref{tab:physics_distribution}), stratified along \textit{three} dimensions of difficulty:
\vspace{-2mm}
\begin{itemize}
    \item \textbf{Interaction Type (The ``What''):} We categorize samples by the material properties involved:
    \begin{itemize}
        \item \textit{Solid-Solid:} Interactions between rigid bodies (testing impulse, friction, and collision response).
        \item \textit{Solid-Fluid:} Interactions between solids and liquids (testing displacement, splashing, and floating).
        \item \textit{Fluid-Fluid:} Dynamics of miscible and immiscible fluids (testing viscosity, mixing, and turbulence).
    \end{itemize}

    \item \textbf{Scenario Context (The ``Where''):} We distinguish between controlled physics experiments (Physical Concepts) and unconstrained, dynamic environments (Sports Scenarios), aiming to test generalization from atomic laws to complex behaviors.

    \item \textbf{Interaction Complexity (The ``How''):} Across all categories, we vary the complexity level:
    \begin{itemize}
        \item \textit{Simple:} Single-object motion or static equilibrium.
        \item \textit{Complex:} Multi-object interactions with simultaneous forces.
        \item \textit{Chain-Reaction:} Causal sequences where an initial action triggers cascading effects.
    \end{itemize}
\end{itemize}

\begin{table*}[h]
\centering
\caption{Distribution of Physical Commonsense evaluation samples. The set is balanced to weigh fundamental physical understanding equally against compositional real-world application.}
\label{tab:physics_distribution}
\small
\begin{tabular}{@{}lc@{}}
\toprule
\textbf{Task Axis} & \textbf{Total Samples} \\
\midrule
Physical Concepts (Atomic Interactions) & 25 \\
Sports Scenarios (Compositional Contexts) & 25 \\
\midrule
\textbf{Total} & \textbf{50} \\
\bottomrule
\end{tabular}
\end{table*}

\subsection{Evaluation and Metrics}
\label{sec:physics_eval}

Modeling physical commonsense inherently requires capturing temporal dynamics such as force propagation, momentum transfer, and continuous motion. Because static image generators lack temporal modeling capabilities and cannot represent causal interactions unfolding over time, they are fundamentally limited in assessing physical plausibility. Therefore, our evaluation focuses exclusively on \emph{video generative models}, capable of producing temporally coherent sequences \citep{cai2025preferences}.\looseness=-1

To ensure consistent and structured evaluation, we employ a Vision–Language Model (VLM) evaluator, Gemini-2.5-Pro \citep{comanici2025gemini}, prompted to act as a physics-aware video analysis expert. For each generated video, the VLM evaluates quality along \textit{four} fine-grained dimensions. To promote comparability and reduce subjectivity, each dimension is operationalized as a binary metric (0/1), where a score of 1 indicates full compliance with the criterion and 0 denotes a clear failure or substantive violation. Based on these assessments, we further derive \textit{one} primary aggregate metric, defined as follows:\looseness=-1
\vspace{-2mm}
\begin{itemize}
    \item \textbf{Physics Accuracy:} Evaluates whether the generated video strictly obeys fundamental physical laws. The model checks if motion adheres to gravity, momentum, and friction, and verifies that object interactions are plausible. A score of 0 is assigned if there are violations such as objects floating, moving at unrealistic speeds for the context, displaying incorrect trajectories, or deviating from the scenario's ``Physics Focus.''
    \item \textbf{Motion Quality:} Assesses the temporal coherence and naturalness of the movement. This metric verifies that the motion follows the expected pattern described in the scenario and remains smooth and continuous. A score of 0 is assigned if the motion is jerky, exhibits unnatural accelerations, or suffers from temporal discontinuities and inconsistency.
    \item \textbf{Visual Realism:} Measures the visual fidelity and believability of the scene. The model checks if objects and materials appear realistic, whether lighting and shadows are consistent, and if the scene composition is plausible. A score of 0 is assigned if there are significant visual artifacts, glitches, or if the scene lacks photorealism.
    \item \textbf{Prompt Adherence:} Determines whether the video semantically matches the user input. This metric verifies that all key elements (objects, setting) are present and that the specific action described actually occurs. A score of 0 is assigned if there are significant mismatches between the generated content and the text prompt.
    \item \textbf{Overall Success:} To provide a holistic measure of generation capability, we compute a strict \emph{Overall Success} score. A generated video is marked as successful (1) \emph{if and only if} it satisfies all four fine-grained metrics simultaneously.
\end{itemize}

This rigorous aggregation ensures that high performance requires a model to generate videos that are not only physically accurate but also visually coherent, smooth, and semantically correct.

\begin{figure*}[h!]
    \centering
    \includegraphics[width=\textwidth, trim=0 0 0 0, clip]{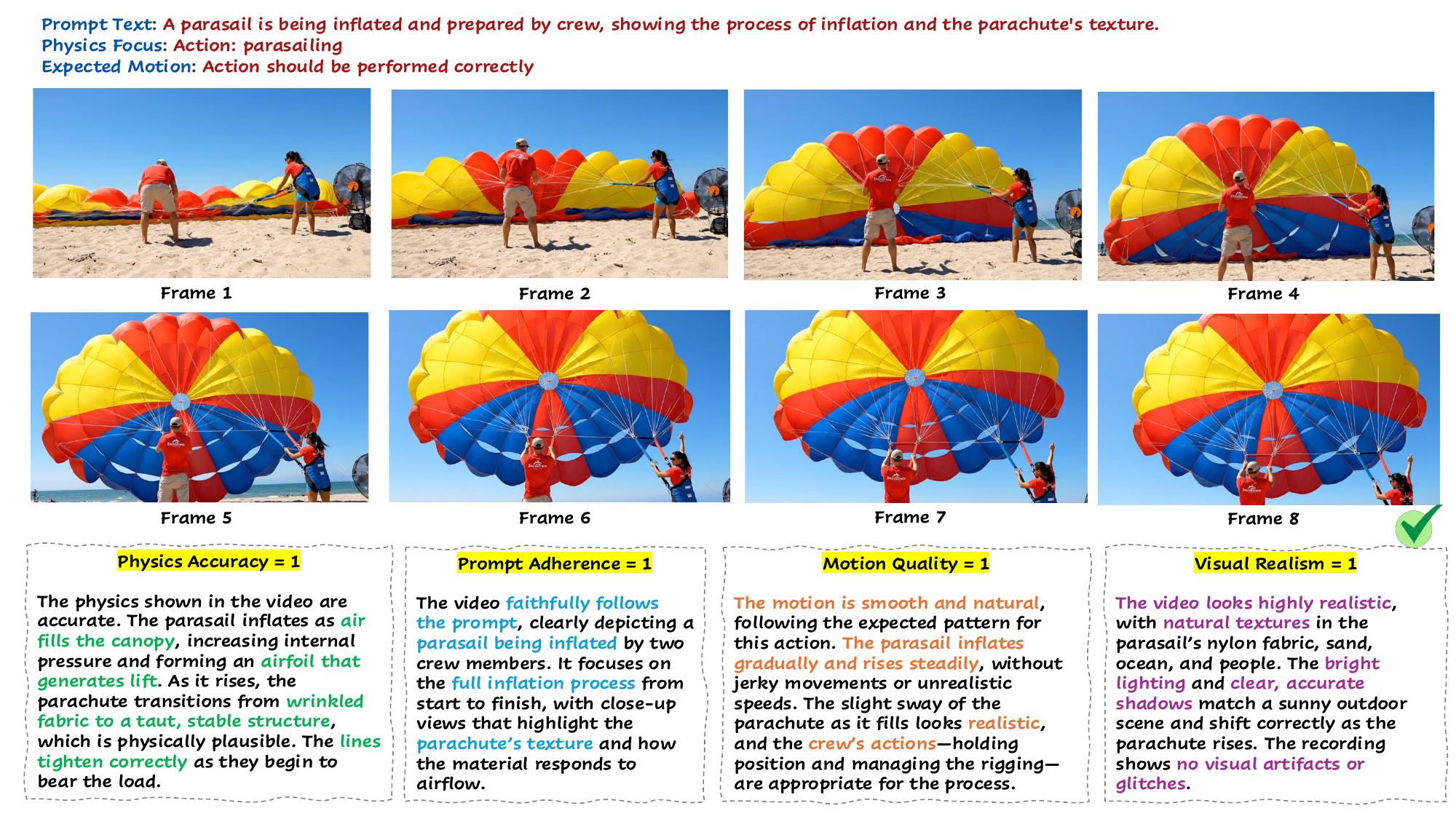}
    \vspace{-3mm}
    \caption{Case Study: Success case generated by \textbf{Veo-3}.  Physically Plausible Parachute Inflation.}
    \label{fig:physics_case_study_1}
\end{figure*}

\begin{figure*}[h!]
    \centering
    \includegraphics[width=\textwidth, trim=0 0 0 0, clip]{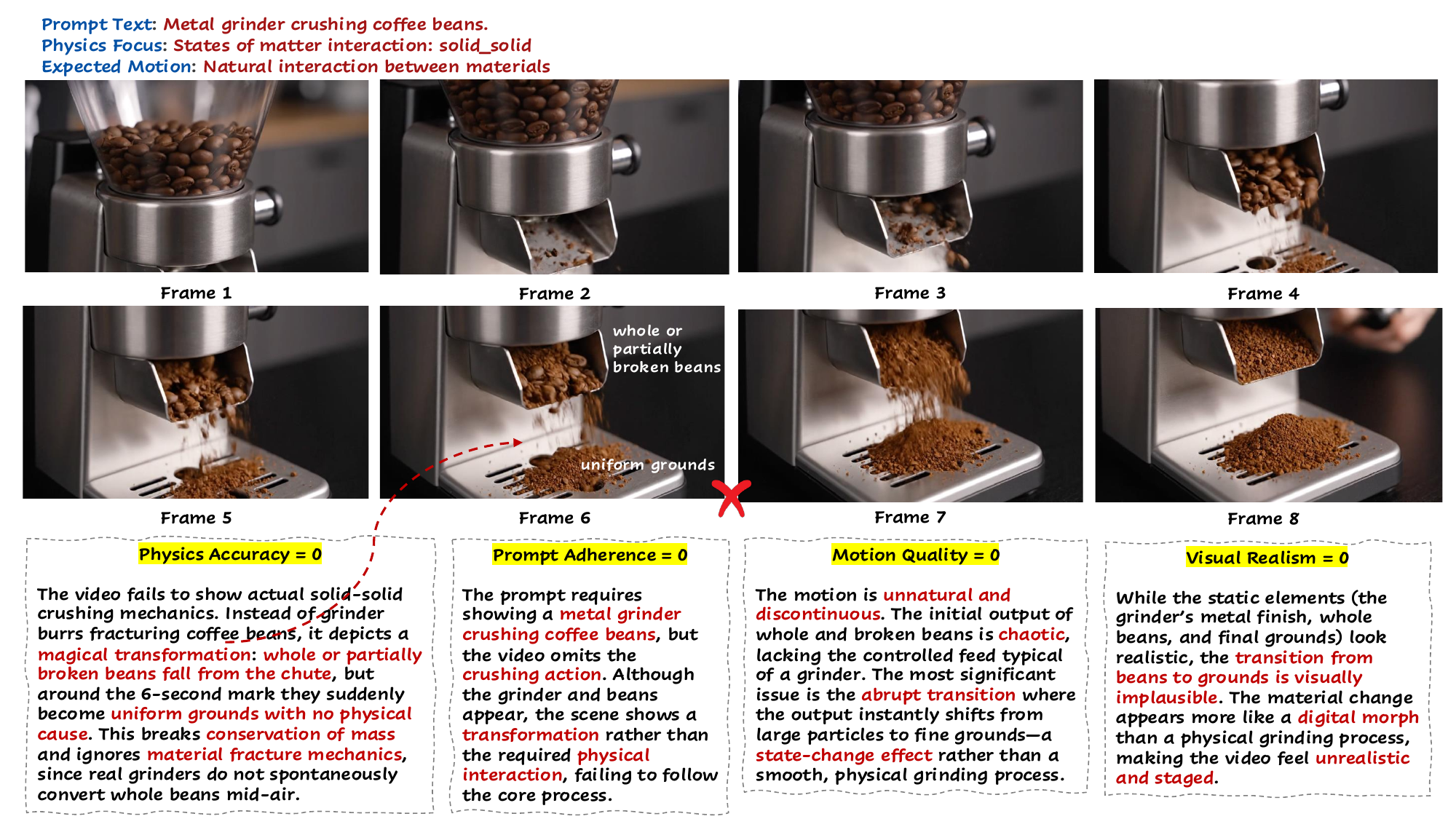}
    \caption{Case Study: Failure case generated by \textbf{Veo-3}. Missing Solid–Solid Interaction in Coffee Grinding.}
    \vspace{-3mm}
    \label{fig:physics_case_study_2}
\end{figure*}

\begin{figure*}[h!]
    \centering
    \includegraphics[width=\textwidth, trim=0 0 0 0, clip]{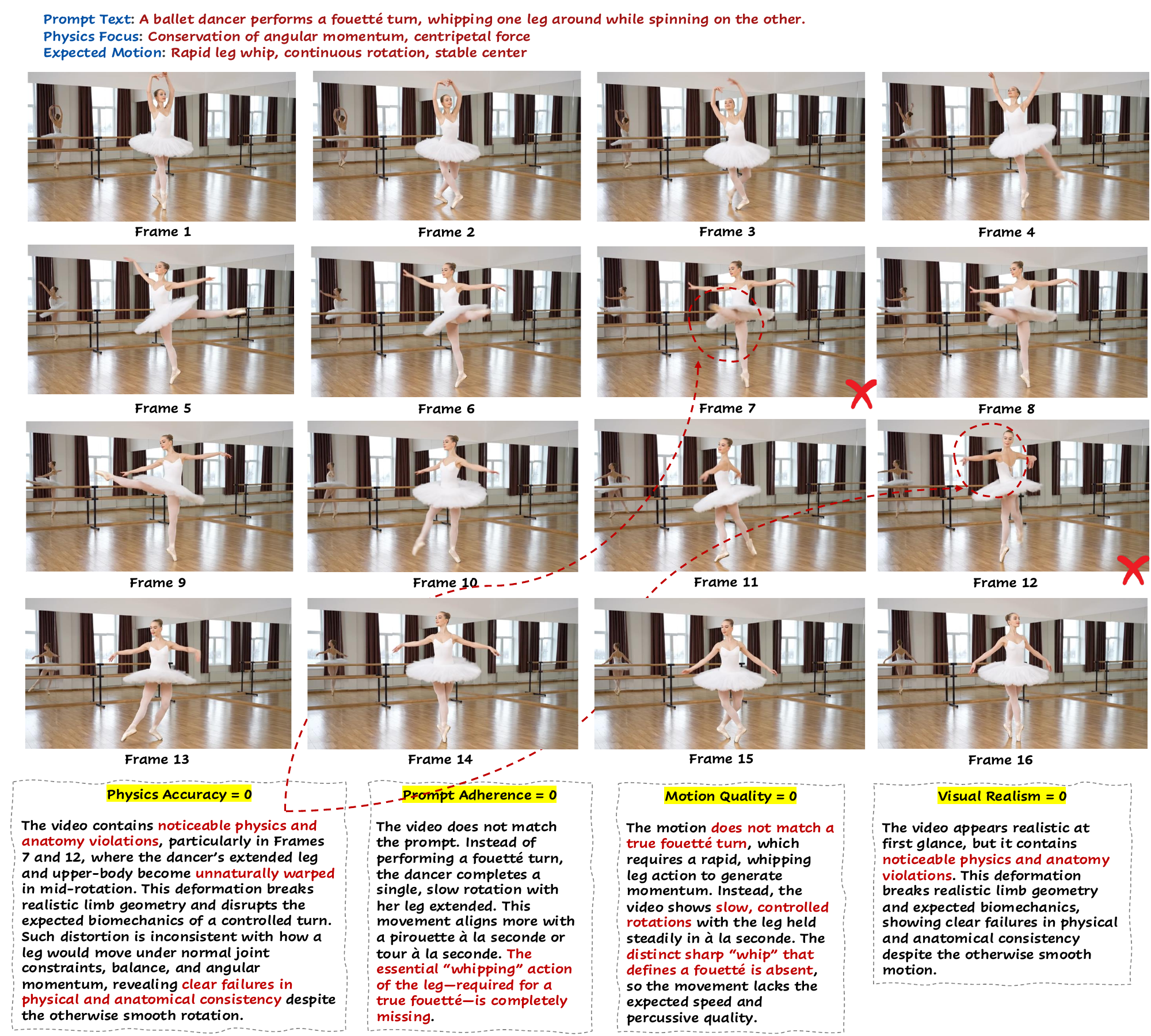}
    \caption{Case Study: Failure case generated by \textbf{Veo-3}. Incorrect Angular Momentum Dynamics in Ballet Rotation.}
    \vspace{-3mm}
    \label{fig:physics_case_study_3}
\end{figure*}

\subsection{Case Study}

To better illustrate the strengths and limitations revealed by our Physical Commonsense evaluation, we present three representative case studies generated by \textbf{Veo-3} (\Cref{fig:physics_case_study_1}, \ref{fig:physics_case_study_2}, and \ref{fig:physics_case_study_3}). Each example highlights how our metrics disentangle distinct dimensions of \textbf{physical reasoning}: \emph{physics accuracy}, \emph{prompt adherence}, \emph{motion quality}, and \emph{visual realism}. These cases further demonstrate that high visual fidelity alone is not indicative of correct physical behavior—underscoring why physics-focused evaluation is necessary for robust real-world video generation.

\textbf{Success Case: Physically Plausible Parachute Inflation.} \Cref{fig:physics_case_study_1} shows a prompt involving the inflation of a parachute by a two-person crew. Veo-3 performs strongly across all evaluation dimensions, achieving full scores on physics accuracy, prompt adherence, motion quality, and visual realism. The model captures the physical mechanics of inflation: air fills the canopy, internal pressure increases, the fabric transitions from wrinkled to taut, and the lines tighten appropriately as load is applied. The motion unfolds smoothly and continuously, without discontinuities or unnatural accelerations. Detailed textures, realistic lighting, and consistent shading further contribute to the clip’s visual plausibility. This example illustrates Veo-3’s ability to correctly represent gradual force buildup, material deformation, and multi-agent coordination.

\textbf{Failure Case I: Missing Solid–Solid Interaction in Coffee Grinding.} \Cref{fig:physics_case_study_2} highlights a failure case involving a metal grinder crushing coffee beans (Solid–Solid interaction). While the static frames resemble a real grinder setup, the temporal dynamics violate fundamental physical laws. Instead of showing a realistic grinding process—where rigid beans fracture into progressively smaller particles—the video abruptly transitions from whole beans to uniform fine grounds, without any mechanical cause. This discontinuous ``state-change'' effect resembles a visual morph rather than a physical transformation, resulting in violations of mass conservation and material fracture mechanics. Although the model adheres partially to the prompt in terms of objects present, it fails to depict the required interaction, leading to zeros in physics accuracy, prompt adherence, motion quality, and visual realism.

\textbf{Failure Case II: Incorrect Angular Momentum Dynamics in Ballet Rotation.} \Cref{fig:physics_case_study_3} presents a ballet scenario requiring a fouetté turn, a physically demanding movement involving rapid leg whipping to generate angular momentum and sustain rotation. Veo-3 fails across all physics-focused dimensions. The dancer’s extended leg and upper body become unnaturally distorted mid-rotation, deviating from realistic human biomechanics. The model also misinterprets the action: instead of producing a sharp, continuous whipping motion that drives a true fouetté, the dancer performs a slow, controlled rotation with no momentum-generating movement. This breaks both prompt adherence and key physical principles such as angular momentum conservation and joint kinematic constraints. Although the visual appearance remains high-quality at a glance, closer inspection reveals anatomy inconsistencies and motion artifacts that undermine physical realism.

\textbf{Takeaways.} These case studies highlight the need to evaluate video models beyond photorealism. \textbf{Veo-3 can produce visually convincing clips, but frequently fails in scenarios requiring nuanced physical reasoning}—especially where material properties, continuous force propagation, or human biomechanics play a central role. The structured metrics in our Physical Commonsense task make these failure modes explicit, providing a robust diagnostic tool for guiding future model improvements.

\subsection{Evaluation Results}

\begin{tcolorbox}[colback=gray!10, colframe=gray!50, title=\textbf{Key Finding: Physical Plausibility Is Decoupled from Visual Realism}]
\textbf{Sora-2 achieves the strongest physical commonsense performance}, leading both final task rows: 76.00\% on Physical Concepts and 64.00\% on Sports. Averaged across the two physical subsets, this corresponds to 70.00\%, ahead of Veo-3 (51.02\%) and Wan-2.2 (24.00\%). Crucially, the results reveal a consistent pattern: \textbf{high visual realism does not imply correct physical reasoning}. Wan-2.2 produces highly photorealistic videos (84--96\% Visual Realism) yet fails to follow prompts and physical constraints (24\% aggregate Overall), indicating a fundamental gap between appearance-level fidelity and physically grounded world modeling. Fine-grained analysis reveals \textbf{Sports Scenarios are systematically easier than Physical Concepts} across most models, while \textbf{Solid-Solid interactions prove most challenging}—Veo-3 achieves 0\% success on rigid body collisions compared to 75\% on Solid-Fluid interactions.
\end{tcolorbox}

\begin{table*}[h!]
\centering
\small
\caption{Quantitative results for the \textbf{Physical Commonsense} task. We evaluate three video generative models (Veo-3, Sora-2, and Wan-2.2) on \textbf{Physical Concepts} and \textbf{Sports Scenarios}.}
\label{tab:physics_results}
\begin{adjustbox}{max width=0.85\textwidth}
{
\begin{tabular}{@{}lccccc@{}}
\toprule
& \multicolumn{4}{c}{\textbf{Fine-grained Metrics}} & \textbf{Primary Metric} \\
\cmidrule(lr){2-5}
\textbf{Model} & \textbf{Physics Accuracy} $\uparrow$ & \textbf{Motion Quality} $\uparrow$ & \textbf{Visual Realism} $\uparrow$ & \textbf{Prompt Adherence} $\uparrow$ & \textbf{Overall} $\uparrow$ \\
\midrule
\multicolumn{6}{@{}l}{\textbf{Scenario Type: Physical Concepts}} \\
\quad Veo-3 & 62.50\% & 54.17\% & 83.33\% & 50.00\% & 41.67\% \\
\quad Sora-2 & 84.00\% & 80.00\% & 96.00\% & 76.00\% & \textbf{76.00\%} \\
\quad Wan-2.2 & 58.67\% & 53.33\% & 72.00\% & 38.67\% & 26.67\% \\
\midrule
\multicolumn{6}{@{}l}{\textbf{Scenario Type: Sports Scenarios}} \\
\quad Veo-3 & 80.00\% & 68.00\% & 92.00\% & 68.00\% & 60.00\% \\
\quad Sora-2 & 88.00\% & 72.00\% & 88.00\% & 68.00\% & \textbf{64.00\%} \\
\quad Wan-2.2 & 42.67\% & 33.33\% & 96.00\% & 21.33\% & 21.33\% \\
\midrule
\multicolumn{6}{@{}l}{\textbf{Average}} \\
\quad Veo-3 & 71.43\% & 61.22\% & 87.76\% & 59.18\% & 51.02\% \\
\quad Sora-2 & 86.00\% & 76.00\% & 92.00\% & 72.00\% & \textbf{70.00\%} \\
\quad Wan-2.2 & 50.67\% & 43.33\% & 84.00\% & 30.00\% & 24.00\% \\
\bottomrule
\end{tabular}
}
\end{adjustbox}
\end{table*}

\subsubsection{VLM-Based Evaluation}

\Cref{tab:physics_results} reports quantitative results on the Physical Commonsense task, evaluated using Gemini-2.5-Pro~\citep{comanici2025gemini} as a VLM-based evaluator. We compare three state-of-the-art video generation models—Veo-3, Sora-2, and Wan-2.2—across two complementary scenario types: \textbf{Physical Concepts} and \textbf{Sports Scenarios}.

\textbf{Overall Performance Trends.} Sora-2 consistently outperforms competing models, achieving the highest aggregate success rate across the two physical subsets (70.00\%) and leading across all fine-grained metrics. The two task-level rows reported in \Cref{tab:main_results} remain Physical Concepts and Sports separately, while the Average row here is a diagnostic aggregate. In contrast, Veo-3 exhibits moderate but uneven performance (51.02\% aggregate), while Wan-2.2 substantially underperforms (24.00\% aggregate) despite strong visual realism. Notably, \textbf{all models score highly on Visual Realism} (84--92\%), yet exhibit large variance in Physics Accuracy (50.67--86.00\%) and Prompt Adherence (30.00--72.00\%), reinforcing that perceptual quality alone is an unreliable indicator of physical correctness.

\textbf{Scenario-Specific Difficulty.} Across all models, Sports Scenarios are consistently easier than Physical Concepts. Veo-3 improves from 41.67\% overall success on Physical Concepts to 60.00\% on Sports Scenarios, while Sora-2 maintains strong performance across both (76.00\% vs.\ 64.00\%). In contrast, Wan-2.2 fails to generalize even in Sports Scenarios, achieving only 21.33\% overall success. These results suggest that contemporary video models benefit from learned biomechanical motion patterns in human-centric activities, while struggling with fine-grained material interactions such as collisions, splashing, and mixing.

\begin{table*}[h!]
\centering
\small
\caption{Fine-grained VLM-based evaluation results by \textbf{Sports Scenarios} attributes. Overall success rate (\%) is reported across sport types (Ballet, Diving, Skiing, Swimming) and difficulty levels (Easy, Medium, Hard).}
\label{tab:physics_sports_finegrained}
\begin{adjustbox}{max width=0.6\textwidth}
\begin{tabular}{@{}l|cccc|ccc@{}}
\toprule
& \multicolumn{4}{c|}{\textbf{By Sport Type}} & \multicolumn{3}{c}{\textbf{By Difficulty}} \\
\textbf{Model} & Ballet & Diving & Skiing & Swimming & Easy & Medium & Hard \\
\midrule
Veo-3 & 33.3\% & 50.0\% & 71.4\% & \textbf{83.3\%} & 60.0\% & 62.5\% & 57.1\% \\
Sora-2 & 33.3\% & 50.0\% & \textbf{85.7\%} & \textbf{83.3\%} & 60.0\% & \textbf{75.0\%} & 57.1\% \\
Wan-2.2 & 44.4\% & 0.0\% & 28.6\% & 11.1\% & 16.7\% & 33.3\% & 14.3\% \\
\bottomrule
\end{tabular}
\end{adjustbox}
\end{table*}

\begin{table*}[h!]
\centering
\small
\caption{Fine-grained VLM-based evaluation results by \textbf{Physical Concepts} attributes. Overall success rate (\%) is reported across states-of-matter interaction types and difficulty levels.}
\label{tab:physics_videophy_finegrained}
\begin{adjustbox}{max width=0.65\textwidth}
\begin{tabular}{@{}l|cccc|cc@{}}
\toprule
& \multicolumn{4}{c|}{\textbf{By States of Matter}} & \multicolumn{2}{c}{\textbf{By Difficulty}} \\
\textbf{Model} & Solid-Solid & Solid-Fluid & Fluid-Fluid & Action/Other & Easy & Hard \\
\midrule
Veo-3 & 0.0\% & 75.0\% & 50.0\% & 40.0\% & 53.3\% & 22.2\% \\
Sora-2 & \textbf{100.0\%} & \textbf{75.0\%} & \textbf{100.0\%} & \textbf{66.7\%} & \textbf{75.0\%} & \textbf{77.8\%} \\
Wan-2.2 & 33.3\% & 25.0\% & 83.3\% & 17.8\% & 35.4\% & 11.1\% \\
\bottomrule
\end{tabular}
\end{adjustbox}
\end{table*}


\textbf{Sport-Specific Patterns.} Fine-grained analysis (\Cref{tab:physics_sports_finegrained}) reveals sport-specific challenges. \textbf{Ballet proves most difficult} for all models (33--44\% success), while Swimming achieves the highest scores (83\% for Veo-3 and Sora-2). Wan-2.2 fails completely on Diving (0\%) yet achieves its best performance on Ballet (44.4\%), suggesting model-specific biases in motion priors. By difficulty level, Sora-2 shows the most consistent performance across Easy (60\%), Medium (75\%), and Hard (57\%) prompts, while Wan-2.2 collapses uniformly across all difficulty levels (14--33\%).

\textbf{States-of-Matter Analysis.} \Cref{tab:physics_videophy_finegrained} reveals interaction-specific challenges. \textbf{Solid-Solid interactions prove most difficult}: Veo-3 achieves 0\% success on rigid body collisions, while Sora-2 achieves perfect 100\%. Fluid-Fluid interactions show more variance, with Wan-2.2 achieving 83.3\%—its highest category score—while Veo-3 scores only 50\%. By difficulty level, all models show degradation on hard cases, with Wan-2.2 exhibiting the most severe collapse (35.4\% Easy → 11.1\% Hard).\looseness=-1

\textbf{The Visual–Physical Disconnect.} Wan-2.2 demonstrates the clearest dissociation between visual quality and physical reasoning. Despite achieving the highest Visual Realism score on Sports Scenarios (96.00\%), it records the lowest Physics Accuracy (42.67\%) and Prompt Adherence (21.33\%) in the same setting. This failure mode suggests that the model prioritizes surface-level appearance over causal and physical consistency, producing videos that ``look right'' but violate core physical principles.

\end{document}